\ifdefined\pdfminorversion
\fi
\documentclass[12pt,a4paper,oneside,openany]{book}

\newcommand{\E}{\mathbb{E}}
\newcommand{\R}{\mathbb{R}}
\newcommand{\Qb}{\mathbb{Q}}
\newcommand{\Pb}{\mathbb{P}}
\newcommand{\Ub}{\mathbb{U}}
\newcommand{\calH}{\mathcal{H}}

\newcommand{\calT}{\mathcal{T}}
\newcommand{\calL}{\mathcal{L}}
\newcommand{\calI}{\mathcal{I}}
\newcommand{\calF}{\mathcal{F}}

\newcommand{\calU}{\mathcal{U}}
\newcommand{\calZ}{\mathcal{Z}}
\newcommand{\calD}{\mathcal{D}}
\newcommand{\calO}{\mathcal{O}}
\newcommand{\calY}{\mathcal{Y}}
\newcommand{\calX}{\mathcal{X}}
\newcommand{\calP}{\mathcal{P}}
\newcommand{\calN}{\mathcal{N}}
\newcommand{\calJ}{\mathcal{J}}

\newcommand{\Gmap}{G}
\newcommand{\dd}{\mathrm{d}}
\newcommand{\md}{\mathrm{d}}
\AtBeginDocument{%

}

\newcommand{\bH}{\mathbf{H}}

\newcommand{\bJ}{\mathbf{J}}
\newcommand{\one}{\mathbf{1}}

\newcommand{\mmd}{\operatorname{MMD}}
\newcommand{\MMD}{\operatorname{MMD}}
\newcommand{\dmmd}{\operatorname{DrMMD}}
\newcommand{\kale}{\operatorname{KALE}}
\newcommand{\Hk}{\mathcal{H}}
\newcommand{\ps}[1]{\langle #1 \rangle}
\newcommand{\PSi}[2]{{\left\langle #1 \right\rangle}_{#2}}

\newcommand{\id}{\iota}
\newcommand{\cP}{\mathcal{P}}
\newcommand{\cF}{\mathcal{F}}
\newcommand{\argmin}{\mathop{\mathrm{arg\,min}}}

\makeatletter
\newcommand{\customlabel}[2]{%
   \protected@write \@auxout {}{\string \newlabel {#1}{{#2}{\thepage}{#2}{#1}{}} }%
   \hypertarget{#1}{}
}
\makeatother
\newcommand{\nkq}{\text{NKQ}}

\newcommand{\kq}{\text{KQ}}
\newcommand{\ckq}{\text{CKQ}}

\newcommand{\Id}{\mathrm{Id}}
\newcommand{\N}{\mathbb{N}}
\newcommand{\kl}{\mathrm{KL}}

\newcommand{\pgd}{\mathrm{PGD}}
\newcommand{\gd}{\mathrm{GD}}

\newcommand{\tildePtheta}[1]{\Pb_{\theta_{#1,m}}^{\pgd}}
\newcommand{\tildePthetanopgd}[1]{\Pb_{\theta_{#1,m}}}
\newcommand{\tildetheta}[1]{\Delta \theta_{#1,m}^\pgd}
\newcommand{\widehattheta}[1]{\widehat{\Delta\theta_{#1, m}^\pgd}}

\newcommand{\acks}[1]{}

\usepackage[T1]{fontenc}
\usepackage[utf8]{inputenc}
\usepackage{lmodern}
\usepackage{CJKutf8}

\usepackage[a4paper,textwidth=140mm,textheight=232mm,centering]{geometry}

\usepackage{emptypage}

\usepackage{graphicx}
\usepackage{microtype}
\graphicspath{{nkq/}{drmmd/}{mmdal/}}
\usepackage{xcolor}
\usepackage{tikz}
\usepackage{wrapfig}
\definecolor{blueviolet}{rgb}{0.541, 0.169, 0.886}
\definecolor{brown}{rgb}{0.647, 0.165, 0.165}
\definecolor{paperboxbg}{HTML}{F7F7F7}
\definecolor{paperboxborder}{HTML}{444444}
\usepackage{float}
\usepackage[chapter]{algorithm}
\usepackage{algorithmic}
\algsetup{indent=1.5em}

\usepackage{amsmath}
\allowdisplaybreaks[3]
\usepackage{mathtools}
\usepackage{amssymb}
\usepackage{amsthm}
\newtheorem{thm}{Theorem}[chapter]
\newtheorem{theorem}[thm]{Theorem}
\newtheorem{cor}[thm]{Corollary}
\newtheorem{prop}[thm]{Proposition}
\newtheorem{lem}[thm]{Lemma}
\newtheorem{defi}[thm]{Definition}
\newtheorem{assumption}[thm]{Assumption}
\newtheorem{ass}[thm]{Assumption}
\newtheorem{example}[thm]{Example}
\newenvironment{talign}{\begin{equation}}{\end{equation}}
\newenvironment{innercustomassumption}{\begin{assumption}}{\end{assumption}}
\theoremstyle{remark}
\newtheorem{rem}[thm]{Remark}

\usepackage{setspace}
\usepackage{enumitem}

\usepackage{array}
\usepackage{booktabs}
\usepackage{multirow}

\usepackage[authoryear,round]{natbib}

\usepackage[format=hang,font=small,labelfont=bf]{caption}
\usepackage{subcaption}

\usepackage{etoolbox}

\definecolor{mydarkblue}{rgb}{0,0.08,0.45}
\usepackage[
    pdfpagemode=UseNone,
    colorlinks=true,
    pdfview=FitH
]{hyperref}
\usepackage[nameinlink,capitalize,noabbrev]{cleveref}
\crefname{thm}{theorem}{theorems}
\Crefname{thm}{Theorem}{Theorems}
\crefname{cor}{corollary}{corollaries}
\Crefname{cor}{Corollary}{Corollaries}
\crefname{prop}{proposition}{propositions}
\Crefname{prop}{Proposition}{Propositions}
\crefname{lem}{lemma}{lemmas}
\Crefname{lem}{Lemma}{Lemmas}
\crefname{rem}{remark}{remarks}
\Crefname{rem}{Remark}{Remarks}
\crefname{assumption}{assumption}{assumptions}
\Crefname{assumption}{Assumption}{Assumptions}
\crefname{algorithm}{algorithm}{algorithms}
\Crefname{algorithm}{Algorithm}{Algorithms}
\makeatletter
\AtBeginDocument{
    \hypersetup{%
        pdftitle={MiniMMD: Minimization of MMD for Sampling, Inference and Integration},
        pdfauthor={Zonghao Chen},
        pdfsubject={},
        pdfkeywords={},
        pdfborder=0 0 0,
        linkcolor=black,
        citecolor=black,
        filecolor=black,
        urlcolor=black
    }
}
\makeatother

\begin{document}

\pagestyle{plain}
\raggedbottom
\hbadness=5000

\title{MiniMMD: Minimization of MMD for Sampling, Inference and Integration}
\author{Zonghao Chen}
\date{PhD thesis\\2026}

\hypersetup{pageanchor=false}
\maketitle
\hypersetup{pageanchor=true}

\chapter*{Abstract}
Many problems in computational statistics and machine learning can be formulated as approximating a target distribution by minimizing a discrepancy over a restricted class of candidate distributions.
This minimum-discrepancy viewpoint appears, for example, in numerical integration, sampling, generative modelling, and statistical inference, but these instances are often studied separately in the literature.
This thesis develops the first unified treatment of these many problems through the minimization of the maximum mean discrepancy (MMD), a kernel-based discrepancy between probability measures.
The central theme is that MMD is not only a statistically meaningful distance between probability distributions, but also a tractable optimization objective to design practical and efficient algorithms.
This thesis summarizes the recent theoretical and empirical advancements of such minimum-MMD based algorithms.




\chapter*{Acknowledgements}
I would like to thank my supervisors, Fran{\c{c}}ois-Xavier Briol and Arthur Gretton, for their guidance on how to do sound research, how to become an independent researcher, and how to develop good taste in scientific problems.
Arthur told me at the start of my PhD that the best way to learn a field is to write a paper on it; this has motivated me to explore many different areas.
FX taught me to write with the audience always in mind; this has also become a habit of mine, although I am still working hard to do it as well as he does.
Beyond research, I have learned a great deal from both of you about academic life more broadly.
You have been the best possible combination of supervisors I could have hoped for.
I am also grateful to Zhu Michael Li and Pierre Glaser, who acted as unofficial supervisors at the beginning of my PhD and taught me how to read and write proper proofs, almost hand by hand. 
Without their tutorials, my understanding of kernel methods are largely limited to linear algebra. 
I am especially grateful to Zhu Michael Li for his introducing me to many different pubs and restaurants around the Gatsby building.
I would also like to thank Anna Korba, Lester Mackey, Chris Oates, Bharath K. Sriperumbudur, and Taiji Suzuki for their advice and support during our collaborations.
Although they were not my official supervisors, they have had a substantial influence on my research taste and outlook.
My three-month internship at RIKEN in Tokyo was one of the most unforgettable experiences of my PhD.
Tokyo is an amazing city, and I hope to visit it again.

I would like to thank my other collaborators, Xiaoyuan Cheng, Ruocheng Guo, Heishiro Kanagawa, Sophia Kang, Toni Karvonen, Yang Liu, Dimitri Meunier, Aratrika Mustafi, Masha Naslidnyk, Atsushi Nitanda, Louis Sharrock, Zikai Steve Shen, Ingo Steinwart, Chenyang Tian, and Jean-Fran{\c{c}}ois Ton.
This thesis would not have been possible without your assistance and contributions.
I am grateful to Kevin Han Huang for introducing me to friends from Warwick and for being willing to talk with me during some of my most difficult moments, and to Zhuo Sun, Jiaye Teng, and Zhiheng Zhang for introducing me to many academic friends in Shanghai.
I am also grateful to Ben Dai, Fanghui Liu, Lei Shi, Weijie Su, and Lei Wu for giving me valuable advice on academic careers.

I would like to thank William Laplante for many wonderful evenings spent over hotpot and Biangbiang noodles.
It has been amazing to meet someone from a completely different cultural background with whom I share so many personal traits.
I am also grateful to the many people at the Gatsby Unit, including Hugh Dance, Peter Orbanz, Yedi Zhang, and others whom I did not always have the courage to reach out and talk to.
It has been a great pleasure to spend time at the Gatsby Unit.
It is a genuinely wonderful place for research, and my time there will remain among my best memories.
From the Foundational AI Centre, I would like to thank William Bankes, David Chanin, Boyu Chen, Tiantian He, Shuofei Qiao, Yuxuan Shu, Seongho Son, Jiayi Wang, Lorenz Wolf, Bin Wu, Tianqi Wu, Moucheng Xu, Ziming Xu, Zhuangjian Yang, Ning Zhang, Mingtian Zhang, and Caroline Zhu for the wonderful events, retreats, and time spent together.
In particular, I would like to thank Bin Wu, Tianqi Wu, Ning Zhang, Zhuangjian Yang, and Ziming Xu for helping me recover my stolen iPhone.
I would like to thank my friends from the Fundamentals of Statistical Machine Learning group (FSML) in the Department of Statistical Science at UCL.
The regular Monday group meetings and lunches helped me keep track of my PhD progress and encouraged me to socialize with others.
Such an atmosphere is very rare.
I would also like to thank my office mates, Wei Bu, Simon Damm, Nicola Gnecco, Zheng He, Christopher Marunoi, and Matthew Tilley, for the inspiring conversations in the visitor office. 
I would like to thank my friends met in Tokyo, Wei Huang, Zhanpeng Zhou, Tom Wakayama, Sho Sonada, Dake Bu, Bingrui Li, and Charles, for all the interesting conversations during lunch and dinners. 
I would also like to thank Tim G. J. Ruder and Weizhong Zhang who have taken me all the way to this path of machine learning back in my undergraduate. 
I would like to thank my roommates, Hengyi Wang, Yuxiang Wu, and Zhengyao Jiang, for their support and for the time we spent together playing VR, Junqi, and watching NBA games.
I hope our friendship will last forever. 

Finally, I would like to thank my family.
I am deeply grateful to my parents, BinWei Chen and Jun Wu, for their support throughout my childhood.
During high school, my obsession with the Mathematical Olympiad placed a considerable financial burden on my family.
Although this path was uncertain and, in many ways, unrealistic, my parents continued to support my dream.
Without their trust and continuing support, I would not have had the confidence and courage to pursue a PhD.
I am also deeply grateful to my grandparents, Guomei Dong and Qifa Wu , for taking such good care of me when I was young.
I wish them good health, and I will always remember their love and care.

I would also like to thank my girlfriend Churong Peng, and soon-to-be wife at the time of writing this thesis, who may already be my wife by the time it is submitted.
Her support has meant a great deal to me throughout my PhD.
Our one-hour WeChat video calls were the happiest part of each day.
I believe a happy future is ahead of us.

\setcounter{tocdepth}{2} 

\tableofcontents
\listoffigures
\listoftables

\chapter{Introduction}
\label{chapterlabel1}

Comparing two probability distributions $\mathbb{P}$ and $\mathbb{Q}$ is a central problem in computational statistics and machine learning.
Historically, such comparisons have been used to test whether two distributions are equal from a collection of finite observations~\citep{gretton2012kernel,sejdinovic2013equivalence,schrab2023mmd,schrab2022efficient,biggs2023mmd,schrab2024robust,schrab2025unified}.
In modern applications, however, the goal is often not merely to decide whether two distributions differ, but to quantify and then minimize the discrepancy between a target distribution and a tractable model or approximation. 
This viewpoint appears in approximate inference~\citep{blei2017variational}, generative modelling~\citep{arbel2019maximum,binkowski2018demystifying}, Bayesian computation~\citep{Robert2004,briol2019statistical}, statistical inference~\citep{basu2011statistical}, and numerical integration~\citep{davis2007methods,briol2019statistical}.
It is also the organizing perspective of this thesis.

Among the many possible choices of divergence or discrepancy $D$, one particularly appealing discrepancy measure is the maximum mean discrepancy (MMD)~\citep{gretton2012kernel}, which is attractive for three main reasons:
1) \textit{flexibility}: different choices of kernel induce different topologies on probability measures~\citep{sriperumbudur2010hilbert,barp2024targeted};
2) \textit{computability}: MMD admits explicit plug-in estimators from finite samples from both distributions~\citep{gretton2012kernel};
3) \textit{sample complexity}: for some commonly-used kernels, its sample complexity does not suffer from the curse of dimensionality~\citep{gretton2012kernel,tolstikhin2016minimax}.
Together, these favourable statistical and computational properties make MMD useful for a wide class of algorithms that approximate or model a target distribution by minimizing a computable discrepancy.

This thesis studies this optimization problem for the squared MMD discrepancy, $\mmd^2(\mathbb{P}, \Qb)$, minimizing with respect to $\mathbb{P}$ while $\Qb$ is a fixed but typically unknown target distribution.
This discrepancy-minimization viewpoint appears in several recurring forms:

\definecolor{mmdGreen}{HTML}{2E7D32}
\definecolor{mmdBlue}{HTML}{1565C0}
\definecolor{mmdRed}{HTML}{C62828}

\begin{itemize}[leftmargin=*]
    \item \emph{Numerical integration}: suppose we wish to approximate an intractable integral $\int f(x)\, \mathbb{Q}(\mathrm{d}x)$, where $\Qb$ is a probability measure, using function evaluations at a fixed collection of points $\{x_i\}_{i=1}^N$. The question is how to choose weights $\{w_i\}_{i=1}^N$ so that the weighted average $\sum_{n=1}^N w_n f(x_n)$ is accurate for a broad class of integrands $f$. 
    From a discrepancy-minimization perspective, this means choosing the weighted empirical measure $\sum_{i=1}^N w_i\delta_{x_i}$ to minimize its MMD distance from the integration measure $\mathbb{Q}$~\citep{karvonen2018fully,briol2019statistical}. 
    Weights chosen in this way can yield integration errors that are much smaller than those of standard Monte Carlo, which uses uniform weights $w_1 = \cdots = w_N = 1/N$, for smooth integrands in low-to-moderate dimensions; this improvement is supported by both theory and empirical evidence~\citep{sommariva2006numerical,Briol2015,Briol2019PI}. 
    This weighted integration rule is known as kernel quadrature or Bayesian quadrature~\citep{OHagan1991,Rasmussen2003,Hennig2022}.

    \begin{center}
    \setlength{\fboxsep}{6pt}
    \fcolorbox{mmdGreen!65!black}{mmdGreen!8}{%
        \begin{minipage}{0.68\linewidth}
        \centering
        {\bfseries Numerical Integration}\\[0.08cm]
        {$
        \min_{w_1, \ldots, w_N \in\R^N}
        \mmd^2\left(\sum_{i=1}^N w_i \delta_{x_i}, \; \Qb\right)
        $}
        \end{minipage}
    }
    \end{center}

    \item \emph{Sampling, generative modelling, and coreset selection}:
    the goal is to construct samples $\{x_i\}_{i=1}^N$ whose empirical distribution $\frac{1}{N}\sum_{i=1}^N\delta_{x_i}$ approximates the target distribution $\Qb$. From a discrepancy-minimization perspective, this amounts to choosing the empirical measure $\frac1N \sum_{i=1}^N \delta_{x_i}$ to minimize its MMD discrepancy from $\Qb$. When $\Qb$ is known up to its normalizing constant, this task is referred to as sampling~\citep{chewi2026logconcave,liu2016stein,korba2020non,li2023sampling}; when $\Qb$ is available only through i.i.d. samples, it is referred to as generative modelling~\citep{li2017mmd,binkowski2018demystifying,arbel2019maximum,song2020score} or coreset selection~\citep{bachem2017practical,chen2010super,mak2018support,teymur2021optimal,shetty2022distribution,dwivedi2024kernel,zhang2023m3d}.   

    \begin{center}
    \setlength{\fboxsep}{6pt}
    \fcolorbox{mmdBlue!70!black}{mmdBlue!8}{%
        \begin{minipage}{0.68\linewidth}
        \centering
        {\bfseries Sampling / Generative Modelling}\\[0.08cm]
        {$
        \min_{x_1, \ldots, x_N} 
        \mmd^2\left(\frac{1}{N} \sum_{i=1}^N \delta_{x_i}, \; \Qb \right)
        $}
        \end{minipage}
    }
    \end{center}

    \item \emph{Statistical inference}: given a model family $\{\Pb_\theta:\theta\in\Theta\}$, the goal is to estimate $\theta$ by matching the model distribution $\Pb_\theta$ to the data-generating distribution $\Qb$. 
    Classical maximum likelihood inference can be viewed as minimizing the Kullback--Leibler divergence between $\Pb_\theta$ and $\Qb$~\citep{wolfowitz1957minimum,basu2011statistical,van2000asymptotic}. 
    Replacing this divergence by the squared MMD leads to the minimum MMD estimator~\citep{gretton2012kernel,muandet2017kernel,briol2019statistical}. 
    This estimator is statistically attractive because it can be robust to model misspecification, including outlier contamination and covariate shift~\citep{briol2019statistical,cherief2020mmd}. It is also computationally useful because it remains well defined when the likelihood is unavailable or intractable, as in likelihood-free inference~\citep{niu2023discrepancy,dellaporta2022robust,Bharti2023}.

    \begin{center}
    \setlength{\fboxsep}{6pt}
    \fcolorbox{mmdRed!70!black}{mmdRed!8}{%
        \begin{minipage}{0.68\linewidth}
        \centering
        {\bfseries Statistical Inference}\\[0.08cm]
        {$
        \min_{\theta \in \Theta}
        \mmd^2(\Pb_\theta,\Qb)
        $}
        \end{minipage}
    }
    \end{center}
\end{itemize}

\section{My contributions}
This thesis develops new methods and theory for MMD minimization in the three settings described above. 
\begin{enumerate}[leftmargin=*]
    \item For numerical integration, $\arg \min_{w_{1},\ldots,w_N}
    \mmd^2 (\sum_{i=1}^N w_i \delta_{x_i}, \; \Qb)$ give rise to the optimal weights,
    which 
    are known to admit a closed-form expression. 
    In \Cref{sec:kernel_quadrature}, I show that this closed-form weighting principle extends to nested expectations, and that the resulting estimators inherit the smoothness-dependent fast convergence rates of kernel quadrature. 
    This chapter is based on a published paper in International Conference on Machine Learning, 2025~\citep{chen2025nested}.
    
    \item For sampling and generative modelling, the finite-dimensional problem $\arg\min_{x_1,\ldots,x_N}
    \mmd^2\left(\frac1N \sum_{i=1}^N \delta_{x_i}, \; \Qb\right)$
    can be viewed as a particle approximation of the lifted optimization problem over the space of probability measures $\calP$, namely
    $\arg\min_{\Pb\in\calP}\mmd^2\left(\Pb, \; \Qb\right)$. 
    This lifted problem admits a natural optimization dynamics, given by the Wasserstein gradient flow of the MMD objective~\citep{ambrosio2005gradient,arbel2019maximum}. 
    The main technical challenge herein is that the map $\Pb\mapsto \mmd^2\left(\Pb, \; \Qb\right)$ is generally non-convex along Wasserstein geodesics~\citep{arbel2019maximum,chen2024regularized}, even for log-concave targets $\Qb$, such as Gaussian distributions. 
    In \Cref{sec:drmmd}, I show that, with suitable regularization, the resulting Wasserstein gradient flow enjoys global convergence for a broad class of target distributions. 
    This chapter is based on a published paper in the Journal of Machine Learning Research,  2025~\citep{chen2024regularized}.
    
    \item For statistical inference, despite the growing popularity of minimum MMD estimators, comparatively little work has addressed the optimization problem, $\arg\min_{\theta\in\Theta} \mmd^2(\Pb_\theta,\Qb)$, itself. 
    Existing methods typically rely on gradient descent on the parameter $\theta$~\citep{briol2019statistical,cherief2020mmd}, including the recent R-package implementation of \cite{alquier2025package}; due to the non-convexity of the objective, these gradient-based methods can at best be expected to converge to stationary points. 
    In \Cref{sec:pgd}, I propose a new descent scheme for minimizing the parametric MMD objective, and prove that it can converge to a (not necessarily unique) \emph{global} minimizer under explicit assumptions.
    This chapter is based on a recently arXived manuscript~\citep{kang2026gradient}.
\end{enumerate}

Beyond the chapters above, I have also contributed as a primary author to the following publications or archival papers, during the course of the PhD:
\begin{itemize}[leftmargin=*]
    \item \underline{Z. Chen}, T. Karvonen, H. Kanagawa, F.-X. Briol, and C. Oates. \emph{Stationary MMD Points for Cubature}. Proceedings of the 43rd International Conference on Machine Learning, 2026~\citep{chen2025stationary};
    \item \underline{Z. Chen}, H. Kanagawa, F.-X. Briol, C. J. Oates, and L. Mackey. \emph{Thinned Mean Field Langevin Dynamics}. Proceedings of the 43rd International Conference on Machine Learning, 2026~\citep{chen2026thinned};
    \item C. Tian, B. K. Sriperumbudur, A. Gretton, and \underline{Z. Chen}. \emph{Sobolev Regularized MMD Gradient Flow}. arXiv, 2026~\citep{tian2026sobolev};
    \item \underline{Z. Chen}, R. Guo, J.-F. Ton, and Y. Liu. \emph{Conformal Counterfactual Inference under Hidden Confounding}. Proceedings of the 30th ACM SIGKDD Conference on Knowledge Discovery and Data Mining, 2024~\citep{chen2024conformal};
    \item \underline{Z. Chen}, A. Nitanda, A. Gretton, and T. Suzuki. \emph{Towards a Unified Analysis of Neural Networks in Nonparametric Instrumental Variable Regression: Optimization and Generalization}. arXiv, 2025~\citep{chen2025towards};
    \item Z. Shen, \underline{Z. Chen}, D. Meunier, I. Steinwart, A. Gretton, and Z. Li. \emph{Nonparametric Instrumental Variable Regression with Observed Covariates}. arXiv, 2025~\citep{shen2025nonparametric};
    \item S. S. Kang, F.-X. Briol, T. Karvonen, and \underline{Z. Chen}. \emph{BayesSum: Bayesian Quadrature in Discrete Spaces}. Proceedings of the Second International Conference on Probabilistic Numerics, 2026~\citep{kang2025bayessum}.
\end{itemize}

\begin{table}[H]
    \centering
    \renewcommand{\arraystretch}{1.25}
    \begin{tabular}{p{0.24\textwidth}p{0.68\textwidth}}
        \toprule
        Notation & Meaning \\
        \midrule
        $\N$ & Natural numbers, $\{0,1,\ldots\}$. \\
        $\Id$ or $\Id_d$ & Identity operator or identity matrix, depending on context; $\Id_d$ denotes the $d\times d$ identity matrix. \\
        $x_{1:N}\in\R^{N\times d}$ & Vector when $d=1$, or matrix $[x_1\;x_2\;\ldots\;x_N]^\top$, for $x_1,\ldots,x_N\in\calX\subseteq\R^d$. \\
        $T_\#\Pb$ & Pushforward of $\Pb$ under a measurable map $T$, defined by $(T_\#\Pb)(A)=\Pb(T^{-1}(A))$ for measurable sets $A$. \\
        $L_2(\Pb)$ & Hilbert space of square-integrable functions under the probability measure $\Pb$, with norm $\|f\|_{L_2(\Pb)}^2=\int |f(x)|^2 \,\dd\Pb(x)$. \\
        $L_2^d(\Pb)$ & Vector-valued $L_2$ space of functions $f=(f_1,\ldots,f_d)$ with $f_i\in L_2(\Pb)$, equipped with $\ps{f,g}_{L_2^d(\Pb)}=\sum_{i=1}^d\int f_i(x)g_i(x)\,\dd\Pb(x)$. \\
        $\calP_2(\R^d)$ & Probability measures on $\R^d$ with finite second moment, equipped with the Wasserstein-2 geometry when a metric is needed. \\
        $\mathcal{H}_k$ & RKHS associated with the kernel $k$. \\
        $\Hk^d$ & Cartesian product RKHS of vector fields $f=(f_1,\dots,f_d)$ with $f_i\in\Hk$, equipped with $\ps{f,g}_{\Hk^d}=\sum_{i=1}^d\ps{f_i,g_i}_{\Hk}$. \\
        $W_2^s(\calX)$ & Sobolev space of order $s$; see \Cref{def:sobolev-space}. \\
        $C^\beta(\calX)$ & H\"older space of regularity $\beta$: functions with $\lfloor\beta\rfloor$ continuous derivatives whose highest-order derivatives are $(\beta-\lfloor\beta\rfloor)$-H\"older continuous. \\
        $m_{\Pb}$ & Kernel mean embedding of $\Pb$, given by $\int k(x,\cdot)\,\dd\Pb(x)$ when the integral is well-defined. \\
        \bottomrule
    \end{tabular}
    \caption[Frequently used notation]{Frequently used function spaces and probability-measure notation.}
    \label{tab:background-notation}
\end{table}

\chapter{Background}

\section{Reproducing Kernel Hilbert Spaces}
Let $\mathcal{X} \subseteq \mathbb{R}^d$ be any non-empty set. A symmetric function $k: \mathcal{X} \times \mathcal{X} \to \mathbb{R}$ is called a positive definite kernel if, for any $N \in \mathbb{N}$ and any $x_{1:N} \in \mathcal{X}^N$, the associated Gram matrix~\citep[Section~2]{Berlinet2004}
\begin{align*}
    k(x_{1:N},x_{1:N})
    :=
    \begin{bmatrix}
        k(x_1,x_1) & \ldots & k(x_1,x_N) \\
        \vdots & \ddots & \vdots \\
        k(x_N,x_1) & \ldots & k(x_N,x_N)
    \end{bmatrix}
\end{align*}
is positive semi-definite. In other words, $\alpha^\top  k(x_{1:N},x_{1:N}) \alpha \geq 0$ for any $\alpha\in\R^N$. 
The associated RKHS $\mathcal{H}_k$ is a Hilbert space of scalar-valued functions on $\mathcal{X}$ equipped with an inner product $\langle \cdot, \cdot \rangle_{\mathcal{H}_k}$ and norm $\|\cdot\|_{\mathcal{H}_k}$ satisfying two key properties~\citep{aronszajn1950theory}:
\begin{enumerate}[leftmargin=*]
    \item For all $x \in \mathcal{X}$, the kernel feature function $k(x, \cdot) \in \mathcal{H}_k$.
    \item The reproducing property holds: for all $h \in \mathcal{H}_k$ and $x \in \mathcal{X}$,
    $h(x) = \langle h, k(x, \cdot) \rangle_{\mathcal{H}_k}$. 
\end{enumerate}
Alternatively speaking, $\mathcal{H}_k$ is the closure, under the RKHS norm, of the linear span of kernel feature maps $\{k(x,\cdot):x\in\mathcal{X}\}$. 
This gives the intuition that functions in the RKHS are built as limits of finite linear combinations of these kernel feature maps. 
Common examples of kernels include the following~\citep[Chapter~7]{bach2024learning}:
\begin{enumerate}[leftmargin=*]
    \item The linear kernel $k(x,y)=x^\top y$ on $\mathbb{R}^d$ has an RKHS consisting of linear functions $f(x)=\theta^\top x$ with $\theta\in\R^d$, with the RKHS norm corresponding to the Euclidean norm of the coefficient $\theta$.
    
    \item The polynomial kernel $k(x,y)=(x^\top y+c)^p$ has an RKHS consisting of polynomial functions of degree at most $p$ when $c>0$, and homogeneous polynomials of degree $p$ when $c=0$~\citep{scholkopf2002learning}. 
    
    \item The Gaussian kernel $k(x,y)=\exp\left(-\frac{\|x-y\|^2}{2\ell^2}\right)$ with any positive lengthscale $\ell$ has an RKHS consisting of functions that are infinitely differentiable and real analytic. 
    
    \item Mat\'ern kernels of order $\nu$ have RKHSs that are norm-equivalent to Sobolev spaces of smoothness $\nu+d/2$~\citep{aronszajn1950theory,Berlinet2004,wendland2004scattered,fasshauer2015kernel,kanagawa2018gaussian}. 
    See \Cref{def:sobolev-space} for definition of Sobolev spaces.
    For half-integer orders $\nu \in \{1/2,3/2,\ldots\}$, the Mat\'ern kernel $k$ takes a convenient form: it is a product of an exponential and a polynomial of degree $\lfloor \nu \rfloor$~\citep{williams2006gaussian}. For example, writing $\rho=\|x-x'\|$, the first three half-integer Mat\'ern kernels are
    \begin{align*}
        \nu=1/2: \quad
        & k_\nu(x,x')=\tau^2 \exp\left(-\frac{\rho}{\ell}\right),\\
        \nu=3/2: \quad
        & k_\nu(x,x')=\tau^2\left(1+\frac{\sqrt{3}\rho}{\ell}\right)
        \exp\left(-\frac{\sqrt{3}\rho}{\ell}\right),\\
        \nu=5/2: \quad
        & k_\nu(x,x')=\tau^2\left(1+\frac{\sqrt{5}\rho}{\ell}+\frac{5\rho^2}{3\ell^2}\right)
        \exp\left(-\frac{\sqrt{5}\rho}{\ell}\right).
    \end{align*}
\end{enumerate}

{\color{black}
For an open set $\calX\subseteq\R^d$ and a multi-index $\alpha\in\mathbb{N}_0^d$, a locally integrable function $v$ is the \emph{weak derivative} $\partial^\alpha f$ of a locally integrable function $f$ if $\int_{\calX} f(x)\partial^\alpha\varphi(x)\,\dd x=(-1)^{|\alpha|}\int_{\calX}v(x)\varphi(x)\,\dd x$ for every smooth, compactly supported test function $\varphi\in C_c^\infty(\calX)$. This extends the integration-by-parts identity to functions that need not have classical derivatives.
\par}

\begin{defi}[Sobolev space]\label{def:sobolev-space}
For an integer $s\geq 0$ and a domain $\calX\subseteq\R^d$, the Sobolev space $W_2^s(\calX)$ consists of functions whose weak derivatives up to order $s$ are square-integrable:
\begin{equation}\label{eq:defi_sobolev}
    W_2^s(\calX)
    :=
    \left\{
        f\in L_2(\calX):
        \partial^\alpha f \in L_2(\calX)
        \text{ for every multi-index } \alpha \text{ with } |\alpha|\leq s
    \right\}.
\end{equation}
Its norm is
\[
    \|f\|_{W_2^s(\calX)}
    :=
    \left(
        \sum_{|\alpha|\leq s}
        \|\partial^\alpha f\|_{L_2(\calX)}^2
    \right)^{1/2}.
\]
\end{defi}

We next review two technical constructions built from kernels and RKHSs: \emph{kernel integral operators} and the \emph{maximum mean discrepancy}. {\color{black}The kernel integral operator and its RKHS counterpart will be used to bound the regression errors underlying nested kernel quadrature in \Cref{sec:nkq}, and to construct the regularized discrepancy and analyse its Wasserstein gradient flow in \Cref{sec:drmmd}.}
The maximum mean discrepancy will be used extensively in all the following chapters of this thesis. 

\vspace{0.3cm}
\paragraph{Kernel integral operator:}
When the kernel $k$ is measurable and satisfies $\int \sqrt{k(x,x)} \;\dd \Pb(x)< \infty$ for a probability measure $\Pb$. 
$\mathcal H$ can be canonically injected into $L_2(\Pb)$ using the operator $\iota_{\Pb} : \Hk \to L_2(\Pb),\,f\mapsto f$ with adjoint $ \iota_\Pb^\ast: L_2(\Pb) \rightarrow \Hk$ given by $\iota_\Pb^\ast f(\cdot) = \int k(x,\cdot)f(x)\dd\Pb(x)$. 
The operator $\iota_{\Pb}$ and its adjoint can be composed to form an $L_2(\Pb)$ endomorphism $\calT_{\Pb} \coloneqq\iota_{ \Pb} \iota_{ \Pb}^\ast$ called the \emph{integral operator}, and a $\mathcal H$ endomorphism  $\Sigma_{\Pb} := \iota_{ \Pb}^\ast \iota_{ \Pb}=\int k(\cdot, x) \otimes k(\cdot, x) d \Pb(x)$
(where $(a\otimes b)c \coloneqq \langle b,c \rangle_{\Hk} a$ for $a,b,c\in \Hk$) called the \emph{covariance operator}.
$\calT_{\Pb}$ is compact, positive, self-adjoint, and can thus be diagonalized into an orthonormal system in $\left\{e_i \right\}_{i \geq 1}$ of $L_2(\Pb)$ with associated eigenvalues $\varrho_1 \geq \cdots \varrho_i \geq \cdots \geq 0$.

\paragraph{Maximum mean discrepancy:} 
When the kernel $k$ is measurable and satisfies $\int \sqrt{k(x,x)} \;\dd \Pb(x)< \infty$ for a probability measure $\Pb$, then $\mathbb{P}$ on $\mathcal{X}$ can be represented in $\mathcal{H}_k$ through its kernel mean embedding
$m_{\Pb} := \int_{\mathcal{X}} k(x,\cdot)\,\dd\Pb(x)$, which satisfies
$\mathbb{E}_{X\sim \mathbb{P}}[h(X)]=
\langle h, m_{\Pb} \rangle_{\mathcal{H}_k}$ for all $h\in \mathcal{H}_k$. 
The maximum mean discrepancy (MMD) between two probability measures $\mathbb{P}$ and $\mathbb{Q}$ is thus defined as the Hilbert-space distance between their respective embeddings,
\[
    \mmd(\mathbb{P},\mathbb{Q})
    :=
    \|m_{\Pb}-m_{\Qb}\|_{\mathcal{H}_k}
    =
    \sup_{\|h\|_{\mathcal{H}_k}\leq 1}
    \left|
    \mathbb{E}_{\mathbb{P}}[h(X)]-\mathbb{E}_{\mathbb{Q}}[h(Y)]
    \right|.
\]
Although the MMD is defined through a supremum, the reproducing property and the Cauchy--Schwarz inequality imply that it admits a closed-form expression in terms of kernel expectations. In particular, for independent $X,X'\sim\mathbb{P}$ and $Y,Y'\sim\mathbb{Q}$,
\[
    \mmd_k^2(\mathbb{P},\mathbb{Q})
    =
    \mathbb{E}[k(X,X')]
    -2\mathbb{E}[k(X,Y)]
    +\mathbb{E}[k(Y,Y')].
\]
{\color{black}A kernel $k$ is called \emph{characteristic} if its mean embedding is injective on the class of probability measures for which the embedding exists: $m_{\Pb}=m_{\Qb}$ implies $\Pb=\Qb$~\citep{muandet2017kernel}. For such a kernel,} $\mmd_k(\mathbb{P},\mathbb{Q})=0$ if and only if $\mathbb{P}=\mathbb{Q}$. This is one reason characteristic kernels are widely used in MMD-based sampling, generative modelling and inference~\citep{gretton2012kernel,muandet2017kernel,chen2024regularized}.

\section{MiniMMD for Integration}\label{sec:nkq_background}
In this section, we provide background of kernel quadrature~\citep{sommariva2006numerical,Rasmussen2003,Briol2019PI}, a kernel-based Monte Carlo method derived by minimizing the MMD between the weighted empirical distribution and the integration measure. 

Consider an intractable expectation $I=\mathbb{E}_{X \sim \Qb}[h(X)]=\int h(x)\dd \Qb(x)$. Standard Monte Carlo (MC) provides a canonical estimator: when $x_{1:N}$ are $N$ independent and identically distributed realisations from $\Qb$, MC approximates $I$ by $I \approx \frac{1}{N} \sum_{n=1}^N h(x_n)$. 
Kernel quadrature (KQ)~\citep{sommariva2006numerical,Rasmussen2003,Briol2019PI} provides an alternative estimator for intractable expectations. 
It exploits additional smoothness of the integrand, assuming that $h$ belongs to a reproducing kernel Hilbert space (RKHS) $\calH_k$ with kernel $k$~\citep{Berlinet2004}. 

Assuming $h \in \calH_k$ and $\mathbb{E}_{X \sim \Qb}[\sqrt{k(X, X)}] < \infty$, 
the KQ estimator $\hat{I}_{\kq} = \sum_{n=1}^N w_n h(x_n)$ uses weights obtained by minimizing an upper bound on the absolute error:
\vspace{-8pt}
\begin{align*}
    \left|I - \hat{I}_{\kq} \right|
    &= \Big| \E_{X \sim \Qb}[h(X)] - \sum_{n=1}^N w_n h(x_n) \Big| \\
    &\leq \|h\|_{\calH_k} \Big\| m_{\Qb}(\cdot) - \sum_{n=1}^N w_n k(x_n,\cdot) \Big\|_{\calH_k} \\ 
    &= \|h\|_{\calH_k} \mmd\left(\sum_{i=1}^N w_i \delta_{x_i}, \; \Qb\right),
\end{align*}
where $m_{\Qb}(\cdot) = \mathbb{E}_{X \sim \Qb}[k(X,\cdot)]$ is the \emph{kernel mean embedding (KME)} of $\Qb$ in the RKHS $\mathcal{H}_k$~\citep{smola2007hilbert}.
Minimizing the right-hand side with respect to the weights, $\arg\min_{w_{1:N}} \mmd^2(\sum_{i=1}^N w_i \delta_{x_i}, \; \Qb)$, leads to the KQ estimator
\begin{align}\label{eq:kq}
     \hat{I}_\kq := m_{\Qb}(x_{1:N}) \boldsymbol{K}^{-1}h(x_{1:N}),
\end{align}
where $\boldsymbol{K} = k(x_{1:N}, x_{1:N}) \in \mathbb{R}^{N \times N}$ is the Gram matrix, $m_{\Qb}(x_{1:N}) = \big(m_{\Qb}(x_1),\allowbreak \ldots,\allowbreak m_{\Qb}(x_N)\big)$, and $h(x_{1:N})=\big(h(x_1),\ldots,h(x_N)\big)^\top$.
The KQ weights are $w_{1:N} = m_{\Qb}(x_{1:N})\boldsymbol{K}^{-1}$.  
In practice, an additional diagonal matrix $\lambda \Id$ is added to the Gram matrix $\boldsymbol{K}$ to improve numerical stability.

KQ uses the structural information $h \in \calH_k$, which can make the absolute error $|I-\hat{I}_{\kq}|$ decay faster than the MC rate as $N \rightarrow \infty$.
Specifically, when the RKHS $\calH_k$ is norm-equivalent to the Sobolev space $W_2^s(\calX)$ with $s > d/2$, KQ achieves the rate $\calO(N^{-\frac{s}{d}})$~\citep{Kanagawa2019adaptive,Kanagawa2017convergence}. This rate is minimax optimal~\citep{novak2006deterministic,novak2016some} and can be significantly faster than the $\calO(N^{-\frac{1}{2}})$ rate of standard MC. 
Existing proof techniques that obtain this rate typically take $\lambda = 0$ in \eqref{eq:kq} and require the Gram matrix $\boldsymbol{K}$ to be invertible. In contrast, the kernel-ridge-regression proof used in this chapter obtains the same optimal rate while allowing positive regularization $\lambda \asymp N^{-\frac{2s}{d}}(\log N)^{\frac{2s+2}{d}}$, which improves numerical stability when inverting $\boldsymbol{K}$; see \Cref{rem:stage_one_error_and_standard_kq}.

Despite the optimality of the KQ convergence rate, the rate constant can be reduced by selecting points $x_{1:N}$ more carefully than by i.i.d. sampling. Strategies include importance sampling~\citep{Bach2015,Briol2017SMCKQ}, QMC point sets~\citep{Briol2019PI,Jagadeeswaran2018,Bharti2023,Kaarnioja2025}, realisations from determinantal point processes~\citep{belhadji2019kernel}, point sets with symmetry properties~\citep{Karvonen2017symmetric, Karvonen2019}, and adaptive designs~\citep{osborne2012active,gunter2014sampling,Briol2015,gessner2020active}. 
Most relevant to our setting is the combination of KQ with MLMC to improve accuracy in multifidelity problems~\citep{li2023multilevel}.

Compared with MC, KQ has two main drawbacks: the worst-case computational cost is $\calO(N^3)$ because of the Gram-matrix solve, and the KME $m_{\Qb}$ must be available in closed form. Fortunately, several approaches can mitigate these drawbacks. To reduce the cost, one can use geometric properties of the point set~\citep{Karvonen2017symmetric,Karvonen2019,Kuo2024}, Nyström approximations~\citep{Hayakawa2022,Hayakawa2023}, randomly pivoted Cholesky~\citep{Epperly2023}, or the fast Fourier transform~\citep{Zeng2009}.
To obtain a closed-form KME, one can use existing derivations, such as Table 1 in \citet{Briol2019PI} or the catalogue in \citet{Wenger2021}, or replace the kernel with a Stein reproducing kernel~\citep{Oates2017,Oates2016CF2,Si2020,Sun2021}. 

In \Cref{sec:nkq}, we tackle both drawbacks through a change-of-variable trick. 
Suppose we can find a continuous transformation $\Phi$ such that $x_{1:N} = \Phi(u_{1:N})$, where $u_{1:N}$ are samples from a simpler distribution $\Ub$ of our choice. 
The change-of-variables theorem~\citep[Section 8.2]{stirzaker2003elementary} gives $\E_{X \sim \Qb} h(X)  = \int_\calU h(\Phi(u))\dd\Ub(u)$, so the integrand changes from $h: \calX \to \R$ to $h \circ \Phi: \calU \to \R$ and the KQ estimator becomes $\hat{I}_{\kq} = \mu_{\Ub}(u_{1:N}) \boldsymbol{K}_{\calU}^{-1} (h \circ \Phi)(u_{1:N})$, where $\boldsymbol{K}_{\calU} = k_{\calU}(u_{1:N}, u_{1:N})$. 
The measure $\Ub$ is chosen so that its KME is known in closed form. The weights $\mu_{\Ub}(u_{1:N})\boldsymbol{K}_{\calU}^{-1}$ can then be precomputed and stored, making KQ a weighted average of function evaluations with $\calO(N)$ computational complexity. 
The main technical challenge is to find a suitable transformation $\Phi$.

Before concluding, we note that the KQ estimator is often called \emph{Bayesian quadrature (BQ)}~\citep{Diaconis1988,OHagan1991,Rasmussen2003,Briol2019PI,Hennig2022}, since it can be derived as the posterior mean obtained by conditioning a Gaussian-process prior on the function values $h(x_{1:N})$~\citep{kanagawa2018gaussian}.
The Bayesian interpretation provides finite-sample uncertainty quantification and enables hyperparameter selection via empirical Bayes.

\section{MiniMMD for Sampling, Generative Modelling and Coreset Selection}\label{sec:drmmd_background}

\label{sec:background}
\label{sec:wass}
In this section, we provide background on MMD gradient flow, which is the Wasserstein gradient flow of the squared MMD discrepancy, and show how it can be used for sampling, generative modelling and coreset selection.

\paragraph{Wasserstein gradient flow}
Let $\calF:\calP_2(\R^d)\to\R$ be a functional with a unique minimizer $\pi$. In many cases, $\calF$ is of the form $\calF(\cdot)=\mathcal{D}(\cdot \| \pi)$, where $\mathcal{D}$ is a probability divergence. Gradient flows are dynamics that use local, differential information about $\calF$ in order to minimize it as fast as possible. In the familiar Euclidean case, the gradient flow of a sufficiently regular $F:\R^d \to \R$ from an initial condition $x_0$ is the solution $(x_t)_{t \ge 0}$ of $\partial_t x_t = -v(x_t)$, where $v$ is the Fréchet subdifferential of $F$, a generalization of the derivative to non-smooth functions~\citep{kruger2003frechet}.

Gradient flows extend from Euclidean spaces to more general metric spaces~\citep{ambrosio2005gradient}. When the metric space is $\calP_2(\R^d)$ endowed with the Wasserstein-$2$ distance, the resulting dynamics are called the Wasserstein gradient flow (WGF). Given an initial distribution $\mu_0\in\calP_2(\R^d)$, the WGF evolves a curve of probability measures $(\mu_t)_{t\ge 0}$ along the steepest descent direction of $\calF$ in the Wasserstein space $(\calP_2(\R^d), W_2)$~\citep{ambrosio2005gradient,villani2009optimal,santambrogio2017euclidean}. Formally, it is characterized by the continuity equation~\citep[Lemma 10.4.1]{ambrosio2005gradient}
\begin{align}\label{eq:cont_eqn_1}
    \partial_t \mu_t + \nabla \cdot \left(\mu_t v_t\right)=0, 
\end{align}
where $v_t$ is the Fr\'{e}chet subdifferential of $\calF$ evaluated at $\mu_t$~\citep[Definition 11.1.1]{ambrosio2005gradient}. Under these dynamics, the mass of $\mu_t$ is transported in the direction $v_t$ that decreases $\calF$ at the fastest rate at each time $t$.
Under mild growth conditions on $v_t$, the trajectory $(\mu_t)_{t\geq 0}$ is a unique absolutely continuous curve in $\calP_2(\R^d)$~\citep[Lemma 8.1.4]{ambrosio2005gradient}. It also admits the characteristic, or Lagrangian, representation
\begin{align}\label{eq:particleode}
    \dd X_t = -v_t(X_t)\dd t, \qquad \mathrm{Law}(X_t)=\mu_t .
\end{align}
Equations~\eqref{eq:cont_eqn_1} and~\eqref{eq:particleode} give complementary viewpoints: the former is Eulerian and describes density evolution, whereas the latter is Lagrangian and follows individual particle trajectories.

In practice, WGF implementations are usually based on a finite-particle discretization of the Lagrangian formulation. Given a fixed time horizon $S\in\N$ and an initial collection of particles $\{x_0^{(i)}\}_{i=1}^N \sim \mu_0$, the forward Euler scheme takes the form
\begin{align}\label{eq:particle_system}
    x_{s+1}^{(i)} = x_s^{(i)} - \gamma v[\hat{\mu}_s](x_s^{(i)}),
    \qquad
    \hat{\mu}_s = \frac{1}{N}\sum_{j=1}^N \delta_{x_s^{(j)}} ,
\end{align}
for $s\in[S]$, where $\gamma>0$ is the step size. Equivalently, in measure form this update is $\mu_{s+1}=(\Id-\gamma v_s)_\#\mu_s$. The scheme therefore involves both a forward Euler discretization in time and a finite-particle discretization in space, and is also known as Wasserstein gradient descent.

{\color{black}
A constant-speed Wasserstein geodesic $(\rho_t)_{t\in[0,1]}$ is a curve satisfying $W_2(\rho_s,\rho_t)=|s-t|W_2(\rho_0,\rho_1)$ for all $s,t\in[0,1]$. A functional $\calF$ is \emph{geodesically convex} if it is convex along every such geodesic, that is, $\calF(\rho_t)\leq(1-t)\calF(\rho_0)+t\calF(\rho_1)$. More generally, it is \emph{geodesically $\Lambda$-convex} if
\begin{align*}
    \calF(\rho_t)\leq(1-t)\calF(\rho_0)+t\calF(\rho_1)
    -\frac{\Lambda}{2}t(1-t)W_2^2(\rho_0,\rho_1).
\end{align*}
This generalizes Euclidean convexity by replacing straight lines with geodesics~\citep{ambrosio2005gradient,villani2009optimal}.
\par}

As with gradient descent in Euclidean spaces, a central tool for characterizing the convergence of Wasserstein gradient descent is the geodesic convexity and smoothness of the functional $\calF$~\citep{villani2009optimal,ambrosio2005gradient}. One way to characterize these properties is through lower and upper bounds on the Wasserstein Hessian of $\calF$~\citep[Proposition 16.2]{villani2009optimal}. Given any $\phi\in C_c^\infty(\R^d)$ that defines a constant speed geodesic\footnote{See \Cref{appsec:wass_discussion} for the definition.} starting at $\mu$, namely $\rho_t=(\Id+t\nabla\phi)_\#\mu$ for $0\leq t\leq1$, the Wasserstein Hessian of $\calF:\calP_2(\R^d)\to\R$ at $\mu$, denoted by $\operatorname{Hess}\calF_{\mid\mu}$, is an operator from $L^2(\mu)$ to $L^2(\mu)$:\footnote{Strictly speaking, $\operatorname{Hess}\calF_{\mid\mu}$ is an operator over the tangent space $\calT_\mu\calP_2(\R^d)$, which is a subset of $L^2(\mu)$~\citep{villani2009optimal}.}
\begin{align*}
    \left\langle \operatorname{Hess}\calF_{\mid\mu}\nabla\phi,\nabla\phi \right\rangle_{L^2(\mu)}
    = \frac{\dd^2}{\dd t^2}\Big|_{t=0} \calF(\rho_t).
\end{align*}
A functional $\calF$ is said to be \emph{geodesically} $M$-smooth at $\mu$ if
\[
    \left\langle \operatorname{Hess}\calF_{\mid\mu}\nabla\phi,\nabla\phi \right\rangle_{L^2(\mu)}
    \leq M\|\nabla\phi\|_{L^2(\mu)}^2,
\]
and \emph{geodesically} $\Lambda$-convex at $\mu$ if
\[
    \left\langle \operatorname{Hess}\calF_{\mid\mu}\nabla\phi,\nabla\phi \right\rangle_{L^2(\mu)}
    \geq \Lambda\|\nabla\phi\|_{L^2(\mu)}^2.
\]
Additionally, $\calF$ is \emph{geodesically semiconvex} if $-\infty<\Lambda<0$ and \emph{geodesically strongly convex} if $\Lambda>0$. A functional that is both smooth and strongly convex is preferred because its Wasserstein gradient descent has an exponential convergence rate under a sufficiently small step size~\citep[Section 9.3.1]{boyd2004convex,bonet2024mirror}.

\paragraph{MMD and KSD gradient flows} 
Given a target probability measure $\pi\in\calP_2(\R^d)$, the $\mmd$ flow is the Wasserstein gradient flow of the squared maximum mean discrepancy functional $\calF_{\mmd}(\cdot)=\frac{1}{2}\mmd^2(\cdot\|\pi)$, which solves the following optimization problem:
\begin{align*}
    \arg\min_{\Pb\in\calP_2(\R^d)} \mmd^2(\Pb, \Qb) . 
\end{align*}
The MMD~\citep{gretton2012kernel} between two probability measures is the RKHS norm of the difference between their kernel mean embeddings.

For $\calF_{\mmd}(\mu)=\frac{1}{2}\mmd^2(\mu,\pi)$, the Fréchet subdifferential is \citet{arbel2019maximum}.
\[
    v_{\mmd}[\mu](\cdot)=\int \nabla_2 k(x,\cdot)\,\dd(\mu-\pi)(x).
\]
This makes the MMD flow well suited to generative modeling because $v_{\mmd}[\mu]$ can be estimated from samples from $\mu$ and $\pi$. Given $M$ i.i.d. target samples $\{y^{(i)}\}_{i=1}^M$ from $\pi$ and $N$ particles $\{x_s^{(j)}\}_{j=1}^N$ at iteration $s$, a direct plug-in estimator is
\[
    v_{\mmd}[\hat{\mu}_s](\cdot)
    \approx
    \frac{1}{N}\sum_{j=1}^N \nabla_2 k(x_s^{(j)},\cdot)
    -
    \frac{1}{M}\sum_{j=1}^M \nabla_2 k(y^{(j)},\cdot).
\]
The MMD flow and closely related MMD-based objectives have therefore been used to construct generative modeling algorithms, including MMD-GAN variants~\citep{li2017mmd,binkowski2018demystifying,arbel2018gradient}, sample-based MMD flows~\citep{arbel2019maximum}, Riesz-kernel and sliced MMD flows~\citep{hertrich2023wasserstein,hertrich2023generative,hertrich2024wasserstein}, and adaptive-kernel flows based on learned features~\citep{galashov2024deep}. However, the Wasserstein Hessian of $\calF_{\mmd}$ for smooth kernels is not positively lower bounded~\citep[Proposition 5]{arbel2019maximum}. Consequently, the MMD flow only converges up to an unknown barrier~\citep[Theorem 6]{arbel2019maximum}, with global convergence only under a strong and unverifiable assumption~\citep[Proposition 7]{arbel2019maximum}. More recent work has demonstrated global convergence of the MMD flow when using the Coulomb kernel~\citep{boufadene2023global}. This kernel is non-smooth, however, which complicates numerical implementations.

The kernel Stein discrepancy (KSD) is a special instance of MMD under a target-dependent kernel $k_\pi$ satisfying the Stein identity $\int k_\pi(x,y)\dd\pi(x)=0$ for all $y\in\R^d$. A common choice is the Langevin Stein kernel, defined through the score $\mathfrak{s}(x)=\nabla\log\pi(x)$ and a base kernel $k:\R^d\times\R^d\to\R$:
\begin{align}\label{eq:steinkernel}
k_{\pi}(x,y)
&:= \mathfrak{s}(x)^\top \mathfrak{s}(y) k(x,y)
 + \mathfrak{s}(x)^\top \nabla_2 k(x,y)
 \nonumber\\
&\quad
 + \nabla_1 k(x,y)^\top \mathfrak{s}(y)
 + \operatorname{Tr}(\nabla_{1}\nabla_2 k(x,y)).
\end{align}
Under suitable tail conditions on the base kernel, the Stein identity makes the KSD flow particularly suitable for sampling from a target $\pi$ known only up to a normalization constant. The corresponding Fréchet subdifferential at iteration $s$ is~\citep{korba2021kernel}
\[
    v_{\operatorname{KSD}}[\hat{\mu}_s](\cdot)
    =
    \frac{1}{N}\sum_{j=1}^N \nabla_2 k_\pi(x_s^{(j)},\cdot).
\]

\section{MiniMMD for Statistical Inference}

In this section, we provide background on minimum MMD inference, which is a novel statistical inference method using squared MMD as the discrepancy measure~\citep{briol2019statistical,alquier2025package,cherief2020mmd,cherief2025parametric}.

Parameter estimation seeks to recover the unknown parameters governing a data-generating mechanism from observed data. Given a parametric family of statistical models $\{\Pb_\theta: \theta \in \Theta\}$, where $\Theta\subseteq\R^p$ is the parameter domain, a standard approach is to estimate these parameters by minimizing a divergence or discrepancy between the model and the data-generating distribution $\Qb$~\citep{wolfowitz1957minimum,parr1980minimum,basu2011statistical}.
Since the latter is unknown, an estimator is typically constructed by replacing it with the empirical distribution consisting of independently and identically distributed (IID) observations.
A canonical choice of such discrepancy is the Kullback--Leibler divergence, for which the resulting minimum-distance estimator coincides with maximum likelihood estimation~\citep{van2000asymptotic}.

Recently, a growing body of work has considered using the maximum mean discrepancy (MMD), a metric based on embeddings of probability measures into a reproducing kernel Hilbert space~\citep{gretton2012kernel,muandet2017kernel}, for minimum-distance estimation~\citep{briol2019statistical}.
Minimum MMD estimators have been studied in simulator-based statistical inference~\citep{briol2019statistical,niu2023discrepancy,Bharti2023}, robust and generalized Bayesian inference~\citep{cherief2020mmd,dellaporta2022robust}, parametric regression and covariate-shift settings~\citep{cherief2020mmd}, and goodness-of-fit or two-sample testing through MMD-based discrepancy statistics~\citep{gretton2012kernel,muandet2017kernel}.
The appeal of the method comes from the particular structure of the MMD objective, which we recall next.

Given observations $Y_1,\ldots,Y_m \sim \Qb$, the empirical distribution $\widehat{\Qb}_m = m^{-1}\sum_{i=1}^m \delta_{Y_i}$, and a model family $\{\Pb_\theta:\theta\in\Theta\}$, the minimum MMD estimator is a solution of
\begin{align*}
    \widehat{\theta}_m \in \operatorname*{arg\,min}_{\theta\in\Theta} \mmd^2(\Pb_\theta,\widehat{\Qb}_m).
\end{align*}
This parallels maximum likelihood in that both procedures choose a model distribution closest to the data, but the discrepancy is now the RKHS distance between kernel mean embeddings rather than the Kullback--Leibler divergence. 
The minimum MMD estimator enjoys the following two desirable properties:
\begin{itemize}[leftmargin=*]
    \item \textbf{Likelihood-free computation.} For simulator-based models of the form $\Pb_\theta = (G_\theta)_\#\rho$, it is enough to generate samples $X_i=G_\theta(U_i)$ with $U_i\sim\rho$ and compare them with the observed data through kernel evaluations. The squared MMD depends only on expectations of $k(X,X')$, $k(X,Y)$, and $k(Y,Y')$, and therefore admits unbiased empirical estimators based on simulated and observed samples. No evaluation of a density $p_\theta(y)$, score $\nabla_y\log p_\theta(y)$, or normalizing constant is required. This makes the estimator applicable to implicit models, stochastic simulators, latent Markov models, and stochastic differential equation models where sampling is feasible but likelihood evaluation is expensive or unavailable.
    \item \textbf{Robustness.} If $k$ is bounded, the effect of replacing one observation in $\widehat{\Qb}_m$ is uniformly controlled in the MMD objective, independently of the magnitude of that observation. Consequently, a small fraction of contaminated samples has limited influence compared with objectives based on unbounded losses or likelihood terms. Standard examples include the Gaussian/RBF kernel $k(x,y)=\exp(-\|x-y\|^2/(2\sigma^2))$, the Laplace kernel $k(x,y)=\exp(-\|x-y\|/\sigma)$, and other bounded kernels satisfying $\sup_{x,y}|k(x,y)|<\infty$.
\end{itemize}

\chapter{Minimum MMD for Integration in Nested Expectations}
\label{sec:nkq}
\label{sec:kernel_quadrature}

\begin{center}
\setlength{\fboxsep}{12pt}
\fcolorbox{paperboxborder}{paperboxbg}{%
\begin{minipage}{0.88\textwidth}
The results in this chapter were published in the following paper:

\begin{itemize}
    \item Zonghao Chen, Masha Naslidnyk, and Fran\c{c}ois-Xavier Briol.

    Nested Expectations with Kernel Quadrature.

    \textit{International Conference of Machine Learning, 2025}.
\end{itemize}
\end{minipage}%
}
\end{center}

\section{Introduction}
{\color{black}
Many decisions under uncertainty require a nonlinear assessment of an expected outcome, averaged over uncertain scenarios. For example, Bayesian experimental design averages the information gained from possible observations, while lookahead Bayesian optimisation evaluates the benefit of a future experiment after averaging over its possible outcomes. These calculations involve nested expectations: an inner average describes the outcome within a scenario, and an outer average accounts for uncertainty across scenarios. Evaluating them efficiently is important because they are often repeated while comparing candidate decisions, and each inner evaluation may require an expensive simulation or model fit.
\par}

We consider the computational task of estimating such a nested expectation, which is the expectation of a function that itself depends on another unknown conditional expectation. More precisely, let $\Qb$ be a Borel probability measure with density $q$ on $\Theta$ and $\Pb_\theta$ a Borel probability measure with density $p_\theta$ on $\calX \subseteq \mathbb{R}^{d_{\calX}}$ which is parameterized by $\theta \in \Theta \subseteq \mathbb{R}^{d_{\Theta}}$.  Given integrable functions $f: \R \to \R$ and $g:\calX \times \Theta \to \R$, we are interested in estimating:
\vspace{-5pt}
\begin{align}\label{eq:nested}
    I & := \E_{\theta \sim \Qb} \left[ f \left(\E_{X \sim \Pb_\theta} \left[ g(X, \theta) \right] \right) \right]\\
    &= \underbrace{\int_{\Theta} f\Bigg(\underbrace{\int_{\mathcal{X}} g(x,\theta) p_{\theta}(x) dx}_{\text{ inner conditional expectation}} \Bigg) q(\theta) d\theta}_{\text{outer expectation}}. \nonumber
\end{align}
Nested expectations arise within a wide range of tasks, such as the computation of objectives in Bayesian experimental design \cite{Beck2020,Goda2020,Rainforth2024}, of acquisition functions in active learning and Bayesian optimisation \cite{Ginsbourger2010,Yang2024}, of objectives in distributionally-robust optimisation \cite{Shapiro2023,Bariletto2024,Dellaporta2024}, and of statistical divergences \citep{Song2020,Kanagawa2019}.
Computing nested expectations is also a key task beyond machine learning, including in fields ranging from simulation-based uncertainty quantification~\cite{Gordy2010,Giles2019} to manufacturing~\cite{Andradottir2016} 
and geology \cite{Goda2018}.

The estimation of nested expectations is particularly challenging since there are two levels of intractability: the inner conditional expectation, and the outer expectation, both of which must be approximated accurately in order to approximate the nested expectation $I$ accurately. The most widely used algorithm for this problem is \emph{nested Monte Carlo (NMC)} \cite{Lee2003,Hong2009,rainforth2018nesting}. It approximates the inner and outer expectations using Monte Carlo estimators with $N$ and $T$ samples respectively. NMC is consistent under mild conditions, but has a relatively slow rate of convergence. Depending on the regularity of the problem, existing results indicate that we require either $\mathcal{O}(\Delta^{-3})$ or $\calO(\Delta^{-4})$ evaluations of $g$ to obtain a root mean squared error smaller or equal to $\Delta$. 
This tends to be prohibitively expensive; for example, we would expect in the order of either $1$ or $100$ million observations to obtain an error of $\Delta=0.01$. This is infeasible for many applications where obtaining samples or evaluating $g$ is expensive.

This issue has led to the development of a number of methods aiming to reduce the cost. \citet{Bartuska2023} proposed replacing the Monte Carlo estimators with quasi-Monte Carlo (QMC) \cite{Dick2013}. This algorithm, called \emph{nested QMC (NQMC)}, requires only $\calO(\Delta^{-2.5})$ function evaluations to obtain an error of size $\Delta$ (so that we only need in the order of $100,000$ observations for an error of $\Delta=0.01$). However, NQMC requires strong regularity assumptions which may not hold in practice (a monotone second and third derivative for $f$). Separately, \citet{bujok2015multilevel,Giles2019, giles2019decision} proposed to use \emph{multi-level Monte Carlo (MLMC)} and showed that this can further reduce the number of function evaluations to $\calO(\Delta^{-2})$ (so that we only need in the order of $10,000$ observations for an error of $\Delta=0.01$). The algorithm has relatively mild assumptions on $f$ and $g$, which makes it broadly applicable but sub-optimal for applications where $f$ and $g$ are smooth and where we might therefore expect further reductions in cost. 

To fill this gap in the literature, we propose a novel algorithm called \emph{nested kernel quadrature (NKQ)}, which is presented in \Cref{sec:methodology}. NKQ replaces the inner and outer MC estimators of NMC with kernel quadrature (KQ) estimators \citep{sommariva2006numerical}. We show in \Cref{sec:theory} that NKQ requires only $\tilde{\calO}(\Delta^{-\frac{d_\calX}{s_\calX} - \frac{d_\Theta}{s_\Theta}})$ function evaluations to guarantee an error smaller or equal to $\Delta$. Here $\tilde{\calO}$ denotes $\calO$ up to logarithmic terms,  $s_{\mathcal{X}}, s_{\Theta}$ are constants relating to the smoothness of $f$ and $g$ in $\mathcal{X}$ and $\Theta$, and we have $s_{\mathcal{X}} > d_{\mathcal{X}}/2$ and $s_{\Theta} > d_{\Theta}/2$. In the least favorable case, we therefore recover the $\calO(\Delta^{-4})$ of NMC, but when the integrand is smooth and the dimension is not too large, we are able to have a cost which scales better than $\calO(\Delta^{-2})$ and the method significantly outperforms all competitors. In those cases, we may only need in the order of a few hundred or thousands observations for an error of $\Delta=0.01$. This fast rate is demonstrated numerically in \Cref{sec:nkq_experiments}, where we show that NKQ can provide significant accuracy gains in synthetic benchmarks and Bayesian optimisation. Moreover, we show that NKQ can be combined with QMC and MLMC, providing an avenue to further accelerate convergence.

\section{Nested Kernel Quadrature}\label{sec:methodology}

We can now present our novel algorithm: \emph{nested kernel quadrature (NKQ)}.
To simplify the formulas, we write
\begin{equation}\label{eq:J_F_defi}
\vspace{-3pt}
    J(\theta) := \E_{X \sim \Pb_\theta} \left[ g(X, \theta) \right], \quad F(\theta) := f(J(\theta)),  
\end{equation}
so that the nested expectation in \eqref{eq:nested} can be written as $I = \mathbb{E}_{\theta \sim \mathbb{Q}}[F(\theta)]$. We will assume that we have access to
\begin{align*}
  \theta_{1:T} & := [\theta_1,\ldots,\theta_T]^\top \in \Theta^T, \\
  x_{1:N}^{(t)} & := \Big[x_{1}^{(t)},\ldots,x_{N}^{(t)} \Big] \in \calX^N,\\
  g\big(x_{1:N}^{(t)},\theta_t \big) & := \Big[ g\big(x_{1}^{(t)},\theta_t \big), \ldots, g\big(x_{N}^{(t)},\theta_t \big)\Big]  \in \mathbb{R}^N,
\end{align*}
for all $t \in \{1, \ldots, T\}$, and $f$ is a function that can be evaluated.
We do not specify how the point sets are generated, although further (mild) assumptions will be imposed for our theory in \Cref{sec:theory}. Using the same number of function evaluations $N$ per $\theta_t$ is not essential, but we assume this as it significantly simplifies our notation. 
Given the above, we are now ready to define NKQ as the following two-stage algorithm.

\paragraph{Stage I}
For each $t \in \{1, \ldots, T\}$, we estimate the inner conditional expectation $J$ evaluated at $\theta_t$ with $N$ observations $x_{1:N}^{(t)}$ and $g(x_{1:N}^{(t)},\theta_t)$ using a KQ estimator: 
\begin{equation}\label{eq:F_J_KQ}
\hat{J}_{\kq} (\theta_t) :=\mu_{\mathbb{P}_{ \theta_t} } \big(x_{1:N}^{(t)} \big) \big( \boldsymbol{K}_\calX^{(t)} + N \lambda_\calX \boldsymbol{I}_N \big)^{-1} g \big(x_{1:N}^{(t)}, \theta_t \big) .
\end{equation}
Here $k_\calX$ is a reproducing kernel on $\calX$, $\mu_{\mathbb{P}_{\theta_t}}(\cdot) = \mathbb{E}_{X \sim \mathbb{P}_{\theta_t}}[k_{\calX}(X, \cdot)]$ is the KME of $\mathbb{P}_{\theta_t}$ and $\boldsymbol{K}_\calX^{(t)} = k_\calX(x_{1:N}^{(t)}, x_{1:N}^{(t)})$ is an $N \times N$ Gram matrix. 
Using the same kernel $k_\calX$ for each $t\in \{1, \ldots, T\}$ is not essential, but we assume this to be the case for simplicity.
Given these KQ estimates, we then we apply the function $f$ to get $\hat{F}_{\kq}(\theta_t) = f(\hat{J}_{\kq} (\theta_t))$. 

\vspace{1mm}
\paragraph{Stage II}
We use a KQ estimator to approximate the outer expectation using the output of Stage I:
\vspace{-3pt}
\begin{align}\label{eq:NKQ_estimator}
    \hat{I}_{\nkq} &:= \mu_{\mathbb{Q}}(\theta_{1:T}) ( \boldsymbol{K}_\Theta + T \lambda_\Theta \boldsymbol{I}_T)^{-1} \hat{F}_{\kq}( \theta_{1:T} ).
\end{align}
Here $k_\Theta$ is a reproducing kernel on $\Theta$,  $\mu_{\mathbb{Q}} = \mathbb{E}_{\theta \sim \mathbb{Q}}[k_{\Theta}(\theta, \cdot)]$ is the embedding of $\mathbb{Q}$ and $\boldsymbol{K}_\Theta = k_\Theta(\theta_{1:T}, \theta_{1:T})$ is a $T\times T$ Gram matrix.

Combining stage I and II, NKQ can be expressed in a single equation as a nesting of two quadrature rules:
\vspace{-5pt}
\begin{align}\label{eq:nkq_weights}
    \hat{I}_{\nkq} = \sum_{t=1}^{T} w^{\Theta}_t f\left(\sum_{n=1}^N w^{\calX}_{n, t} g(x_n^{(t)}, \theta_t) \right),
\end{align}
where $w^{\calX}_{1, t},\ldots,w^{\calX}_{N, t}$ are the KQ weights used in stage I for $\hat{J}_{\text{KQ}}(\theta_t)$ and $w^{\Theta}_1,\ldots, w^{\Theta}_T$ are the KQ weights used in stage II.
Although these weights are stage-wise optimal when $\lambda_{\mathcal{X}} = \lambda_\Theta = 0$ thanks to the optimality of KQ weights, it is unclear whether they are globally optimal due to the non-linearity of $f$. Note that NMC can be recovered by taking all stage I weights to be $1/N$ and all stage II weights to be $1/T$, which is sub-optimal. 
In addition to the algorithmic simplicity of our proposed estimator NKQ, we demonstrate its superior performance in terms of both rate of convergence (\Cref{sec:theory}) and empirical performances (\Cref{sec:nkq_experiments}). 

NKQ inherits the two main drawbacks of KQ. Firstly, solving the linear systems to obtain the stage I and II weights has a worst-case  computational complexity of $\calO(T N^3 + T^3)$. 
Secondly, NKQ requires closed-form KMEs at both stages: $\mu_{\mathbb{P}_{ \theta_t}}$ for all $t \in \{1,\ldots,T\}$ in stage I, and $\mu_{\mathbb{Q}}$ in stage II. {\color{black}A change of variables can provide tractable kernel embeddings and, after precomputing the weights, reduce the cost of subsequent evaluations to $\calO(TN+T)$, as described below.}

NKQ requires the selection of hyperparameters, including for the kernels in both stage I and II. 
We typically take $k_{\mathcal{X}}$ and $k_{\Theta}$ to be Mat\'ern kernels whose orders are determined by the smoothness of $f$ and $g$ (as justified by \Cref{thm:main}; see \Cref{sec:theory} for details).
This leaves us with a choice of kernel hyperparameters which include lengthscales $\gamma_{\mathcal{X}}, \gamma_{\Theta}>0$ and amplitudes $A_{\mathcal{X}}, A_{\Theta}>0$. 
The lengthscales are selected via the median heuristic. 
The regularizers are set to $\lambda_{\mathcal{X}} = \lambda_{0, \calX} N^{-\frac{2s_\calX}{d_\calX}} (\log N)^{\frac{2s_\calX+2}{d_\calX}}$ and $ \lambda_{\Theta} = \lambda_{0,\Theta} T^{-\frac{2s_\Theta}{d_\Theta}} (\log T)^{\frac{2s_\Theta+2}{d_\Theta}}$ following \Cref{thm:main}, where $\lambda_{0,\calX}, \lambda_{0,\Theta}$ are selected with grid search over $\{0.01, 0.1, 1.0\}$. Finally, we standardise our function values (by subtracting the empirical mean then dividing by the empirical standard deviation), and then set the amplitudes to $A_{\mathcal{X}}=A_{\Theta}=1$. This last choice could further be improved using a grid search, but we do not do this as we do not notice significant improvements when doing so in experiments and this tends to increase the cost.

Before presenting our theoretical results, we briefly comment on the connection with existing KQ methods.
If we could evaluate the exact expression for the inner conditional expectation $J(\theta)$ pointwise, then the KQ estimator for $I$ would be $\bar{I}_{\kq} = \mu_{\mathbb{Q}}(\theta_{1:T}) ( \boldsymbol{K}_\Theta + T \lambda_\Theta \boldsymbol{I}_T)^{-1} F( \theta_{1:T})$. Comparing with \eqref{eq:NKQ_estimator}, NKQ can thus be seen as KQ with noisy function values $\hat{F}_{\kq}( \theta_{1:T} )$ (replacing the exact values $F( \theta_{1:T})$ in \eqref{eq:NKQ_estimator}). 
Although it is proved in \citet{Cai2023} that noisy observations make KQ converge at a slower rate, we prove that the stage II observation noise is of the same order as the stage I error, and consequently, we can still treat stage II KQ as noiseless kernel ridge regression and the additional error caused by the stage II observation noise would be \emph{subsumed} by the stage I error (See \Cref{rem:noise_stage_2}).
NKQ is also closely related to a family of regression-based methods for estimating conditional expectations~\cite{longstaff2001valuing,chen2024conditional}. Indeed, with a slight modification of Stage II in \eqref{eq:NKQ_estimator}, we can obtain an estimator of $J(\theta)$ that we call \emph{conditional kernel quadrature (CKQ)} 
\begin{align}\label{eq:ckq}
    \hat{J}_{\text{CKQ}}(\theta) := k_\Theta(\theta, \theta_{1:T}) ( \boldsymbol{K}_\Theta + T \lambda_\Theta \boldsymbol{I}_T)^{-1} \hat{J}_{\kq}( \theta_{1:T}) .
\end{align}
CKQ highly resembles \textit{conditional BQ (CBQ)}~\citep{chen2024conditional}; the difference is in stage II, where CBQ uses heteroskedastic Gaussian process regression whilst CKQ uses kernel ridge regression. Interestingly, the proof in this paper leads to a much better rate for CKQ than the best known rate for CBQ (see \Cref{rem:ckq_rate}).

{\color{black}
\subsection{\texorpdfstring{``Change of Variable'' Trick for Kernel Quadrature}{Change of Variable Trick for Kernel Quadrature}}\label{sec:reparam}
As discussed above, the two major bottlenecks of KQ/NKQ are:
\begin{itemize}
    \item The need for a closed-form KME $\E_{X \sim \Pb}[k(X, x)]$.
    \item The $\calO(N^3)$ computational cost of inverting the Gram matrix $k(x_{1:N}, x_{1:N})$.
\end{itemize}
The ``change of variable'' trick can address both challenges. Here, we only present the idea for KQ but the same holds for NKQ in both stages.

The integral of interest is $I = \int_\calX h(x) \Pb(dx)$.
Suppose we can find a transformation $\Phi$ such that $X = \Phi(U)\sim\Pb$, where $U\sim\Ub$ is uniform on $\calU=[0,1]^d$. In one dimension, $\Phi$ can be chosen as the inverse cumulative distribution function of $\Pb$.
Then the integral $I$ can be equivalently expressed as $I = \int_\calU h(\Phi(u)) d\Ub(u)$, by the change-of-variables formula~\citep[Section 8.2]{stirzaker2003elementary}.
Now the integrand changes from $h: \calX \to \R$ to $h \circ \Phi: \calU \to \R$ and the kernel quadrature estimator becomes
\begin{align*}
    \hat{I}_{\kq} = \E_{U \sim \Ub} [k_\calU(U, u_{1:N})] \left( k_\calU(u_{1:N}, u_{1:N}) + N \lambda \Id_N \right)^{-1} (h \circ \Phi)(u_{1:N}) .
\end{align*}
Here $u_{1:N}$ are uniform samples on $\calU$, and $k_\calU$ is a reproducing kernel on $\calU$.
For suitable kernels, closed-form KMEs under $\Ub$ are available in Table 1 of \citet{Briol2019PI} or the \texttt{ProbNum} package~\citep{Wenger2021}, which addresses the first challenge.
Additionally, notice that both the Gram matrix $k_\calU(u_{1:N}, u_{1:N})$ and the KME $\E_{U \sim \Ub} [k_\calU(U, u_{1:N})]$ are independent of $h$ and $\Phi$, so the KQ weights $w_{1:N}^{\kq} = \E_{U \sim \Ub} [k_\calU(U, u_{1:N})] \left( k_\calU(u_{1:N}, u_{1:N}) + N \lambda \Id_N \right)^{-1}$ can be precomputed and stored when the points, kernel and regularizer are fixed.
As a result, KQ becomes a simple weighted average of function evaluations $\sum_{i=1}^N w_i^{\kq} h(x_i)$. Here $x_i=\Phi(u_i)$. After the initial $\calO(N^3)$ computation of the weights, each subsequent integral evaluation has linear cost $\calO(N)$.
For NKQ, we apply this construction at both stages: write $\theta=\Psi(V)\sim\Qb$ and $X=\Phi_\theta(U)\sim\Pb_\theta$, with independent uniform variables $V$ and $U$ on the corresponding unit cubes, and apply the two KQ rules to $g(\Phi_{\Psi(v)}(u),\Psi(v))$. Reusing the same inner uniform points for each outer point makes the inner KQ weights independent of $\theta$, so both sets of weights can be precomputed at cost $\calO(N^3+T^3)$. Subsequent evaluations of the nested estimator then cost $\calO(NT+T)$.
}

\section{Theoretical Results}\label{sec:theory}

In this section, we derive a convergence rate for the absolute error $|\hat{I}_{\nkq} - I|$ as the number of samples $N, T \to \infty$. Before doing so, we recall the connection between RKHSs and Sobolev spaces. A kernel $k$ on $\mathbb{R}^d$ is said to be translation invariant if $k(x, x^\prime) = \Psi(x-x^\prime)$ for some positive definite function $\Psi$ whose Fourier transform $\hat{\Psi}(\omega)$ is a finite non-negative measure on $\mathbb{R}^d$~\citep[Theorem 6.6]{wendland2004scattered}. 
Suppose $\calX$ has a Lipschitz boundary, if $k$ is translation invariant and its Fourier transform $\hat{\Psi}(\omega)$ decays as $\mathcal O (1 + \|\omega \|_2^2)^{-s}$ when $\omega \to \infty$ for $s > d/2$, then its RKHS $\calH_k$ is norm equivalent to the Sobolev space $W_2^s(\calX)$~\citep[Corollary 10.48]{wendland2004scattered}.
More specifically, it means that their set of functions coincide and there are constants $c_1,c_2>0$ such that $c_1\|h\|_{\calH_k} \leq\|h\|_{W_2^s(\calX)} \leq c_2\|h\|_{\calH_k}$ holds for all $h \in \calH_k$. 
In this paper, we call such kernel a \emph{Sobolev reproducing kernel of smoothness $s$}.
An important example of Sobolev kernel is the Matérn kernel--- the RKHS of a Matérn-$\nu$ kernel is norm-equivalent to $W_2^s(\calX)$ with $s=\nu + d/2$.
All Sobolev kernels are bounded, i.e. $\sup_{x \in \calX} k(x, x) \leq \kappa$ for some positive constant $\kappa$. 
When the context is clear, we use $\|f\|_{s,2} \coloneq \|f\|_{W_2^s(\calX)}$ to denote the Sobolev space norm. 
\begin{thm}\label{thm:main}
Let $\calX =[0,1]^{d_\calX}$ and $\Theta =[0,1]^{d_\Theta}$. {\color{black}Let $q$ and $p_\theta$ denote the densities of $\Qb$ and $\Pb_\theta$, respectively, with respect to Lebesgue measure on $\Theta$ and $\calX$.} Suppose $\theta_{1:T}$ are i.i.d. samples from $\Qb$ and $x_{1:N}^{(t)}$ are i.i.d samples from $\Pb_{\theta_t}$ for all $t \in \{1, \cdots, T\}$.
Suppose further that $k_\calX$ and $k_\Theta$ are Sobolev kernels of smoothness $s_\calX > d_\calX / 2$ and $s_\Theta > d_\Theta/2$, and that the following conditions hold
\begin{enumerate}
[itemsep=5.0pt,topsep=0pt,leftmargin=*]
    \item [(1)] There exist $G_{0, \Theta}, G_{1, \Theta}, G_{0, \calX}, G_{1, \calX} > 0$ such that $G_{0, \Theta} \leq q(\theta) \leq G_{1, \Theta}$ and $G_{0, \calX} \leq p_\theta(x) \leq G_{1, \calX}$ for any $\theta \in \Theta$ and $x \in \calX$. 
    \customlabel{as:equivalence}{(1)} 
    \item [(2)] There exists $S_1 >0$ such that for any $\theta \in \Theta$ and any $\beta \in \N^{d_\Theta}$ with $|\beta| \leq s_\Theta$, $\| D_\theta^\beta g(\cdot, \theta) \|_{ s_\calX,2 } \leq S_1$.
    \customlabel{as:app_true_g_smoothness}{(2)} 
    \item[(3)] There exist $S_2,S_3 >0$ such that for any $x \in \calX$, $\| g(x, \cdot) \|_{ s_\Theta ,2} \leq S_2$ and $\|\theta \mapsto p_\theta(x) \|_{ s_\Theta ,2} \leq S_3 \leq 1$. 
    \customlabel{as:app_true_J_smoothness}{(3)} 
    \item[(4)] 
    There exists $S_4 >0$ such that derivatives of $f$ up to and including order $s_\Theta + 1$ are bounded by $S_4$.
    \customlabel{as:app_lipschitz}{(4)} 
\end{enumerate}
Then, there exists $N_0, T_0 \in \mathbb{N}^{+}$ such that for $N>N_0, T>T_0$, we can take $\lambda_{\calX} \asymp N^{-2\frac{s_\calX}{d_\calX}} (\log N)^{\frac{2s_\calX+2}{d_\calX}}$ and $\lambda_{\Theta} \asymp T^{-2\frac{s_\Theta}{d_\Theta}} (\log T)^{\frac{2s_\Theta+2}{d_\Theta}}$ to obtain the following bound
\begin{align*}
\vspace{-5pt}
    \left| I - \hat{I}_{\nkq} \right| \leq \tau \left( C_1 N^{- \frac{ s_\calX}{d_\calX} } (\log N)^{\frac{s_\calX+1}{d_\calX}} + C_2 T^{- \frac{ s_\Theta}{d_\Theta} } (\log T)^{\frac{s_\Theta+1}{d_\Theta}} \right), 
\end{align*}
which holds with probability at least $1 - 8 e^{-\tau}$.
$C_1, C_2$ are two constants independent of $N, T, \tau$.
\end{thm}
\begin{cor}\label{cor:nkq}
    Suppose all assumptions in \Cref{thm:main} hold. If we set $N = \tilde{\calO} (\Delta^{-\frac{d_\calX}{s_\calX} })$ and $T = \tilde{\calO}(\Delta^{- \frac{d_\Theta}{s_\Theta}})$, then $N \times T = \tilde{\calO}(\Delta^{-\frac{d_\calX}{s_\calX} - \frac{d_\Theta}{s_\Theta}})$ samples are sufficient to guarantee that $|I - \hat{I}_{\nkq}| \leq \Delta$ holds with high probability. 
\end{cor}
To prove these results, we can decompose $| I - \hat{I}_{\nkq}|$ into the sum of stage I and stage II errors, which can be bounded by terms of order $
N^{- \frac{ s_\calX}{d_\calX}} (\log N)^{\frac{s_\calX+1}{d_\calX}}$ and $T^{- \frac{ s_\Theta}{d_\Theta}} (\log T)^{\frac{s_\Theta+1}{d_\Theta}}$ respectively; 
see  \Cref{sec:proof}. 
Interestingly, note that the stage II error does not suffer from the fact that we are using noisy observations $\hat{F}_{\kq}(\theta_{1:T})$ and we maintain the standard KQ rate up to logarithm terms (see \Cref{rem:noise_stage_2}).
We emphasize that our bound indicates that the tail of $|I - \hat{I}_{\nkq}|$ is sub-exponential. This contrasts
with existing work on Monte Carlo methods for nested expectation, which only provides upper bounds on the expectation of error with no
constraints on its tails~\citep{giles2018MLMC,rainforth2018nesting,Bartuska2023}.

We now briefly discuss our assumptions.
Assumption \ref{as:equivalence} is mild and allows $L_2(\Pb_\theta)$ (resp. $L_2(\Qb)$) to be norm equivalent to $L_2(\calX)$ (resp. $L_2(\Theta)$), which is widely used in statistical learning theory that involves Sobolev spaces \cite{fischer2020sobolev, suzuki2021deep}.
Since our proof essentially translates quadrature error into generalization error of  kernel ridge regression, Assumptions \ref{as:app_true_g_smoothness}, \ref{as:app_true_J_smoothness}, \ref{as:app_lipschitz} ensure that functions $f, g$ and the density $p$ have enough regularity so that the regression targets in both stage I and stage II belong to the correct Sobolev spaces. These are more restrictive, but are essential to obtain our fast rate and are common assumptions in the KQ literature.
Assumptions \ref{as:app_true_g_smoothness}, \ref{as:app_true_J_smoothness}, \ref{as:app_lipschitz} can be relaxed if mis-specification is allowed; see e.g.~\citet{fischer2020sobolev,Kanagawa2019, zhang2023optimality}. 
\Cref{thm:main} shows that for NKQ to have a fast convergence rate, one ought to use Sobolev kernels which are as smooth as possible in both stages.
Furthermore, when $s_\calX = s_\Theta = \infty$ (e.g. when the integrand and kernels belong to Gaussian RKHSs), our proof could be modified to show an exponential rate of convergence in a similar fashion as \citet[Theorem 10]{briol2018statistical} or \citet{Karvonen2020}.

\begin{table}[t]
\centering
\renewcommand{\arraystretch}{1.8}
\setlength{\tabcolsep}{6pt}
\resizebox{\textwidth}{!}{%
\begin{tabular}{|c|c|c|c|}
\hline
\textbf{Method} & \textbf{Cost} & \textbf{Method} & \textbf{Cost} \\ \hline
NMC & $\calO(\Delta^{-3})$ or $\calO(\Delta^{-4})$ &
NQMC & $\calO(\Delta^{-2.5})$ \\ \hline
MLMC & $\calO(\Delta^{-2})$ &
\textbf{NKQ (\Cref{cor:nkq})} & $\tilde{\calO}\Big(\Delta^{-\frac{d_\calX}{s_\calX} - \frac{d_\Theta}{s_\Theta}} \Big)$ \\ \hline
\end{tabular}
}
\caption[Cost of nested-expectation methods]{Cost of methods for nested expectations, measured through the number of function evaluations required to ensure $|I - \hat{I}| \leq \Delta$. 
NMC rate is taken from Theorem 3 of \citet{rainforth2018nesting}, NQMC rate is taken from Proposition 4 of \citet{Bartuska2023}, MLMC rate is taken from Section 3.1 of \citet{giles2018MLMC}. Smaller exponents $r$ in $\Delta^{-r}$ indicate a cheaper method. } \label{tab:table}
\end{table}

\begin{rem}[Noisy observations in Stage II of NKQ]\label{rem:noise_stage_2}
    Note that NKQ employs noisy observations $\{\theta_t, \hat{F}_{\kq}(\theta_t)\}_{t=1}^T$ in stage II KQ rather than the ground truth observations $\{\theta_t, F(\theta_t)\}_{t=1}^T$. 
    Although \citet{Cai2023} establishes that KQ with noisy observations converges at a slower rate than KQ with noiseless observations, a key distinction in our setting is that, as shown in \eqref{eq:bernstein_noise}, the observation noise in stage II KQ is of order $\tilde{\calO}(N^{-\frac{s_\calX}{d_\calX}})$, whereas the noise in \citet{Cai2023} remains at a constant level.
    As a result, we can still use KQ in stage II as if the observations $\{\hat{F}_{\kq}(\theta_t)\}_{t=1}^T$ are noiseless, and the additional error it introduces happens to be of the same order as the stage I error $\tilde{\calO}(N^{-\frac{s_\calX}{d_\calX}})$ and is therefore subsumed by it.
\end{rem}

\begin{rem}[Convergence rate for CKQ]\label{rem:ckq_rate}
    For the CKQ estimator $\hat{J}_{\text{CKQ}}$ (defined in \eqref{eq:ckq}) that approximates the parametric / conditional expectation $J(\theta)$ uniformly over all $\theta \in \Theta$, the error can be upper bounded in the same way as NKQ. 
    The \emph{stage I error} and can be shown to be $\tilde{\calO}(N^{-\frac{s_\calX}{d_\calX}})$ using the same analysis from \eqref{eq:lipschitz_F} to \eqref{eq:high_prob_hat_g_g}; and the \emph{stage II error} and can be shown to be $\tilde{\calO}(T^{-\frac{s_\Theta}{d_\Theta}})$ using the same analysis from \eqref{eq:stage_1} to \eqref{eq:stage_2}. Combining the two error terms, we have $\| J - \hat{J}_{\ckq} \|_{L_2(\Qb)} = \tilde{\calO}(N^{-\frac{s_\calX}{d_\calX}} + T^{-\frac{s_\Theta}{d_\Theta}})$ holds with probability at least $1 - 8 e^{-\tau}$.
    The rate is better than the best known rate $\calO(N^{-\frac{s_\calX}{d_\calX}} + T^{-\frac{1}{4}})$ of CBQ proved in Theorem 1 of \cite{chen2024conditional} since $\frac{s_\Theta}{d_\Theta} > \frac{1}{2} > \frac{1}{4}$.
    The intuition behind the faster rate is that CKQ benefits from the extra flexibility of choose regularization parameters $\lambda_\calX, \lambda_\Theta$; while CBQ, as a two stage Gaussian Process based approach, is limited to choose $\lambda_\Theta$ equal to the heteroskedastic noise from the first stage. It may be possible to modify the proof of \cite{chen2024conditional} to improve the rate further, but this has not been explored to date.
\end{rem}

In \Cref{tab:table}, we compare the \emph{cost} of all methods evaluated by the number of  evaluations required to ensure $|\hat{I} - I| \leq \Delta$.
We can see that NKQ is the only method that explicitly exploits the smoothness of $g, p, f$ in the problem so that it outperforms all other methods when $\frac{d_{\calX}}{s_{\calX}} + \frac{d_\Theta}{s_\Theta} <2$.

\section{Experiments}\label{sec:nkq_experiments}

We now illustrate NKQ over a range of applications, including some where the theory does not hold but where we still observe significant gains in accuracy. 
The code to reproduce all experiments is available at
 \url{https://github.com/hudsonchen/nest_kq}.

\vspace{-2mm}

\paragraph{Synthetic Experiment}
We start by verifying the bound in \Cref{thm:main} using the following synthetic example: $\Qb = \operatorname{U}[0,1]$, $\Pb_\theta = \operatorname{U}[0,1]$, $g(x, \theta) = x^{\frac{5}{2}} + \theta^{\frac{5}{2}}$, and $f(z) = z^2$, in which case $I=0.4115$ can be computed analytically. We estimate $I$ with i.i.d. samples $\theta_{1:T} \sim \operatorname{U}[0,1]$ and i.i.d. samples $x_{1:N}^{(t)} \sim \operatorname{U}[0,1]$ for $t \in \{1, \ldots, T\}$.
The assumptions from \Cref{thm:main} are satisfied with $s_\calX = s_\Theta = 2$ and $d_\calX = d_\Theta = 1$.
Therefore, to reach the absolute error threshold $\Delta$, we choose $N = T = \Delta^{-0.5}$ for NKQ  following \Cref{cor:nkq}. 
On the other hand, based on Theorem 3 of \citet{rainforth2018nesting}, the optimal way of assigning samples for NMC is to choose $N = \sqrt{T} = \Delta^{-1}$.

\begin{figure}[t]
\vspace{-5pt}
\centering
\begin{subfigure}{0.48\linewidth}
    \centering
    \includegraphics[width=\linewidth]{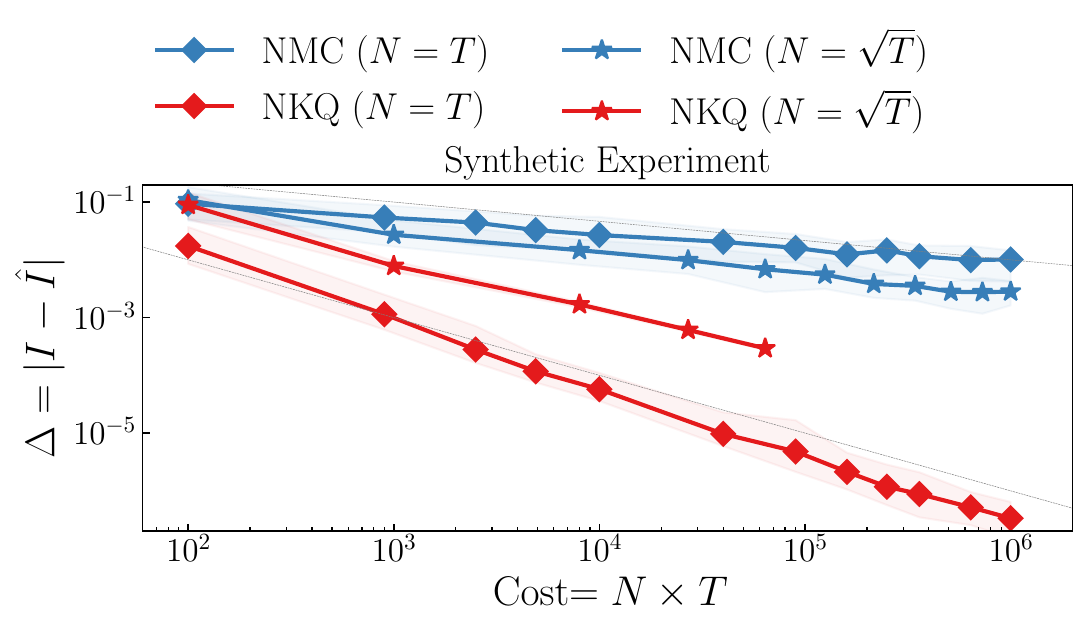}
\end{subfigure}
\begin{subfigure}{0.48\linewidth}
    \centering
    \includegraphics[width=\linewidth]{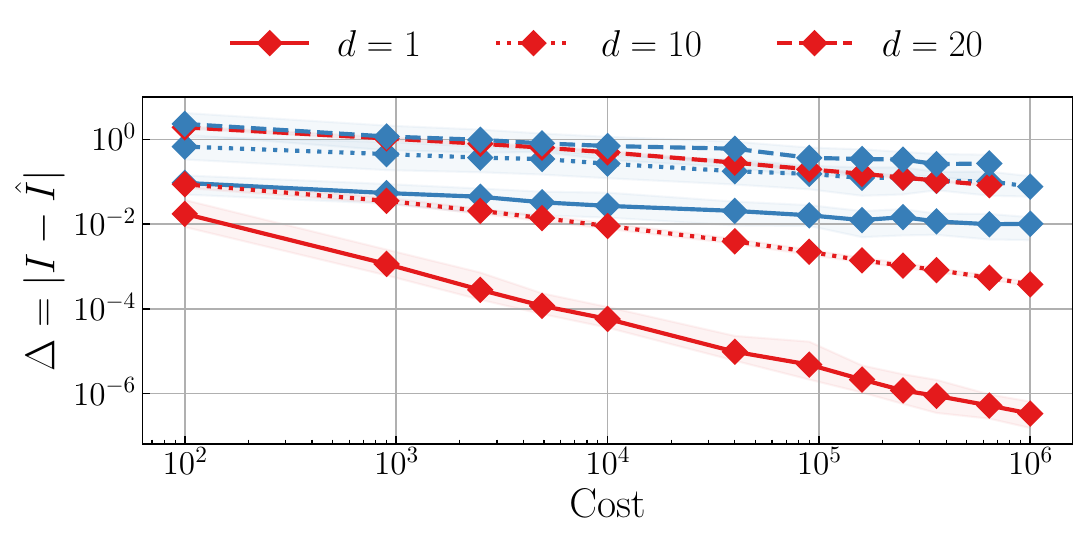}
\end{subfigure}
\caption[Synthetic NKQ experiment]{\emph{Synthetic experiment.}
\textbf{Top:} Verification of theoretical results. The thin grey lines are theoretical rates of 
$\Delta = \text{Cost}^{-1}$ and $\Delta = \text{Cost}^{-1/3}$.
\textbf{Bottom:} Comparison of NKQ and NMC as dimension $d$ increases.
Results are averaged over 1000 independent runs, while shaded regions 
give the 25\%-75\% quantiles.}
\label{fig:toy_experiment}
\vspace{-10pt}
\end{figure}

In \Cref{fig:toy_experiment} \textbf{Top}, we see that the optimal choice of $N$ and $T$ suggested by the theory indeed results in a faster rate of convergence for both NMC and NKQ. 
For this synthetic problem, we confirm that both the theoretical rates of NKQ ($\text{Cost} = \Delta^{-1}$) and NMC ($\text{Cost} = \Delta^{-3}$) from \Cref{thm:main} and \citet[Theorem 3]{rainforth2018nesting} are indeed realized. We also adapt the synthetic problem to higher dimensions ($d_\calX=d_\Theta=d$) and observe in \Cref{fig:toy_experiment} \textbf{Bottom} that the performance gap between NKQ and NMC closes down as dimension grows. 
Such behaviour is expected because the cost of NKQ is $\tilde{\calO}(\Delta^{- \frac{d_\calX}{s_\calX} - \frac{d_\Theta}{s_\Theta}})$ and therefore degrades as the dimensions $d_\calX$ and $d_\Theta$ increase; whilst the cost of NMC remains the same.

\vspace{-1mm}

\begin{figure*}[t]
\vspace{-5pt}
    \centering
    \includegraphics[width=1.0\linewidth]{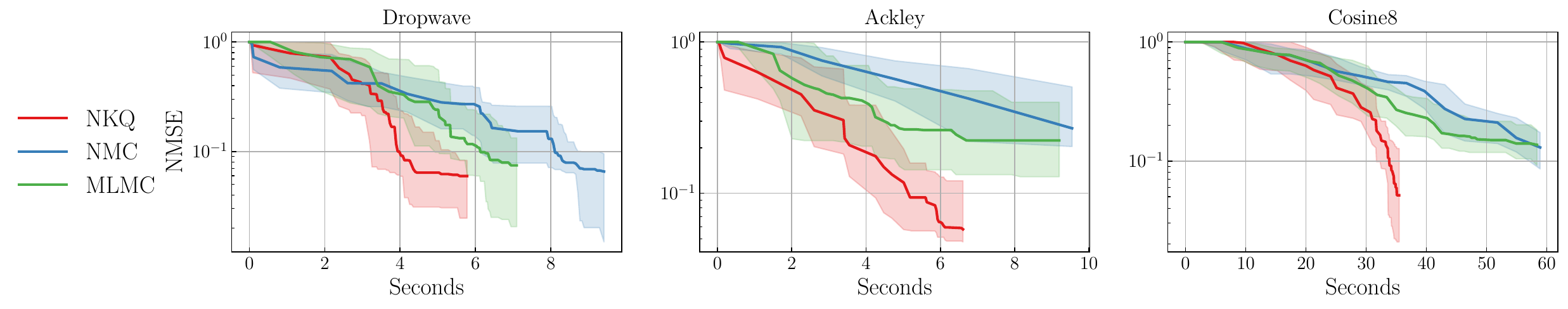}
    \vspace{-20pt}
    \caption[Bayesian optimization with look-ahead acquisition]{\emph{Bayesian optimization with look ahead acquisition function}. The plots are NMSE against accumulative wall clock time.
    Results are averaged over 100 independent runs, while shaded regions give the 25\%-75\% quantiles. 
    }
    \label{fig:bo}
    \vspace{-5pt}
\end{figure*}

\vspace{-1mm}
\paragraph{Bayesian Optimization}
We conclude with an application in Bayesian optimization. Typical acquisition functions are greedy approaches that maximize the immediate reward, while look-ahead acquisition functions optimize accumulated reward over a planning horizon, which results in reduced number of required function evaluations~\citep{Ginsbourger2010,gonzalez2016glasses,Wu2019,Yang2024}. 
The utility of a two-step look ahead acquisition functions can be written as the following nested expectation.
\vspace{-5pt}
\begin{align*}
    \alpha(z ; \calD) := \E_{f_{\mid \calD} } \left[g(f_{\mid \calD}, z) + \max _{z^\prime} \E_{f_{\mid \calD^\prime}  } \left[g \left(f_{\mid \calD^\prime}, z^\prime \right)\right]\right],
\end{align*}
where $f_{\mid \calD}, f_{\mid \calD^\prime}$ are the posterior distributions given data $\calD$ and $\calD^\prime:= \calD \cup (z , f_{\mid \calD}(z))$. 
In this experiment, the prior is a Gaussian process with zero mean and Mat\'{e}rn-$0.5$ covariance so the posterior $f_{\mid \calD}$ remain a Gaussian process.
The initial starting data $\calD_0$ consists of $2$ points sampled uniformly from a prespecified interval.
Here, $g$ is the reward function and we use q-expected improvement~\cite{wang2020parallel} with $q=2$ so $z=(z_1, z_2)$ and
$g(f_{\mid \calD}, z) = \max _{j=1, 2} ( f_{\mid \calD}(z_j) - r_{\max} ), 0)$. The constant 
$r_{\max}$ is the maximum reward obtained from previous queries.
Although $f_{\mid \calD} $ (resp. $f_{\mid \calD^\prime}$) is a Gaussian process, we only ever consider its evaluation on $z$ (resp. $z^\prime$), and we therefore only have to integrate against two-dimensional Gaussians. 
Notationally speaking, $f_{\mid \calD^\prime}(z_1, z_2)$ correspond to $x$ and $f_{\mid \calD}(z_1, z_2)$ correspond to $\theta$ in \eqref{eq:nested} (i.e. $d_\calX = d_\Theta = 2$), but we use the notation of $f_{\mid \calD^\prime}, f_{\mid \calD}$ to stay consistent with the GP literature. 
As a result of the \textit{max} operation, $s_\calX = 1$ but we do not have sufficient smoothness in $\Theta$.

We benchmark NKQ, NMC and MLMC on three synthetic tasks from BoTorch~\citep{balandat2020botorch}. 
For NKQ, both $k_\calX$ and $k_\Theta$ are Mat\'ern-$\frac{1}{2}$ kernels since we want to be conservative about the smoothness. 
Although both $\Qb$ and $\Pb_\theta$ are Gaussian so closed-form KMEs are available, we use the ``change of variable trick'' which maps Gaussian distributions to two uniform distributions over $[0,1]^d$ to reduce the computational complexity of NKQ to $\calO(T \times N)$.
To reach a specific error threshold $\Delta=0.01$, following \Cref{tab:table}, we choose $N=T=\Delta^{-2}$ for NMC and $N=T=\Delta^{-1}$ for NKQ. 
For MLMC, we use the same code as \citet{Yang2024}.
The normalized mean squared error (NMSE) $\frac{\|\max _{z \in \calD_{\mathcal{S}} } f_{\text{BB}}(z)-f_{\text{BB}}(z^*)\|^2}{\|\max_{z \in \calD_0} f_{\text{BB}}(z) - f_{\text{BB}}(z^*)\|^2}$ is used as  performance metric, where $\calD_0$ (resp. $\calD_{\mathcal{S}}$) is queried data at initialization (resp. after $\mathcal{S}$ iterations),  $f_{\text{BB}}$ is the black box function to be optimized and $f_{\text{BB}}(z^*)$ is the maximum reward. 

In \Cref{fig:bo}, we compare the efficiency of each method by plotting their NMSE against cumulative computational time in wall clock. We can see that NKQ achieves the lowest NMSE among all methods under a fixed amount of computational time in all three datasets, even though the assumptions of \Cref{thm:main} are not all satisfied.
Since the \texttt{Dropwave}, \texttt{Ackley}, and \texttt{Cosine8} functions are synthetic and computationally cheap, we expect the advantages of NKQ to be more pronounced for Bayesian optimization on real-world expensive problems.
Furthermore, many other utility functions in Bayesian optimization—such as predictive entropy search—involve nested expectations~\citep{balandat2020botorch}. We leave the empirical evaluation of NKQ on these utility functions to future work.

\vspace{-1mm}
\section{Conclusion}
This paper introduces a novel estimator for nested expectations based on kernel quadrature. 
We prove in \Cref{thm:main} that our method has a faster rate of convergence than existing methods provided that the problem has sufficient smoothness. 
This theoretical result is consistent with the empirical evidence in several numerical experiments. 
Additionally, even when the problem is not as smooth as the theory requires, NKQ can still outperform baseline methods potentially due to the use of non-equal weights.

Following our work, there remain a number of interesting future problems and we now highlight two main ones. Firstly, we propose a combination of KQ and MLMC that we call MLKQ. However, we believe our current theoretical rate for MLKQ is sub-optimal due to the sub-optimal allocation of samples at each level. Further work will therefore be needed to determine whether this is a viable approach in some cases.
Secondly, for applications where function evaluations are extremely expensive, NKQ could be extended to its Bayesian counterpart. This would allow us to use the finite sample uncertainty quantification for adaptive selection of samples, which could further improve performance.
Finally, establishing minimax lower bounds for nested expectation remains an open and compelling problem. The main difficulty lies in its two-stage structure. To the best of our knowledge, existing minimax lower bounds for two-stage problem typically reduce the problem to a one-stage problem before deriving the bound; see, for example, \citet[Chapter 3]{chen2011rate}, \citet[Appendix F.1]{meunier2024nonparametricinstrumentalregressionkernel}, and \citet{zhang2025minimax}. However, it remains unclear how to directly establish meaningful minimax lower bounds of genuinely two-stage problems, such as our nested expectation.

\chapter{Minimum MMD for Generative Modelling}
\label{sec:drmmd}

\begin{center}
\setlength{\fboxsep}{12pt}
\fcolorbox{paperboxborder}{paperboxbg}{%
\begin{minipage}{0.88\textwidth}
The results in this chapter were published in the following paper:

\begin{itemize}
    \item Zonghao Chen, Aratrika Mustafi, Pierre Glaser, Anna Korba, Arthur Gretton, and Bharath K. Sriperumbudur.

    (De)-regularized Maximum Mean Discrepancy Gradient Flow.

    \textit{Journal of Machine Learning Research, 2025}.
\end{itemize}

\end{minipage}%
}
\end{center}

\section{Introduction}
Many applications in computational statistics and machine learning involve approximating 
a probability distribution $\pi$ on $\R^d$ (in terms of samples) when only partial information on $\pi$ is accessible.
For example, in Bayesian inference, $\pi$ is known up to an intractable normalizing constant for complex models. 
The setting of interest in this work is the so-called generative modeling setting \citep{brock2019large,ho2020denoising,song2020score,franceschi2023unifying}  where one assumes access to a set of samples from the target distribution $\pi $, with the goal being to generate new samples from $\pi$. 
In some cases of the generative modelling,  the target distributions $\pi$ do not even admit a density with respect to Lebesgue measure on $\R^d$. 
This is motivated by the \emph{manifold hypothesis}: high-dimensional observations often lie on or near a manifold of much lower dimension~\citep{fefferman2016testing}.

Recently, a popular framework to perform this task involves solving a minimization problem in
$\cP(\R^d)$, the space of probability distributions over $\R^d$, by choosing the objective function to be a \emph{dissimilarity} function $\calD(\cdot \| \pi)$ (a distance or divergence) between probability distributions that satisfies: $\calD(\nu\| \pi)=0$ if and only if $\nu=\pi$. 
Since only samples from $\pi$ are available, this problem is solved approximately---yet, the approximate minimizers may
converge to $\pi$ as the number of available samples increases. 
In particular, in the space of probability distributions with bounded second moment $\cP_2(\R^d)$, a common approach is to solve this optimization problem by running a (sample-based) approximation of the Wasserstein gradient flow of the functional $\cF = \calD(\cdot \| \pi)$, which defines a path of distributions with steepest descent for $\cF$ with respect to the Wasserstein-2 distance.

In generative modeling, the choice of $\calD(\cdot \| \pi) $ depends on two crucial aspects: First, its flow should admit consistent and preferably tractable numerical implementations using only samples from $\pi$, and second, under reasonable assumptions, it should guarantee convergence of its flow to $\pi$, the unique \emph{global} minimizer. Combined, these two properties ensure that, in the large sample limit, this algorithm generates new samples from the target.
Unfortunately, verifying these two properties simultaneously has proved to be a surprisingly challenging task. For instance, recent approaches based on Maximum Mean Discrepancy (MMD)
\citep{arbel2019maximum,hertrich2023generative}, the sliced-Wasserstein distance \citep{liutkus2019sliced} and the Sinkhorn divergence \citep{genevay2018learning} typically admit consistent finite-sample implementations,
but their global convergence guarantees---when they exist---do not apply to most practical targets $\pi$ of interest. 
To guarantee global convergence, one could instead choose $\calD$ as an $f$-divergence.
For instance, the (reverse) KL divergence 
and $\chi^2$-divergence are geodesically convex \citep[Definition 16.5]{villani2009optimal} when the target is log-concave (i.e. $\pi \propto e^{-V}$ with $V$ convex) \citep{ohta2011displacement}, and hence their flows enjoy better convergence behaviour. 
However, while the population Wasserstein gradient flows of the $\chi^2$ and KL divergences are well-defined \citep{jordan1998variational, chewi2020svgd}, they do not come naturally with consistent and tractable sample-based implementations. 
Multiple approaches propose to solve a surrogate optimization problem with samples at each iteration of the flow \citep{gao2019deep,ansari2020refining,simons2022variational,birrell2022f, gu2022lipschitz,liu2023variational}; however, it remains to be formally established whether these surrogate problems preserve the desirable convergence guarantees of $f$-divergence flows.

In the face of the trade-offs present in the current approaches, a natural question arises: \emph{Does there exist a divergence functional $\calD(\cdot \| \pi)$ whose gradient flow both globally converges, and admits a tractable, consistent sample-based implementation?}
In this work, we take a step towards a positive answer by constructing a ``de-regularized'' variant of the Maximum Mean Discrepancy ($\dmmd$) and its associated Wasserstein gradient flow. 
We prove that the $\dmmd$ gradient flow converges exponentially to the global minimum up to a controllable barrier term for targets $\pi$ that satisfy a Poincar\'{e} inequality, in both continuous and discrete time regimes. To do so, we establish and leverage a connection between the $\dmmd$ and the $\chi^2$ divergence, an $f$-divergence whose gradient flow benefits from strong convergence guarantees.
By alternatively viewing $\dmmd$ as MMD with a regularized kernel, $\dmmd$ flow comes with a consistent and tractable implementation when only samples from the target $\pi$ are available. 
In addition, given the empirical success of using adaptive kernels in MMD-based generative models~\citep{galashov2024deep, li2017mmd,arbel2018gradient}, 
our paper shows theoretically that using adaptive kernels through adaptive regularization indeed improves the convergence of MMD gradient flow.

In this paper, we make the following assumption on our kernel.
\begin{innercustomassumption}\label[assumption]{assumption:universal}
$k: \R^d \times \R^d \to \R$ is a  continuous and $c_0$-universal kernel, and there exists $K>0$ such that $\sup_{x} k(x,x) \leq K$.
\end{innercustomassumption}

\begin{innercustomassumption}\label[assumption]{assumption:bounded_kernel}
$k: \R^d \times \R^d \to \R$ is twice differentiable in the sense of \cite[Definition 4.35]{steinwart2008support}, i.e., for $i, j \in\{1, \cdots, d\}$, both $ \partial_i \partial_{i+d} k$ and $\partial_i \partial_j \partial_{i+d} \partial_{j+d} k$ exist and are continuous. 
There exist constants $K_{1d}, K_{2d} > 0$ such that $\left\|\nabla_1 k(x,\cdot) \right\|_{\calH^d} := \sum_{i=1}^{d} \left\|\partial_i k(x,\cdot) \right\|_\calH \leq \sqrt{ K_{1 d} }$ and $\left\|\bH_1 k(x, \cdot) \right\|_{\calH^{d \times d}} := \sum_{i,j=1}^{d} \left\|\partial_i \partial_j k(x,\cdot) \right\|_\calH  \leq \sqrt{ K_{2 d} }$ for all $x\in\mathbb{R}^d$.
\end{innercustomassumption}

Many kernels satisfy both {\color{black}\Cref{assumption:universal,assumption:bounded_kernel}}, including the class of bounded, continuous, and translation invariant kernels on $\mathbb{R}^d$ whose Fourier transforms have finite second and fourth moments. This is easy to verify by employing the Fourier transform representation of the RKHS \citep[Theorem 10.12]{wendland2004scattered} and noting that the finiteness of the RKHS norm of $\nabla_1 k(\cdot,x)$ and $\mathbf{H}_1k(\cdot,x)$ for all $x$ corresponds to the existence and finiteness of the second and fourth moments of the Fourier transform of the kernel, respectively. This condition is satisfied by the Gaussian kernel, Mat\'{e}rn kernels of order $\nu$ with $\nu + \frac{d}{2} \geq 2$ and the inverse multiquadratic kernel.


\section[\texorpdfstring{(De)-regularized Maximum Mean Discrepancy ($\dmmd$)}{(De)-regularized Maximum Mean Discrepancy (DrMMD)}]{(De)-regularized Maximum Mean\\Discrepancy ($\dmmd$)}\label{sec:chard}
In this section, we introduce a (de)-regularized version of maximum mean discrepancy, or $\dmmd$ in short. 
%
The $\dmmd$ is rooted in a unified representation of
the MMD and the $\chi^2$-divergence, given in the following proposition, which is proved in \Cref{appsec:mmd_regularized_chi2}.
\begin{prop}[$\mmd$ and $\chi^2$-divergence]\label{prop:mmd_chi2}
Suppose
$\frac{d \mu}{d \pi} -1 \in L^2(\pi)$ for $\mu, \pi \in \calP_2(\R^d)$. Then 
\begin{align*}
    \mmd^2(\mu \| \pi) = \left \|  \mathcal  T_{\pi}^{1 / 2}\left(\frac{d \mu}{d \pi}-1\right) \right \|^2_{L^2(\pi)} & \text{  and  }\,\,
    \chi^2(\mu \| \pi) = \left\| \Id \left(\frac{d \mu}{d \pi}-1\right) \right\|^2_{L^2(\pi)}. 
\end{align*}
\end{prop}
\begin{rem}\label{rem:mmd-chi2}
The $ \chi^{2}$ identity follows from the definition but is provided for comparison purposes. 
Together, these identities express both the MMD and the $\chi^2$-divergence as functionals of the (centered) density ratio $\frac{d \mu}{d \pi} -1$.
While the $\chi^2$-divergence directly
computes the $ L^2(\pi)$ norm of the centred ratio, the $\mmd^2$ first computes the image by the operator $ \mathcal  T_{\pi}^{1 / 2} $ before taking the $ L^2(\pi) $ norm. 
The smoothing effect of the compact operator $ \mathcal  T_{\pi}^{1 / 2}$---note that $\mathcal{T}_\pi$ is compact if $k$ is bounded as assumed in {\color{black}\Cref{assumption:universal}}---has both positive and negative consequences:
$\calF_{\mmd}(\cdot) = \mmd^2(\cdot \| \pi)$ admits finite sample estimators but is not geodesically convex, making the first-order optimization of MMD objective (as done in generative modeling) challenging~\citep{arbel2019maximum}. 
In contrast, $ \calF_{\chi^2}(\cdot) = \chi^2(\cdot \| \pi) $ is geodesically convex for log-concave targets $\pi$~\citep{ohta2011displacement} but is hard to estimate with samples.
\end{rem}

With these facts in mind, we introduce a divergence whose purpose is to combine the beneficial properties of both the $\chi^2$-divergence and $ \operatorname{MMD} $. 
To do so, this divergence computes the $ L^2(\pi) $ norm of the image of $ \frac{ d \mu }{  d \pi } - 1 $ by an alternative operator which interpolates between $ \Id $ and $ \mathcal  T_{\pi}^{1 / 2} $. We set this operator to be $ ((\mathcal  T_{\pi} + \lambda \Id)^{-1} \mathcal  T_{\pi})^{1 / 2} $, where $ \lambda > 0 $ is a regularization parameter. 
The operator $ ((\mathcal  T_{\pi} + \lambda \Id)^{-1} \mathcal  T_{\pi})^{1 / 2} $ can be seen as a (de)-regularization of the operator $ \mathcal  T_{\pi}^{1 / 2} $ used by the MMD---- a similar idea has been used in kernel Fisher discriminant
analysis~\citep{mika99fisher}, goodness-of-fit testing~\citep{balasubramanian2017optimality,hagrass2023spectralgof}, and two-sample testing~\citep{eric2007testing, hagrass2022spectral}. We call the resulting divergence the (De)-regularized Maximum Mean Discrepancy ($\dmmd$).

\begin{defi}[$\dmmd$]\label{def:chard}
Suppose $\frac{d \mu}{d \pi} -1 \in L^2(\pi)$ where $ \mu, \pi \in \calP_2(\R^d) $. 
Then the (de)-regularized maximum mean discrepancy ($ \dmmd $) between $ \mu, \pi \in \calP_2(\R^d) $ is defined as
\begin{align}\label{eq:dmmd_with_inverse_operator}
    \dmmd (\mu || \pi) &= ( 1+\lambda) \left\|\left((\mathcal  T_{\pi} + \lambda \Id)^{-1}\mathcal  T_{\pi} \right )^{1 / 2} \left(\frac{d \mu}{d \pi}-1\right) \right\|_{L^2(\pi)}^2,
\end{align}
where $\lambda>0$.
\end{defi}
While all three operators,
$ \Id $, $ \mathcal  T_{\pi} $ and $ (\mathcal  T_{\pi} + \lambda \Id)^{-1}\mathcal  T_{\pi}$ are diagonalizable in the same eigenbasis of $ L^{2}(\pi)$, the key difference between them lies in the behavior of their eigenvalues. The identity operator  $\Id$ has all eigenvalues $1$, the integral operator $\mathcal{T}_\pi$ has eigenvalues $(\varrho_i)_{i \geq 1}$ which decay to zero as $i\rightarrow\infty$ and the (de)-regularized integral operator $(\mathcal{T}_\pi+\lambda \Id)^{-1}\mathcal{T}_\pi$ has eigenvalues $\left(\varrho_i/(\varrho_i+\lambda)\right)_{i \geq 1}$ which either decay to zero or converge to 1 depending on the choice of $\lambda$ as $i\rightarrow\infty$. 
$ (\mathcal  T_{\pi} + \lambda \Id)^{-1}  \mathcal  T_{\pi} $ is known in the statistical estimation literature as Tikhonov regularization; alternative definitions of $\dmmd$ could be obtained by using other regularizing operators, such as Showalter regularization in \cite{engl1996regularization}. In this paper, we primarily focus on Tikhonov regularization and leave other types of regularization for future work. 


One of the stated purposes of the $\dmmd$ is to retain the computational benefits of the MMD, which are crucial for its use in particle algorithms for generative modeling. 
To this end,
we provide an alternative representation of $\dmmd$ which does not involve the density ratio $ \frac{ d \mu }{ d \pi } $ directly, but only kernel expectations.

\begin{prop}[Density ratio--free and variational formulations]\label{prop:drmmd_representation_no_ratio}
$\dmmd$ can be alternately represented as 
\begin{align} 
\dmmd(\mu || \pi) &= ( 1+\lambda) \left\|\left(\Sigma_\pi + \lambda \Id\right)^{-\frac{1}{2}}\left(m_\mu - m_\pi \right) \right\|_{\mathcal{H}}^2 \label{eq:drmmd_representation_no_ratio} \\
&= (1+\lambda) \sup _{h \in \mathcal{H}}\left\{\int h \mathrm{~d} \mu-\int\left(\frac{h^2}{4}+h\right) \mathrm{d} \pi-\frac{\lambda}{4}\|h\|_{\mathcal{H}}^2\right\}
\label{eq:drmmd_variational}
\end{align}
    with $h_{\mu, \pi}^\ast = 2\left(\Sigma_{\pi} + \lambda \Id\right)^{-1} (m_{\mu} - m_{\pi})$ being the witness function. 
\end{prop}
The proof is in \Cref{appsec:proof_chard_also}.
The density ratio-free representation \eqref{eq:drmmd_representation_no_ratio} contains three expectations under $ \mu $ and $ \pi$: the mean embeddings $ m_{\mu} $ and $ m_{\pi} $, and the covariance operator $\Sigma_{\pi}$. 
Given samples $ \{ x_i \}_{i=1}^{M} \sim \mu $ and $ \{ y_i \}_{i=1}^{N} \sim \pi $, we can construct a plug-in finite sample estimator in \eqref{appeq:dmmd_samples} by replacing $ \mu $ and $ \pi $ with their empirical counterparts: $ \hat{\mu} = \frac{1}{M} \sum_{i=1}^{M} \delta_{x_i} $ and $ \hat{\pi} = \frac{1}{N} \sum_{i=1}^{N} \delta_{y_i} $. 

The density ratio-free representation \eqref{eq:drmmd_representation_no_ratio} frames $\dmmd$ as acting on the difference of $m_\mu $ and $m_\pi$ similarly to MMD. 
In fact, up to a multiplicative factor $(1+\lambda)$, $\dmmd$ is $\mmd$ computed with respect to another kernel $ \tilde{k} $ defined as 
\begin{align}\label{eq:tilde_k}
    \tilde{k}(x, x^\prime) = \left\langle \left(\Sigma_\pi + \lambda \Id\right)^{-\frac{1}{2} } k(x,\cdot),  \left(\Sigma_\pi + \lambda \Id\right)^{-\frac{1}{2} } k(x^\prime, \cdot) \right\rangle_{\calH}.
\end{align}
The derivations are provided in \Cref{appsec:drmmd_is_mmd}. The kernel $\tilde{k}$ is symmetric and positive semi-definite by construction, and its associated reproducing kernel Hilbert space is $\tilde{\calH}$. The density ratio-free representation of $\dmmd$ in \Cref{prop:drmmd_representation_no_ratio} is already known in~\citep{balasubramanian2017optimality, eric2007testing,hagrass2022spectral,hagrass2023spectralgof} in the context of non-parametric hypothesis testing.

\subsection{\texorpdfstring{Properties of $\dmmd$}{Properties of DrMMD}}
In this section, we establish various properties of $\dmmd$. 
As discussed earlier, $\dmmd$ is constructed to interpolate between $\chi^{2}$-divergence and MMD to exploit the advantages associated with each. The following result formalizes the interpolation property.
\begin{prop}[Interpolation property]\label{prop:chard_interpolation}
Let $\mu, \pi \in \mathcal{P}_2(\mathbb{R}^d)$. 
If {\color{black}\Cref{assumption:universal}} holds and $\frac{d \mu}{d \pi} -1 \in L^2(\pi)$, then
\begin{align*}
    \lim _{\lambda \rightarrow 0} \dmmd(\mu \| \pi) = \chi^2(\mu \| \pi), \quad\text{and}\quad \lim _{\lambda \rightarrow \infty} \dmmd(\mu \| \pi) =\MMD^2(\mu \| \pi). 
\end{align*}
\end{prop}
\Cref{prop:chard_interpolation}, whose proof can be found in \Cref{appsec:proof_chard_interpolation}, shows that $\dmmd$ asymptotically becomes a probability divergence in the small and large $ \lambda $ regimes.
We seek to use $\dmmd$ as a minimizing objective in generative modeling algorithms; however, we need to ensure that $\dmmd$ is a probability divergence for \emph{any} fixed value of $\lambda$. This result holds, as shown next.

\begin{prop}[$\dmmd$ is a probability divergence]\label{prop:topology_chard}
Under {\color{black}\Cref{assumption:universal}},
for any $ \lambda \in \left( 0, \infty \right ) $, $\dmmd$ is a probability divergence, i.e., $\dmmd (\mu \| \pi) \geq  0 $, with  equality iff $ \mu = \pi $.
Moreover, $\dmmd$ metrizes the weak topology between probability measures, i.e., $ \dmmd (\mu_n \| \pi) \to 0 $ iff $\mu_n $ converges weakly to $\pi \in \cP_2(\R^d)$  as $n\rightarrow\infty$.
\end{prop} 
As MMD with $c_0$-universal kernels metrizes the weak convergence of distributions~\citep{sriperumbudur2016optimal, simon2023metrizing}, \Cref{prop:topology_chard}, whose proof can be found in \Cref{appsec:proof_of_topology_chard}, shows that for any $ \lambda > 0 $, $\dmmd$ is ``MMD--like'' topologically speaking, and is different from the $\chi^2$-divergence which induces a strong topology~\citep{agrawal2021optimal}. 
\begin{rem}
$\dmmd$ is a specific case of a so-called ``Moreau envelopes of  $f$-divergences in reproducing kernel Hilbert spaces'' introduced in \cite{neumayer2024wasserstein}, when the $f$-divergence is taken as the $\chi^2$-divergence. This connection is uncovered by the variational formulation of $\dmmd$ in \eqref{eq:drmmd_variational}. 
In contrast to general $f$-divergences, the Moreau envelope of the $\chi^2$-divergence enjoys a closed-form expression, as we highlight in this paper with various analytical formulas for $\dmmd$. 
The interpolation property (\Cref{prop:chard_interpolation}) and the metrization of weak convergence (\Cref{prop:topology_chard}) are proved concurrently in Corollaries 12 and 13 of \cite{neumayer2024wasserstein}, relying on formulation via Moreau envelopes.
In our case, we use direct computations thanks to the closed form of $\dmmd$.\footnote{We would like to clarify that \cite{neumayer2024wasserstein} appeared on arxiv when our paper was already under review at ICML 2024.}
\end{rem}

\section{\texorpdfstring{Wasserstein Gradient Flow of $\dmmd$}{Wasserstein Gradient Flow of DrMMD}}\label{sec:chard_flow}
Having introduced the $\dmmd$ in the previous section, we now construct and analyze its \emph{Wasserstein Gradient Flow} (WGF). 
As discussed in \Cref{sec:background}, WGFs define dynamics $ (\mu_t)_{t \geq 0} $ in Wasserstein-2 space that minimize a given functional $ \mathcal  F $ by transporting $ \mu_t $ in the direction of steepest descent, given by the \emph{Fréchet subdifferential} of $ \mathcal  F $ evaluated at $ \mu_t $. 
Given some target distribution $ \pi $ from which we wish to sample, the WGF of $\calF_{\dmmd}(\cdot) = \dmmd(\cdot \| \pi)$, called $\dmmd$ flow, has the potential to form the basis of a generative modeling algorithm, since $ \mu_t $ progressively minimizes its distance (in the DrMMD sense) to the target $ \pi$. 

To fulfill this potential, we formally establish that $ \mu_t $ reaches the global minimizer $ \pi $ and does not get stuck in a spurious local minimum of $ \mathcal F $.
In this section, we show how $\dmmd$ benefits from its interpolation towards $\chi^2$-divergence such that the $\dmmd$ flow achieves near-global convergence for a large class of target distributions.

\subsection{\texorpdfstring{$\dmmd$ flow: Definition, existence, and uniqueness}{DrMMD flow: Definition, existence, and uniqueness}}
To prove that the $\dmmd$ flow is well-defined and admits solutions, the key is to show that the $\calF_{\dmmd}$ admits Fréchet subdifferentials, as formalized in the following proposition.
\begin{prop}[$\dmmd$ gradient flow]\label{prop:chard_gradient_flow}
Let $\lambda>0$, and $\mu_0, \pi \in \mathcal{P}_2\left(\R^d\right)$. Under {\color{black}\Cref{assumption:universal,assumption:bounded_kernel}}, the functional $\calF_{\dmmd}$ admits Fréchet subdifferential of the form $(1 + \lambda) \nabla h_{\mu_t, \pi}^\ast$, where $ h_{\mu_t, \pi}^\ast $ is the witness function defined in \Cref{prop:drmmd_representation_no_ratio}. Consequently, the $\dmmd$ flow is well-defined and is the solution to the following equation
\begin{align}\label{eq:cont_eqn}
   \partial_t \mu_t - \nabla \cdot \left(\mu_t(1+\lambda) \nabla  h_{\mu_t, \pi}^\ast \right)=0.
\end{align}
In addition, the $\dmmd$ flow starting at $\mu_0$ is unique because $\calF_{\dmmd}$ is 
semiconvex, i.e., for any $\phi \in C_c^\infty(\R^d)$ and $\mu \in \calP_2(\R^d)$,
\begin{align}\label{eq:dmmd_semi_convex}
    \left| \left\langle \operatorname{Hess} {\calF_{\dmmd} }_{\mid \mu} \nabla \phi, \nabla \phi \right\rangle_{L^2(\mu)} \right| \leq 2(1+\lambda) \frac{ 2 \sqrt{ K K_{2d}} + K_{1d} }{\lambda} \|\nabla \phi \|_{L^2(\mu)}^2 .
\end{align}
\end{prop}
The proof is in \Cref{appsec:proof_chard_gradient_flow}.
By recalling the discussion of Wasserstein Hessian in \Cref{sec:wass}, \eqref{eq:dmmd_semi_convex} indicates that  $\calF_{\dmmd}$ is both \emph{geodesically smooth} and \emph{geodesically semiconvex}, which is expected because $\dmmd$ is equivalent to $\mmd$ with a regularized kernel $\tilde{k}$ defined in \eqref{eq:tilde_k} and $\calF_{\mmd}$ is both \emph{geodesically smooth} and \emph{geodesically semiconvex}~\citep[Proposition 5]{arbel2019maximum}.
\Cref{prop:chard_gradient_flow} is proved concurrently in Corollaries 14 and 20 of \cite{neumayer2024wasserstein} relying on formulation via Moreau envelopes, while our proof uses the closed-form expression for $\dmmd$.

\subsection{\texorpdfstring{Near-global convergence of $\dmmd$ flow}{Near-global convergence of DrMMD flow}}
Having defined the $\dmmd$ flow, we are now concerned with its convergence to the target $ \pi $. 
Since $\dmmd$ is constructed to interpolate between the $\mmd$ and the $ \chi^2 $-divergence, $\dmmd$ flow is expected to recover the convergence properties of the $\chi^2$ flow.
With this goal in mind, we first study the Wasserstein Hessian of $\dmmd$ and prove that it becomes \emph{asymptotically} positive as $ \lambda \to 0$ for strongly log-concave targets $\pi$. 
Next, to obtain a \emph{non-asymptotic} convergence rate, we take another route and show that the $\dmmd$ flow converges to $ \pi $ exponentially fast in KL divergence up to a barrier term that vanishes in the small $ \lambda $ regime, provided that $ \pi $ satisfies a Poincar\'{e} inequality.

\subsubsection{\texorpdfstring{Near-Geodesic Convexity of $\calF_{\dmmd}$}{Near-Geodesic Convexity of the DrMMD Functional}}\label{sec:near-geodesic-convexity}
One popular approach to proving that the $\dmmd$ flow $(\mu_t)_{t \geq 0}$ converges to the target $\pi$ in terms of $\dmmd$ is to show that $\calF_{\dmmd}$ is \emph{geodesically convex}~\citep[Theorem 4.0.4]{ambrosio2005gradient}, or, equivalently in our definition in \Cref{sec:wass}, its Wasserstein Hessian is positive definite. 
In the next proposition, we show that the Wasserstein Hessian of $\calF_{\dmmd}$ is indeed positive definite for small enough $\lambda$; however, as we will see below (\cref{lambda_difficult}), the form of the result will not allow us to show convergence besides in the limit $\lambda \rightarrow 0$, which leads us to take a different approach in subsequent sections.
\begin{prop}[Near-geodesic convexity of $\calF_{\dmmd}$] \label{prop:geometric_prop}
Let $\mu, \pi \in \calP_2(\R^d)$, $\mu, \pi \ll \calL^d$ and $\phi \in C_c^\infty(\R^d)$. 
Under {\color{black}\Cref{assumption:universal,assumption:bounded_kernel}},
let $\pi$ be $\alpha$ strongly log-concave, i.e., $\pi \propto \exp(-V)$, $\bH V \succeq \alpha \Id$, and assume additionally that $x \mapsto \bH V(x)$ is continuous. Then for all $ \mu $ such that $ x \mapsto \nabla \log \mu(x)$ is continuous and $\frac{d \mu}{d \pi} -1 \in \mathcal  H$, 
\begin{align}\label{eq:hessian_lower_bound}
    \left\langle \operatorname{Hess} {\calF_{\dmmd}}_{\mid \mu} \nabla \phi,
    \nabla \phi \right\rangle_{L^2(\mu)}
    &\geq \alpha (1 + \lambda)
    \int \frac{d\mu}{d \pi}(x) \|\nabla \phi(x) \|^2  d \mu(x) \nonumber\\
    &\quad - R(\lambda, \mu, \nabla \phi), 
\end{align}
where $\lim_{\lambda \to 0} R(\lambda, \mu, \nabla \phi) = 0$. 
\end{prop}
The proof can be found in \Cref{appsec:proof_geometric}. 
To obtain this result, we relate the Wasserstein Hessian of $\calF_{\dmmd}$ with that of $\calF_{\chi^2}$. 
When $ \frac{  d \mu }{ d \pi } - 1 \in \mathcal  H $, they coincide asymptotically as $ \lambda \to 0$, showing that the interpolation properties of  $\dmmd$ to the $\chi^2$-divergence hold at the level of Wasserstein derivatives.
Together with the fact that the Wasserstein Hessian of $\calF_{\chi^2}$ is positive definite for $ \alpha $-strongly log-concave $ \pi $~\citep{ohta2011displacement}, we obtain the lower bound in \eqref{eq:hessian_lower_bound}.
It is noteworthy that although $\dmmd$ can be viewed as squared $\mmd$ with a regularized kernel $\tilde{k}$, the near-geodesic convexity in \Cref{prop:geometric_prop} is not observed for standard $\mmd$ with a fixed kernel $k$ because the latter does not interpolate towards $\chi^2$-divergence.

\begin{rem}[Geodesic convexity/smoothness trade-off in $\calF_{\dmmd}$]\label{rem:convex_smooth_trade_off} 
The geodesic smoothness and near-convexity of the functional $\calF_{\dmmd}$ are characterized by \eqref{eq:dmmd_semi_convex} and \eqref{eq:hessian_lower_bound} respectively via upper and lower bounds on the  Wasserstein Hessian of $\calF_{\dmmd}$. However, \eqref{eq:dmmd_semi_convex} and \eqref{eq:hessian_lower_bound}
impose \emph{contradictory} conditions on the (de)-regularization parameter $\lambda$: \eqref{eq:dmmd_semi_convex} indicates that $\dmmd$ is smoother if $\lambda$ is larger while \eqref{eq:hessian_lower_bound} indicates that $\dmmd$ is more convex if $\lambda$ is small enough.
Consequently, there is a trade-off between the geodesic convexity and smoothness of $\calF_{\dmmd}$.
\end{rem}
\begin{rem}\label{lambda_difficult}
\Cref{prop:geometric_prop} shows that for fixed $\mu$ and $ \nabla \phi $, there exists $\lambda$ small enough yet positive such that 
$\left\langle \operatorname{Hess} {\calF_{\dmmd}}_{\mid \mu} \nabla \phi, \nabla \phi \right\rangle_{L^2(\mu)} > 0$ at $\mu$.  
The remainder term $R(\lambda, \mu, \nabla \phi)$ is only controlled in the limit as $\lambda \rightarrow 0$, however, which complicates the use of \Cref{prop:geometric_prop} to show global convergence of the $\dmmd$ flow. In the next section, we employ a different set of techniques that rely on the Poincar\'{e} condition on $\pi$, a condition which, as we show, will ensure a sufficient dissipation of KL divergence along the $\dmmd$ flow to obtain \emph{non-asymptotic} near-global convergence. 
\end{rem}

\subsubsection{\texorpdfstring{Near-Global Convergence of $\dmmd$ flow via Poincar\'{e} inequality}{Near-Global Convergence of DrMMD flow via Poincare inequality}}
Even when a functional $ \mathcal  F $ is not geodesically convex, convergence guarantees for its Wasserstein gradient flow $(\mu_t)_{t \geq 0}$ can still be obtained if the target $\pi$ satisfies certain functional inequalities. 
Consider the $\chi^2$ flow $(\nu_t)_{t \geq 0}$, for example: if $ \pi $ satisfies the Poincar\'{e} inequality, then $(\nu_t)_{t \geq 0}$ converges exponentially fast to $\pi$ in terms of KL divergence~\citep[Theorem 1]{chewi2020svgd}, i.e., 
\begin{align}\label{eq:chi2_converge}
    \kl(\nu_T \| \pi) \leq \exp \left( -\frac{2T}{C_P}\right) \kl(\nu_0 \| \pi) .
\end{align}
Recall that $ \pi $ satisfies a Poincar\'{e} inequality~\citep[Definition 1]{pillaud2020statistical} if
for all functions $f \colon \R^d \to \R$ such that $f , \nabla f \in L^2(\pi)$, there exists a constant $C_P > 0$ such that
\begin{equation}\label{eqn:poincare}
\int f(x)^2 d \pi(x) - \left( \int f(x) d \pi(x) \right)^2 \leq C_P\|\nabla f\|^2_{L^2(\pi)}.
\end{equation}
The smallest constant $C_P$ for which \eqref{eqn:poincare} holds is called the Poincar\'{e} constant.
The Poincar\'{e} inequality is widely used for studying the convergence of  Langevin diffusions~\citep{chewi2024analysis} and $\chi^2$ flow~\citep{chewi2020svgd,garcia2020bayesian}.
The Poincar\'{e} condition is implied by the strong log-concavity of $\pi$, and is weaker than strong log-concavity because it also allows for nonconvex potentials. 
The set of probability measures satisfying a Poincar\'{e} inequality includes distributions with sub-gaussian tails or with exponential tails. This set is also closed under bounded perturbations and finite mixtures (see \cite{vempala2019rapid} and \cite{chewi2024analysis} for a more detailed discussion).

Given the interpolation property of DrMMD to $ \chi^2 $-divergence, ~\eqref{eq:chi2_converge} suggests investigating the convergence of the $\dmmd$ flow $(\mu_t)_{t \geq 0}$ in KL divergence under a Poincar\'{e} inequality. 
To this end,  we first derive an upper bound on $\kl(\mu_t \| \pi)$ along the $\dmmd$ flow.
\begin{thm}[KL control of the $\dmmd$ flow] \label{thm:continuous_time_convergence}
Suppose $k$ satisfies Assumptions \ref{assumption:universal} and \ref{assumption:bounded_kernel}, and the target $\pi$ and $\dmmd$ gradient flow $\left(\mu_t\right)_{t \geq 0}$ satisfy the following conditions:
\begin{enumerate}[itemsep=5.0pt,topsep=5pt,leftmargin=*]
\item $\pi$ satisfies a Poincar\'{e} inequality with constant $C_P$.
\item $\mu_t, \pi \ll \calL^d$.
\item $\frac{ d\mu_t}{d \pi} - 1 \in \operatorname{Ran} ( \mathcal{T}_\pi^{r} )$ with $r > 0$, i.e., there exists $ q_t \in L^2(\pi) $ such that $\frac{d\mu_t}{d\pi} - 1 = \calT_\pi^{r} q_t$.
\item $\left\| \nabla \left(\log \pi\right)^\top \nabla \left( \frac{d \mu_t}{d\pi} \right) \right\|_{L^2(\pi)} \leq \mathcal{J}_t$ and $\left\| \Delta \left( \frac{d \mu_t}{d\pi} \right) \right\|_{L^2(\pi)} \leq \mathcal{I}_t $.
\item For all $i = 1, \ldots, d$, $\lim\limits_{x\to \infty} \left( h_{\mu_t, \pi}^\ast(x) - 2 \frac{d \mu_t}{ d \pi} (x) \right) \left( \partial_i \frac{d \mu_t}{ d \pi} (x) \right) \pi(x) \to 0$.
\end{enumerate}
Then, for any $T\ge 0$,
\begin{align}\label{eq:KL_continuous}
    \kl(\mu_T \| \pi) & \leq \exp\left( -\frac{ 2 (1+\lambda)}{ C_P} T\right) \kl(\mu_0 \| \pi) \nonumber\\
    &+ 4 (1+\lambda) \lambda^r \int_0^T \exp\left( -\frac{ 2(1+\lambda)}{C_P} (T-t) \right)\left\| q_t \right\|_{L^2(\pi)} (\mathcal{J}_t + \mathcal{I}_t)  dt.
\end{align}
\end{thm}
The proof, which can be found in \Cref{appsec:proof_continuous_time_convergence}, leverages the fact that the $\dmmd$ can approximate not only the $\chi^2$-divergence, but also its \emph{Wasserstein gradient}. The $\dmmd$'s approximation properties can be combined with functional inequalities to obtain an upper-bound for the continuous-time dissipation of KL divergence along the flow, given by:
\begin{align}\label{eq:kl_d_dt_final_main}
    \frac{d}{dt} \kl(\mu_t \| \pi) \leq  - \frac{ 2 (1+\lambda)}{ C_P} \kl(\mu_t \| \pi) + \underbrace{ 4 (1+\lambda)\lambda^{r} \left\| q_t \right\|_{L^2(\pi)} \left( \mathcal{J}_t +\mathcal{I}_t \right)}_{ \text{Approximation error}},
\end{align}
from which \Cref{thm:continuous_time_convergence} follows upon applying the Growall's lemma~\citep{gronwall1919note}.
The first term is strictly negative, while the second term is an approximation error term arising from $\dmmd$ not perfectly matching the $\chi^2$-divergence for $\lambda > 0$.
When $\lambda = 0$, \Cref{thm:continuous_time_convergence} recovers the exponential decay of KL divergence along $\chi^2$ flow in \eqref{eq:chi2_converge}.
\begin{rem}\label{rem:continous_time}
(i) The second condition assumes that $\mu_t$ and $\pi$ have densities.

(ii) The third condition that $\frac{ d\mu_t}{d \pi} - 1 \in \operatorname{Ran} ( \mathcal{T}_\pi^{r} )$ is a regularity condition on the density ratio so that it can be well approximated by the witness function $h_{\mu_t, \pi}^\ast$. 
This assumption is known as the range assumption in the literature of kernel ridge regression~\citep{cucker2007learning, fischer2020sobolev}. 
We posit that this assumption can be relaxed to $\frac{ d\mu_t}{d \pi} - 1 \in L^2(\pi)$ as in \Cref{prop:chard_interpolation} if only asymptotic convergence is needed with no explicit rate.

(iii) The fourth condition is another regularity condition on the density ratio along the flow. This condition is automatically satisfied under a stronger range condition ($r = \frac{1}{2}$) in the third condition, i.e., $\frac{ d\mu_t}{d \pi} - 1 \in \calH$, along with a moment condition on the score function $\nabla  \log \pi$. 
To see this, notice that we can further write (derivations are provided in \Cref{appsec:proof_continuous_time_convergence})
\begin{align}\label{eq:fourth_condition}
\begin{aligned}
    \left\| \nabla \left( \log \pi \right)^\top \nabla \left( \frac{d \mu_t}{d \pi} \right) \right\|_{L^2(\pi)} &\leq \sqrt{K_{1d}} \| q_t \|_{L^2(\pi)} \left\| \nabla  \log \pi \right\|_{L^2(\pi)} \\
     \left\| \Delta \left(\frac{d \mu_t}{d \pi} \right)  \right\|_{L^2(\pi)} & \leq \sqrt{K_{2d} } \| q_t \|_{L^2(\pi)} .
\end{aligned}
\end{align}
For Gaussian kernels and Gaussian targets $\pi$, explicit upper bounds on $\mathcal{I}_t$ and $\mathcal{J}_t$ can be derived under a Gaussian approximation of the DrMMD flow, following \citet{lambert2022variational,liu2024towards}.

(iv) The fifth condition is a boundary condition that allows integration by parts equality in the proof. We highlight that many works (e.g., Theorem 1 of \cite{he2022regularized}, Theorem 2 of \cite{nitanda2022convex}, Lemma 6 of \cite{vempala2019rapid}) on Wasserstein gradient flow apply integration by parts without explicitly stating this condition. 
\end{rem}
Additionally, if $\| q_t \|_{L^2(\pi)} \leq Q$,\,$\mathcal{J}_t \leq \mathcal{J},\, \mathcal{I}_t \leq \mathcal{I}$  for all $0 \leq t \leq T$, then \Cref{thm:continuous_time_convergence} will ensure KL convergence of the $\dmmd$ flow up to a controllable barrier term, e.g. \emph{near global convergence}.
\begin{cor}[Near global convergence of the $\dmmd$ flow]\label{cor:continuous_time_convergence}
In addition to the assumptions of \Cref{thm:continuous_time_convergence}, if $\left\| q_t \right\|_{L^2(\pi)} \leq Q$,  $\mathcal{J}_t \leq \mathcal{J}, \mathcal{I}_t \leq \mathcal{I}$ for all $0 \leq t \leq T$, where $Q, \mathcal{J},$ and $\mathcal{I}$ are universal constants independent of $\lambda$, then for any $T\ge 0$,
\begin{align*}
    \kl(\mu_T \| \pi) \leq \exp\left( -\frac{ 2 (1+\lambda)}{C_P} T\right) \kl(\mu_0 \| \pi) + 2 \lambda^{r}  C_P Q \left( \mathcal{J} + \mathcal{I} \right) .
\end{align*}
\end{cor}
The proof of \Cref{cor:continuous_time_convergence} follows directly from upper-bounding the second term of \eqref{eq:KL_continuous} with universal constants $\mathcal{Q}$, $\mathcal{J}$, and $\mathcal{I}$, and using the closed-form expression of the resulting integral. 
\Cref{cor:continuous_time_convergence} provides a condition under which the $\dmmd$ flow will exhibit an exponential rate of convergence (linear convergence) in terms of KL divergence up to an extra approximation error term which vanishes as $\lambda \to 0$. 
If $\left\| q_t \right\|_{L^2(\pi)} \leq Q, \mathcal{J}_t \leq \mathcal{J}, \mathcal{I}_t \leq \mathcal{I}$ for all $0 \leq t \leq T$ as shown in \Cref{cor:continuous_time_convergence}, the approximation error is of explicit order $\calO(\lambda^r)$. Therefore, in the continuous time regime, to have a smaller approximation error, it is beneficial to use a small (de)-regularization parameter $\lambda$, so that  $\dmmd$ flow operates closer to the regime of $\chi^2$ flow.
However, as we will see in the next section, when it comes to time-discretized $\dmmd$ flow, i.e., $\dmmd$ gradient descent, there is a trade-off between the approximation error and the time discretization error such that the selection of $\lambda$ would require more careful analysis to strike a good balance.
Finally, a smaller Poincar\'{e} constant $C_P$ results in both a faster rate of convergence and a smaller barrier.

Previously, \cite{arbel2019maximum} established (sublinear) global convergence of the $\mmd$ flow in terms of $\mmd$ distance by assuming that a Lojasiewicz inequality (or a variant of it if additionally performing noise injection, see \citealp[Proposition 8]{arbel2018gradient}) holds along the flow. 
Our result thus complements that of \cite{{arbel2019maximum}} by showing that MMD-type functionals can achieve near-global convergence for targets satisfying a Poincar\'{e} inequality regardless of whether such inequalities hold, by studying their behavior in the $f$-divergence interpolation regime.


\begin{rem}\label{rem:one}
    Since $\dmmd$ is asymmetric in its arguments, the reader may wonder why we focus on the gradient flow of $\dmmd(\cdot||\pi)$ instead of $\dmmd(\pi||\cdot)$. 
    From the convergence standpoint, \Cref{thm:continuous_time_convergence} shows that $\dmmd(\cdot||\pi)$ converges globally with an exponential rate up to a small barrier when $\pi$ satisfies Poincar\'{e} inequality along with an extra regularity condition on the density ratio. 
    This favorable convergence property is no longer true for $\dmmd(\pi||\cdot)$. 
    Practically speaking, the Wasserstein gradient of $\dmmd$ requires inverting a kernel integral operator.
    It is more efficient to do this just once with a kernel integral operator with respect to $\pi$ as in \eqref{eq:dmmd_with_inverse_operator}, rather than with respect to $\mu_t$, which would happen at each time (as done by \citealt{he2022regularized}). 
\end{rem}

Computing the $\dmmd$ flow exactly is intractable since the dynamics are in continuous time, and in practice, we often only have samples from $\pi$.

\section{\texorpdfstring{Time-discretized $\dmmd$ flow}{Time-discretized DrMMD flow}} \label{sec:chard_particle_flow}
In this section, we first construct and analyze the forward Euler scheme of \eqref{eq:cont_eqn}, a simple time-discretization of
the $\dmmd$ flow which we call $\dmmd$ Descent.  Compared to the $\dmmd$ flow, the convergence of  $\dmmd$ descent is affected by an additional smoothness-related time-discretization error that blows up as $\lambda$ approaches 0, thus preventing near-global convergence. 
To address this issue, we propose in \Cref{subsec:time_dis_adaptive} an alternative discrete-time scheme that adapts the value of the regularization coefficient $\lambda$  across the descent iterates, which we call \emph{Adaptive $\dmmd$ Descent}. We show that this scheme converges in the KL divergence up to a barrier term that vanishes as the discretization step size goes to zero.

\subsection{\texorpdfstring{$\dmmd$ Descent}{DrMMD Descent}}\label{subsec:time_dis}
The forward Euler discretization of the $\dmmd$ flow (or $\dmmd$ \emph{Descent} in short) with step size $\gamma >0$ consists of sequence of probabilities $(\mu_n)_{n \in \N}$ defined by the recursion
\begin{align}\label{eq:time_discretized_flow}
\mu_{n+1} = \left(\Id - \gamma(1+\lambda) \nabla h_{\mu_n,\pi}^\ast \right)_{\#} \mu_n , \quad \mu_{n=0} = \mu_0,
\end{align}
where $h_{\mu_n,\pi}^\ast = 2\left(\Sigma_\pi + \lambda \Id\right)^{-1} \left(m_{\mu_n} - m_{\pi} \right)$ is the $\dmmd$ witness function.
This scheme was previously considered in \citet{arbel2019maximum}; in particular, Proposition 4 of \citet{arbel2019maximum} shows that the discrete-time $\mmd$ dissipation rate along the $\mmd$ Descent iterates follows the (continous time) rate of the $\mmd$ flow up to an error term proportional to the step size $\gamma$ and the smoothness parameters of the problem\footnote{In the $\mmd$ flow case, the main smoothness parameters is the Lipschitz constant of the kernel.}. 

Next, we turn to study the convergence of $\dmmd$ descent. One way is to treat $\dmmd$ as $\mmd$ with a regularized kernel $\tilde{k}$ and follow Proposition 4 of \cite{arbel2019maximum}, however this does not quantitatively take into account the role of (de)-regularization parameter $\lambda$ that balances the trade-off between geodesic convexity and smoothness of $\calF_{\dmmd}$.
Instead, we adopt the same strategy of \Cref{sec:chard_flow} that exploits the interpolation property of $\dmmd$ towards $\chi^2$-divergence: in the following proposition, we study the dissipation of KL divergence along the $\dmmd$ Descent when the target $\pi$ satisfies a Poincar\'{e} inequality, in which the role of (de)-regularization parameter $\lambda$ is highlighted in the approximation and discretization errors.

\begin{prop}[Descent lemma in KL]\label{prop:descent_lemma_KL}
Suppose $k$ satisfies {\color{black}\Cref{assumption:universal,assumption:bounded_kernel}}, and suppose the target $\pi$ and $\dmmd$ gradient descent iterates $\left(\mu_n \right)_{n \in \N}$ satisfy the following:
\begin{enumerate}[itemsep=5.0pt,topsep=5pt,leftmargin=*]
\item $\pi$ satisfies a Poincar\'{e} inequality with constant $C_P$  and its potential is $\beta$-smooth, i.e.,  $\pi \propto \exp(-V)$ with $\bH V \preceq \beta \Id$.
\item $\mu_n, \pi \ll \calL^d$.
\item $\frac{ d\mu_n}{d \pi} - 1 \in \operatorname{Ran} ( \mathcal{T}_\pi^{r} )$ with $r > 0$, i.e., there exists $ q_n \in L^2(\pi) $ such that $\frac{d\mu_n }{ d\pi} - 1 = \calT_\pi^{r} q_n$.
\item $\|q_n\|_{L^2(\pi)} \leq Q$, $\left\| \nabla V^\top \nabla \left( \frac{d \mu_n}{d\pi} \right) \right\|_{L^2(\pi)} \leq \mathcal{J}$, $\left\| \Delta \left( \frac{d \mu_n}{d\pi} \right) \right\|_{L^2(\pi)} \leq \mathcal{I}$ for all $n = 1, \cdots, n_{max}$. 
\item For all $i = 1, \ldots, d$, $\lim\limits_{x\to \infty} \left( h_{\mu_n, \pi}^\ast(x) - 2 \frac{d \mu_n}{ d \pi} (x) \right) \left( \partial_i \frac{d \mu_n}{ d \pi} (x) \right) \pi(x) \to 0$.
\item There exists a constant $1 < \zeta < 2$ such that for all $n = 1, \ldots, n_{\max}$, the step size $\gamma$ satisfies 
\begin{align}\label{appeq:gamma_condition_1}
\gamma  \leq \frac{\zeta-1}{2\zeta (1+\lambda) \sqrt{\chi^2\left(\mu_n \| \pi\right) \frac{K_{2 d}}{\lambda } } } .
\end{align}
\end{enumerate}
Then for all $ 0 \leq n \leq n_{\max}$ and $0 < \lambda \leq 1$,
\begin{align}\label{eq:KL_descent_1}
\begin{aligned}
\kl(\mu_{n+1} \| \pi) - \kl(\mu_{n} \| \pi) &\leq -\frac{2}{C_P} \chi^2(\mu_n \| \pi) \gamma \\
&+  \underbrace{ 4 \gamma \lambda^{r} Q \left( \mathcal{J} +\mathcal{I} \right)}_{\text{Approximation error}} \\
&+ \underbrace{ 8 \gamma^2 (\beta + \zeta^2) \chi^2(\mu_n \| \pi) \frac{K_{1d}+K_{2d} }{\lambda}}_{\text{Discretization error}}.
\end{aligned}
\end{align}
\end{prop}
The proof is provided in \Cref{appsec:proof_kl_descent_lemma}. 
Conditions 1-5 of \Cref{prop:descent_lemma_KL} are similar to  conditions 1-5 of \Cref{thm:continuous_time_convergence}, which assumes a functional inequality on the target $\pi$ and  regularity on the density ratio $\frac{d \mu_n}{d \pi} - 1$. 
In a similar spirit to \Cref{thm:continuous_time_convergence}, the fourth regularity condition is automatically satisfied under a stronger range assumption on $ \frac{ d\mu_n}{d \pi} - 1 $.
For the sake of brevity, we directly assume uniform upper bounds in the fourth condition rather than writing it out in a separate corollary like \Cref{cor:continuous_time_convergence}.
Compared to the continuous time regime, two extra conditions are necessary. The first one is a smoothness condition on the potential, $\bH V \preceq \beta \Id$, which is commonly used in the convergence analysis of discrete-time Langevin-based samplers 
~\citep{dalalyan2017further,dalalyan2019user,durmus2019analysis, dalalyan2022bounding,vempala2019rapid}.
It can be relaxed to $\nabla V$ being Hölder-continuous with exponent $s \in [0, 1]$~\citep{chatterji2020langevin}. 
The second one, \eqref{appeq:gamma_condition_1}, is an upper bound on the step size $\gamma$, aligning with the principle that step size should be small enough for the discrete-time scheme to inherit the properties of its continuous analog. This condition will be more thoroughly discussed in \Cref{rem:step_size} when all the conditions on $\gamma$ in \Cref{thm:discrete_time_KL} and \Cref{thm:discrete_time_KL_2} are presented.
%
\begin{rem}[Approximation-discretization trade-off of $\dmmd$ Descent]
If we compare the discrete-time KL dissipation of \eqref{eq:KL_descent_1} with its continuous-time counterpart in \eqref{eq:kl_d_dt_final_main}, we see that the first two terms on the RHS of \eqref{eq:KL_descent_1} admit continuous-time analogs present in \eqref{eq:kl_d_dt_final_main}. The discrete-time KL dissipation contains an additional (positive) term representing the \emph{time discretization error}:  unlike the approximation error term that vanishes as $\lambda $ approaches 0, this term actually diverges as $\lambda $ approaches 0. Therefore, replicating the arguments of the continuous-time result of \Cref{thm:continuous_time_convergence} in the discrete-time regime would yield a barrier that does not vanish as $\lambda \to 0$, hinting at a trade-off between approximation and discretization similar to that of \Cref{rem:convex_smooth_trade_off}. In the next section, we propose a refined adaptive discrete-time descent scheme that addresses the convergence issues.
\end{rem}

\subsection{\texorpdfstring{Adaptive $\dmmd$ Descent}{Adaptive DrMMD Descent}}\label{subsec:time_dis_adaptive}
The $\dmmd$ flow and descent dynamics are defined for a value of $\lambda$ that remains fixed throughout time.
The KL dissipation provided in \Cref{prop:descent_lemma_KL} is
a function of $\lambda$, however; thus, to obtain a sequence of measures with better convergence guarantees than the $\dmmd$ descent, we now construct and analyze a sequence of iterates obtained by selecting, at each iteration, the value of $\lambda$ minimizing the sum of the approximation error and time-discretization error presented in \eqref{eq:KL_descent_1}. This sequence, which we term \emph{Adaptive $\dmmd$ descent}, is given by
\begin{align}\label{eq:adaptive_time_discretized_flow}
\mu_{n+1}
&= \left(\Id - \gamma(1+\lambda) \nabla h_{\mu_n,\pi}^\ast \right)_{\#} \mu_n, \nonumber\\
\lambda_n
&= \left(
2 \gamma \chi^2(\mu_n \| \pi)
\frac{( \beta + \zeta^2 ) (K_{1d} + K_{2d})}
{ Q (\mathcal{J} + \mathcal{I}) }
\right)^{\frac{1}{r + 1}}.
\end{align}
The particular choice of $\lambda_n$ above minimizes the sum of the approximation error and time-discretization error presented in \eqref{eq:KL_descent_1}. 
This optimal choice indicates that $\lambda_n$ should shrink towards $0$ as $\chi^2(\mu_n\|\pi)$ decreases along $\dmmd$ gradient descent: at the early stages of the scheme, it is desirable to have a larger $\lambda_n$, corresponding to a smoother objective functional and enabling larger step sizes; then as $\dmmd$ gradient descent iterates $\mu_n$ get closer to $\pi$, a smaller $\lambda_n$ enables the scheme to operate closer to the $\chi^2$ flow regime, which metrizes a stronger topology and can better witness the difference between $\mu_n$ and $\pi$. 
Additionally, as the potential $V$ becomes less smooth, i.e., $\beta$ gets larger, then $\lambda_n$ should also increase to account for the loss of smoothness from $V$. 
Unlike the $\dmmd$ descent in \Cref{subsec:time_dis}, since the Wasserstein gradient updating $\mu_n$ comes from a different $\dmmd$ at each iteration, the adaptive scheme of \eqref{eq:adaptive_time_discretized_flow} constitutes a significant departure from related works in Wasserstein gradient descent \citep{glaser2021kale, arbel2019maximum, korba2021kernel, hertrich2023generative, hertrich2023wasserstein, chewi2020svgd}. 

\begin{rem}[Adaptive kernel]
Recent applications of MMD-based generative modeling algorithms with adaptive kernels (in particular, time-dependent kernel hyperparameters) demonstrate improved empirical performance over fixed kernels in both Wasserstein gradient flow on $\mmd$~\citep{galashov2024deep} and generative adversarial networks with an $\mmd$ critic~\citep{li2017mmd,arbel2018gradient}: the latter can be related to gradient flow on the critic where $\mu_n$ is restricted to the output of a generator network \citep[see e.g.][]{franceschi2023unifying}. As the $\dmmd$ is an $\mmd$ with a regularized kernel $\tilde{k}$ that depends on $\lambda_n$, the Adaptive $\dmmd$ Descent thus falls into the former category.
\citet[Proposition 3.1]{galashov2024deep} demonstrates faster convergence for $\mmd$ gradient flow with an adaptive kernel, for the parametric setting of Gaussian distributions $\pi$ and $\mu_t$.
Our analysis is the first to prove theoretically that adaptive kernels can result in improved convergence for more general nonparametric settings. We believe that the theoretical analysis of adaptivity by varying other hyperparameters (such as the kernel bandwidth for RBF kernels) remains an interesting avenue for future work.
\end{rem}


By leveraging the quasi-descent lemma in KL divergence in \Cref{prop:descent_lemma_KL}, we are able to establish the following theorem, which provides a near-global convergence result of the Adaptive $\dmmd$ gradient descent iterates in KL divergence. 

\begin{thm}[Near-global convergence of adaptive $\dmmd$ gradient descent]\label{thm:discrete_time_KL}
Suppose $k$ satisfies {\color{black}\Cref{assumption:universal,assumption:bounded_kernel}} and $K \leq 1$, and suppose the target $\pi$ and adaptive $\dmmd$ gradient descent iterates $\left(\mu_n \right)_{n\in \N}$ satisfy the following conditions:
\begin{enumerate}[itemsep=3.0pt,topsep=5pt,leftmargin=*]
\item $\pi$ satisfies a Poincar\'{e} inequality with constant $C_P$  and its potential is $\beta$-smooth, i.e.  $\pi \propto \exp(-V)$ with $\bH V \preceq \beta \Id$.
\item $\mu_n, \pi \ll \calL^d$.
\item $\frac{ d\mu_n}{d \pi} - 1 \in \operatorname{Ran} ( \mathcal{T}_\pi^{r} )$ with $r > 0$, i.e., there exists $ q_n \in L^2(\pi) $ such that $\frac{d\mu_n }{ d\pi} - 1 = \calT_\pi^{r} q_n$.
\item $\|q_n\|_{L^2(\pi)} \leq Q$, $\left\| \nabla V^\top \nabla \left( \frac{d \mu_n}{d\pi} \right) \right\|_{L^2(\pi)} \leq \mathcal{J}$, $\left\| \Delta \left( \frac{d \mu_n}{d\pi} \right) \right\|_{L^2(\pi)} \leq \mathcal{I}$ for all $n = 1, \cdots, n_{max}$. 
\item For all $i = 1, \ldots, d$, $\lim\limits_{x\to \infty} \left( h_{\mu_n, \pi}^\ast(x) - 2 \frac{d \mu_n}{ d \pi} (x) \right) \left( \partial_i \frac{d \mu_n}{ d \pi} (x) \right) \pi(x) \to 0$.
\item There exists a constant $1 < \zeta < 2$ such that the step size $\gamma$ satisfies 
\hspace{-100pt}
\begin{align}\label{eq:gamma_condition}
    \gamma
    &\leq
    \frac{1}{8}
    \left( \frac{\zeta - 1}{\zeta} \right)^{\frac{2r+ 2}{2r+1} }
    \left( \frac{1}{Q} \right)
    \left( \frac{1}{\mathcal{J} + \mathcal{I}} \right)^{\frac{1}{2r + 1}}
    \left( \frac{1}{K_{2d}} \frac{1}{ \beta + \zeta^2} \right)^{\frac{r}{2r + 1}}
    \nonumber\\
    &\quad \wedge \frac{1}{4} \frac{\zeta - 1}{\zeta} \frac{1}{Q^2 K_{2d}}
    \wedge \frac{C_P}{2} \wedge 1 .
\end{align}
\end{enumerate}
Then by taking parameter $\lambda_n = \left( 2\gamma \chi^2(\mu_n \| \pi) \frac{( \beta + \zeta^2 ) (K_{1d} + K_{2d})}{ Q (\mathcal{J} + \mathcal{I}) } \right)^{\frac{1}{r + 1}} \wedge 1$, we have
\begin{align}\label{eq:discrete_time_rate_1}
    \kl(\mu_{n_{max} } \| \pi) &\leq \exp\left(- \frac{ 2 n_{max} \gamma}{C_P} \right) \kl(\mu_0 \| \pi) \nonumber \\
    &+ 4 \gamma^{ \frac{ r }{ r + 1 } } C_P Q^{\frac{2r+1}{r+1}} \Big( (K_{1d} + K_{2d})(\beta + \zeta^2) \Big)^{\frac{ r }{ r + 1 }} ( \mathcal{J} +\mathcal{I})^{\frac{ 1 }{ r + 1 }}  .
\end{align}
\end{thm}

The proof can be found in \Cref{appsec:proof_discrete_time_convergence}. 
The conditions 1-5 of \Cref{thm:discrete_time_KL} are the same as conditions 1-5 of \Cref{prop:descent_lemma_KL}. Since the RHS of \eqref{eq:gamma_condition} are all constants, condition 6 is satisfied when the step size $\gamma$ is small enough.

The implication of \Cref{thm:discrete_time_KL} is that $\dmmd$ gradient descent exhibits an exponential rate of convergence (linear convergence) in terms of KL divergence up to an extra barrier of order $\calO (\gamma^{\frac{r}{r+1}})$.
The barrier term shows up as the result of picking the optimal regularization parameter $\lambda_n$ that best trades off the approximation error and discretization error in \Cref{prop:descent_lemma_KL}.
\Cref{thm:discrete_time_KL} is reminiscent of the convergence result of the Langevin Monte Carlo sampling algorithm, whose KL divergence also decreases exponentially up to an extra barrier, but of order $\calO(\gamma)$ ~\citep[Theorem 2]{vempala2019rapid}.
Unlike the continuous-time result of \Cref{thm:continuous_time_convergence}, in which the barrier can be made arbitrarily small by taking small enough regularization, taking the step size $\gamma$ in \Cref{thm:discrete_time_KL} to be arbitrarily small will significantly impact the rate of convergence, even though it is exponential in terms of $n_{\max}$. 
By making the step sizes adaptive with the number of iterations and imposing an extra condition, the barrier term actually vanishes, as demonstrated in the following theorem. 

\begin{thm}[Global convergence of $\dmmd$ gradient descent]\label{thm:discrete_time_KL_2}
Suppose that $k$ satisfies Assumptions \ref{assumption:universal} and \ref{assumption:bounded_kernel}, and that the conditions in \Cref{thm:discrete_time_KL} on $\dmmd$ gradient descent iterates $(\mu_n)_{n\in \N}$, target distribution $\pi$, regularization coefficient $\lambda_n$ and step size $\gamma_n$ are satisfied.
If additionally, the step size $\gamma_n$ satisfies
\begin{align}\label{eq:gamma_condition_2}
    &\gamma_n \leq  \frac{1}{(K_{1d} + K_{2d})(\beta + \zeta^2)} \left( \frac{1}{Q( \mathcal{J} +\mathcal{I}) } \right)^{\frac{1}{r}} \left( \frac{1}{8 C_P} \right)^{\frac{r+1}{r}} \chi^2(\mu_n \| \pi)^{\frac{1}{r}},
\end{align}
for all $n = 1, \cdots, n_{max}$,
then
\begin{align}\label{eq:discrete_time_rate_2}
\kl(\mu_{n_{max}} \| \pi) \leq \prod_{n=1}^{n_{max}} \left(1 -\frac{1}{C_P} \gamma_n \right) \kl(\mu_{0} \| \pi) . 
\end{align}
\end{thm}
The proof can be found in \Cref{appsec:proof_discrete_time_convergence_2}. 
Compared with \eqref{eq:discrete_time_rate_1}, \eqref{eq:discrete_time_rate_2} provides a cleaner upper bound without the barrier term and leads to global convergence. 

\begin{rem}[Iteration complexity]
We now turn to analyze the iteration complexity of $\dmmd$ gradient descent from Theorems \ref{thm:discrete_time_KL} and \ref{thm:discrete_time_KL_2}.

(i) From \Cref{thm:discrete_time_KL}, given an error threshold $\delta > 0$, $\dmmd$ descent would reach $\mathrm{KL}(\mu_{n_{\max}} \| \pi) \leq \delta$ after $n_{max} \geq \frac{C_P}{2 \gamma} \log \frac{\mathrm{KL}(\mu_0 \| \pi)}{\delta} = \calO ( (\frac{1}{\delta})^{\frac{r+1}{r} } \log \frac{1}{\delta} )$ iterations. By comparison, when $\pi$ satisfies a Poincar\'{e} inequality, Langevin Monte Carlo (LMC) has an iteration complexity of $\mathcal{O}(\frac{1}{\delta})$ up to logarithmic terms~\citep[Theorem 7]{chewi2024analysis}. 
    As an approximation to the $\chi^2$ flow, the iteration complexity of DrMMD flow is worse than LMC, because at each iterate of DrMMD flow, there is an extra approximation error in addition to time-discretization error (see our Proposition 5.1). With the optimal choice of regularization $\lambda$ that balances these two errors, the overall per-step error is of order $\calO(\gamma^{1+ \frac{r}{r+1}})$. 
    In contrast, at each iteration, LMC only incurs the time-discretization error, which is of order $\calO(\gamma^2)$, smaller than that of DrMMD flow. 
    As a result, LMC exhibits better iteration complexity than our DrMMD flow; however, LMC requires knowledge of the score of $\pi$, while DrMMD flow only requires samples from $\pi$.

(ii) From \Cref{thm:discrete_time_KL_2}, we consider two cases. On the one hand, if there exists a threshold $N_0$ such that $\chi^2(\mu_n \| \pi) \geq n^{-r}$ holds for all $n \geq N_0$, then we select step size $\gamma_n \asymp C_P n^{-1}$ for all $n \geq N_0$ such that both \eqref{eq:gamma_condition} and \eqref{eq:gamma_condition_2} are satisfied, and consequently $\prod_{n=N_0}^{n_{max}} \left( 1 -\frac{1}{C_P} \gamma_n \right) = \calO \left( \frac{1}{n_{\max}} \right) \to 0$ so the iteration complexity of $\dmmd$ gradient descent is $\calO\left( \frac{1}{\delta} \right)$. On the other hand, if such a threshold $N_0$ does not exist, then there exists a subsequence $n_1, n_2, \ldots, n_S, \ldots$ such that $ \chi^2(\mu_{n_s} \| \pi) \leq n_s^{-r}$ for all $s \geq 1$. Since KL divergence is smaller than $\chi^2$-divergence~\citep{van2014renyi}, we have $ \kl(\mu_{n_s} \| \pi) \leq n_s^{-r}$ for all $s \geq 1$. Notice that KL divergence is monotonically decreasing based on \eqref{eq:kl_monotone}, we have $ \kl(\mu_n \| \pi) \leq n_s^{-r}$ for all $n_s \leq n \leq n_{s+1}$ so that $\lim_{n \to \infty} \kl(\mu_n \| \pi) = 0$. 
Unfortunately, we are not able to derive iteration complexity in this case because the growth rate of $\{n_s\}_{s \geq 1}$ is unknown. 
\end{rem}

\begin{rem}[Step size $\gamma$]\label{rem:step_size}
\Cref{thm:discrete_time_KL} imposes a condition on the step size $\gamma$ in \eqref{eq:gamma_condition} and \Cref{thm:discrete_time_KL_2} imposes an additional condition in \eqref{eq:gamma_condition_2}.
These conditions subsume the condition \eqref{appeq:gamma_condition_1} on step size in \Cref{prop:descent_lemma_KL}. (See derivations in \Cref{appsec:proof_discrete_time_convergence}.) The conditions \eqref{eq:gamma_condition} and \eqref{eq:gamma_condition_2} become more stringent as the potential $V$ becomes less smooth, i.e., when $\beta$ gets larger, similar to the analysis in Langevin Monte Carlo~\citep{balasubramanian2022towards, vempala2019rapid} and Stein Variational Gradient Descent~\citep{korba2020non}.
The condition also becomes more stringent as the density ratio becomes less regular, i.e., when $r$ gets closer to $0$ and $Q, \mathcal{J}, \mathcal{I}$ get larger, similar to \cite{he2022regularized}. 
\end{rem}

{\color{black}
\subsection{\texorpdfstring{Practical adaptive $\dmmd$ particle descent}{Practical adaptive DrMMD particle descent}}\label{subsec:drmmd_practical}
In practice, we are given fixed target samples $\{x^{(i)}\}_{i=1}^M$ and initial source samples $\{y_0^{(i)}\}_{i=1}^N$, and use the empirical measures $\hat{\pi}=M^{-1}\sum_{i=1}^M\delta_{x^{(i)}}$ and $\hat{\mu}_n=N^{-1}\sum_{i=1}^N\delta_{y_n^{(i)}}$. The witness $h_{\hat{\mu}_n,\hat{\pi}}^\ast$ is computed by the Gram-matrix formula in \eqref{appeq:empirical_h}, and its spatial gradient gives the particle update in \Cref{alg:drmmd_particle_descent}.

As indicated in \eqref{eq:adaptive_time_discretized_flow}, the regularization parameter $\lambda_n$ should be chosen to be proportional to $\chi^2(\mu_n \| \pi)^{\frac{1}{r+1}}$. In practice, however, both $\chi^2(\mu_n \| \pi)$ and $r$ are not accessible and $\chi^2(\mu_n \| \pi)$ is not even well-defined for the particle descent algorithm. To address this, we use $\mathrm{DrMMD}(\hat{\mu}_n \| \hat{\pi})$ as a proxy for $\chi^2(\mu_n \| \pi)$  (see \Cref{alg:drmmd_particle_descent}), which admits a closed-form expression with particles. The parameter $r$ is picked via a search over a pre-defined set $\{0.1, 0.5, 1.0\}$. The step size $\gamma$ should be chosen to satisfy the upper bound \eqref{eq:gamma_condition}, which contains several constants that cannot generally be computed. In practice, our approach has been to select $\gamma$ to be sufficiently small for the flow to converge empirically.
The final algorithm is summarized in \Cref{alg:drmmd_particle_descent}.

This proxy avoids density estimation and permits adaptive particle updates even when the empirical measures have different supports. Regularization also stabilizes the kernel linear system; in the experiments, we enforce the positive lower bound $\tilde{\lambda}=10^{-3}$ to prevent excessively small regularization.

The substitution is nevertheless a practical approximation to the theoretical schedule. At fixed positive $\lambda$, $\dmmd$ measures a weaker notion of discrepancy than $\chi^2$, so a small empirical $\dmmd$ need not imply a small population $\chi^2$. Regularization bias and sampling error can therefore distort the choice of $\lambda_n$: reducing it too quickly can amplify sampling noise and discretization error, whereas keeping it too large retains smoothing bias and may slow convergence. A positive lower bound improves numerical stability but prevents the limit $\lambda_n\to0$ used to remove the approximation barrier. Consequently, the KL guarantees in \Cref{thm:discrete_time_KL,thm:discrete_time_KL_2}, which assume the population schedule and additional regularity, do not automatically apply to this empirical adaptive rule. Its effectiveness in the experiments is empirical evidence rather than a guarantee for arbitrary singular targets.
\par}

\begin{algorithm}[tbp]
\color{black}
\captionsetup{labelfont={bf,color=black},textfont={color=black}}
\caption[DrMMD particle descent]{\texorpdfstring{$\dmmd$}{DrMMD} particle descent (Algorithm~1 of \citet{chen2025regularized})}
\label{alg:drmmd_particle_descent}
\renewcommand{\algorithmicrequire}{\textbf{Input:}}
\renewcommand{\algorithmicensure}{\textbf{Output:}}
\begin{algorithmic}[1]
    \REQUIRE Target samples $\{x^{(i)}\}_{i=1}^M\sim\pi$ and initial source samples $\{y_0^{(i)}\}_{i=1}^N\sim\mu_0$. Hyperparameters: step size $\gamma$, initial (de)-regularization coefficient $\lambda_0$, maximum number of iterations $n_{\max}$ and regularity $r$.
    \FOR{$n=0$ to $n_{\max}-1$}
        \STATE Compute witness function $h_{\hat{\mu}_n,\hat{\pi}}^\ast$ from \eqref{appeq:empirical_h}.
        \STATE Compute $\dmmd(\hat{\mu}_n,\hat{\pi})$ with $h_{\hat{\mu}_n,\hat{\pi}}^\ast$ from \eqref{appeq:dmmd_samples}.
        \STATE Rescale regularization coefficient $\lambda_n\propto\dmmd(\hat{\mu}_n,\hat{\pi})^{1/(r+1)}$.
        \STATE Update particles, for $i=1,\ldots,N$:
        \begin{align*}
            y_{n+1}^{(i)}=y_n^{(i)}-\gamma(1+\lambda_n)
            \nabla h_{\hat{\mu}_n,\hat{\pi}}^\ast(y_n^{(i)}).
        \end{align*}
    \ENDFOR
    \ENSURE $\{y_{n_{\max}}^{(i)}\}_{i=1}^N$.
\end{algorithmic}
\end{algorithm}

\section{Experiments}\label{sec:drmmd_experiments}
In this section, we demonstrate the superior empirical performance of the proposed $\dmmd$ descent in various experimental settings.

\tikzset{example/.style={draw, circle, fill=orange, inner sep=1pt}}

\subsection{Three ring experiment}
We follow the experimental set-up in \cite{glaser2021kale} in which the target distribution $\pi$ ($\textcolor[rgb]{0.121, 0.467, 0.706}{\bullet}$) is defined on a manifold in $\R^2$ consisting of three non-overlapping rings. 
The initial source distribution $\mu_0$ ($\textcolor{orange}{\bullet}$) is a Gaussian distribution close to the vicinity of the first ring.
In this setting, all $f$-divergence gradient flows, including Langevin diffusion, are ill-defined because the target $\pi$ is not absolutely continuous with respect to the initial source $\mu_0$.
Nevertheless, we will simulate $\chi^2$ flow with an existing implementation of \cite{liu2023variational} that estimates the velocity field with a local linear estimator as one of the baseline methods.
In contrast, kernel-based gradient flows like $\mmd$, $\kale$, and $\dmmd$ gradient flows are well-defined in this setting, and are also used as baseline methods for comparison.

We sample $N=M=300$ samples from the initial source and the target distributions and run $\dmmd$ descent with adaptive $\lambda$ for $n_{max} = 100,000$ iterations, at which point all methods have converged. As in \cite{glaser2021kale}, we use a Gaussian kernel $ k(x, x^\prime)=\exp \left(- 0.5 \|x-x^\prime \|^2 / l^2 \right)$ with  bandwidth $l=0.3$. The step size for $\mmd$ descent is $\gamma = 10^{-2}$ and the step size for $\kale$ and $\dmmd$ descent is $\gamma = 10^{-3}$. We enforce a positive lower bound $\tilde{\lambda} = 10^{-3}$ for numerical stability and the regularity hyperparameter $r$ is optimized over the set of $\{0.1, 0.5, 1.0\}$.

\begin{figure}
    \centering
    \includegraphics[width=\linewidth]{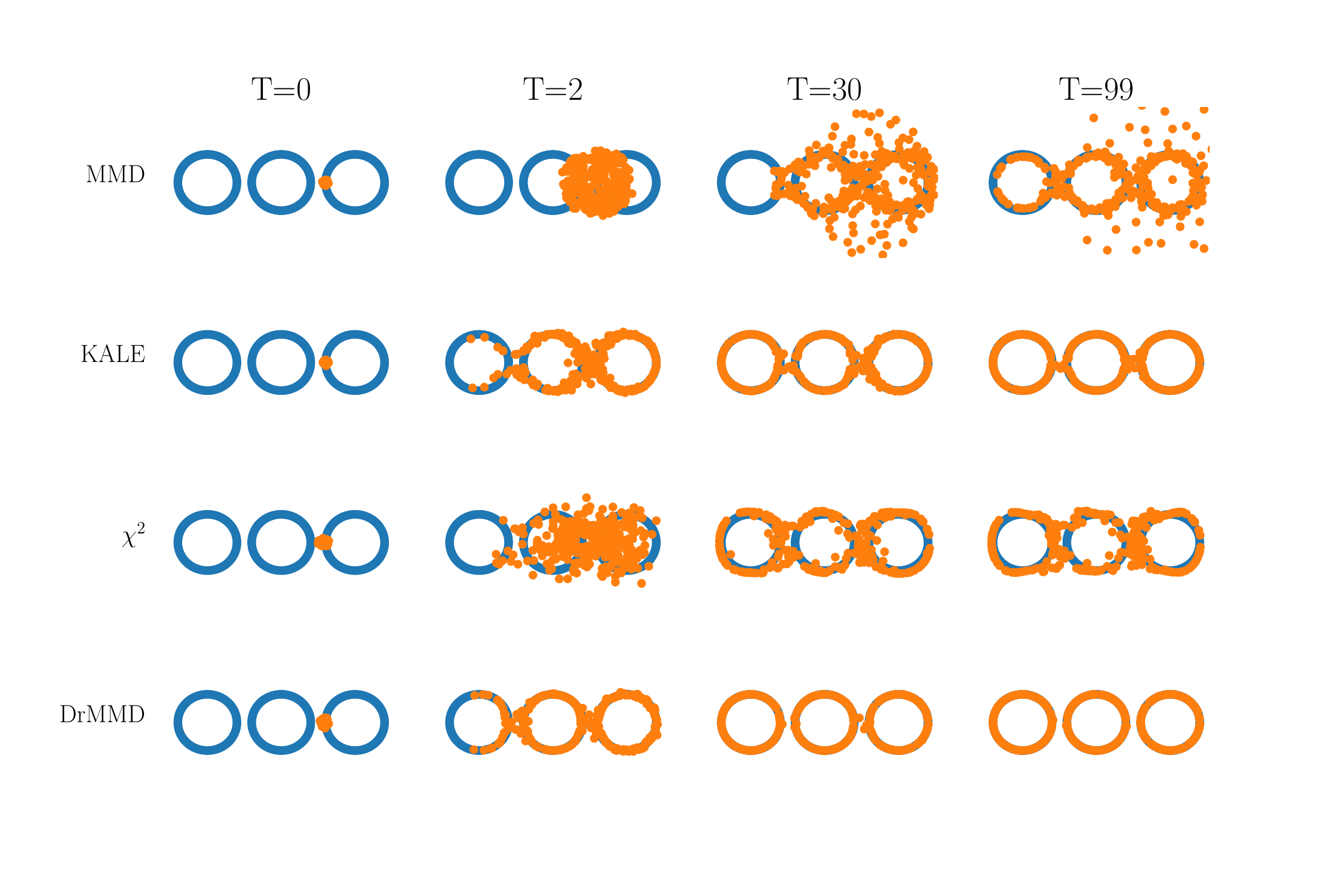}
    \vspace{-25pt}
    \caption[Three-ring gradient descent animation]{Animation of MMD, KALE, $\chi^2$ and DrMMD gradient descent on the Three-ring dataset.}
    \label{fig:three_ring_animation}
\end{figure}

From \Cref{fig:three_ring} \textbf{Left} and \textbf{Middle}, we can see that $\dmmd$ descent outperforms $\mmd$, $\kale$, and $\chi^2$ descent in terms of all dissimilarity metrics with respect to the target $\pi$: $\mmd$ and Wasserstein-2 distance.
\Cref{fig:three_ring_animation} is an animation plot visualizing the evolution of particles under these descent schemes, which demonstrates that both $\kale$ and $\dmmd$ descent are sensitive to the mismatch of support and stay concentrated in the support of the target $\pi$, while particles of $\mmd$ descent can diffuse outside the support of $\pi$. 
Note that "$\chi^2$" denotes an alternate estimate of the $\chi^2$ divergence due to ~\citet{liu2023variational}: being an $f$-divergence, we would expect "$\chi^2$ descent" to match the support of the target (as in $\kale$ and $\dmmd$). This is not the case due to bias in the velocity field being learned from samples.
Compared to $\kale$ descent, $\dmmd$ descent does not suffer from the numerical approximation error of the optimization routine when solving the velocity field of $\kale$, which explains its improved performance.

\subsection{Gradient flow for training student/teacher networks}\label{sec:student_teacher}
Next, we consider a large-scale setting following \cite{arbel2019maximum}, where a student network is trained to imitate the outputs of a
teacher network. 
We consider a two-layer neural network of the form
\begin{align*}
    \psi(z, x) = G \left(b^1+W^1 \sigma\left(W^0 z+b^0\right)\right) ,
\end{align*}
where $\sigma$ is the ReLU non-linearity and $x$ is the concatenation of all network parameters $\left(b^1, W^1, b^0, W^0\right) \in \R^d$. $G$ is an element-wise non-linear function $G : \R \to \R, x \mapsto \exp(-\frac{1}{4} x^2)$.
The teacher network is of the form: $\Psi^{T}(z, \pi) = \int \psi(z, x) d\pi(x)$ where $\pi$ denotes the teacher distribution, and the student network is $\Psi^{S}(z, \mu) = \int \psi(z, x) d\mu(x)$ where $\mu$ denotes the student distribution.
Here we consider Gaussian distributed $\mu$ and $\pi$  for simplicity. 
The student network can imitate the behavior of the teacher network by minimizing the  objective\footnote{Note that our setting is slightly different from \cite{chizat2018global} in which $\mu,\pi$ are measures over the hidden neurons, while our setting follows \cite{arbel2019maximum} in which $\mu,\pi$ are measures over all the network parameters.} 
\begin{align}\label{eq:student_teacher_1}
    \min_{\mu \in \calP_2(\R^d)} \E_{z \sim \Pb_\text{data}} \left(\Psi^{T}(z, \pi) - \Psi^{S}(z, \mu) \right)^2,
\end{align}
where $\Pb_\text{data}$ is the distribution of the input data.
If we define the kernel as the inner product of the neural network feature maps,
\begin{align*}
    k(x, x^{\prime}) = \E_{z \sim \Pb_\text{data}} [\psi(z, x)^\top \psi(z, x^{\prime})], 
\end{align*}
then the objective of \eqref{eq:student_teacher_1} can be equivalently expressed as
\begin{align*}
    \min_{\mu \in \calP_2(\R^d)} \iint_\calX k(x, x^\prime) d (\pi - \mu)(x) d (\pi - \mu)(x^\prime) ,
\end{align*}
which is precisely the $\mmd(\mu \| \pi)^2$  under the kernel $k$. Since $G(x) = \exp(-\frac{1}{4} x^2)$, the kernel is bounded and so the $\mmd$ is well-defined. Also, since the kernel $k$ has bounded first and second-order derivatives, it satisfies {\color{black}\Cref{assumption:bounded_kernel}}.

\begin{figure}[t]
\vspace{-8pt}
    \begin{minipage}{0.60\textwidth}
    \centering
    \includegraphics[width=\linewidth]{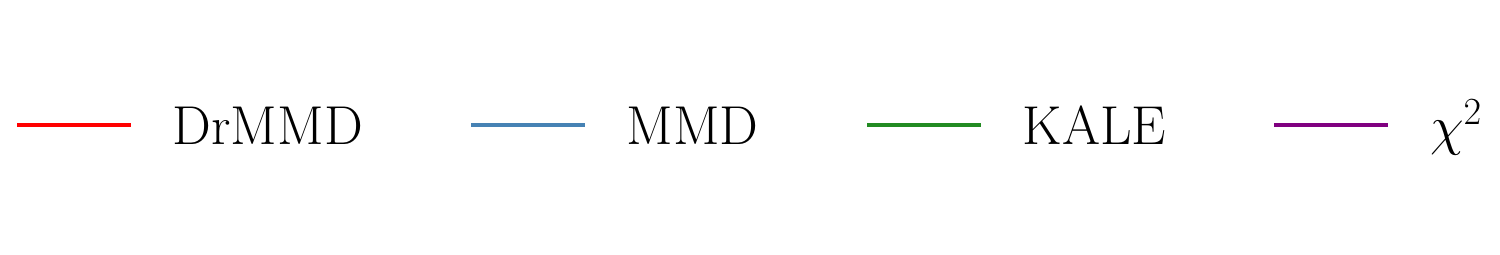}
    \vspace{-10pt}
    \end{minipage}
    \hfill
    \begin{minipage}{0.38\textwidth}
    \centering
    \includegraphics[width=\linewidth]{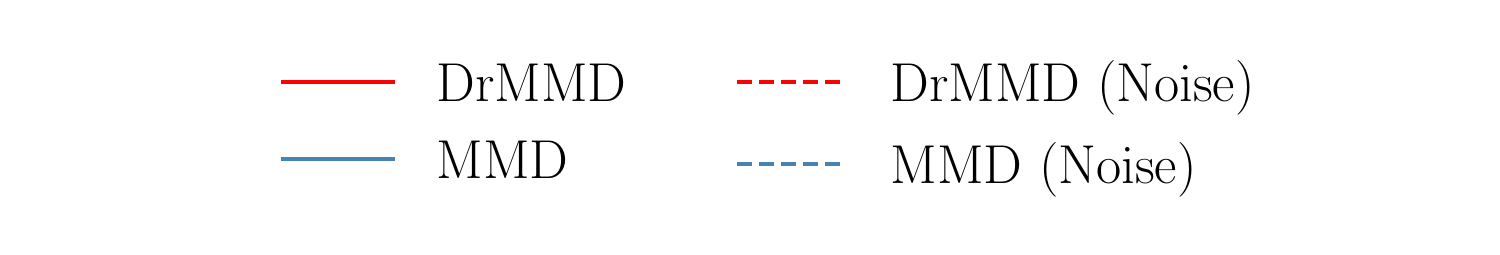}
    \vspace{-20pt}
    \end{minipage}
    \centering
    \begin{subfigure}{0.66\textwidth}
        \centering
        \includegraphics[width=1.0\linewidth]{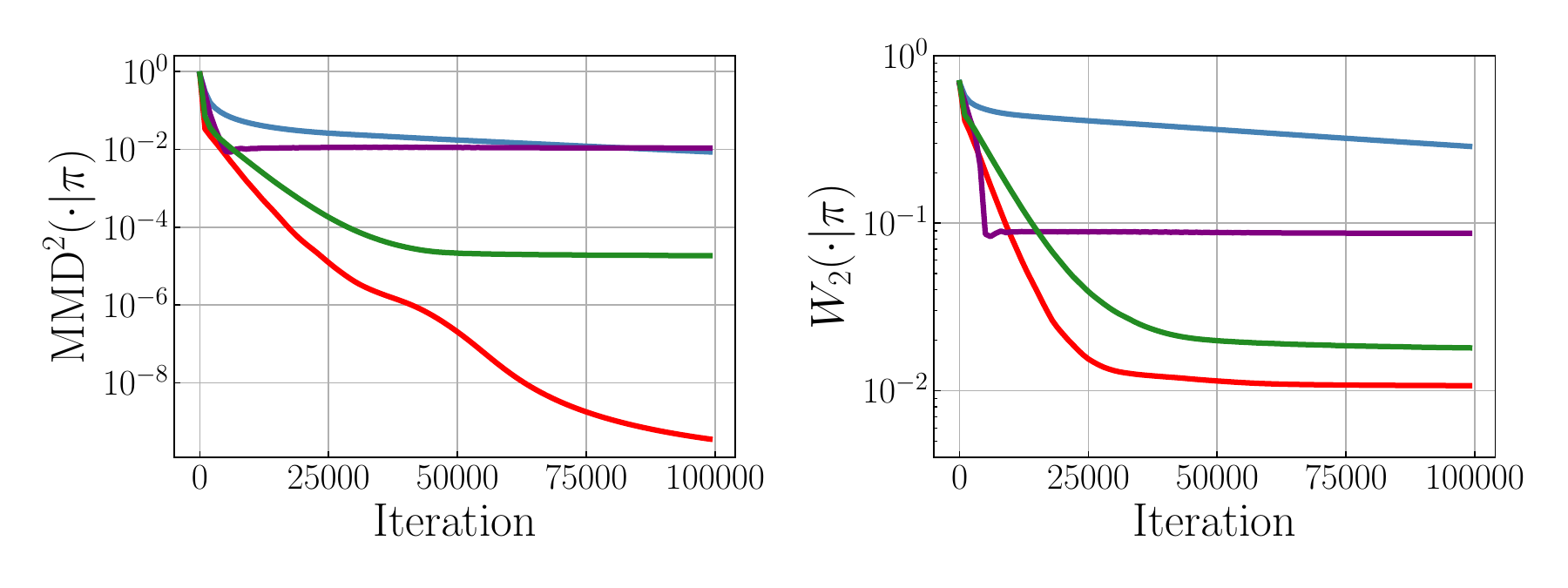}
    \end{subfigure}%
    \hfill
    \begin{subfigure}{0.33\textwidth}
        \centering
        \includegraphics[width=1.0\linewidth]{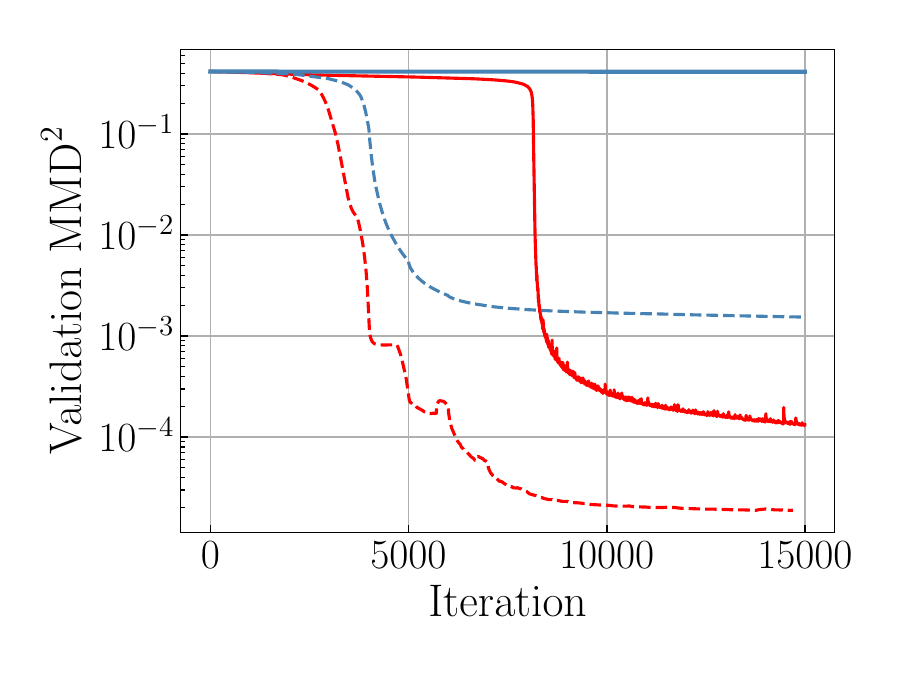}
    \end{subfigure}%
    \vspace{-20pt}
    \caption[Three-ring and student-teacher experiments]{\textbf{Left \& Middle:} Comparison of $\dmmd$ descent with adaptive $\lambda$, $\mmd$ and $\kale$ descent on three-ring synthetic data in terms of $\mmd$ and Wasserstein-2 distance with respect to the target $\pi$. \textbf{Right:} Comparison of MMD descent with and without noise injection, $\dmmd$ descent with and without noise injection on training
    student/teacher networks in terms of validation $\mmd^2$ distance.}
    \label{fig:three_ring}
    \vspace{-15pt}
\end{figure}

Therefore, the training of the student network with objective \eqref{eq:student_teacher_1} can be treated as an optimization problem of $\mmd^2$ distance in $\calP_2(\R^d)$, i.e., as $\mmd$ gradient flow. 
It is shown in \cite{arbel2019maximum} that $\mmd$ flow and its descent scheme will generally get stuck in  local optima because 
$\mmd$ is not \emph{geodesically} convex; therefore, noise injection has been proposed to escape these local optima. 

With support from Theorems \ref{thm:continuous_time_convergence} and \Cref{thm:discrete_time_KL} on the convergence of $\dmmd$ flow and its descent scheme,  we propose to minimize $\dmmd(\mu\|\pi)$ and use $\dmmd$ descent rather than minimizing  $\mmd(\mu\|\pi)^2$ directly. 
Although this does not directly minimize the objective in \eqref{eq:student_teacher_1}, the favorable convergence performance of $\dmmd$ descent should result in a smaller $\mmd(\mu\|\pi)^2$ at convergence. 

In our experimental setting, we are given $M=10$ particles $x^{(1)}, \cdots, x^{(M)}$ from the teacher distribution $\pi = \calN(0,\Id)$ and $N=1000$ particle $y_0^{(1)}, \cdots, y_0^{(N)}$ from the initial student distribution $\mu_0 = \calN(0, 10^{-3} \Id)$. The teacher particles are fixed while
the student particles are updated by {\color{black}\Cref{alg:drmmd_particle_descent}} at each time step.
The initial (de)-regularization parameter is $\lambda_0 = 0.1$, the step size is $\gamma = 0.1$, we apply a lower bound $\tilde{\lambda} = 10^{-3}$, and the regularity $r$ is optimized over the set of $\{ 0.1, 0.5, 1.0\}$.
For the architecture of the neural network, there are $3$ neurons in the hidden layer, and the output dimension is $1$. 
The data distribution $\Pb_{\text{data} }$  is a uniform distribution on the sphere in $\R^p$ with $p = 50$.
$2000$ data are sampled from $\Pb_{\text{data} }$ with $1000$ as training dataset and another $1000$ as validation dataset.
The kernel $k(x, x^\prime) = \mathbb{E}_{z \sim \Pb_{\text{data} } } \left[\psi(z, x)^{\top} \psi\left(z, x^{\prime}\right)\right]$ is estimated by the average over $100$ randomly selected samples from the training dataset at each iteration. 
$\dmmd$ and $\mmd$ descent stop after $n_{max}=15, 000$ iterations when both converge.
The final performance is evaluated in terms of $\mmd^2(\mu_{n_{max}} \| \pi)$ with kernel $k$ estimated by the average of $1000$ samples in the validation dataset.

In \Cref{fig:three_ring} \textbf{Right}, we report the performance of $\mmd$ descent (with and without noise injection) along with the $\dmmd$ descent (with and without noise injection) in terms of $\mmd$ distance on the validation dataset.
We can see that the $\dmmd$ descent does not get stuck in a local optimum, and leads to much lower validation $\mmd(\mu_{n_{max}} \| \pi)^2$ even without noise injection. 
We also run $\dmmd$ descent with the noise injection scheme and find that noise injection can further improve the performance of $\dmmd$ descent and outperforms $\mmd$ descent with noise injection.
Although it is unclear whether the density ratio has enough regularity to meet the condition of \Cref{thm:discrete_time_KL}, the kernel $k$ satisfies the boundedness and smoothness conditions of {\color{black}\Cref{assumption:bounded_kernel}} and the target $\pi$ satisfies the Poincar\'{e} inequality since it is Gaussian. The $\dmmd$ descent benefits from more favorable convergence properties, which explains its superior performance.


The code to reproduce all the experiments can be found in the following GitHub repository. \url{https://github.com/hudsonchen/DrMMD}.

\section{Discussion}
In this paper, we introduced (de)-regularization of the MMD (called $\dmmd$) and its associated Wasserstein gradient flow. 
As an interpolation between the $\mmd$ and $\chi^2$-divergence, the $\dmmd$ gradient flow inherits strengths from both sides: it has a closed-form velocity field and an exponential rate of convergence towards the global minimum up to a controllable barrier term when the target $\pi$ satisfies a Poincar\'{e} inequality.
Additionally, we provide the optimal adaptive selection of a regularization coefficient
that best balances the approximation and time discretization errors in $\dmmd$ gradient descent. 
Our work is the first to prove theoretically that an adaptive kernel through adaptive regularization can result in improved convergence of MMD gradient flow. The theoretical results are consistent with the empirical evidence in several numerical experiments. 

Following our work, there remain a number of interesting open problems. For example, 
(i) Since the kernel bandwidth has been known to play an important role in the performance of kernel-based algorithms, it is of interest to study the adaptive choice of kernel bandwidth in the context of $\dmmd$ gradient flow. (ii) To generalize our convergence analysis to the Wasserstein gradient flow of all Moreau envelopes of $f$-divergences in reproducing kernel Hilbert space, even when they do not have a closed-form expression as $\dmmd$. (iii) While the current work proposes an approximation to the $\chi^2$-squared flow in the generative modeling setting, i.e., where the target distribution $\pi$ is known only through samples, it will be interesting to construct approximations to $\chi^2$-flow in the sampling setting, i.e., where $\pi$ is known in closed form (at least up to normalization).





\chapter{Minimum MMD for Statistical Inference}
\label{sec:pgd}

\begin{center}
\setlength{\fboxsep}{12pt}
\fcolorbox{paperboxborder}{paperboxbg}{%
\begin{minipage}{0.88\textwidth}
The results in this chapter were published in the following paper:

\begin{itemize}
    \item Sophia Seulkee Kang, Louis Sharrock, Xiaoyuan Cheng, Fran\c{c}ois-Xavier Briol, and Zonghao Chen.

    A Gradient Flow Perspective on Minimum MMD Estimation.

    \href{https://arxiv.org/abs/2607.03871}{\texttt{arXiv:2607.03871}}.
\end{itemize}
All theoretical results in this chapter are due to the author. 
Experiments are carried out by Sophia Seulkee Kang and are included to support the
theory. 
As the last author, the author also played a supervisory role in guiding the overall research direction and presentation of the work. 

\end{minipage}%
}
\end{center}

\section{Introduction}

Parameter estimation seeks to recover the unknown parameters governing a data-generating mechanism from observed data. Given a parametric family of statistical models $\{\Pb_\theta: \theta \in \Theta\}$, where $\Theta\subseteq\R^p$ is the parameter domain, a standard approach is to estimate these parameters by minimizing a divergence or discrepancy between the model and the data-generating distribution $\Qb$~\citep{wolfowitz1957minimum,parr1980minimum,basu2011statistical}.
Since the latter is unknown, an estimator is typically constructed by replacing it with the empirical distribution consisting of independently and identically distributed (IID) observations.
A canonical choice of such discrepancy is the Kullback--Leibler divergence, for which the resulting minimum-distance estimator coincides with maximum likelihood estimation~\citep{van2000asymptotic}.

Recently, a growing body of work has considered using the maximum mean discrepancy (MMD), a metric based on embeddings of probability measures into a reproducing kernel Hilbert space~\citep{gretton2012kernel,muandet2017kernel}, leading to the so-called \emph{minimum MMD} estimator~\citep{briol2019statistical}.
This choice is attractive for both statistical and computational reasons. Statistically, minimum MMD estimators are robust to model misspecification, such as outlier contamination and covariate shift~\citep{briol2019statistical,cherief2020mmd,dellaporta2023robust}. Computationally, they remain well defined even when the likelihood is unavailable or intractable; such settings are commonly referred to as likelihood-free inference~\citep{niu2021discrepancy,dellaporta2022robust,bharti2023optimally}. These favorable properties have also led to numerous applications of minimum MMD estimators, including for hypothesis testing~\citep{key2025composite,schrab2023mmd,bruck2025distribution,fazeli2023semi}, parametric regression~\citep{alquier2024universal}, time-series modeling~\citep{alquier2026timeseries}, distributionally robust optimization~\citep{dellaporta2025decision}, generalizations of Bayesian inference \citep{dellaporta2022robust,dellaporta2023robust,chen2025total}, calibration of stochastic simulation models in operations research~\citep{su2025differentiable}, copula modeling~\citep{alquier2023estimation}, inference under missing data~\citep{cherief2025parametric}, and learning of summary statistics~\citep{khoo2026minimum}.

\begin{figure*}[t]
    \centering
    \begin{minipage}[t]{0.66\linewidth}
        \centering
        \includegraphics[width=\linewidth]{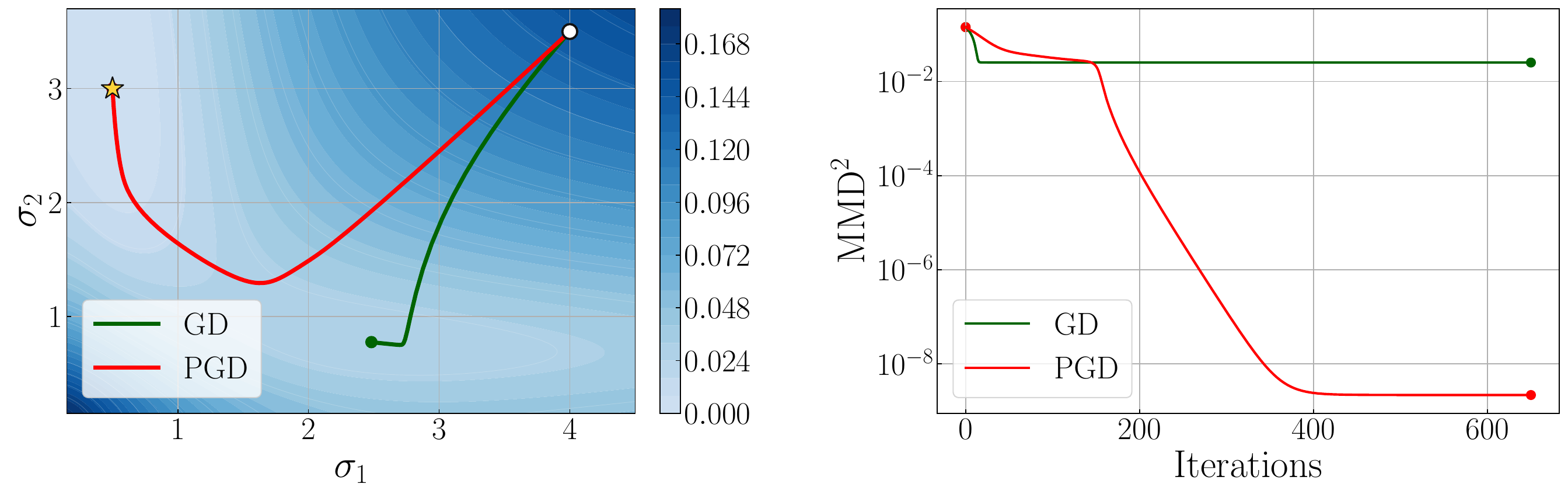}
    \end{minipage}
    \vspace{-10pt}
    \caption[Illustrative minimum MMD example]{\textit{Illustrative example.} Both the target distribution $\Qb$ and the model distribution $\Pb_\theta$ are two-mode Gaussian mixtures, where $\theta=(\sigma_1,\sigma_2)$ denotes the standard deviations of the two modes. We compare minimum MMD estimators trained by standard gradient descent (GD) and our preconditioned gradient descent (PGD). The left panel shows the optimization trajectories, with the yellow star marking the global minimizer, while the right panel shows the evolution of the MMD objective values.
    The objective landscape is non-convex.}
    \label{fig:illustration}
\vspace{-10pt}
\end{figure*}

Despite the growing popularity of these estimators, very little work has focused on solving the associated optimization problem
\[
    \arg\min_{\theta\in\Theta} \mmd^2(\Pb_\theta,\Qb),
\]
compared to its accompanying statistical theory.
Existing methods typically rely on (stochastic) gradient descent~\citep{briol2019statistical,cherief2020mmd}, including the recent R-package implementation of \cite{alquier2025package}, and this can at best be expected to converge to stationary points due to the non-convexity of the objective.
To the best of our knowledge, the only existing optimization guarantee is Proposition~5.2 of \cite{cherief2022finite}, which assumes convexity of the objective, a condition that rarely holds in practice; see \Cref{fig:illustration} for an illustration on a simple two-mode Gaussian mixture model with unknown variances.
Therefore, an open problem remains:
\begin{center}
\emph{``Can we design a gradient-based optimization scheme for minimum MMD estimation with convergence guarantees beyond convex objectives?''}
\end{center}

\definecolor{darkgreen}{RGB}{0,100,0}

In this paper, we make progress on this problem by proposing a new descent scheme for minimizing the MMD objective, and prove that it can converge to a (not necessarily unique) \emph{global} minimizer under explicit assumptions.
Our approach builds on MMD gradient flow, a descent scheme for minimizing the squared MMD directly over the space of probability measures, without imposing a parametric model representation~\citep{arbel2019maximum}.
Existing MMD gradient flows have been studied primarily in the contexts of generative modelling~\citep{galashov2024deep,hertrich2023generative,neuralwgf,hertrich2024wasserstein,chen2025regularized,tian2026sobolev} and sampling~\citep{korba2021kernel,chen2025stationary,tian2026sobolev}, and are typically run with a fixed kernel throughout the dynamics.
We depart from this fixed-kernel formulation by introducing an \emph{adaptive} nonparametric MMD descent scheme $(\Pb_t)_{t\in\N}$, in which the kernel lengthscale $(\ell_t)_{t\in\N}$ is varied during optimization according to a schedule converging to a prescribed limit $\ell>0$~\citep{chen2021deterministic}.
Our first main result, \Cref{thm:np_rate}, proves that this adaptive scheme satisfies $\lim_{t\to\infty}\mmd(\Pb_t,\Qb)=0$, under a trajectory-wise gradient-dominance condition.

Motivated by this global convergence result, we then design a descent scheme  such that the parametric distributions $(\Pb_{\theta_t})_{t\in\N}$ closely approximate $(\Pb_t)_{t\in\N}$ at each iteration $t$.
This is achieved via projecting the nonparametric descent direction onto the space of directions realizable by the parametric model.
Compared with the standard gradient, the resulting update on $\theta$ involves a preconditioning matrix. We therefore refer to it as \emph{preconditioned} gradient descent (PGD).
We next prove in \Cref{thm:p_rate} that, under gradient dominance and projection residual conditions, the parametric models optimized under PGD, denoted as $(\Pb_{\theta_t}^\pgd)_{t\in\N}$, satisfy $\lim_{t\to\infty}\mmd(\Pb_{\theta_t}^\pgd, \Qb)=\min_{\theta\in\Theta} \mmd(\Pb_{\theta}^\pgd, \Qb)$, even in non-convex settings.
Therefore, our results help bridge the gap between the statistical theory of minimum MMD estimation and its practical computation.
Finally, the effectiveness of the proposed PGD descent scheme is empirically verified on several statistical inference benchmarks where we show that the PGD scheme not only converges to a global minimum, but can also accelerate convergence relative to gradient descent.

\section{Nonparametric MMD Descent with Adaptive Lengthscale}
In this section, we first establish a convergence guarantee for nonparametric MMD descent with \emph{adaptive} kernel lengthscales. We propose to initiate the MMD gradient flow algorithm with a large lengthscale, then to monotonically reduce it towards some target value as the number of iterations increases.
This is known empirically to improve the convergence behavior of MMD flows~\citep{chen2021deterministic,galashov2024deep,chen2025regularized}, and to even be useful for other statistical applications of the MMD~\citep{schrab2023mmd,biggs2023mmd}. 
The intuition for adaptive lengthscale is analogous to that of annealing methods: adapting the objective function throughout the optimisation routine can significantly improve exploration. We also note that adaptive lengthscale will be straightforward to use directly in the context of parametric updates. 
Hence, the convergence guarantee proved here provides one foundation for the optimization analysis of parametric MMD descent in \Cref{sec:parametric}.

Let $f_{\ell, \Pb, \Qb}$ denote the witness function between $\Pb$ and $\Qb$ under kernel $k_\ell$; and  $\mmd_{\ell}(\Pb, \Qb)$ the corresponding MMD. 
We now add subscripts to emphasize the dependence on the kernel lengthscale $\ell$. 
Given a sequence of lengthscales $(\ell_t)_{t\in\mathbb{N}}$, the adaptive MMD descent scheme is defined by 
\begin{align}\label{eq:adaptive_mmd_descent}
    \Pb_{t+1} = (\Id - \gamma_t \nabla f_{\ell_t, \Pb_t,\Qb})_\# \Pb_t .  
\end{align}
In the remainder of this paper, we focus explicitly on the Gaussian and Mat\'ern kernels,  due to their widespread use for minimum MMD estimation~\citep{briol2019statistical,cherief2020mmd}; see \Cref{rem:extension_general_kernel} for an extension to other kernels. 

\begin{ass}[Kernel]\label[assumption]{ass:kernel}
Let \(k_\ell\) be either a Gaussian kernel
\begin{align*}
    k_\ell(x,y) = \exp\left(-\frac{\|x-y\|^2}{2\ell^2} \right), 
\end{align*} 
or the Mat\'ern kernel with smoothness parameter \(s>1\),
\begin{align*}
    k_\ell(x,y)
    =
    \frac{2^{1-s}}{\Gamma(s)}
    \left(
        \frac{\sqrt{2s}\|x-y\|}{\ell}
    \right)^s 
    K_s \left( \frac{\sqrt{2s}\|x-y\|}{\ell}
    \right),
\end{align*}
Here, $l>0$ is a lengthscale, and $\Gamma$ and $K_s$ denote the Gamma function and the modified Bessel function of the second kind respectively.
\end{ass}

\begin{theorem}[Global convergence of nonparametric MMD descent] \label{thm:np_rate} 
Let kernel $k$ satisfy \textcolor{black}{\Cref{ass:kernel}}.
Suppose $\Qb\in\calP_2(\R^d)$. 
Let $\Pb_0\in\calP_2(\R^d)$ and let 
$(\Pb_t)_{t\in\N}$ be the adaptive MMD descent defined in Eq.~\eqref{eq:adaptive_mmd_descent} with lengthscale $\ell_t$. 
Let $C>0$ be a constant independent of $t$ and $(\gamma_t)_{t\in\N}$ a monotonically decreasing sequence of step sizes such that $\gamma_t\in(0,\frac{1}{2C})$, $\lim_{t\to\infty} \gamma_t=0$ and $\sum_{t=0}^\infty \gamma_t=\infty$. 
Let $(\ell_t)_{t\in\N}$ be a monotonically decreasing sequence of lengthscales such that $\lim_{t\to\infty} \ell_t = \ell > 0$ and that $\left(\ell_t \ell^{-1}\right)^d \|\nabla f_{\ell, \Pb_t,\Qb} \|_{L_2^d(\Pb_t)}^2 \geq C \mmd_{\ell}^2(\Pb_{t}, \Qb)$ for all $t\in\N$. Then $\lim_{t\to\infty} \mmd_{\ell}^2(\Pb_t, \Qb) = 0$. 
\end{theorem}
The proof is in \Cref{sec:proof_np_descent}. 
\Cref{thm:np_rate} shows that, under its step-size and trajectory-wise gradient-dominance assumptions, an appropriate kernel lengthscale sequence ensures that $\Pb_t$ converges to $\Qb$ in MMD despite the potential non-convexity of the objective.
A non-asymptotic upper bound on $\mmd_{\ell}^2(\Pb_t, \Qb)$ is provided in Eq.~\eqref{eq:term_1}; however, its interpretation is less transparent without further assumptions on $(\ell_t)_{t\in\N}$. We therefore choose to present only the asymptotic result. 

The condition of a diminishing yet cumulatively unbounded step size is standard in convex optimization \cite[e.g.,][]{moulines2011non}. The lengthscale condition 
can be interpreted as a trajectory-wise gradient dominance condition, and can be verified \emph{a posteriori}; see Section~\ref{sec:mmdal_experiments}. 
Since $\ell_t$ appears only on the left-hand side of the inequality, this suggests that larger kernel lengthscales can make the sufficient condition easier to satisfy. 
Increasing the lengthscale can yield a more `benign' optimization landscape via more smoothing~\citep{starnes2026gaussian}. 
Therefore, it is natural to initialize with a relatively large lengthscale and gradually decrease it toward $\ell$ as optimization proceeds. 
As $\Pb_t$ approaches $\Qb$, the relevant local optimization landscape may behave more like a quadratic, and global convergence can be expected even with smaller lengthscales. 
Furthermore, since $\ell_t/\ell > 1$, the prefactor on the left-hand side increases with $d$, and therefore the condition is likely less restrictive in higher dimensions. 
A smaller value of $C$ makes the condition easier to satisfy, but at the cost of a slower convergence rate from Eq.~\eqref{eq:term_1}. 
Similar conditions have been used in the convergence analysis of kernel-based gradient flows; see Proposition 8 of \cite{arbel2019maximum}, Proposition 5 of \cite{glaser2021kale}, Theorem 4.1 of \cite{chen2025stationary}, and also in graduated optimization~\citep{hazan2016graduated}. 
Their assumptions involve expectations with respect to an additional injected noise, whereas our condition is deterministic and only needs to be checked along the current \emph{realized} trajectory. 




\section{Parametric MMD Descent with Adaptive Lengthscale} \label{sec:parametric}
In this section, we propose a new approach for training parametric MMD models. The guiding principle is to design a parametric update that remains as faithful as possible to nonparametric MMD descent, and leverage the convergence result in \Cref{thm:np_rate}.

Given an unknown data generating distribution $\Qb$ and a parametrized
family of model distributions $\{\Pb_\theta : \theta \in \Theta\}$, where $\Theta\subseteq\R^p$, minimum MMD estimation seeks
$\arg \min _{\theta\in\Theta} \mmd_{\ell}^2(\Pb_\theta, \Qb)$ for some lengthscale $\ell > 0$.
We consider in this paper parametric models $\Pb_\theta=(\Gmap_\theta)_{\#} \rho$, where $\rho\in\calP_2(\calZ)$ is a simple base probability measure on a latent space $\calZ\subseteq\R^q$, such as a Gaussian or uniform distribution, and $\Gmap_\theta: \calZ \rightarrow \R^d$ is a generator parameterized by $\theta \in \Theta$.
For the analysis, we take $\Theta$ to be convex so that line segments between successive parameter iterates remain in the parameter domain.
The latent dimension $q$ need not equal the data dimension $d$.
A natural way to evolve the parametric model $\Pb_\theta$ is to update the parameter $\theta$ such that the distribution moves through the pushforward map. In particular, we consider an iterative scheme given by $\theta_{t+1}=\theta_t-\gamma_t\Delta \theta_t \in \Theta$ for any $t\in\N$; if the unconstrained update leaves $\Theta$, it is projected back to $\Theta$. Hence,
\begin{align}\label{eq:mmd_descent}
    \Pb_{\theta_{t+1}} = (\Gmap_{\theta_{t+1}})_\# \rho = (\Gmap_{\theta_t - \gamma_t \Delta \theta_t})_\# \rho,  \quad \forall t \in\N .
\end{align}
Here, $\gamma_t>0$ is a step size and $\Delta \theta_t \in \R^p$ is a parameter update to be specified.
A standard choice is to take $\Delta \theta_t$ as the gradient of $\frac{1}{2} \mmd_{\ell}^2(\Pb_\theta, \Qb)$ with respect to $\theta$.
This gives,
via the chain rule,
\begin{align}\label{eq:parametric_mmd_update_old}
    \Delta \theta_t^{\mathrm{GD}}  = \nabla_\theta \left[ \frac{1}{2} \mmd_{\ell}^2(\Pb_\theta, \Qb) \right] =  \E_{z\sim\rho}\left[ \bJ_\theta \Gmap_{\theta_t}(z)^\top \nabla f_{\ell, \Pb_{\theta_t}, \Qb}(\Gmap_{\theta_t}(z)) \right],  
\end{align}
where $\bJ_\theta \Gmap_{\theta_t}(z) \in\R^{d\times p}$ represents the Jacobian matrix of the mapping $\theta\mapsto \Gmap_\theta(z)$ evaluated at $\theta_t$.
We call the resulting sequence $(\Pb_{\theta_{t}}^\gd)_{t\in\N}$ \emph{parametric MMD descent}.
Although this straightforward update scheme has been used in the majority of the literature, it remains unclear whether it can guarantee $\lim_{t\to\infty} \mmd_{\ell}(\Pb_{\theta_t}^\gd, \Qb) = {\min_{\theta\in\Theta}}\mmd_{\ell}(\Pb_{\theta}, \Qb)$.

\begin{wrapfigure}{r}{0.50\textwidth}
    \centering
    \vspace{-0.2em}
    \includegraphics[width=0.40\textwidth]{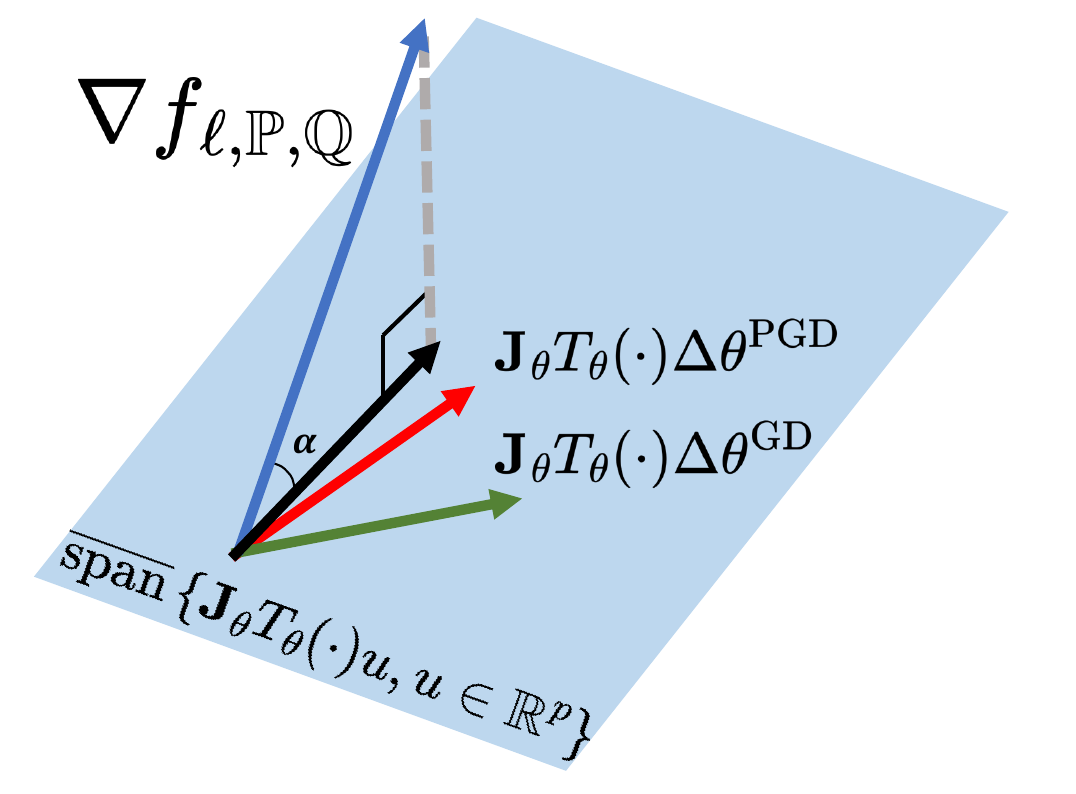}
    \caption[Geometric interpretation of projection]{A geometric interpretation of the projection.
    }
    \label{fig:residual_geometry}
\end{wrapfigure}

Motivated by the global convergence of nonparametric MMD descent established in \Cref{thm:np_rate}, we introduce an alternative parameter update $\Delta \theta$, such that the squared MMDs at the next iterate under the parameterid and nonparametric updates, $\mmd_{\ell}^2(\Pb_{t+1},\Qb)$ and $\mmd_{\ell}^2(\Pb_{\theta_{t+1}},\Qb)$, match up to first order in $\gamma_t$.
Informally, suppose that $\Pb_t \approx \Pb_{\theta_t} = (\Gmap_{\theta_t})_{\#}\rho$ at iteration $t$. Under mild regularity assumptions, both squared MMD objectives at the next iteration $t+1$ admit the following Taylor expansions in $\gamma_t$ up to second-order remainders:
\begin{align*}
    \mmd_{\ell}^2(\Pb_{t+1}, \Qb)
    &= \mmd_{\ell}^2(\Pb_{t}, \Qb)
    - 2\gamma_t \E_{z\sim\rho}\left[
    \begin{aligned}
    &\nabla f_{\ell,\Pb_{t},\Qb} \!\left(\Gmap_{\theta_{t}}(z)\right)^\top\\
    &\quad \nabla f_{\ell_t,\Pb_{t},\Qb} \!\left(\Gmap_{\theta_t}(z)\right)
    \end{aligned}
    \right] + \calO(\gamma_t^2) \\
    \mmd_{\ell}^2(\Pb_{\theta_{t+1}}, \Qb)
    &= \mmd_{\ell}^2(\Pb_{\theta_t}, \Qb)
    - 2\gamma_t \E_{z\sim\rho}\left[
    \begin{aligned}
    &\nabla f_{\ell,\Pb_{\theta_t},\Qb} \!\left(\Gmap_{\theta_t}(z)\right)^\top\\
    &\quad \bJ_\theta \Gmap_{\theta_t}(z) \Delta\theta
    \end{aligned}
    \right] + \calO(\gamma_t^2).
\end{align*}
Comparing the first-order term in $\gamma_t$ above, we are motivated to pick the update $\Delta \theta$ such that $\bJ_\theta \Gmap_{\theta_t}(\cdot) \Delta\theta \approx \nabla f_{\ell_t,\Pb_{t},\Qb}(\Gmap_{\theta_t}(\cdot))$.
This naturally leads to a \emph{projection} of the vector field $\nabla f_{\ell_t,\Pb_{t},\Qb}(\Gmap_{\theta_t}(\cdot))$ onto
$\mathrm{span}\{ \bJ_\theta \Gmap_{\theta_t}(\cdot) u : u\in \R^p \}\subset L_2^d(\rho)$ (see \Cref{fig:residual_geometry}).
This subspace consists of the directions that can be realized by infinitesimal updates of the parametric model at $\theta_t$.
The parameter update realizing this projection is therefore obtained by solving the following least-squares problem:
\begin{align}\label{eq:calJ}
    \calJ_t(\Delta \theta_t)
    := \E_{z\sim \rho} \left[
    \left\| \bJ_{\theta} \Gmap_{\theta_t}(z) \Delta \theta_t
    - \nabla f_{\ell_t, \Pb_{\theta_t},\Qb}(\Gmap_{\theta_t}(z)) \right\|^2
    \right] .
\end{align}
Since $\calJ_t(\Delta \theta_t)$ is quadratic in $\Delta \theta_t\in\R^p$, its minimizer admits a closed-form expression
\begin{align}\label{eq:parametric_mmd_update_new}
    \Delta \theta_t^\pgd
    &=
    \left\{ \E_{z\sim\rho}\left[
    \bJ_\theta \Gmap_{\theta_t}(z)^\top
    \bJ_\theta \Gmap_{\theta_t}(z)\right] + \lambda_t \Id \right\}^{-1}
    \nonumber\\
    &\quad \times
    \E_{z\sim\rho}\left[
    \bJ_\theta \Gmap_{\theta_t}(z)^\top
    \nabla f_{\ell_t, \Pb_{\theta_t}, \Qb}(\Gmap_{\theta_t}(z))
    \right] .
\end{align}
where we include a ridge regularization $\lambda_t \geq 0$.
Taking $\lambda_t=0$ corresponds to an exact projection of $\nabla f_{\ell_t,\Pb_{\theta_t},\Qb}\circ \Gmap_{\theta_t}$, with respect to the $L_2(\rho)$ norm, onto the linear subspace $\mathrm{span}\{ \bJ_\theta \Gmap_{\theta}(\cdot) u : u\in \R^p \}$.
However, this would require inverting $\E_{z\sim\rho} \left[ \bJ_\theta \Gmap_{\theta_t}(z)^\top \bJ_\theta \Gmap_{\theta_t}(z)\right]$, which can be numerically unstable when the matrix has small eigenvalues; this instability is further amplified when the expectations with $\rho$ are estimated from finite samples.
Including a small diagonal matrix $\lambda_t \Id$ avoids this issue.

At each iterate $t$, the choice of $\lambda_t$ balances a trade-off between an \emph{approximation error} of $\Delta\theta^\pgd_t$ towards the exact projection; and another \emph{approximation error} from using a second-order Taylor expansion.
See \Cref{rem:tradeoff} for details.
Compared with the standard update $\Delta \theta_t^\gd$ in Eq.~\eqref{eq:parametric_mmd_update_old}, our update $\Delta \theta_t^\pgd$ includes a \emph{preconditioning} matrix. For this reason, we refer to the resulting sequence $(\Pb_{\theta_{t}}^\pgd)_{t\in\N}$ as \emph{preconditioned} parametric MMD descent.

Before we proceed to present an upper bound  that controls the gap between the nonparametric and preconditioned parametric updates after one step, we first introduce the following assumptions.

\begin{ass}[Regularity of $\Gmap_\theta$]\label[assumption]{ass:regularity_G}
Suppose that $\theta\mapsto \Gmap_\theta(z)$ is twice continuously differentiable. 
For all $\theta\in\Theta$ and for all \(i,j\in\{1,\dots,p\}\), suppose that $\E_{z\sim\rho}[\|\Gmap_\theta(z)\|^2] \leq \mathfrak{T}_1$,
$\E_{z\sim\rho}[\|\partial_{\theta_i} \Gmap_\theta(z)\|^2] \le \mathfrak{T}_2$ and $\E_{z\sim\rho}[\|\partial_{\theta_j}\partial_{\theta_i} \Gmap_\theta(z)\|^2] \le \mathfrak{T}_3$.
\end{ass}
For notational simplicity, we set $\mathfrak{T}_1=\mathfrak{T}_2=\mathfrak{T}_3=1$ in the proofs.
\begin{ass}[Projection residual]\label[assumption]{ass:residual}
For each \(t\in\mathbb{N}\), suppose that there exists a minimizer
$u_t^\star \in \arg\min_{u\in\mathbb{R}^p} \calJ_t(u)$ such that, for some constants \(0\leq \mathfrak{R}<1\) and \(0<\calU<\infty\) independent of \(t\),
\begin{align*}
    \calJ_t(u_t^\star)
    \leq \mathfrak{R}^2
    \|\nabla f_{\ell_t,\Pb_{\theta_t},\Qb}\|_{L_2^d(\Pb_{\theta_t})}^2,
    \qquad
    \|u_t^\star\| \leq \calU .
\end{align*}
\end{ass}
\noindent
\textcolor{black}{\Cref{ass:regularity_G} serves} two purposes. First, the condition $\E_{z\sim\rho}[\|\Gmap_\theta(z)\|^2]\leq 1$ ensures that the parametric model family satisfies $\{\Pb_\theta:\theta\in\Theta\} \subseteq \calP_2(\R^d)$.
Second, the bounds
$\E_{z\sim\rho} [\|\partial_{\theta_i}\Gmap_\theta(z)\|^2] \leq 1$
and $\E_{z\sim\rho}[\|\partial_{\theta_j} \partial_{\theta_i} \Gmap_\theta(z)\|^2\big]\leq 1$ provide the regularity needed to control first- and second-order variations of the model $\Pb_\theta$ with respect to $\theta$, required for the remainder terms in the Taylor expansion to be well controlled.
It is slightly stronger than the integrability condition imposed by \cite{briol2019statistical} for only the well-posedness of the gradient update in Eq.~\eqref{eq:parametric_mmd_update_old}. 
\par
\textcolor{black}{\Cref{ass:residual}} admits a natural geometric interpretation.
For the first inequality of \textcolor{black}{\Cref{ass:residual}}, $\mathfrak{R}$ can be interpreted as a uniform-in-$t$ upper bound on $|\sin(\alpha)|$, where $\alpha$ denotes the projection angle shown in  \Cref{fig:residual_geometry}.
Hence the condition $\Re<1$ rules out the degenerate case in which $\nabla f_{\ell_t, \mathbb{P}_{\theta_t}, \mathbb{Q}} \circ \Gmap_{\theta_t}(\cdot)$ is orthogonal to the \emph{tangent subspace} $\operatorname{span}\left\{\mathbf{J}_\theta \Gmap_{\theta_t}(\cdot) u: u \in \mathbb{R}^p\right\}$, the span of all admissible parametric directions.
In that case, the projection is necessarily zero, so the preconditioned descent scheme is not expected to make progress in reducing the MMD. 
The second inequality of \textcolor{black}{\Cref{ass:residual}} requires the exact-projection coefficient, displayed by the black vector in \Cref{fig:residual_geometry}, to have a uniform-in-$t$ Euclidean upper bound.
This is a potentially necessary condition for the projection to be uniformly controlled.
\textcolor{black}{\Cref{ass:regularity_G}} can be checked, whereas \textcolor{black}{\Cref{ass:residual}} is difficult to verify in practice.

\begin{example}[Sufficient conditions and examples for \textcolor{black}{\Cref{ass:residual}}]\label{example:ass}
    A sufficient condition for \textcolor{black}{\Cref{ass:residual}} to hold, is that $\nabla f_{\ell, \Pb_\theta, \Qb}(\Gmap_\theta(\cdot )) \in \mathrm{span}\{ \bJ_\theta \Gmap_\theta(\cdot) u, u\in\R\}$ with coefficients uniformly-in-$t$ bounded.
    Then the residual is zero, so the first inequality in \textcolor{black}{\Cref{ass:residual}} holds with $\mathfrak{R}=0$, and the second inequality holds from uniformly bounded coefficients.
    We give a simple example in which the tangent-span condition is satisfied. 
    Let the target distribution $\Qb$ be any rotationally invariant probability measure on $\mathbb{R}^d$, such as an isotropic Gaussian, an isotropic Student-t distribution, or the uniform distribution on a sphere or annulus. Consider the parametric family $\Pb_\theta=(\Gmap_\theta)_\# \operatorname{Unif}(\mathbb{S}^{d-1})$, where $\Gmap_\theta: z \mapsto \theta z$ for $\theta\in\mathbb{R}$, and take $k$ to be a Gaussian kernel. Then the tangent-span condition holds; see \Cref{lem:example} for the proof. 
\end{example}

\begin{rem}[Further interpretation of \textcolor{black}{\Cref{ass:residual}}]
     \textcolor{black}{\Cref{ass:residual}} implies that whenever parametric descent becomes stationary, i.e., zero-gradient, the induced distribution $\Pb_\theta$ is also stationary for the objective \(\Pb\mapsto \mmd_\ell(\Pb,\Qb)\); see \Cref{lem:residual_excludes_spurious_minima} for a proof.
     This is a strong condition, but it is needed for our analysis. 
     The adaptive lengthscale scheme is designed to address the nonconvexity of the nonparametric MMD descent, as established in \Cref{thm:np_rate}. 
     However, it does not by itself remove additional nonconvexity introduced by the parametrization \(\theta\mapsto\Pb_\theta\). 
     This limitation is also apparent from the PGD update. 
     If the parametric objective reaches a spurious stationary point satisfying $\nabla_\theta \left[ \frac12 \mmd_\ell^2(\Pb_\theta,\Qb) \right] =0$, then preconditioning cannot create a descent direction, since the preconditioned gradient is zero as well.  
\end{rem}

\begin{prop}[Closeness of $\Pb_{\theta_t^\pgd}$ to $\Pb_t$] \label{prop:delta_theta_ast}
Let kernel $k$ satisfy \textcolor{black}{\Cref{ass:kernel}}.
Suppose \textcolor{black}{\Cref{ass:regularity_G,ass:residual}} hold.
Let $\Qb\in\calP_2(\R^d)$ and $\rho\in\calP_2(\calZ)$. 
Let both the nonparametric and parametric MMD descent schemes share the same initialization: $\Pb_0 = \Pb_{\theta_0}$.
Let the step size $\gamma_0 > 0$. 
Denote $\Pb_1$ as the next iterate under nonparametric update, and $\Pb^\pgd_{\theta_{1}}$ as the next iterate under our preconditioned parametric update. Then, taking $\lambda_0 \propto (\gamma_0 p^2 (1+\ell^{-1})  \ell_0^{-2})^{\frac{2}{5}}$, we have
\begin{align*}
    \mmd_{\ell}^2 (\Pb_{\theta^\pgd_{1}}, \Qb)
    -\mmd_{\ell}^2(\Pb_{1}, \Qb)
    &\leq 2\gamma_0 \mathfrak{R}
    \|\nabla f_{\ell,\Pb_0,\Qb}\|_{L_2^d(\Pb_0)}^2 \\
    &\quad + C_k\gamma_0 \mathfrak{R} \ell^{-4}(\ell_0^2 - \ell^2)
    + o(\gamma_0) .
\end{align*}
Here, $C_k$ is a constant that depends only on the kernel profile.
\end{prop}
The proof is given in \Cref{sec:proof_delta_theta}.
\Cref{prop:delta_theta_ast} gives a \emph{one-step} comparison between the PGD update and nonparametric MMD descent, measured by their squared MMD distance to $\Qb$.
The leading term captures the residual component of the nonparametric descent that \emph{cannot} be represented, to first order, by any parametric update on the parameter $\theta$.
The second term comes from the mismatch induced by the adaptive kernel lengthscale: the MMD objective is evaluated at $\ell$, whereas the current iterate uses lengthscale $\ell_0$.
The residual term $o(\gamma_0)$ vanishes faster than linear; see Eq.~\eqref{eq:precise_proposition} for a more precise characterization.
In contrast, the standard gradient descent scheme does not
admit the same first-order comparison with the nonparametric MMD descent 

\begin{rem}[Tradeoff with $\lambda_0$]\label{rem:tradeoff}
A smaller $\lambda_0$ yields a smaller projection-approximation error, of order $\gamma_0 \sqrt{\lambda_0}\,\sqrt{d}\,\ell^{-1}$, relative to the exact projection.
By contrast, a larger $\lambda_0$ reduces the error, of order $\lambda_0^{-2}\gamma_0^2 \sqrt{d}\,\ell^{-2}\ell_0^{-2}$, arising from the second-order Taylor expansion.
The choice of $\lambda_0$ in \Cref{prop:delta_theta_ast} optimally balances these two terms.
A precise characterization is given in Eq.~\eqref{eq:tradeoff}.
\end{rem}

\begin{rem}[Connection to natural gradient]\label{rem:connection_ngd}
Natural gradient descent is another preconditioned gradient method requiring fewer iterations than standard gradient descent to reach convergence~\citep{amari1998natural,hoffman2013stochastic,pascanu2014revisiting,zhang2019fast,martens2020new}.
It can be interpreted as the steepest descent direction when the parameter space is endowed with the local geometry induced by the KL divergence~\citep{amari2000methods}.
For the MMD objective, a natural analogue is to endow the parameter space with the local geometry induced by the squared MMD distance~\citep[Section 2.3]{briol2019statistical}.
This leads to the preconditioning matrix (cf. \Cref{sec:ngd})
\begin{talign}\label{eq:precondition_ngd}
    \begin{aligned}
    \Big\{ \E_{z,z'\sim\rho}\Big[
    &\bJ_\theta \Gmap_{\theta}(z)^\top
    \nabla_{1,2} k_{\ell}(\Gmap_{\theta}(z), \Gmap_{\theta}(z')) \bJ_\theta \Gmap_{\theta}(z')
    \Big] \Big\}^{-1}.
    \end{aligned}
\end{talign}
This matrix is different from our preconditioning matrix
\[
\left\{
\E_{z\sim\rho}\big[
\bJ_\theta \Gmap_{\theta_t}(z)^\top
\bJ_\theta \Gmap_{\theta_t}(z)
\big]
\right\}^{-1}
\]
in Eq.~\eqref{eq:parametric_mmd_update_new}, which arises instead from minimizing, after one step, the discrepancy between parametric and nonparametric descent.
\end{rem}

In practice, the target distribution $\Qb$ is only known with $m$ IID drawn samples. Denote $\widehat{\Qb}_m$ as the corresponding empirical distribution.
We therefore consider the PGD scheme in which the expectation with respect to $\Qb$ is approximated by an empirical average with $\widehat{\Qb}_m$.
Let $(\tildePtheta{t})_{t\in\N}$ be the resulting sequence of parametric distributions, defined by:  $\theta_{t+1} = \theta_t - \gamma_t \tildetheta{t}$, where
\begin{align}\label{eq:parametric_mmd_update_new_tilde}
    \tildetheta{t}
    &=
    \left\{ \E_{z\sim\rho}\left[
    \bJ_\theta \Gmap_{\theta_t}(z)^\top
    \bJ_\theta \Gmap_{\theta_t}(z)\right] + \lambda_t \Id \right\}^{-1}
    \nonumber\\
    &\quad \times
    \E_{z\sim\rho}\left[
    \bJ_\theta \Gmap_{\theta_t}(z)^\top
    \nabla f_{\ell_t, \Pb_{\theta_t}, \widehat{\Qb}_m}(\Gmap_{\theta_t}(z))
    \right] .
\end{align}


\begin{theorem}[Global convergence of PGD scheme]\label{thm:p_rate}
Let kernel $k$ satisfy \textcolor{black}{\Cref{ass:kernel}}.
Suppose \textcolor{black}{\Cref{ass:regularity_G,ass:residual}} hold.
Suppose $\Qb\in\calP_2(\R^d)$.
Let $C>0$ be a constant independent of $t$.
Let $(\gamma_t)_{t\in\N}$ be a monotonically decreasing sequence of step sizes such that $\gamma_t\in(0,\frac{1}{2C(1-\mathfrak{R})})$, $\lim_{t\to\infty} \gamma_t=0$, and $\sum_{t=0}^\infty \gamma_t=\infty$.
Let $(\ell_t)_{t\in\N}$ be a monotonically decreasing sequence of lengthscales satisfying $\lim_{t\to\infty} \ell_t = \ell > 0$.
Take $\lambda_t \propto (\gamma_t p^2 (1 + \ell^{-1}) \ell_t^{-2})^{\frac{2}{5}}$.
For a fixed horizon $T \in \mathbb{N}$, let $m=m(T) \in \mathbb{N}$ be the empirical target sample size, chosen such that $\sqrt{m(T)} \geq (\sum_{s=0}^T \gamma_s^{0.6}) \cdot \log(T+1)$.
\begin{enumerate}[leftmargin=6mm,labelsep=0.5em]
    \item[\textup{(i)}] \textbf{Well-specified case.} Suppose that, for all \(t\in[T]\),
    \[
    \ell_t^d \ell^{-d}
    \|\nabla f_{\ell, \tildePtheta{t},\Qb} \|_{L_2^d(\tildePtheta{t})}^2
    \geq C \mmd_{\ell}^2(\tildePtheta{t}, \Qb).
    \]
    Then, there exists a remainder $r_T$ such that $\E_{m} \left[\mmd^2_{\ell}(\tildePtheta{T}, \Qb)\right] \leq r_T$, and $r_T\rightarrow 0$ as $T\to\infty$. 

    \item[\textup{(ii)}] \textbf{Mis-specified case.} Suppose that, for all \(t\in[T]\),
    \[
    \begin{aligned}
    &\ell_t^d \ell^{-d}
    \|\nabla f_{\ell, \tildePtheta{t},\Qb} \|_{L_2^d(\tildePtheta{t})}^2 \geq
    C \Big(\mmd_{\ell}^2(\tildePtheta{t}, \Qb)
    - \min_{\theta\in\Theta} \mmd_{\ell}^2(\Pb_{\theta}, \Qb) \Big).
    \end{aligned}
    \]
    Then, there exists a remainder $r_T$ such that
    \[
    \E_{m} [\mmd^2_{\ell}(\tildePtheta{T}, \Qb)]
    - \min_{\theta\in\Theta} \mmd^2_{\ell}(\Pb_{\theta}, \Qb)
    \leq r_T,
    \]
    and $r_T\rightarrow 0$ as $T\to\infty$. 
\end{enumerate}
\end{theorem}

The proof is given in \Cref{sec:proof_p_descent}.
The conditions on the step size $\gamma_t$ and lengthscales $\ell_t$ are analogous to the nonparametric MMD descent in \Cref{thm:np_rate}.
As the regularization parameter $\lambda_t$ depends on unknown quantities such as $\calU$, in practice it is selected by a search over a pre-specified candidate set.
The degenerate case $\mathfrak{R}=1$ is likewise manifested in the admissible range of step sizes.
The expectation $\E_m$ is taken over the empirical target sample $\widehat{\Qb}_m$ used in the PGD update. 
The condition on the target sample size $m$ ensures that the statistical error from replacing $\Qb$ by its empirical approximation
\(\widehat{\Qb}_{m}\) remains negligible compared with the descent achieved by the PGD updates.
In the case when $\gamma_t\asymp t^{-1}$, this condition becomes $m(T)\gtrsim T^{4 / 5}(\log T)^2$. 

\Cref{thm:p_rate} gives two convergence guarantees in expectation over the empirical target sample. 
In the well-specified case, the model class is rich enough to represent the target distribution in MMD, and the bound shows that our PGD drives terminal value
$\mmd_{\ell}^2(\tildePtheta{T},\Qb)$ tending to $0$ in expectation. 
In the misspecified case, however, exact recovery of $\Qb$ is not possible within the model class; the appropriate benchmark is therefore the best attainable MMD value,
$\min_{\theta\in\Theta} \mmd_{\ell}^2(\Pb_{\theta},\Qb)$.
The theorem above shows that our PGD drives the terminal PGD objective towards this oracle value under a different gradient-dominance condition;
this condition is harder to verify than its well-specified counterpart, since it involves the unknown best approximation within the model class.

A related result is Proposition 5.2 of \cite{cherief2022finite}, which establishes convergence under \emph{convexity} of the map $\theta \mapsto \mmd^2(\Pb_\theta, \Qb)$, a condition rarely satisfied in practice; see \Cref{fig:illustration}.
By contrast, our result does \emph{not} rely on convexity of the MMD objective. Instead, it uses two key ingredients: 1) adaptive choice of kernel lengthscale, and 2) a close approximation of nonparametric MMD descent.
Our expected gradient-dominance condition is nevertheless compatible with the classical convex setting.
In particular, if $\theta \mapsto \mmd^2(\Pb_\theta,\Qb)$ is $\alpha$-strongly convex and \textcolor{black}{\Cref{ass:regularity_G}} holds, then our expected gradient-dominance condition is satisfied with $C=(2p)^{-1}\alpha$; see \Cref{lem:strong_convex}.
To the best of our knowledge, although our result is only \emph{asymptotic} and relies on the stated gradient-dominance and projection-residual assumptions, it is among the first global convergence guarantees for finding minimum MMD estimators without assuming convexity of the MMD objective.
Extension to a non-asymptotic convergence rate, with respect to both $m$ and $t$, would likely require stronger conditions on $\gamma_t$, $\ell_t$, and $\mathfrak{R}$.

Below we outline a quick sketch of the proof of \Cref{thm:p_rate}. 
The key step is to prove the following \emph{descent lemma}: for any $t\in\N$,
\begin{align}\label{eq:descent_lemma_p_main}
    &\quad \E_m\left[
    \mmd_{\ell}^2 (\tildePtheta{t+1}, \Qb)
    - \mmd_{\ell}^2 (\tildePtheta{t}, \Qb)
    \right]
    \nonumber \\
    &\leq -2(1-\mathfrak{R}) \gamma_t 
    \|\nabla f_{\ell, \tildePtheta{t}, \Qb}\|_{L_2^d(\tildePtheta{t})}^2
    + o(\gamma_t)
\end{align}
To establish this, we define two auxiliary distributions: 1) $\widehat{\Pb}_{t+1}:=(\Id - \gamma_t \nabla f_{\ell_t, \tildePtheta{t},\Qb})_\# \tildePtheta{t}$, which corresponds to the next distribution under the \emph{nonparametric} MMD update; and 2)  $\Pb_{\theta_{t+1}}^\pgd:=\Pb_{\theta_t -\gamma_t \Delta \theta_t^\pgd}$ which corresponds to the next distribution under the parametric update scheme with the \emph{true} target distribution $\Qb$ in Eq.~\eqref{eq:parametric_mmd_update_new_tilde}.
The proof of Eq.~\eqref{eq:descent_lemma_p_main} is then a combination of the following inequalities.
\begin{align*}
    \mmd_{\ell}^2 (\widehat{\Pb}_{t+1}, \Qb)
    -\mmd_{\ell}^2(\tildePtheta{t}, \Qb)
    &\leq -2\gamma_t \ell_t^d \ell^{-d}
    \|\nabla f_{\ell, \tildePtheta{t}, \Qb}\|_{L_2^d(\tildePtheta{t})}^2
    + o(\gamma_t) \\
    \mmd_{\ell}^2 (\Pb_{\theta_{t+1}}^\pgd, \Qb)
    -\mmd_{\ell}^2(\widehat{\Pb}_{t+1}, \Qb)
    &\leq 2\gamma_t \mathfrak{R}
    \|\nabla f_{\ell, \tildePtheta{t}, \Qb}\|_{L_2^d(\tildePtheta{t})}^2
    + o(\gamma_t) \\
    \E_m\left[ \mmd_{\ell}^2 (\tildePtheta{t+1}, \Qb)\right]
    &\leq \mmd_{\ell}^2(\Pb_{\theta_{t+1}}^\pgd, \Qb) \\
    &\quad + C_k \gamma \lambda^{-1} p \ell^{-1}\ell_t^{-1}
    \sqrt{d} \cdot \frac{1}{\sqrt{m}} .
\end{align*}
The first inequality is proved in \Cref{prop:descent_np} and controls the decrease of squared MMD under the nonparametric MMD descent after one step. The second inequality is proved in \Cref{prop:delta_theta_ast} and controls the difference of squared MMD between two schemes after one step.
The third inequality is proved in \Cref{prop:tilde_no_tilde}, and bounds the finite-sample error from using $\widehat{\Qb}_m$ instead of the unknown true distribution $\Qb$. 
$C_k$ is a constant that depends only on the kernel profile.
Combining the three inequalities yields Eq.~\eqref{eq:descent_lemma_p_main}.
The proof of \Cref{thm:p_rate} is concluded with a discrete Gronwall's lemma.

\paragraph{Computational complexity}
We now turn to how to compute our PGD update $\tildetheta{t}$ in Eq.~\eqref{eq:parametric_mmd_update_new_tilde} in practice.
Given $n$ i.i.d samples $\{z_i\}_{i=1}^n$ from $\rho$, which may either be fixed throughout the algorithm or resampled at each iteration, and $m$ fixed i.i.d samples $\{y_i\}_{i=1}^m$ from $\Qb$, we can approximate $\tildetheta{t}$ with $\widehattheta{t}:= ( \widehat{H_{t}} + \lambda_t \Id)^{-1} \widehat{g_t}$, where
\begin{align*}
    \widehat{H_t}
    &= \frac{1}{n} \sum_{i=1}^n
    \bJ_\theta \Gmap_{\theta_t}(z_i)^\top
    \bJ_\theta \Gmap_{\theta_t}(z_i), \\
    \widehat{g}_t
    &= \nabla_{\theta_t} \left\{
    \frac{\sum_{i,j=1}^n k(\Gmap_{\theta_t}(z_i), \Gmap_{\theta_t}(z_j))}{n^2}
    - \frac{2 \sum_{i,j=1}^{n,m} k(\Gmap_{\theta_t}(z_i), y_j)}{nm}
    \right\} .
\end{align*}

Relative to Eq.~\eqref{eq:parametric_mmd_update_new_tilde}, this implementation additionally replaces the expectation with respect to $\rho$ by an empirical average over samples $\{z_i\}_{i=1}^n$. 
Here, $\widehat{g}_t$ before taking the derivative is the finite-sample V-statistic estimator of $\mmd_\ell^2(\Pb_\theta,\widehat{\Qb})$. We use the V-statistic rather than the U-statistic because the former is always nonnegative~\citep{gretton2012kernel,huang2023high}.
The derivative in $\widehat g_t$ can be computed by automatic differentiation, e.g., \cite{jax2018github}.
If $\widehat H_t$ is formed explicitly, its cost is $\calO(n \min(d,p) +ndp^2)$, since each Jacobian $\bJ_\theta \Gmap_{\theta_t}(z_i)\in\R^{d\times p}$ must be evaluated and multiplied to form a $p\times p$ matrix. The matrix inversion $(\widehat{H_{t}} + \lambda_t \Id)^{-1}$ incurs an additional $\calO(p^3)$ cost.
The cost of computing the preconditioning matrix is the same as that of Fisher information matrix in natural gradient descent~\citep{martens2020new}, and is cheaper than the preconditioner in Eq.~\eqref{eq:precondition_ngd} of \cite{briol2019statistical}, which involves a double expectation.
In practice, this extra cost is worthwhile as it gives a more faithful update direction that leads to global convergence, as proved in \Cref{thm:p_rate} and verified empirically next in \Cref{sec:mmdal_experiments}.
The cost can be reduced by computing both $\widehat g_t$ and $\widehat{H_{t}}$ via mini-batches, although this introduces additional biases.
Moreover, when $p$ is large, the cost of explicitly forming $\widehat H_t$ can be avoided by using Jacobian-vector products. Since the experiments in \Cref{sec:mmdal_experiments} are relatively small-scale, we leave these implementations to future work.

Our convergence result in \Cref{thm:p_rate} accounts only for the finite-sample error arising from replacing $\Qb$ by $\widehat{\Qb}_m$, and does not explicitly analyze the additional approximation error induced by using finitely many samples $\{z_i\}_{i=1}^n$ from $\rho$.
This additional error can be controlled by standard arguments of biased gradient descent~\citep{devolder2014first,ajalloeian2020convergence}, aided by the regularization parameter $\lambda_t$, which ensures stability of the empirical inverse.
We omit these details to streamline the presentation.


\begin{rem}[Population and empirical minimum MMD estimators]
Throughout the paper, our PGD scheme targets the population minimum MMD estimator $\theta^\ast \in \arg\min_{\theta\in\Theta}  \mmd^2(\Pb_\theta,\Qb)$.
This differs from the more common empirical minimum MMD estimator $\theta^\ast_m \in \arg\min_{\theta\in\Theta}  \mmd^2(\Pb_\theta,\widehat{\Qb}_m)$ with $\widehat{\Qb}_m := \frac{1}{m}\sum_{i=1}^m \delta_{y_i}$, where $y_1,\ldots,y_m$ are IID samples from $\Qb$; see, for example,
\cite[Section~2.2]{briol2019statistical} and
\cite[Section~2.2]{cherief2022finite}.
We focus on the population objective, because, for an empirical measure
$\widehat{\Qb}_m$, the well-specified setting is typically unachievable: in general,
$\widehat{\Qb}_m \notin \{\Pb_\theta:\theta\in\Theta\}$.
Consequently, the final estimator obtained from our PGD scheme should be viewed as a numerical approximation to the population estimator $\theta^\ast$ rather than $\theta^\ast_m$.
Nevertheless, both \cite{briol2019statistical,cherief2020mmd} also target the population quantity since $\theta^\ast_m$ approximates $\theta^\ast$ through the empirical measure $\widehat{\Qb}_m$ (see Proposition 1 of \cite{briol2019statistical} and Theorem 3.2 of \cite{cherief2025parametric}), whereas our implementation approximates $\theta^\ast$ through a finite-sample implementation of the PGD scheme.
\end{rem}

\vspace{1mm}
\textbf{A practical guide to choosing $\ell_t$, $\lambda_t$, and $\gamma_t$.}
The lengthscale schedule $(\ell_t)_{t\in\mathbb{N}}$ is the most important hyperparameter in PGD, with the limiting value $\ell$ chosen based on statistical considerations~\citep{briol2019statistical,cherief2020mmd}.
We use an exponentially decaying schedule of the form
$\ell_t=\max\{\ell, 100\ell\cdot \eta^t\}$,
for some $\eta\in[0.99,1)$ and find that it performs very well in our experiments.
For the step size $\gamma_t$, we use a polynomial decay $\gamma_t = \gamma_0 \cdot (1+t)^{-\beta}$ with $\beta\in(0,1)$ such that it satisfies the conditions in \Cref{thm:np_rate,thm:p_rate}.
For the regularization parameter $\lambda_t$, our theory suggests an adaptive schedule, while in the experiments we find fixed choices to be reasonably stable and use them to reduce tuning cost, in a similar spirit to regularization in 
kernel ridge regression~\citep{fischer2020sobolev,shen2025nonparametric}. 
Compared to standard gradient descent, our method introduces two additional parameters: $\ell_t$ and $\lambda_t$. In our experiments, the performance of PGD is relatively insensitive to $\lambda_t$ and requires only modest tuning of the step size $\gamma_t$. 

\begin{figure}[t]
    \centering
    \includegraphics[width=1.0\linewidth]{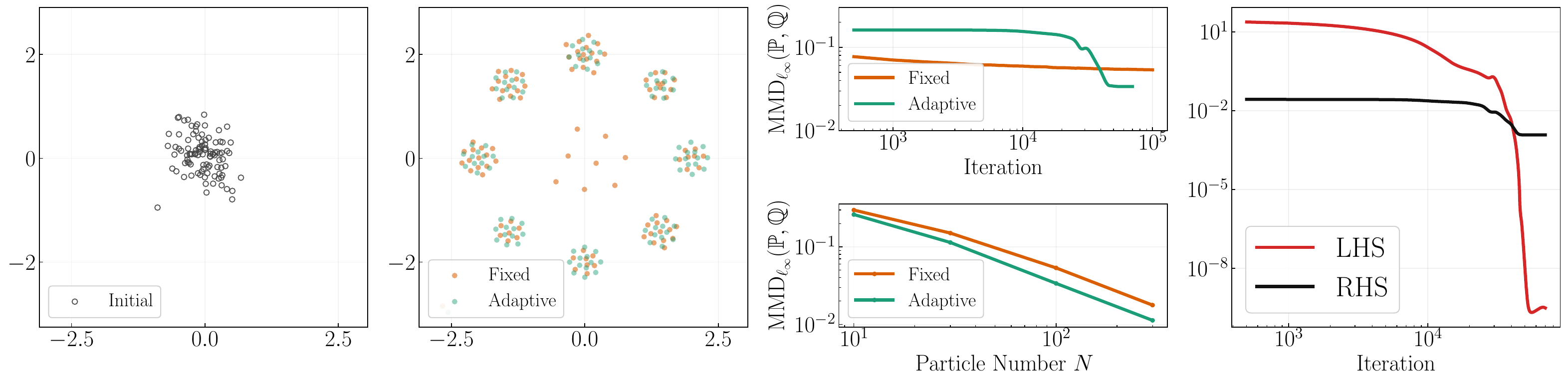}
    \vspace{-20pt}
    \caption[Mixture of Gaussians MMD descent]{\textit{Mixture of Gaussians.} From left to right: first and final locations of $100$ particles; MMD versus iterations;
    MMD versus particle number $N$; the left- and right-hand sides of the condition in \Cref{thm:np_rate}. Results are over 10 seeds; curves show medians and bands the 25--75\% percentiles. }
    \label{fig:mmd_flow}
    \vspace{-15pt}
\end{figure}

\section{Experiments}\label{sec:mmdal_experiments}
In this section, we provide empirical evidence to illustrate the effectiveness of our \emph{preconditioned} gradient descent (PGD) scheme. 
For comparison, we also consider standard gradient descent (\textsc{GD}) and \textsc{NGD}~\citep{briol2019statistical}, which uses Eq.~\eqref{eq:precondition_ngd} as the preconditioner.
\textcolor{black}{Our PGD preconditioner is computationally cheaper to compute than the NGD preconditioner, but both methods are more expensive per iteration than GD.}
\textcolor{black}{All three methods are run until observed convergence, when the difference between two consecutive iterations falling below $10^{-4}$.}
Code for reproducing all experiments is available at \url{https://github.com/seulkang0518/MDF_AL}. 

\vspace{1mm} 
\textbf{Nonparametric MMD descent.} 
We first consider a synthetic experiment where the target distribution is a mixture of 8 two-dimensional $(d=2)$ Gaussian distributions \citep{chen2025stationary}. 
This simple multi-modal setting is a good benchmark to evaluate the ability of sampling methods to capture all modes of a distribution. 
This experiment is intended to show that adaptive kernel lengthscales can improve convergence for the nonparametric MMD descent scheme, supporting the qualitative message of \Cref{thm:np_rate}. 
We use a particle implementation of nonparametric MMD descent~\citep{arbel2019maximum,chen2025stationary}. 
Let $\{y_i\}_{i=1}^m$ be $m$ i.i.d.\ samples from $\Qb$, and let $\{x_i^{(0)}\}_{i=1}^N$ be $N$ particles drawn IID from an initial distribution $\calN(0, 1)$. 
Then, at iteration $t\in\N$, the particles are updated according to
\begin{equation*}
    x_i^{(t+1)} = x_i^{(t)} - \gamma_t \nabla f_{\ell_t,\hat{\Pb}_{N,t},\widehat{\Qb}_m}\bigl(x_i^{(t)}\bigr), 
    \qquad \forall i\in[N],
\end{equation*}
where $\widehat{\Qb}_m=\frac{1}{m}\sum_{i=1}^m \delta_{y_i}$ and $\hat{\Pb}_{N,t}=\frac{1}{N} \sum_{i=1}^N \delta_{x_i^{(t)}}$.  
We compare two versions of the method: one with a fixed kernel lengthscale, $\ell_t=\ell=0.1$, and one with an adaptive kernel lengthscale $\ell_t=\max\{\ell, 10.0 \times 0.9999^t\}$. 
The decay factor $0.9999$ was selected by tuning over the prespecified grid $\{0.99,0.995,0.999,0.9995,0.9999\}$.
Both use an adaptive step size $\gamma_t = 0.01 / (1+t)^{0.1}$ which satisfies the step size conditions in \Cref{thm:np_rate}. 
The two plots in the third panel of \Cref{fig:mmd_flow} show that the adaptive scheme attains a smaller final $\mmd_{\ell}(\Pb_t,\Qb)$. 
Interestingly, the advantage of adaptive MMD descent over fixed-lengthscale MMD descent becomes apparent only once $\ell_t$ starts to approach its limiting value $\ell$. 
Although the initial decrease is slower because the algorithm minimizes a different MMD objective, the MMD decreases more rapidly as $\ell_t$ approaches $\ell$, avoiding the local traps observed in the fixed-lengthscale scheme; see the orange particles near the center of the second panel.
In the fourth panel of \Cref{fig:mmd_flow}, we also plot the left- and right-hand sides of the key condition in \Cref{thm:np_rate}, with constant $C=0.01$. 
We observe that this condition is satisfied throughout most of the dynamics until the final stage of the evolution. 

\begin{figure}
    \centering
    \includegraphics[width=1.0\linewidth]{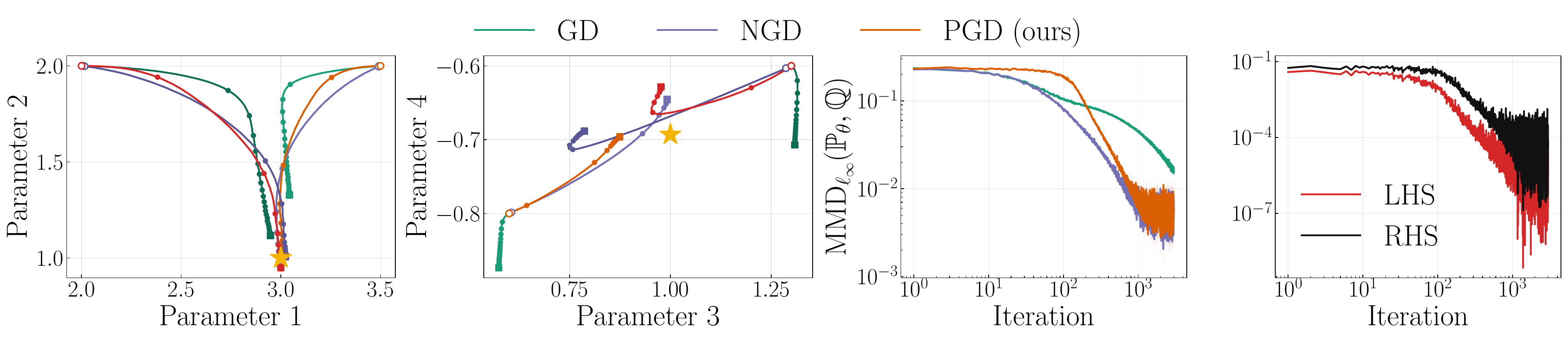}
    \vspace{-19pt}
    \caption[G-and-k distribution experiment]{\textit{G-and-k distribution.} 
    From left to right: two optimization trajectories toward the global optimum (yellow star), MMD versus iteration, and left- and right-hand sides of the condition in \Cref{thm:p_rate}. 
    In the left two plots, two trajectories for each method are shown. 
    Results are over 10 seeds; curves show medians and bands the 25--75\% percentiles. } 
    \label{fig:g_and_k}
\end{figure}

\vspace{1mm}
\textbf{G-and-k distribution.} 
We next evaluate our PGD scheme on the g-and-k distribution, a commonly used benchmark for minimum distance estimators~\citep{briol2019statistical,dellaporta2022robust}. 
This distribution has widespread utility, including in the pricing of AirBnB rentals~\citep{rodrigues2020likelihood}, air pollution modelling~\citep{rayner2002numerical} and non-life
insurance~\citep{peters2016estimating}. Despite being a univariate model, it does not admit a closed-form likelihood: specifically, the statistical model is given by $(\Gmap_\theta)_{\#} \rho$, with $\rho$ being a uniform distribution over $[0,1]$, and $\Gmap_\theta$ the standard g-and-k quantile map. 
The parameter of interest is $\theta=(a,b,c,\exp(k))$, whose four components represent the location, scale, skewness and tail-heaviness of the model distribution.
For this model, $\Gmap_\theta$ and its first- and second-order parameter derivatives have finite second moments for each $\theta$, and \textcolor{black}{\Cref{ass:regularity_G}} holds.

For this experiment, we draw $m=1000$ IID samples from the target distribution $\Qb=(\Gmap_{\theta^\ast})_{\#}\rho$ with $\theta^\ast=(3,1,1,\exp(-\log 2))$ following the setting of \cite{briol2019statistical}. 
This corresponds to a \emph{well-specified} setting. For GD and NGD, we use fixed step sizes $\gamma=0.1$; for our PGD, we use $\gamma_t = 0.1 / (1+t)^{0.1}$ which satisfies the step size condition in \Cref{thm:p_rate}. 
GD and NGD use a fixed lengthscale, $\ell_t=\ell=2$, whereas PGD uses an adaptive lengthscale, $\ell_t=\max\{\ell, 10.0 \times 0.99^t\}$. 
The decay factor $0.99$ was selected by tuning over the prespecified grid $\{0.99,0.995,0.999,0.9995,0.9999\}$. 
As is shown in \Cref{fig:g_and_k}, both PGD and NGD outperform standard GD. In \Cref{fig:g_and_k}, both PGD and NGD perform better than standard GD over the reported optimization horizon. In the first panel, both preconditioned methods reach the optimum in fewer iterations; and in the second panel, standard GD remains far from the global minimum over the optimization horizon, especially in the more challenging third and fourth parameters. 
These results suggest that preconditioning not only accelerates convergence, but also helps overcome the non-convexity of the MMD objective. 
NGD performs slightly better than our PGD scheme, albeit at a larger cost of computing the preconditioner due to the double expectation. 
More empirical studies are thus needed to better understand which preconditioner is preferable in practice. 

\begin{figure}[t]
\vspace{-10pt}
    \centering
    \includegraphics[width=1.0\linewidth]{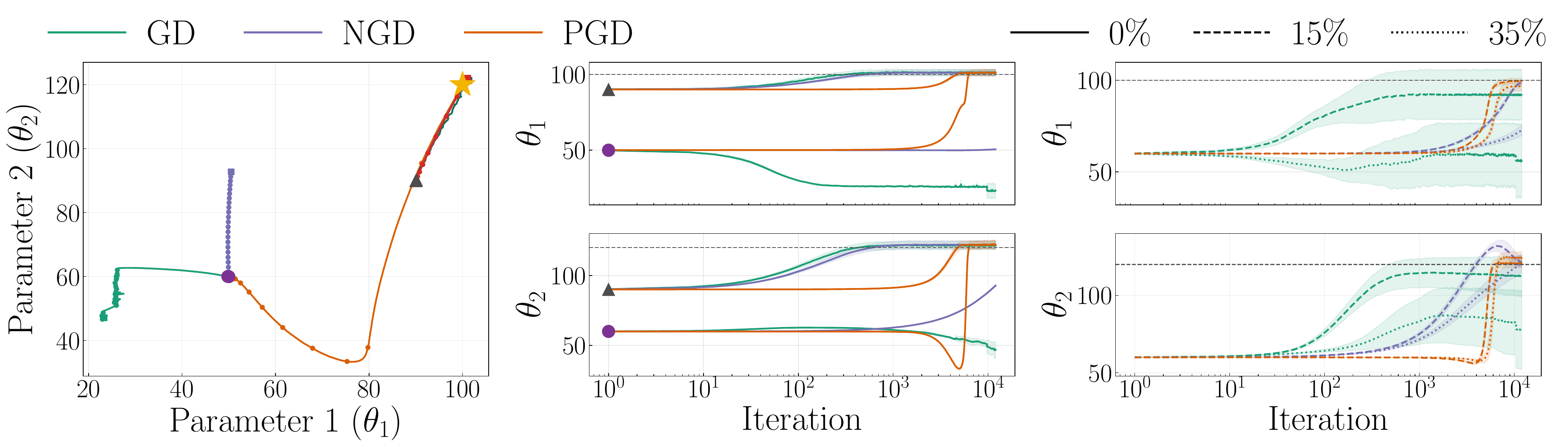}
    \vspace{-15pt}
    \caption[Lotka-Volterra model experiment]{\textit{Lotka-Volterra model.} 
    \textbf{Left:} Two optimization trajectories toward the global optimum (yellow star) under two initializations. 
    \textbf{Middle:} Evolution of $\theta_1, \theta_2$ under different initializations.
    \textbf{Right:} Evolution of $\theta_1, \theta_2$ under different corruption levels. Results are over 10 seeds; curves show medians and bands the 25--75\% percentiles.} 
    \vspace{-10pt}
    \label{fig:lv}
\end{figure}

\vspace{1mm} 
\textbf{Inference for Lotka-Volterra model under corruptions.} 
We next consider an example where the model $\Pb_\theta$ is the output of a stochastic Lotka-Volterra model which describes the evolution of two species $(X_{1, t}, X_{2, t})$ through time via a pair of stochastic differential equations~\citep{lotka1927fluctuations,shen2025prediction,chen2026thinned}. 
This is a prototypical model
in population ecology.  
Following \cite{briol2019statistical}, the parameter of interest $\theta=(X_{1,0},X_{2,0})$ corresponds to the initial conditions.
For this simulator, we are not able to verify the uniform first- and second-order derivative bounds in \textcolor{black}{\Cref{ass:regularity_G}}; the square-root coefficients in the Euler--Maruyama recursion make such a verification delicate near the boundary of the positive orthant.

We study robustness to data corruption, a setting where minimum MMD estimators are particularly relevant. Specifically, following the setting of \cite{briol2019statistical}, we consider inferring the true initial condition $\theta^\ast=(100,120)$ from observations in which $\mathfrak{p}\%$ of the data is corrupted by samples generated from $\theta^\dagger=(50,50)$. 
For GD and NGD, we use fixed step sizes $\gamma=100$; for our PGD, we use $\gamma_t = 300 / (1+t)^{0.1}$ which satisfies the step size condition in \Cref{thm:p_rate}. 
GD uses a fixed lengthscale, $\ell_t=\ell=30$, whereas PGD uses an adaptive lengthscale, $\ell_t=\max\{30, 3000 \times 0.9995^t\}$. 
The decay factor $0.9995$ was selected by tuning over the prespecified grid $\{0.95, 0.99,0.995,0.999,0.9995,0.9999\}$. 

Our results are shown in \Cref{fig:lv}. In particular, the left and middle panels show that, when initialized at $\theta_0=(90,90)$, close to the global minimizer, all methods converge to the correct solution. In contrast, under a less favorable initialization at $\theta_0=(50,60)$, our PGD still recovers the global minimizer, whereas GD and NGD fail within the given number of iterations. 
The last panel of \Cref{fig:lv} compares the methods across different corruption levels, $\mathfrak{p}\in\{0,15,35\}$. As $\mathfrak{p}$ increases, recovering the ground truth becomes more challenging, since a larger fraction of the data is generated from the corrupted distribution rather than the true distribution. Consequently, GD becomes increasingly unreliable in identifying the ground truth. By contrast, our PGD scheme consistently recovers the global minimizer across all corruption levels. 
These results suggest that PGD exhibits superior convergence behavior, even in mis-specified settings. 
\begin{table}
\centering
\renewcommand{\arraystretch}{1.15}
\begin{tabular}{lcc|cc}
\toprule
Method & Type-I error ($\downarrow$) & Time & Power ($\uparrow$) & Time \\
\midrule
PGD ($\calI=1$) & 0.050 $\pm$ 0.018 & 8.6s & 0.829 $\pm$ 0.032 & 3.6s \\
\midrule
GD ($\calI=400$) & 0.107 $\pm$ 0.026 & 10.4s & 0.600 $\pm$ 0.041 & 4.9s \\
GD ($\calI=1$) & 0.071 $\pm$ 0.022 & 2.6s & 0.279 $\pm$ 0.038 & 1.1s \\
\bottomrule
\end{tabular}
\caption[Composite goodness-of-fit test]{\textit{Composite goodness-of-fit test.} Critical values are computed by parametric bootstrap with $B=50$ bootstrap samples. The nominal Type-I error is $\alpha=0.05$; lower Type-I error and higher power indicate better performance. All experiments are repeated over $140$ random seeds; results report the mean $\pm$ standard error.}
\label{tab:toggle_switch_gof}
\vspace{-0.5em}
\end{table}

\vspace{1mm} 
\textbf{Composite goodness-of-fit test.}
Finally, we consider a toggle-switch model: a stochastic dynamical model for interacting gene-expression levels over a fixed time horizon $\calT\in\N$~\citep{bonassi2011toggleswitch}. The model describes the coupled temporal evolution of two genes through a dynamical system parametrized by $\theta$. It poses a challenging inference problem, since each parameter evaluation requires simulating a discretized stochastic dynamical system, and the resulting trajectories can be highly sensitive to the parameter values. 
We denote by $\Pb_{\theta;\calT}$ the distribution of the final observations generated by this simulator. 
The explicit dependence on the time horizon $\calT$ is reflected in the notation. 
In this setting, $\Pb_{\theta;\calT} = (\Gmap_{\theta;\calT})_{\#} \rho$; for a fixed $\calT$, the pushforward map $\Gmap_{\theta;\calT}$ satisfies the requirements in \textcolor{black}{\Cref{ass:regularity_G}}.

We evaluate a composite goodness-of-fit test for the model class
$\{\Pb_{\theta;\calT}:\theta\in\Theta\}$, with null hypothesis
$H_0:\Qb\in\{\Pb_{\theta;\calT}:\theta\in\Theta\}$.
Following~\cite{key2025composite}, we use the test statistic
$\Delta_m := m \min_{\theta\in\Theta} \mmd_{\ell}^2(\Pb_{\theta;\calT},\widehat{\Qb}_m)$, 
where $\widehat{\Qb}_m$ denotes the empirical distribution of $m$ IID samples from $\Qb$.
Computing $\Delta_m$ is precisely a minimum-MMD estimation problem over the parameter space.
First, we assess calibration by testing under the null, where
$\Qb=\Pb_{\theta^\ast;50}$ and the model class is $\{\Pb_{\theta;50}:\theta\in\Theta\}$. 
This corresponds to the well-specified setting where $\Qb\in\{\Pb_{\theta;\calT}:\theta\in\Theta\}$ is in the model class. 
Next, we assess testing power under the
alternative, where $\Qb=\Pb_{\theta^\ast;20}$ and the model class is $\{\Pb_{\theta;300}:\theta\in\Theta\}$. 
This corresponds to the mis-specified setting where $\Qb\notin\{\Pb_{\theta;\calT}:\theta\in\Theta\}$ is not in the model class. 
Critical values are computed by parametric bootstrap with
$B=50$ bootstrap samples at a significance level $\alpha=0.05$. 
For GD, we use a fixed step size $\gamma=0.2$ and a fixed lengthscale $\ell=120$; for our PGD, we use an adaptive step size $\gamma_t = 0.2 / (1+t)^{0.1}$ which satisfies the step size condition in \Cref{thm:p_rate}, and an adaptive lengthscale, $\ell_t=\max\{120, 2000 \times 0.95^t\}$. The decay factor $0.95$ was selected by tuning over the prespecified grid $\{0.95, 0.99,0.995,0.999,0.9995,0.9999\}$.

In \Cref{tab:toggle_switch_gof}, we compare PGD and GD in terms of the Type-I errors for testing under the null, and powers for testing under the alternative. 
Following Appendix~D.3.3 of \cite{key2025composite}, GD is run with a multi-start scheme: $\calI=400$ initial parameter values are sampled, and the run attaining the smallest final MMD is retained. 
This multi-start procedure is designed specifically to mitigate the non-convexity of the optimization problem. 
Accordingly, GD with only a single initialization, $\calI=1$, performs poorly, as shown in \Cref{tab:toggle_switch_gof}. 
By contrast, PGD is run from a single initialization, $\calI=1$, and achieves well-calibrated Type-I error and higher power at lower computational cost in this experiment, despite the additional step of preconditioning.
These results provide evidence that our PGD scheme can be efficient for locating the minimum-MMD estimator in this benchmark. 
Since Initializations are run in parallel on GPU rather than sequentially, so increasing $I=1$ to $I=400$ in GD cost only 4x in time rather than 400x.

\begin{figure}
    \centering
    \includegraphics[width=0.75\linewidth]{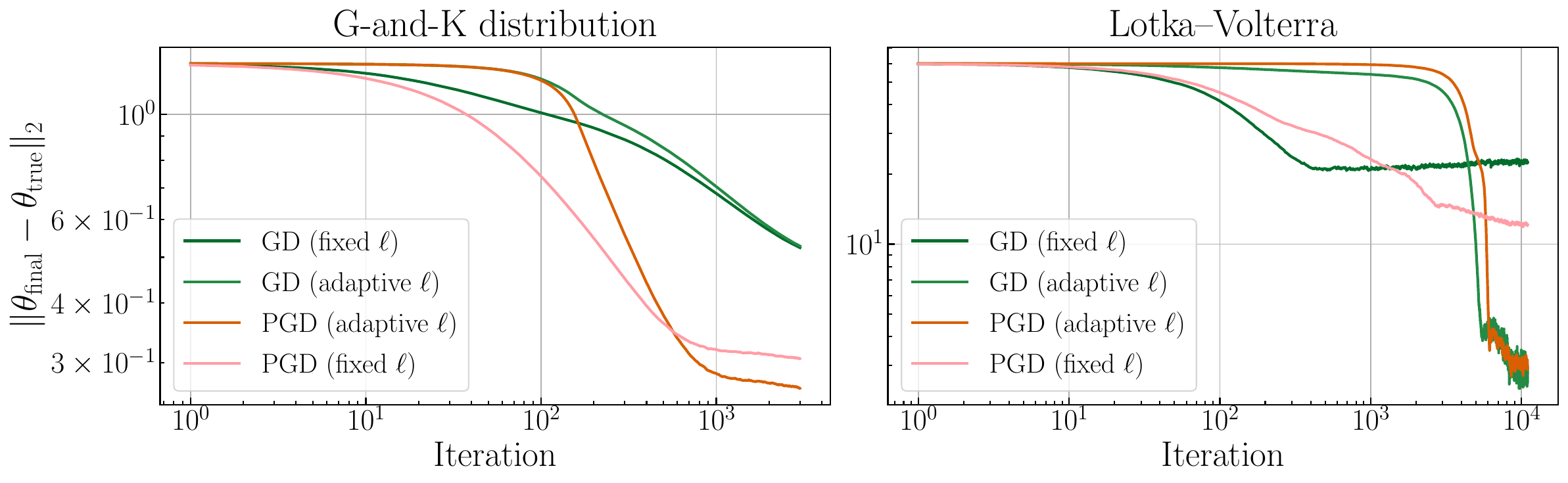}
    \vspace{-10pt}
    \caption[PGD ablation study]{\textit{Ablation study:} Standard GD with fixed/adaptive kernel lengthscales and our PGD with fixed/adaptive kernel lengthscales. \textbf{Left}: G-and-K distribution and \textbf{Right}: Lotka-Volterra model in well-specified setting with $\theta_0=(90,90)$.}
    \label{fig:cross_method}
    \vspace{-10pt}
\end{figure}

\vspace{1mm} 
\textbf{Ablation study.}
Our proposed PGD scheme introduces two main components: an adaptive lengthscale and a preconditioner. To assess their respective contributions, we conduct an ablation study in \Cref{fig:cross_method}, comparing GD and PGD under both fixed and adaptive lengthscales, respectively. 
The results suggest that adaptive lengthscale plays a more prominent role in improving convergence for the Lotka--Volterra model, whereas the preconditioner is the more important component for the g-and-k distribution. 
The relative importance of the two components may be problem-dependent; in practice, we therefore recommend using both.

\section{Conclusions}
We propose a new preconditioned gradient descent scheme for computing minimum MMD estimators, and establish an asymptotic convergence guarantee to a global minimizer under explicit non-convex gradient-dominance and projection-residual conditions.
A key limitation of our current analysis is that it relies on an \emph{a priori} condition on the adaptive choice of kernel lengthscales. Although such conditions are common in the analysis of kernel-based gradient flows~\citep{arbel2019maximum,chen2025stationary}, removing or relaxing this requirement would make our analysis more appealing. 
Another limitation is the projection-residual condition in \textcolor{black}{\Cref{ass:residual}}, which we posit is necessary, as it reflects the unavoidable error incurred when approximating the nonparametric update by a parametric update.

Looking ahead, it would be interesting to extend our techniques to the optimization of other popular minimum-distance estimation methods~\citep{basu2011statistical}, such as minimum Stein discrepancy estimation~\citep{barp2019minimum}, minimum Fisher divergence estimation~\citep{hyvarinen2005estimation}, minimum Wasserstein distance estimation~\citep{bassetti2006minimum,zhang2012minimum,lang2026minimum}, and many others. 
A main challenge is that, unlike the MMD, these discrepancies typically do not yield a simple computable finite-sample expression for the associated nonparametric Wasserstein gradient. Minimum kernel Stein discrepancy estimation~\citep{barp2019minimum} is the closest to our minimum MMD estimation; however, it requires access to the model score, whereas in our pushforward setting $\Pb_\theta = (\Gmap_\theta)_\# \rho$ is specified only through simulation. 
It remains unclear how to adapt our projection-based gradient method to this objective.

\chapter{Conclusions and Future Work}
\label{sec:conclusion}

This thesis has studied the minimization of squared maximum mean discrepancy (MMD) as a unifying framework for comparing, approximating, and sampling from probability distributions.
The central message is that MMD is not only a statistically meaningful discrepancy between probability measures, but also an optimization objective whose structure can be exploited to design practical algorithms for computational statistics and machine learning.
The preceding chapters of this thesis develop this perspective in three settings: numerical integration, sampling and generative modelling, and statistical inference.

\section{Future Work}

\subsection{Causal inference}
One promising future direction is the intersection between minimum MMD methods and causal inference.
This direction is a natural continuation of my recent work on causal inference~\citep{chen2024conformal,shen2025nonparametric,chen2025towards}. 
Causal problems are fundamentally distributional: interventions change the data-generating distribution, counterfactual questions compare distributions across parallel worlds.
This makes them a natural domain for MMD-based discrepancies, especially in settings where likelihoods are unavailable but samples from observational, simulated, or interventional distributions can be obtained.
An important future direction is to combine MMD-based distribution matching with doubly robust causal estimation, building on recent doubly robust methods for proxy causal learning and counterfactual policy mean embeddings~\citep{bozkurt2025density,zenati2025doubly}.

\subsection{Scalable generative models}
Another promising direction for future research is the application of MMD to large-scale generative models. 
MMD and the minimization of which has once been popular in the literature of generative adversarial networks (GANs), e.g., the MMD GANs~\citep{li2017mmd,arbel2018gradient,binkowski2018demystifying}. 
Although diffusion models have largely superseded GANs since 2020~\citep{song2020score,ho2020denoising}, recent work on one-step, high-quality generation—including drifting models~\citep{deng2026drifting,gretton2026wasserstein} and W-Flow~\citep{han2026one}—suggests that Wasserstein gradient flows provide a promising framework for
developing one-step generative models that achieve state-of-the-art performance on ImageNet256 dataset. Based on the striking similarity of their approaches and MMD gradient flows~\citep{arbel2019maximum}, I believe MMD flows have the potential to achieve state-of-the-art performances as well, provided with a good latent encoder.


\subsection{MMD with adaptive kernels and distributions}
Beyond the relatively direct extensions to causal inference and generative modelling, a more
fundamental question arises concerning the continued relevance of MMD:

\begin{quote}
\emph{Looking back from 2026, one might wonder: why should we still care about maximum mean
discrepancy (MMD)?}
\end{quote}

MMD is a kernel-based discrepancy introduced around 2010, before the widespread rise of modern
deep learning. At first glance, MMD
may therefore appear outdated, particularly when instantiated with a Gaussian kernel. Its fixed,
non-adaptive feature representation may seem poorly suited to high-dimensional structured data
such as images and text.

\emph{But who says that kernels must rely on fixed features?} Growing empirical evidence suggests
that, when equipped with features learned by neural networks, kernel methods can achieve
competitive performance in applications ranging from hypothesis testing~\citep{liu2020learning}, regression~\citep{galashov2025closed} to generative modelling~\citep{galashov2024deep}.
Indeed, a neural network can be viewed as learning an
adaptive feature representation, upon which a kernel-like notion of similarity is implicitly
constructed~\citep{jacot2018neural,takakura2024meanfield,ba2022highdimensional,suzuki2023feature}. 
From this broader perspective of MMD with adaptive kernels, many exciting open questions remain.
  How should kernel features be adapted to optimize downstream performance? Should the adaptation
  mechanism differ across regression, generative modelling, statistical inference, and causal
  inference? Although the preceding chapters have provided partial answers to some of these
  questions, a general theory of adaptive kernels for MMD remains largely undeveloped.

\emph{And who says that the distributions for which the MMD is being computed must themselves be also fixed?} 
As this thesis has demonstrated, minimizing MMD provides a versatile framework for designing
practical algorithms in computational statistics and machine learning~\citep{key2025composite,chen2025nested,chen2024conditional,kang2025bayessum,chen2025regularized,chen2025stationary,tian2026sobolev,chen2026thinned,kang2026gradient,chen2024conformal,shen2025nonparametric,chen2025towards}. 
The methods developed here consider settings in which one distribution remains fixed while the other evolves along a
trajectory toward the fixed reference. Many other areas remain to be explored, including causal inference, optimal transport, inverse problems, where allowing one of the two distributions to evolve may offer new algorithmic possibilities.

\addcontentsline{toc}{chapter}{Appendices}

\appendix
\chapter{\texorpdfstring{Supplementary Material for \Cref{sec:kernel_quadrature}}{Supplementary Material for Minimum MMD for Integration in Nested Expectations}}
\label{app:nkq}
\begingroup
\paragraph{Additional notations}
For two normed vector spaces $A, B$, $A \cong B$ means that $A$ and $B$ are norm equivalent, i.e. their sets coincide and the corresponding norms are equivalent. In other words, there are constants $c_1,c_2>0$ such that $c_1\|h\|_{A} \leq\|h\|_{B} \leq c_2\|h\|_{A}$ holds for all $h \in A$, written as $\| \cdot \|_A \cong \| \cdot \|_B$.
For $A \subseteq B$, $A$ is said to be continuously embedded in $B$ if the inclusion map between them is continuous, written as $A \hookrightarrow B$.
$\|T \|$  denotes the norm of an operator $T: A\to B$.
For a function $f:\calX \subseteq \R^d \to \R$ and $\alpha \in \mathbb{N}^d$, we use $\partial_x^\alpha f$ to denote the standard derivative $\partial x_1^{\alpha_1} \cdots \partial x_n^{\alpha_n} f$ and $D_x^\alpha f$ to denote the weak derivative.
For $f\in W_2^s(\calX)$, we use $\|f\|_{s,2} \coloneq \|f\|_{W_2^s(\calX)}$ to denote its Sobolev space norm.
$\lesssim$ means $\leq$ up to some positive multiplicative constants.

\section{Existing Results on Kernel Ridge Regression}

In this section, we present \Cref{prop:krr_bias} to \ref{prop:krr_all} which are adaptation of theorems from \citet{fischer2020sobolev} applied to Sobolev spaces. These propositions are foundations of the proof of \Cref{thm:main} in \Cref{sec:proof}.

In the standard regression setting, we are given $N$ observations $\{x_i, y_i\}_{i=1}^N$ which are i.i.d sampled from an unknown joint distribution $\mathbf{P}$ on $\calX \times \R$.
Here, $\calX \subset \R^d$ is a compact domain. 
The marginal distribution of $\mathbf{P}$ on $\calX$ is $\Qb$, and the conditional distribution $\mathbf{P}(\cdot \mid x)$ satisfies the Bernstein moment condition~\citep{fischer2020sobolev}. In other words, there exists constants $\sigma, L > 0$ independent of $x$ such that
\begin{align}\label{eq:mom}
    \int_{\mathbb{R}} \left|y -h^*(x)\right|^m \mathbf{P}(d y \mid x) \leq \frac{1}{2} m!\sigma^2 L^{m-2}
\end{align}
is satisfied for $\Qb$-almost all $x \in \calX$ and all $m \geq 2$. For example,  \eqref{eq:mom} is satisfied with $\sigma=L=\sigma_0$ when $\mathbf{P}(\cdot \mid x)$ is a Gaussian distribution with bounded variance $\sigma_0$. 
Additionally, \eqref{eq:mom} is also satisfied when there is no noise in the observation so $\sigma=L=0$, which will be discussed in \Cref{sec:noiseless_krr}.

In a regression problem, the target of interest is the Bayes predictor $h^\ast : \calX \to \R$, $x \mapsto \E[Y \mid X = x]$. 
One way of estimating $h^\ast$ is through kernel ridge regression~\citep{fischer2020sobolev}: given a reproducing kernel $k:\calX \times \calX \to \R$, the kernel ridge regression estimator $\hat{h}_\lambda: \calX \to \R$ is defined as the solution to the following optimization problem ($\lambda > 0$):
\begin{align}\label{eq:opt}
    \hat{h}_\lambda
    = \argmin_{h \in \calH_k} \left\{ \lambda\|h\|_{\calH_k}^2 + \frac{1}{N} \sum_{i=1}^N \left( y_i - h (x_i) \right)^2 \right\}.
\end{align}
$\calH_k$ is the reproducing kernel Hilbert space (RKHS) associated with a kernel $k$.
Fortunately, it has the following closed-form expression~\citep[Section 7]{gretton2013introduction} 
\begin{align*}
    \hat{h}_\lambda = k(\cdot, x_{1:N}) \left( k(x_{1:N}, x_{1:N}) + N \lambda \Id_N \right)^{-1} y_{1:N} .
\end{align*}
We also introduce an auxiliary function $h_\lambda: \calX \to \calY$ which is the solution to another optimization problem:
\begin{align}\label{eq:opt_dummy}
    h_\lambda = \argmin_{f \in \calH_k } \left\{ \lambda\|f\|_\calH^2 + \int_{\calX \times \R} ( y - f(x))^2 \mathbf{P}(dx, dy) \right\} .
\end{align}
In regression setting, it is of interest to study the generalization error between the estimator $\hat{h}_\lambda$ and the Bayes optimal predictor $h^\ast$, $\| \hat{h}_\lambda - h^\ast \|_{L_2(\Qb)}$, and particularly its asymptotic rate of convergence towards $0$ as the number of samples $N$ tend to infinity.
The generalization error can be decomposed into two terms, through a triangular inequality,
\begin{align}\label{eq:bias_variance_decompose}
    \| \hat{h}_\lambda - h^\ast \|_{L_2(\Qb)} \leq \| \hat{h}_\lambda - h_\lambda \|_{L_2(\Qb)} + \left\| h_\lambda  - h^\ast \right\|_{L_2(\Qb)},
\end{align}
where the first term $\| \hat{h}_\lambda - h_\lambda \|_{L_2(\Qb)}$ is known as the estimation error and the second term $ \left\| h_\lambda  - h^\ast \right\|_{L_2(\Qb)} $ is known as the approximation error.
Next, we are going to present propositions that study these two terms separately under the following list of conditions.
\begin{enumerate}[leftmargin=1.0cm]
    \item [(S1)] $k$ is a Sobolev reproducing kernel of smoothness $s > \frac{d}{2}$.
    \customlabel{as:kernel}{(S1)} 
    \item [(S2)] $\Qb$ is a probability measure on $\calX$ with density $p: \calX \to \R$. There exist positive constants $G_0, G_1$ such that $G_0 \leq p(x) \leq G_1$ for any $x \in \calX$. 
    \customlabel{as:density}{(S2)} 
    \item [(S3)] The Bayes predictor $h^\ast \in W_2^s(\calX)$.
    \customlabel{as:bayes_predictor}{(S3)} 
    \item [(S4)] There exists universal constants $\sigma, L > 0$ such that \eqref{eq:mom} holds. 
    \customlabel{as:noise}{(S4)} 
\end{enumerate}

\begin{prop}[Approximation error]\label{prop:krr_bias}
Under Assumptions \ref{as:kernel}-\ref{as:noise}, 
\begin{align*}
    \left\| h_\lambda - h^\ast \right\|_{L_2(\Qb)} \leq \left\| h^\ast \right\|_{s,2} \lambda^{\frac{1}{2}} .
\end{align*}
\end{prop}
\begin{proof}
    This is direct application of Lemma 14 of \cite{fischer2020sobolev} with $\beta = 1$ and $ \gamma = 0$.
\end{proof}
\begin{prop}[Estimation error]\label{prop:krr_variance}
Suppose Assumptions \ref{as:kernel}-\ref{as:noise} hold. Let $\calN(\lambda)$ be the effective dimension defined in \Cref{lem:dof},
and $k_{\alpha}$ be defined in \Cref{lem:embedding}.
If $N > A_{\lambda, \tau}$, then with probability at least $1 - 4e^{-\tau}$,
\begin{align}
    \| h_\lambda - \hat{h}_\lambda \|_{L_2(\Qb)}^2 \leq \frac{576 \tau^2}{N} \left( L^2 D \lambda^{ -\frac{d}{2s} } + M^2 \lambda^{1 - \frac{d}{2s}} \left\| h^\ast \right \|_{s,2}^2 + M^2 \frac{L_\lambda^2}{N} \lambda^{-\frac{d}{2s}} \right),
\end{align}
where $D$ and $M$ are constants independent of $N$, and $g_\lambda, A_{\lambda, \tau}, L_\lambda$ are defined as follows
\begin{align*}
    g_\lambda &:=\log \left(2 e \mathcal{N}(\lambda) \frac{ \|\Sigma_\Qb \| + \lambda}{\|\Sigma_\Qb \|} \right), 
    \quad A_{\lambda, \tau}:= 8 k_{\alpha}^2 \tau g_\lambda \lambda^{-\frac{d}{2s} },\\
    L_\lambda &:= \max \left \{
    L,\,
    \lambda^{\frac{1}{2} - \frac{d}{4s} }
    \left( \| h^\ast \|_{L_\infty(\Qb)} + k_{\alpha} \|  h^\ast  \|_{s,2} \right)
    \right\} .
\end{align*}

\end{prop}
\begin{proof}
This proposition is a special case of Theorem 16 in \citet{fischer2020sobolev} under the following adaptations towards our Sobolev space setting: 1) \Cref{lem:dof} proves that $\calN(\lambda) \leq D \lambda^{ -\frac{d}{2s} }$ and \Cref{lem:embedding} proves that $k_{\alpha} \leq M$ for $\alpha = \frac{d}{2s}$.
2) $\|h^\ast- h_\lambda \|_{L_{\infty}(\Qb)}$ is upper bounded by $\lambda^{\frac{1}{2} - \frac{d}{4s} } \left( \| h^\ast \|_{L_\infty(\Qb)} + k_{\alpha} \| h^\ast \|_{s,2} \right) $ proved in Corollary 15 of \citet{fischer2020sobolev}. $\| \Sigma_\Qb\|$ is the norm of the covariance operator defined in \eqref{eq:covariance_operator}.
\end{proof}
\begin{prop}\label{prop:krr_all}
Suppose Assumptions \ref{as:kernel}-\ref{as:noise} hold. For $A_{\lambda, \tau}$ and $L_\lambda$ defined above in \Cref{prop:krr_variance}, if $N > A_{\lambda, \tau}$, then with probability at least $1 - 4e^{-\tau}$,
\begingroup\small
\begin{align*}
    \left\| \hat{h}_\lambda - h^\ast \right\|_{L_2(\Qb)}^2
    &\leq \frac{576 \tau^2}{N} \left(
    L^2 D \lambda^{-\frac{d}{2s}}
    + M^2 \lambda^{1 - \frac{d}{2s}} \left\| h^\ast \right\|_{s,2}^2
    \right. \\
    &\hspace{85pt}\left.
    + 2 M^2 \frac{L_\lambda^2}{N} \lambda^{-\frac{d}{2s}}
    \right)
    + \left\|h^\ast\right\|_{s,2}^2 \lambda .
\end{align*}
\endgroup
\end{prop}
\begin{proof}
By the triangle inequality in \eqref{eq:bias_variance_decompose}, combining \Cref{prop:krr_bias} and \Cref{prop:krr_variance} finishes the proof.
\end{proof}

\section{Noiseless Kernel Ridge Regression (Kernel Quadrature)}\label{sec:noiseless_krr}
In this section, we present the upper bound on the generalization error $\| h^\ast - \hat{h}_{\lambda} \|_{L_2(\Qb)}$ in \Cref{prop:krr_all} adapted to the noiseless regression setting, which will be employed in the proof of \Cref{thm:main} in the next section.
Our proof follows the outline of the proof for Theorem 1 in \cite{fischer2020sobolev}, modified for our choice of regularization parameter $\lambda$.
Note that this section is of independent interest to some readers as it presents the first standalone proof on the convergence rate of kernel quadrature that 1): it allows positive regularization parameter $\lambda > 0$ and 2): it provides convergence in high probability rather than in expectation.
The closely-related work is \citet{Bach2015} which requires i.i.d samples from an intractable distribution; and
\citet{long2024duality} which provides a more general analysis on noiseless kernel ridge regression in both well-specified and mis-specified setting.

Suppose we have $N$ observations $x_{1:N}$ which are i.i.d sampled from an unknown distribution $\Qb$ on $\calX$ along with $N$ \emph{noiseless} function evaluations $h^\ast(x_{1:N})$ where $h^\ast : \calX \subset \R^d \to \R$. 
The setting appears for instance when the measurement of the output
values is very accurate, or when the output values are obtained as a result of computer experiments. 

\begin{prop}\label{prop:noiseless_krr}
Let $\mathcal{X} \subset \mathbb{R}^d$ be compact, and $x_{1:N}$ be $N$ i.i.d. samples from $\Qb$. Define
$\hat{h}_{\lambda_N}(\cdot) \coloneq k(\cdot, x_{1:N}) \left( k(x_{1:N}, x_{1:N}) + N \lambda_N \Id_N \right)^{-1} h^\ast(x_{1:N})$, and suppose conditions \ref{as:kernel}-\ref{as:noise} are satisfied. Then, if $\lambda_N \asymp N^{-\frac{2 s}{d}} (\log N)^{\frac{2s+2}{d}}$, there exists an $N_0>0$ such that for all $N > N_0$, 
\begingroup\small
\begin{align}
\| h^\ast - \hat{h}_{\lambda_N} \|_{L_2(\Qb)} \leq \mathfrak{C} \tau N^{-\frac{s}{d}} (\log N)^{\frac{s+1}{d}} \left\| h^\ast \right \|_{s,2}
\end{align} 
\endgroup
holds with probability at least $1 - 4e^{-\tau}$, for a constant $\mathfrak{C}=\mathfrak{C}(\calX, G_0, G_1)$ that only depends on $\calX, G_0, G_1$.
\end{prop}
\begin{proof}
Notice that $ \hat{h}_{\lambda_N} $ is precisely the solution to the optimization problem defined in  \eqref{eq:opt} only with $y_i$ replaced by $h^\ast(x_i)$. 
Similarly, we define $h_{\lambda_N}$ as the solution to the optimization problem defined in \eqref{eq:opt_dummy} only with $y$ replaced by $h^\ast(x)$. 
Note that Assumption \ref{as:noise} is instantly satisfied with $L = 0$. 

Similar to the proof of \Cref{prop:krr_all}, we decompose the generalization error into an estimation error term $\| h_{\lambda_N} - \hat{h}_{\lambda_N} \|_{L_2(\Qb)}$ and an approximation error term $\| h_{\lambda_N} - h^\ast \|_{L_2(\Qb)}$. 

\paragraph{Approximation error}
Take $\lambda_N \asymp N^{-\frac{2s}{d}} (\log N)^{\frac{2s+2}{d}}$, then from \Cref{prop:krr_bias}, we have 
\begin{align*}
    \left\| h_{\lambda_N} - h^\ast \right\|_{L_2(\Qb)} \leq \lambda_N^{\frac{1}{2}} \| h^\ast \|_{s,2} \asymp N^{-\frac{s}{d}}  (\log N)^{\frac{s+1}{d}} \| h^\ast \|_{s,2}.
\end{align*}
\paragraph{Estimation error}
Recall all the constants $g_{\lambda_N}$, $A_{\lambda_N, \tau}$ and $ L_{\lambda_N}$ defined in \Cref{prop:krr_variance}.
Since $L=0$, we know the constant $L_{\lambda_N} = \lambda_N^{\frac{1}{2} - \frac{d}{4s} }\allowbreak \big( \| h^\ast \|_{L_\infty(\Qb)} +\allowbreak k_{\alpha} \| h^\ast \|_{s,2} \big)$.
In order to apply \Cref{prop:krr_variance}, we need to check there indeed exists $N_0$ such that $N \geq A_{\lambda_N, \tau}$ is satisfied for all $N \geq N_0$. 
To this end, we are going to verify that $\lim_{N \to \infty} A_{\lambda_N, \tau} / N \to 0$. 
Notice that
\begingroup\small
\begin{align*}
    \lim_{N \to \infty} \frac{A_{\lambda_N, \tau}}{N} = \frac{8 k_{\alpha}^2 \tau g_{\lambda_N} \lambda_N^{-\frac{d}{2s} }}{N} &= 8 (\log N)^{- \frac{s+1}{s}} k_{\alpha}^2 \tau \log \left( 2 e \mathcal{N}(\lambda_N) \frac{\| \Sigma_\Qb \| + \lambda_N}{ \| \Sigma_\Qb \| } \right) 
\end{align*}
\endgroup
where $\calN(\lambda_N)$ and $k_{\alpha}^2$ are defined in \Cref{lem:dof} and \Cref{lem:embedding}.
Since $\lim_{N \to \infty} \lambda_N = \lim_{N \to \infty} N^{-\frac{2s}{d}} (\log N)^{\frac{2s+2}{d}} = 0$, there exists $N^\prime$ such that $\lambda_N \leq \| \Sigma_\Qb \|$ for all $N \geq N^\prime$. 
Therefore, as $N$ tends to infinity,
\begingroup\small
\begin{align}\label{eq:ratio_N_A}
\begin{aligned}
    \lim_{N \to \infty} \frac{A_{\lambda_N, \tau}}{N} &\leq \lim_{N \to \infty} 8 (\log N)^{- \frac{s+1}{s}} k_{\alpha}^2 \tau \log \left(4 e \mathcal{N}(\lambda_N) \right) \\
    &\leq \lim_{N \to \infty} 8 (\log N)^{- \frac{s+1}{s}} k_{\alpha}^2 \tau \log \left(4 e D \lambda_N^{-\frac{d}{2s}} \right) \\
    &= \lim_{N \to \infty} 8 (\log N)^{- \frac{s+1}{s}} k_{\alpha}^2 \tau \log \left(4 e D \right)\\
    &\quad + \lim_{N \to \infty} 8 (\log N)^{- \frac{s+1}{s}} k_{\alpha}^2 \tau \log \left( N (\log N)^{-\frac{s+1}{s}} \right)  \\
    &\leq \lim_{N \to \infty} 16 (\log N)^{- \frac{s+1}{s}} k_{\alpha}^2 \tau \log \left( N \right) \\
    &= 0,
\end{aligned}
\end{align}
\endgroup
where $M$ and $D$ are constants defined in \Cref{lem:dof} and \Cref{lem:embedding}. 
So there exists $N''$ such that $N \geq A_{\lambda_N, \tau}$ for all $N \geq N''$. Taking $N_0 = \max \{ N^\prime, N'' \}$, then we have $N \geq A_{\lambda_N, \tau}$ for all $N \geq N_0$. Set $B_\ast := \|h^\ast\|_{L_\infty(\Qb)} + M\|h^\ast\|_{s,2}$. From \Cref{prop:krr_variance}, we know that with probability at least $1 - 4e^{-\tau}$,
\begingroup\small
\begin{align*}
    \| h_{\lambda_N} - \hat{h}_{\lambda_N} \|_{L_2(\Qb)}^2
    &\leq \frac{576 \tau^2}{N} \\
    &\quad {}\times
    \left(
    M^2 \lambda_N^{1 - \frac{d}{2s}} \left\| h^\ast \right \|_{s,2}^2
    + M^2 \lambda_N^{1 - \frac{d}{2s} } B_\ast^2
    \frac{1}{N} \lambda_N^{-\frac{d}{2s}}
    \right) \\
    &\asymp \frac{576 \tau^2}{N}
    \left(
    M^2 N^{1 - \frac{2s}{d}}
    (\log N)^{\frac{s+1}{s} \frac{2s-d}{d}}
    \left\| h^\ast \right \|_{s,2}^2
    \right.\\
    &\hspace{55pt}\left.
    + M^2
    B_\ast^2
    N^{1 -\frac{2s}{d}}
    (\log N)^{\frac{s+1}{2s} \frac{4s-d}{d}}
    \right) \\
    &\leq 576 \tau^2 N^{-\frac{2s}{d}} (\log N)^{\frac{2s+2}{d} } \\
    &\quad {}\times
    \left(
    M^2 \left\| h^\ast \right\|_{s,2}^2
    \right. \\
    &\hspace{100pt}\left.
    + M^2 B_\ast^2
    \right) .
\end{align*}
\endgroup
So we have, with probability at least $1 - 4e^{-\tau}$, 
\begingroup\small
\begin{align*}
    \| h_{\lambda_N} - \hat{h}_{\lambda_N} \|_{L_2(\Qb)}
    &\leq 24 \tau N^{-\frac{s}{d}} (\log N)^{\frac{s+1}{d} }
    \left( (M + M^2) \left\| h^\ast \right\|_{s,2} \right. \\
    &\hspace{110pt}\left. {} + M \|h^\ast \|_{L_\infty(\Qb)} \right).
\end{align*}
\endgroup
\paragraph{Combine approximation and estimation error}
Combining the above two inequalities on approximation error $\| h_{\lambda_N} - h^\ast \|_{L_2(\Qb)}$ and estimation error $\| h_{\lambda_N} - \hat{h}_{\lambda_N} \|_{L_2(\Qb)}$, we have with probability at least $1 - 4e^{-\tau}$, 
\begingroup\small
\begin{align*}
    \| h^\ast - \hat{h}_{\lambda_N} \|_{L_2(\Qb)}
    &\leq 24 \tau N^{-\frac{s}{d}} (\log N)^{\frac{s+1}{d}}
    \left( (1 + M + M^2) \left\| h^\ast \right \|_{s,2} + M \|h^\ast \|_{L_\infty(\Qb)} \right).
\end{align*}
\endgroup
Finally, following the arguments of \Cref{lem:embedding} that the operator norm of $W_2^s(\calX) \hookrightarrow L_\infty(\calX) $ is bounded, we have $ \|h^\ast \|_{L_\infty(\Qb)} \leq R \left\| h^\ast \right \|_{s,2}$ where $R$ is a constant that depends on $\calX, G_0, G_1$. With probability at least $1 - 4e^{-\tau}$,
\begingroup\small
\begin{align*}
    \| h^\ast - \hat{h}_{\lambda_N} \|_{L_2(\Qb)}
    &\leq 24 \tau N^{-\frac{s}{d}} (\log N)^{\frac{s+1}{d}}
    (1 + (1 + R) M + M^2) \left\| h^\ast \right \|_{s,2} \\
    &= \mathfrak{C} \tau N^{-\frac{s}{d}} (\log N)^{\frac{s+1}{d}} \| h^\ast\|_{s,2},
\end{align*}
\endgroup
for $\mathfrak{C} := 24 (1 + (1 + R) M + M^2)$, which concludes the proof.
\end{proof}
\begin{cor}\label{cor:kq_rate}
    Let $\mathcal{X}$ be a compact domain in $\mathbb{R}^d$ and $x_{1:N}$ are $N$ i.i.d samples from $\Qb$. 
    $\hat{I}_\kq \coloneq \E_{X \sim \Qb}[k(X, x_{1:N})] \left( k(x_{1:N}, x_{1:N}) + N \lambda_N \Id_N \right)^{-1} h^\ast(x_{1:N})$ is the KQ estimator defined in \eqref{eq:kq}.
    Suppose conditions (A1)-(A3) are satisfied. Take $\lambda_N \asymp N^{-\frac{2 s}{d}} (\log N)^{\frac{2s+2}{d}}$, then there exists $N_0>0$ such that for $N > N_0$, 
    \begin{align}
    \left| \hat{I}_\kq - \int_\calX h^\ast(x) d\Qb(x) \right| \leq \mathfrak{C} \tau N^{-\frac{s}{d}} (\log N)^{\frac{s+1}{d}} 
    \end{align} 
holds with probability at least $1 - 4e^{-\tau}$. 
Here $\mathfrak{C}$ is a constant that is independent of $N$. 
\end{cor}
The proof of \Cref{cor:kq_rate} is a direct application of \Cref{prop:noiseless_krr} after observing the following.
\begingroup\small
\begin{align*}
    \left| \hat{I}_\kq - \int h^\ast(x) d\Qb(x) \right|
    &\leq \int_\calX \left| \hat{h}_{\lambda_{N}}(x) - h^\ast(x) \right| d \Qb(x) \\
    &= \| h^\ast - \hat{h}_{\lambda_N} \|_{L_1(\Qb)}
    \leq \| h^\ast - \hat{h}_{\lambda_N} \|_{L_2(\Qb)} .
\end{align*}
\endgroup

\begin{rem}
    We prove in \Cref{prop:noiseless_krr} that the generalization error of $\hat{h}_{\lambda_N}$ in noiseless regression setting is $\tilde{\calO}(N^{-\frac{s}{d}})$, which is faster than the minimax optimal rate $\calO(N^{-\frac{s}{2s+d}})$ in standard regression setting. The fast rate is expected because we are in the noiseless regime so ``overfitting" is not a problem --- hence our choice of regularization parameter $\lambda_N \asymp N^{-\frac{2s}{d}} (\log N)^{\frac{2s+2}{d}}$ decays to $0$ at a faster rate than $\lambda_N \asymp N^{-\frac{2s}{2s + d}}$ in standard kernel ridge regression~\citep[Corollary 5]{fischer2020sobolev}. 
    The $\tilde{\calO}(N^{-\frac{s}{d}})$ rate is also optimal (up to logarithm terms) and cannot be further improved because it matches the lower bound of interpolation (Sections 1.3.11
    and 1.3.1 of \citet{novak2006deterministic},  Section 1.2, Chapter V of \citet{ritter2000average}). 
\end{rem}

\begin{rem}[Comparison to existing upper bound of kernel (Bayesian) quadrature] \label{rem:stage_one_error_and_standard_kq}
The upper bound in \Cref{cor:kq_rate} matches existing analysis based on scattered data approximation in the literature of both kernel quadrature and Bayesian quadrature~\citep{sommariva2006numerical, Briol2019PI, wynne2021convergence} and is known to be minimax optimal~\citep{novak2016some, novak2006deterministic}. 
Existing analysis takes $\lambda = 0$ and requires the Gram matrix $ k(x_{1:N}, x_{1:N})$ to be invertible, in contrast, our result allows a positive regularization parameter $\lambda_N \asymp N^{-\frac{2s}{d}} (\log N)^{\frac{2s+2}{d}}$ which improves numerical stability of matrix inversion in practice. 
One closely-related work is \citet{Bach2015}, but it requires i.i.d samples from an intractable distribution.
\end{rem}

\section{\texorpdfstring{Proof of \Cref{thm:main}}{Proof of the main theorem}}\label{sec:proof}
\begin{rem}
    In this section, we use $p(x; \theta)$ to denote the density $p_\theta(x)$ so that we can use $p(x; \cdot)$ to denote the mapping $\theta \mapsto p_\theta(x)$. Although we introduce a shorthand notation of kernel mean embedding in the main text, $m_{\Qb} = \mathbb{E}_{X \sim \Qb}[k(X,\cdot)]$, in this section we are going to write it out with its explicit formulation.
\end{rem}

For any $\theta \in \Theta$, $\hat{F}_{\kq}: \Theta \to \R$  and $\hat{J}_{\kq}: \Theta \to \R$ are two functions that generalize the definition of $\hat{F}_{\kq}(\theta_t)$ and $\hat{J}_{\kq}(\theta_t)$ in \eqref{eq:F_J_KQ} to all $\theta \in \Theta$. 
To be more specific, for any $\theta \in \Theta$, given samples $x_{1:N}^{(\theta)} := \big[ x_1^{(\theta)}, \ldots, x_N^{(\theta)} \big]^\top$ consisting of $N$ i.i.d. samples from $\Pb_\theta$,
\begingroup\small
\begin{align}\label{eq:hat_F_KQ_all_theta}
    \hat{J}_{\kq} (\theta; x_{1:N}^{(\theta)})
    &:= \left( \int_\calX k_\calX (x, x_{1:N}^{(\theta)}) d \Pb_{\theta}(x) \right) \nonumber \\
    &\quad \times \left( k_\calX (x_{1:N}^{(\theta)}, x_{1:N}^{(\theta)})
    + N \lambda_\calX \Id_N \right)^{-1} g(x_{1:N}^{(\theta)}, \theta), \\
    \quad \hat{F}_{\kq}(\theta; x_{1:N}^{(\theta)}) &:= f(\hat{J}_{\kq} (\theta; x_{1:N}^{(\theta)})),
\end{align}
\endgroup
where we explicitly specify the dependence of samples $x_{1:N}^{(\theta)}$ on $\theta$ in the above two equations. 
Next, we define 
\begin{align}\label{eq:bar_F_KQ_all_theta}
\begin{aligned}
    \bar{J}_{\kq} (\theta) &:= \E_{x_{1:N}^{(\theta)} \sim \Pb_\theta} \left[\hat{J}_{\kq} (\theta; x_{1:N}^{(\theta)})\right] = \int \hat{J}_{\kq} (\theta; x_{1:N}) \prod_{i=1}^N p(x_i; \theta) dx_{1:N}, \\ 
    \bar{F}_{\kq} (\theta) &:= \E_{x_{1:N}^{(\theta)} \sim \Pb_\theta} \left[\hat{F}_{\kq} (\theta; x_{1:N}^{(\theta)})\right] = \int \hat{F}_{\kq} (\theta; x_{1:N}) \prod_{i=1}^N p(x_i; \theta) dx_{1:N},
\end{aligned}
\end{align}
which marginalize out the dependence on samples $x_{1:N}^{(\theta)}$. 
We can see that $\bar{J}_{\kq} \in L_2(\Qb)$ since $g(x, \cdot) \in W_2^{s_\Theta}(\Theta) \subset L_2(\Theta) \cong L_2(\Qb)$ from Assumption \ref{as:equivalence} and \ref{as:app_true_J_smoothness}; and $p(x_i; \cdot) \in L_2(\Qb)$. Also $\bar{F}_{\kq} \in L_2(\Qb)$ because $f$ is Lipschitz continuous from Assumption \ref{as:app_lipschitz}.
Therefore, the absolute error $| I - \hat{I}_{\nkq}|$ can be decomposed as follows:
\begingroup\small
\begin{align}\label{eq:stage_i_stage_ii_decomposition}
& \quad \left| I - \hat{I}_{\nkq} \right| \nonumber \\
& = \left| \int_\Theta F(\theta) q(\theta) d\theta
 - \left( \int_\Theta k_\Theta(\theta, \theta_{1:T}) q(\theta) d\theta \right) \right. \nonumber \\
& \hspace{75pt}\left. {}
 \times \left( k_\Theta(\theta_{1:T}, \theta_{1:T}) + T \lambda_\Theta \Id_T \right)^{-1}
 \hat{F}_{\kq}(\theta_{1:T}) \right| \nonumber \\
&\leq \left| \int_\Theta F(\theta) q(\theta) d\theta  - \int_\Theta \bar{F}_{\kq}(\theta) q(\theta) d\theta  \right| \nonumber \\
& \quad + \left| \int_\Theta \bar{F}_{\kq}(\theta) q(\theta) d\theta
 - \left( \int_\Theta k_\Theta(\theta, \theta_{1:T}) q(\theta) d\theta \right) \right. \nonumber \\
& \hspace{75pt}\left. {}
 \times \left( k_\Theta(\theta_{1:T}, \theta_{1:T}) + T \lambda_\Theta \Id_T \right)^{-1}
 \hat{F}_{\kq}(\theta_{1:T}) \right| \nonumber \\
&\leq \underbrace{\E_{\theta \sim \Qb}
 \left[\left| F(\theta) - \bar{F}_{\kq}(\theta) \right|\right]}_{\text{Stage I error}} \nonumber \\
&\quad + \underbrace{\left\| \bar{F}_{\kq}(\cdot)
 - k(\cdot, \theta_{1:T})
 \left(k_\Theta(\theta_{1:T}, \theta_{1:T}) + T \lambda_\Theta \Id_T\right)^{-1}
 \hat{F}_{\kq}(\theta_{1:T}) \right\|_{L_2(\Qb)}}_{\text{Stage II error}} .
\end{align}
\endgroup
The last inequality holds because $\|\cdot\|_{L_1(\Qb)} \leq \|\cdot\|_{L_2(\Qb)}$. Next, we analyze Stage I error and Stage II error separately.

\paragraph{Stage I Error}
From Assumption \ref{as:app_lipschitz}, $f$ is Lipschitz continuous and the Lipschitz constant is bounded by $S_4$,
\begin{align}\label{eq:lipschitz_F}
     \left| \bar{F}_{\kq}(\theta) - F(\theta) \right| &= \left| \E_{x_{1:N}^{(\theta)} \sim \Pb_\theta} \hat{F}_{\kq} (\theta; x_{1:N}^{(\theta)}) - F(\theta) \right| \nonumber \\
     &\leq \E_{x_{1:N}^{(\theta)} \sim \Pb_\theta} \left| \hat{F}_{\kq} (\theta; x_{1:N}^{(\theta)}) - F(\theta) \right| \nonumber \\
     &\leq S_4 \E_{x_{1:N}^{(\theta)} \sim \Pb_\theta} \left| \hat{J}_{\kq}(\theta; x_{1:N}^{(\theta)}) - J(\theta) \right| ,
\end{align}
where the first inequality holds by Jensen inequality and the last inequality holds by Lipschitz continuity of $f$.
Define 
\begin{align}
\label{eq:defn_of_g}
    \hat{g}(x, \theta; x_{1:N}^{(\theta)}) = k_\calX(x, x_{1:N}^{(\theta)}) (k_\calX(x_{1:N}^{(\theta)}, x_{1:N}^{(\theta)}) + N \lambda_\calX \Id_N)^{-1} g(x_{1:N}^{(\theta)}, \theta) .
\end{align}
Here $\hat{g}(\cdot, \theta; x_{1:N}^{(\theta)}) \in L_2(\Pb_\theta)$ because the Sobolev reproducing kernel $k_\calX$ is bounded and measurable; and $g(\cdot, \theta) \in L_2(\Pb_\theta)$ by Assumption \ref{as:app_true_J_smoothness}. Thus, 
\begingroup\small
\begin{align}\label{eq:F_L_g}
    \left| \hat{J}_{\kq}(\theta; x_{1:N}^{(\theta)}) - J(\theta) \right|
    &= \left| \int \left( \hat{g}(x, \theta; x_{1:N}^{(\theta)})
    - g(x, \theta) \right) p(x; \theta) dx \right| \nonumber \\
    &\leq \left\| \hat{g}(\cdot, \theta; x_{1:N}^{(\theta)})
    - g(\cdot, \theta) \right\|_{L_2(\Pb_\theta)} .
\end{align}
\endgroup
Based on Assumption~\ref{as:app_true_g_smoothness}, $g(\cdot, \theta) \in W_2^{s_\calX}(\calX)$ for any $\theta \in \Theta$.
Therefore, based on \Cref{prop:noiseless_krr}, if one takes $\lambda_{\calX, N} \asymp N^{-2 \frac{s_\calX}{d_\calX}} (\log N)^{\frac{2s_\calX+2}{d_\calX}}$, then there exists $N_0$ such that for $N > N_0$,
\begin{align}\label{eq:high_prob_hat_g_g}
    \left\| \hat{g}(\cdot, \theta; x_{1:N}^{(\theta)}) - g(\cdot, \theta) \right\|_{L_2( \Pb_\theta ) } \leq \mathfrak{C} \tau N^{-\frac{s_\calX}{d_\calX}} (\log N)^{\frac{s_\calX+1}{d_\calX}} \| g(\cdot, \theta)\|_{s_\calX,2},
\end{align}
holds with probability at least $1 - 4 e^{-\tau}$. The probability is taken over the distribution of $x_{1:N}^{(\theta)}$, i.e $\Pb_\theta$.
Here $\mathfrak{C}$ is a constant independent of $N$. 
Hence, with \Cref{lem:prob_to_expectation}, we have
\begin{align}\label{eq:F_KQ_F}
     \E_{x_{1:N}^{(\theta)} \sim \Pb_\theta} \left\| \hat{g}(\cdot, \theta; x_{1:N}^{(\theta)}) - g(\cdot, \theta) \right\|_{L_2( \Pb_\theta ) } \leq \mathfrak{C} N^{-\frac{s_\calX}{d_\calX}} (\log N)^{\frac{s_\calX+1}{d_\calX}} \| g(\cdot, \theta)\|_{s_\calX,2}.
\end{align}
By plugging the above inequality back into \eqref{eq:F_L_g}, we obtain
\begin{align*}
    \E_{x_{1:N}^{(\theta)} \sim \Pb_\theta} \left| \hat{J}_{\kq}(\theta; x_{1:N}^{(\theta)}) - J(\theta) \right| \leq \mathfrak{C} N^{-\frac{s_\calX}{d_\calX}} (\log N)^{\frac{s_\calX+1}{d_\calX}} \| g(\cdot, \theta)\|_{s_\calX,2}.
\end{align*}
Therefore, the Stage I error can be upper bounded by
\begin{align}\label{eq:stage_1}
 \E_{\theta \sim \Qb} \left| F(\theta) - \bar{F}_{\kq}(\theta) \right| 
&\leq S_4 \E_{x_{1:N}^{(\theta)} \sim \Pb_\theta} \left| \hat{J}_{\kq}(\theta; x_{1:N}^{(\theta)}) - J(\theta) \right| \| g(\cdot, \theta)\|_{s_\calX,2} \nonumber \\
&\leq S_4 S_1 \mathfrak{C} N^{-\frac{ s_\calX}{d_\calX}} (\log N)^{\frac{s_\calX+1}{d_\calX}} \nonumber \\
&= C_3 N^{-\frac{ s_\calX}{d_\calX}} (\log N)^{\frac{s_\calX+1}{d_\calX}},
\end{align}
where $C_3 := S_4 S_1 \mathfrak{C}$ is a constant independent of $N$.

\paragraph{Stage II Error}
The upper bound on the stage II error is done in five steps. In step one, we prove that $\hat{J}_{\kq}(\cdot; x_{1:N}^{(\theta)}) \in W_2^{s_\Theta} (\Theta)$ given fixed samples $x_{1:N}^{(\theta)}$. In step two, we show that $J \in W_2^{s_\Theta} (\Theta)$. 
In step three, we upper bound $\|\hat{J}_{\kq}(\cdot; x_{1:N}^{(\theta)})\|_{s_\Theta,2}$ through the triangular inequality that
$\|\hat{J}_{\kq}(\cdot; x_{1:N}^{(\theta)})\|_{s_\Theta,2} \leq \|J\|_{s_\Theta,2} + \|J - \hat{J}_{\kq}(\cdot; x_{1:N}^{(\theta)})\|_{s_\Theta,2}$.
In step four, we upper bound $\bar{F}_{\kq}(\theta) = \E_{x_{1:N}^{(\theta)}}\left[f\left( \hat{J}_{\kq}(\cdot; x_{1:N}^{(\theta)}) \right)\right]$ through marginalizing out the samples $x_{1:N}^{(\theta)}$. In the last step, we use kernel ridge regression bound proved in \Cref{prop:krr_all} to upper bound the stage II error.

\underline{\emph{Step One.}} In this step, we are going to show that $\hat{J}_{\kq}$ lies in the Sobolev space $W_2^{s_\Theta} (\Theta)$ given fixed samples $x_{1:N}^{(\theta)}$. 
Notice that the dependence of $\hat{J}_{\kq}(\theta)$ on $\theta$ is through two mappings: $\theta \mapsto \int_\calX k_\calX (x, x_{1:N}^{(\theta)}) p(x; \theta) dx$ and $\theta \mapsto g(x_{1:N}^{(\theta)}, \theta)$. 
We are going to show that $ \theta \mapsto \int_\calX k_\calX (x, x_i^{(\theta)}) p(x; \theta) dx$ lies in the Sobolev space $W_2^{s_\Theta} (\Theta)$ for any $i \in \{1, \ldots, N\}$. 
To this end, we are going to demonstrate it possesses weak derivatives up to and including order $s_\Theta$ that lie in $\calL^2(\Theta)$.
Take $\varphi : \Theta \to \R$ to be any infinitely differentiable function with compact support in $\Theta$ (commonly denoted as $\varphi \in C_c^\infty(\Theta)$), with its standard, non-weak derivative of order $\beta$ denoted by $\partial^\beta \varphi$. Since $\theta \mapsto p(x; \theta) \in W_2^{s_\Theta}(\Theta)$, for any $| \beta| \leq s_\Theta$ it has a weak derivative $\theta \mapsto D_\theta^\beta p(x; \theta) \in \calL^2(\Theta)$. Then,
\begingroup\small
\begin{align}\label{eq:weak_derivative}
\begin{aligned}
    &\quad \int_\Theta \varphi(\theta)
    \int_\calX k_\calX (x, x_i^{(\theta)}) D_\theta^\beta p(x; \theta) dx \\
    &\stackrel{(i)}{=} \int_\calX k_\calX (x, x_i^{(\theta)})
    \int_\Theta \varphi(\theta) D_\theta^\beta p(x; \theta) d\theta dx \\
    &\stackrel{(ii)}{=} (-1)^{|\beta|} \int_\calX k_\calX (x, x_i^{(\theta)})
    \int_\Theta \partial^\beta \varphi(\theta) p(x; \theta) d\theta dx \\
    &\stackrel{(iii)}{=} (-1)^{|\beta|} \int_\Theta \partial^\beta \varphi(\theta)
    \int_\calX k_\calX (x, x_i^{(\theta)}) p(x; \theta) dx d\theta.
\end{aligned}
\end{align}
\endgroup
In the above chain of derivations, we are allowed to swap the integration order in $(i)$ by the Fubini theorem~\citep{rudin1964principles} because $k_\calX$ is bounded and the fact that $\theta \mapsto \varphi(\theta) \cdot D_\theta^\beta p(x; \theta) \in L_1(\Theta)$ since $D_\theta^\beta p(x; \cdot) \in L_2(\Theta)$ (Assumption \ref{as:app_true_J_smoothness}) and $\varphi \in L_2(\Theta)$; $(ii)$ holds by definition of weak derivatives for $D_\theta^\beta p(x; \theta)$; and $(iii)$ holds again by the Fubini theorem. By definition of weak derivatives,~\eqref{eq:weak_derivative} shows that $\int_\calX k_\calX (x, x_i^{(\theta)}) p(x; \theta) dx$ has a weak derivative of order $\beta$ of the form
\begin{equation*}
    D_\theta^\beta \left[\int_\calX k_\calX (x, x_i^{(\theta)}) p(x; \theta) dx \right] = \int_\calX k_\calX (x, x_i^{(\theta)}) D_\theta^\beta p(x; \theta) dx 
\end{equation*}
Also, since $k_\calX$ is bounded and $\theta \mapsto D_\theta^\beta p(x; \theta) \in \calL^2(\Theta)$, the weak derivative above is in $\calL^2(\Theta)$.
Define $f_{1,i}(\theta) := \int_\calX k_\calX(x,x_i^{(\theta)})p(x;\theta)dx$.
Consequently, we have
\begingroup\small
\begin{align*}
    \sum_{ | \beta| \leq s_\Theta }
    \left\| D_\theta^\beta f_{1,i} \right\|_{L_2(\Theta)}^2
    &= \sum_{ | \beta| \leq s_\Theta }
    \left\| \int_\calX k_\calX (x, x_i^{(\theta)}) D_\theta^\beta p(x; \cdot) dx \right\|_{L_2(\Theta)}^2 \\
    &\stackrel{(i)}{\leq} \text{Vol}(\calX)
    \sum_{ | \beta| \leq s_\Theta } \int_\Theta \int_\calX
    \left| k_\calX (x, x_i^{(\theta)}) D_\theta^\beta p(x; \theta) \right|^2 dx d\theta \\
    &\stackrel{(ii)}{\leq} \text{Vol}(\calX) \kappa^2
    \sum_{ | \beta| \leq s_\Theta } \int_\Theta \int_\calX
    \left| D_\theta^\beta p(x; \theta) \right|^2 dx d\theta \\
    &\stackrel{(iii)}{=} \text{Vol}(\calX) \kappa^2 \int_\calX \left\| p(x; \cdot) \right\|_{s_\Theta, 2}^2 dx .
\end{align*}
\endgroup
In the above chain of derivations, $(i)$ holds because $| \int_\calX f(x) dx |^2 \leq \text{Vol}(\calX) \int_\calX | f(x) |^2 dx $ for compact $\calX$, $(ii)$ holds because $k_\calX$ is upper bounded by $\kappa$ and $\int_\Theta |D_\theta^\beta p(x; \theta)|^2 d\theta < \infty$ from Assumption \ref{as:app_true_J_smoothness}, $(iii)$ holds because 
$p(x; \cdot) \in W_2^{s_\Theta}(\Theta)$ for any $x \in \calX$ based on Assumption \ref{as:app_true_J_smoothness}. Also, one can interchange the order of integration in $(iii)$ by the Fubini's theorem~\citep{rudin1964principles}.

As a result, for any $i, j \in \{1, \ldots, N\}$, the functions $f_{1, i}: \theta \mapsto \int_\calX k_\calX (x, x_i^{(\theta)}) p(x; \theta) dx$ and $f_{2, j}: \theta \mapsto g(x_j^{(\theta)}, \theta)$ both belong to $W_2^{s_\Theta}(\Theta)$ from Assumption \ref{as:app_true_J_smoothness}.
Therefore, we know from \Cref{lem:sobolev_algebra} that their product $f_{1, i} \cdot f_{2,j} \in W_2^{s_\Theta} (\Theta)$ hence $\hat{J}_{\kq}$ as a linear combination of $f_{1, i} \cdot f_{2,j}$ is in $W_2^{s_\Theta} (\Theta)$.

\underline{\emph{Step Two.}} In this step, we are going to show that $J: \theta \mapsto \int_\calX g(x, \theta)p(x; \theta) dx$ is also in the Sobolev space $W_2^{s_\Theta} (\Theta)$. 
Since both $g(x, \cdot)\in W_2^{s_\Theta}(\Theta)$ and $p(x; \cdot)\in W_2^{s_\Theta}(\Theta)$, we know from \Cref{lem:sobolev_algebra} that $\theta \mapsto g(x, \theta) \cdot p(x,\theta) \in W_2^{s_\Theta}(\Theta)$. 
By following the same steps as in \eqref{eq:weak_derivative}, we obtain that for any $| \beta| \leq s_\Theta$, 
\begin{align}\label{eq:interchange_integral_derivative}
    D_\theta^\beta \int_\calX g(x, \theta) p(x; \theta) dx = \int_\calX D_\theta^\beta \Big( g(x, \theta) p(x; \theta) \Big) dx .
\end{align}
We are now ready to study the Sobolev norm of $J$,
\begin{align}\label{eq:J}
    \left\| J \right\|_{s_\Theta, 2}^2 
    &:= \sum_{ | \beta| \leq s_\Theta } \int_\Theta \left| D_\theta^\beta \int_\calX p(x; \theta) g(x, \theta) dx \right|^2 d\theta \nonumber \\
    &\stackrel{(i)}{=} \sum_{ | \beta| \leq s_\Theta } \int_\Theta \left| \int_\calX D_\theta^\beta \Big( p(x; \theta) g(x, \theta) \Big) dx \right|^2 d \theta \nonumber \\
    &\stackrel{(ii)}{\leq} \text{Vol}(\calX) \int_\calX \sum_{ | \beta| \leq s_\Theta } \int_\Theta \left| D_\theta^\beta \Big( p(x; \theta) g(x, \theta) \Big) \right|^2 d \theta dx \nonumber \\
    &\stackrel{(iii)}{=} \text{Vol}(\calX) \int_\calX \left\| p(x; \cdot) g(x, \cdot)  \right\|_{s_\Theta, 2}^2 dx \nonumber \\
    &\stackrel{(iv)}{\leq} \text{Vol}(\calX)^2  S_2^2 S_3^2
\end{align}
Here, $(i)$ holds by \eqref{eq:interchange_integral_derivative}, $(ii)$ holds since $| \int_\calX f(x) dx |^2 \leq \text{Vol}(\calX) \int_\calX | f(x) |^2 dx $ for compact $\calX$, 
$(iii)$ follows from the definition of Sobolev norm, and $(iv)$ holds by \Cref{lem:sobolev_algebra} and Assumption \ref{as:app_true_J_smoothness} that $\left\| g(x, \cdot) \right\|_{s_\Theta, 2} \leq S_2$, $\left\| p(x; \cdot) \right\|_{s_\Theta, 2} \leq S_3$. 

\underline{\emph{Step Three.}}
In this step, we study the Sobolev norm of $\hat{J}_{\kq}$ for some fixed $x_{1:N}^{(\theta)}$, by upper bounding it with $\| J - \hat{J}_{\kq}(\cdot ; x_{1:N}^{(\theta)}) \|_{s_\Theta, 2} + \| J \|_{s_\Theta, 2}$. 
Since $g(x_i^{(\theta)}, \cdot) \in W_2^{s_\Theta}(\Theta)$ by Assumption \ref{as:app_true_J_smoothness}, the function
\begin{align*}
    \hat{g}(x, \cdot; x_{1:N}^{(\theta)})
    &= k_\calX(x, x_{1:N}^{(\theta)})
    \big(k_\calX(x_{1:N}^{(\theta)}, x_{1:N}^{(\theta)}) + N \lambda_\calX \Id_N\big)^{-1} \\
    &\qquad {}\times g(x_{1:N}^{(\theta)}, \cdot)
\end{align*}
is in $W_2^{s_\Theta}(\Theta)$ for any fixed $x_{1:N}^{(\theta)}$. Therefore,
\begin{align}\label{eq:p_g_hat_g}
    \left\| p(x; \cdot) \Big( g(x, \cdot) - \hat{g}(x, \cdot; x_{1:N}^{(\theta)}) \Big) \right\|_{s_\Theta, 2} 
    &\leq  \left\| p(x; \cdot) \right\|_{s_\Theta, 2} \left\| g(x, \cdot) - \hat{g}(x, \cdot; x_{1:N}^{(\theta)}) \right\|_{s_\Theta, 2} \nonumber \\
    &\leq  S_3 \left\| g(x, \cdot) - \hat{g}(x, \cdot; x_{1:N}^{(\theta)}) \right\|_{s_\Theta, 2},
\end{align}
where the first inequality holds by \Cref{lem:sobolev_algebra} and the second inequality holds by Assumption \ref{as:app_true_J_smoothness} that $\left\| p(x; \cdot) \right\|_{s_\Theta, 2} \leq S_3$.
For brevity, write $r(x,\theta):=g(x,\theta)-\hat g(x,\theta;x_{1:N}^{(\theta)})$.
Now, we consider the Sobolev norm of $\hat{J}_{\kq} - J$,
\begingroup\small
\begin{align}\label{eq:hat_J_J}
    \left\| \hat{J}_{\kq}(\cdot ; x_{1:N}^{(\theta)}) - J \right\|_{s_\Theta, 2}^2 
    &= \sum_{ | \beta| \leq s_\Theta }
    \left\| D_\theta^\beta \int_\calX p(x; \cdot) r(x,\cdot) dx
    \right\|_{L_2(\Theta)}^2 \nonumber \\
    &\stackrel{(i)}{=} \sum_{ | \beta| \leq s_\Theta }
    \left\| \int_\calX D_\theta^\beta \Big( p(x; \cdot) r(x,\cdot) \Big) dx
    \right\|_{L_2(\Theta)}^2 \nonumber \\
    &\stackrel{(ii)}{\leq} \text{Vol}(\calX) 
    \int_\calX \sum_{ | \beta| \leq s_\Theta } \int_\Theta
    \left| D_\theta^\beta \Big( p(x; \theta) r(x,\theta) \Big) \right|^2
    d\theta dx \nonumber \\
    &\stackrel{(iii)}{=} \text{Vol}(\calX) 
    \int_\calX \left\| p(x; \cdot) r(x,\cdot) \right\|_{s_\Theta, 2}^2 dx \nonumber \\
    &\stackrel{(iv)}{\leq} \text{Vol}(\calX) S_3^2 \, \nonumber \\
    &\quad {}\times \int_\calX
    \left\| r(x,\cdot) \right\|_{s_\Theta, 2}^2 dx,
\end{align}
\endgroup
where the above chain of derivations (i) --- (iv) follow the exact same reasoning as \eqref{eq:weak_derivative} and \eqref{eq:J}.
Next, notice that
\begin{align}\label{eq:g_g_hat_sobolev}
    \int_\calX \left\| r(x, \cdot) \right\|_{s_\Theta, 2}^2 dx
    &= \sum_{ | \beta| \leq s_\Theta } \int_\Theta \int_\calX
    \left| D_\theta^\beta r(x, \theta) \right|^2 dx d\theta \nonumber \\
    &= \sum_{ | \beta| \leq s_\Theta }
    \left\| D_\theta^\beta r \right\|_{L_2(\calX \times \Theta)}^2 .
\end{align}
By Assumption \ref{as:app_true_g_smoothness}, $D_\theta^\beta  g(\cdot, \theta) \in W_2^{s_\calX}(\calX)$ for any $|\beta| \leq s_\Theta$. Therefore, by applying~\Cref{prop:noiseless_krr} with $h^\ast(\cdot):= D_\theta^\beta g(\cdot, \theta)$, and $\hat{h}_{\lambda}(\cdot) := D_\theta^\beta \hat{g}(\cdot, \theta; x_{1:N}^{(\theta)}) = k_\calX(\cdot, x_{1:N}^{(\theta)}) (k_\calX(x_{1:N}^{(\theta)}, x_{1:N}^{(\theta)}) + N \lambda_\calX \Id_N)^{-1} D_\theta^\beta g(x_{1:N}^{(\theta)}, \theta; x_{1:N}^{(\theta)})$, we get that
\begingroup\small
\begin{align}\label{eq:D_g_D_g_hat}
    \left\| D_\theta^\beta  g(\cdot, \theta) - D_\theta^\beta \hat{g}(\cdot, \theta; x_{1:N}^{(\theta)}) \right\|_{L_2(\Pb_\theta)} \leq \mathfrak{C} \tau N^{-\frac{s_\calX}{d_\calX}} (\log N)^{\frac{s_\calX+1}{d_\calX}} \left\| D_\theta^\beta g(\cdot, \theta) \right \|_{s_\calX,2}
\end{align}
\endgroup
holds with probability at least $ 1 - 4e^{-\tau}$, for a $\mathfrak{C}$ that only depends on $\calX, G_{0, \calX}, G_{1, \calX}$. 
From Assumption 
\ref{as:equivalence}, we know that $L_2(\Pb_\theta) \cong L_2(\calX)$ (they are norm equivalent) and $\|f\|_{L_2(\calX)} \leq \text{Vol}(\calX)^{-1} G_{0, \calX}^{-1} \|f\|_{L_2(\Pb_\theta)}$ for any $f \in L_2(\Pb_\theta)$. 
Therefore, for any $\theta \in \Theta$ and any $|\beta| \leq s_\Theta$, with probability at least $ 1 - 4e^{-\tau}$,
\begin{align}\label{eq:D_g_D_g_hat_lebesgue}
    \left\| D_\theta^\beta g(\cdot, \theta)
    - D_\theta^\beta \hat{g}(\cdot, \theta; x_{1:N}^{(\theta)}) \right\|_{L_2(\calX)}
    &\leq \mathfrak{C} \tau \text{Vol}(\calX)^{-1} G_{0, \calX}^{-1}
    N^{-\frac{s_\calX}{d_\calX}} \nonumber \\
    &\quad \times (\log N)^{\frac{s_\calX+1}{d_\calX}}
    \left\| D_\theta^\beta g(\cdot, \theta) \right\|_{s_\calX,2} \nonumber \\
    &\leq \mathfrak{C} \tau \text{Vol}(\calX)^{-1} G_{0, \calX}^{-1}
    N^{-\frac{s_\calX}{d_\calX}} (\log N)^{\frac{s_\calX+1}{d_\calX}} S_1.
\end{align}
By plugging \eqref{eq:D_g_D_g_hat_lebesgue} into \eqref{eq:g_g_hat_sobolev}, and then plugging the result into \eqref{eq:hat_J_J}, we get that with probability at least $ 1 - 4e^{-\tau}$, 
\begin{align}\label{eq:hat_j_j_sobolev}
    \left\| \hat{J}_{\kq}(\cdot; x_{1:N}^{(\theta)}) - J \right\|_{s_\Theta, 2}
    &\leq \left(\sum_{|\beta| \leq s_\Theta } 1 \right) \mathfrak{C} \tau
    G_{0, \calX}^{-1} \text{Vol}(\calX)^{-1} S_1 S_3 \nonumber \\
    &\quad \times N^{-\frac{s_\calX}{d_\calX}}
    (\log N)^{\frac{s_\calX+1}{d_\calX}} \nonumber \\
    &= \binom{s_\Theta + d_\Theta -1}{d_\Theta -1} \mathfrak{C} \tau
    G_{0, \calX}^{-1} \text{Vol}(\calX)^{-1} S_1 S_3 \nonumber \\
    &\quad \times N^{-\frac{s_\calX}{d_\calX}}
    (\log N)^{\frac{s_\calX+1}{d_\calX}}.
\end{align}
By combining this result with the bound $\left\| J \right\|_{s_\Theta, 2} \leq \text{Vol}(\calX)  S_2 S_3$ proven in \eqref{eq:J}, we get that with probability at least $ 1 - 4e^{-\tau}$ and any $N>N_0$ it holds that
\begin{align}\label{eq:hat_J_sobolev_norm}
    \left\| \hat{J}_{\kq}(\cdot; x_{1:N}^{(\theta)}) \right\|_{s_\Theta, 2} &\leq \left\| \hat{J}_{\kq}(\cdot; x_{1:N}^{(\theta)}) - J \right\|_{s_\Theta, 2} + \left\| J \right\|_{s_\Theta, 2} \nonumber \\
    &\leq \binom{s_\Theta + d_\Theta -1}{d_\Theta -1} \mathfrak{C} \tau
    G_{0, \calX}^{-1} \text{Vol}(\calX)^{-1} S_1 S_3 \nonumber \\
    &\quad \times N^{-\frac{s_\calX}{d_\calX}}
    (\log N)^{\frac{s_\calX+1}{d_\calX}}
    + \text{Vol}(\calX) S_2 S_3 \nonumber \\
    &\leq 2 \text{Vol}(\calX)  S_2 S_3,
\end{align}
where $N_0$ is defined as the smallest integer for which the first term is subsumed by the second term.

\underline{\emph{Step Four.}} 
In this step, we are going to upper bound the Sobolev norm of $\bar{F}_{\kq}$.
From Chapter 5, Exercise 16 of \cite{evans2022partial}, we have 
$\hat{F}_{\kq} = f \circ \hat{J}_{\kq} $ is in $W_2^{s_\Theta} (\Theta)$ because $f$ has bounded derivatives up to including $s_\Theta+1$ and $\|\hat{J}(\cdot; x_{1:N}^{(\theta)})\|_{s_\Theta, 2} \leq \text{Vol}(\calX)  S_2 S_3$ with probability at least $1- 4e^{-\tau}$ proved in \eqref{eq:hat_J_sobolev_norm}. 
Hence, $\|\hat{F}_{\kq}(\cdot; x_{1:N}^{(\theta)})\|_{s_\Theta, 2} \leq C_6$ holds with probability at least $1- 4e^{-\tau}$.
Next, recall the definition of $\bar{F}_{\kq}(\theta)$ in \eqref{eq:bar_F_KQ_all_theta},
\begin{align*}
    \bar{F}_{\kq}(\theta)
    &= \mathbb{E}_{x_{1: N}^{(\theta)} \sim \mathbb{P}_\theta}
    \left[\hat{F}_{\mathrm{KQ}}\left(\theta ; x_{1: N}^{(\theta)}\right)\right] \\
    &= \int_{\calX} \cdots \int_{\calX}
    \hat{F}_{\kq}(\theta; x_{1:N}^{(\theta)})
    \prod_{n=1}^N p(x_n^{(\theta)}; \theta)
    d x_1^{(\theta)} \cdots d x_N^{(\theta)} .
\end{align*}
For any $i = 1, \ldots, N$, we know that $\| p(x_i^{(\theta)}; \cdot) \|_{s_\Theta, 2} \leq S_3$ from Assumption \ref{as:app_true_J_smoothness} and $\| \hat{F}_{\kq}(\cdot; x_{1:N}^{(\theta)}) \|_{s_\Theta, 2} \leq C_6$ proved above.
Therefore, from \Cref{lem:sobolev_algebra} we have $\| p(x_i^{(\theta)}; \cdot) \hat{F}_{\kq}(\cdot; x_{1:N}^{(\theta)})\|_{s_\Theta, 2}$ is bounded, so $x_i^{(\theta)} \mapsto p(x_i^{(\theta)}; \cdot) \hat{F}_{\kq}(\cdot; x_{1:N}^{(\theta)})$ is Bochner integrable with respect to the Lebesgue measure $\calL_\calX$. 
From \Cref{lem:integral_in_hilbert}, we have,
\begin{align}\label{eq:bar_F_norm}
    \left\| \bar{F}_{\kq} \right\|_{s_\Theta, 2}
    &\leq \int_{\calX} \cdots \int_{\calX}
    \left\| \hat{F}_{\kq}(\cdot; x_{1:N}^{(\theta)})
    \prod_{n=1}^N p(x_n^{(\theta)}; \cdot) \right\|_{s_\Theta, 2}
    d x_{1:N}^{(\theta)} \nonumber \\
    &\leq \int_{\calX} \cdots \int_{\calX}
    \left\| \hat{F}_{\kq}(\cdot; x_{1:N}^{(\theta)}) \right\|_{s_\Theta, 2}
    \left( \prod_{n=1}^N \| p(x_n^{(\theta)}; \cdot) \|_{s_\Theta, 2} \right)
    d x_{1:N}^{(\theta)} \nonumber \\
    &\leq C_6 S_3^N \text{Vol}(\calX)^N \nonumber \\
    &\leq C_6 .
\end{align}
The last inequality holds by $S_3\leq1$ from Assumption \ref{as:app_true_J_smoothness} and $\calX=[0,1]^{d_\calX}$ so $\text{Vol}(\calX)=1$.

\underline{\emph{Step Five.}} 
We are now ready to upper bound the stage II error, which was defined as
\begingroup\small
\begin{align*}
    \text {Stage II error }
    &= \left\|
    \begin{aligned}
    &\bar{F}_{\kq}(\cdot) - k\left(\cdot, \theta_{1: T}\right) \\
    &\quad {}\times
    \left(k_{\Theta}\left(\theta_{1: T}, \theta_{1: T}\right)
    +T \lambda_{\Theta} \Id_T\right)^{-1}
    \hat{F}_{\kq}\left(\theta_{1: T}\right)
    \end{aligned}
    \right\|_{L_2(\Qb)} .
\end{align*}
\endgroup
The idea is to treat the stage II error as the generalization error of kernel ridge regression---which can be bounded via \Cref{prop:krr_all}. Given i.i.d. observations $(\theta_1, \hat{F}_{\kq}(\theta_1, x_{1:N}^{(\theta_1)}) ), \ldots, (\theta_T, \hat{F}_{\kq}(\theta_T, x_{1:N}^{(\theta_T)}) )$, the target of interest in the context of regression is the conditional mean, which in our case is precisely $ \bar{F}_{\kq}(\theta) = \E_{x_{1:N}^{(\theta)} \sim \Pb_\theta} \hat{F}_{\kq} (\theta; x_{1:N}^{(\theta)})$ defined in \eqref{eq:bar_F_KQ_all_theta}. 
Alternatively, $\hat{F}_{\kq}(\theta; x_{1:N}^{(\theta)})$ can be treated as noisy observation of the target function $\bar{F}_{\kq}(\theta)$ where the observation noise is defined as $ r: \Theta \to \R $ with $r(\theta;x_{1:N}^{(\theta)}) = \hat{F}_{\kq}(\theta; x_{1:N}^{(\theta)}) - \bar{F}_{\kq}(\theta)$. So we automatically have $\E_{x_{1:N}^{(\theta)} \sim \Pb_{\theta} }[r(\theta)] = 0$. For any positive integer $m \geq 2$,
\begingroup\small
\begin{align}\label{eq:bernstein_noise}
    \E_{x_{1:N}^{(\theta)} \sim \Pb_{\theta}} [|r(\theta)|^m] &= \E_{x_{1:N}^{(\theta)} \sim \Pb_{\theta}} \left| \hat{F}_{\kq}(\theta ; x_{1:N}^{(\theta)} ) - \bar{F}_{\kq}(\theta) \right|^m \nonumber \\
    &\stackrel{(i)}{\leq} 2^{m-1} \E_{x_{1:N}^{(\theta)} \sim \Pb_{\theta}} \left| \hat{F}_{\kq}(\theta ; x_{1:N}^{(\theta)} ) - F(\theta) \right|^m \nonumber\\
    &\quad + 2^{m-1} \left| \bar{F}_{\kq}(\theta) - F(\theta) \right|^m \nonumber \\
    &= 2^{m-1} \E_{x_{1:N}^{(\theta)} \sim \Pb_{\theta}} \left| \hat{F}_{\kq}(\theta ; x_{1:N}^{(\theta)}) - F(\theta) \right|^m \nonumber\\
    &\quad + 2^{m-1} \left| \E_{x_{1:N}^{(\theta)} \sim \Pb_{\theta}} \hat{F}_{\kq}(\theta; x_{1:N}^{(\theta)}) - F(\theta) \right|^m \nonumber \\
    &\leq 2^{m-1} \E_{x_{1:N}^{(\theta)} \sim \Pb_{\theta}} \left| \hat{F}_{\kq}(\theta; x_{1:N}^{(\theta)}) - F(\theta) \right|^m \nonumber\\
    &\quad + 2^{m-1} \E_{x_{1:N}^{(\theta)} \sim \Pb_{\theta}} \left| \hat{F}_{\kq}(\theta; x_{1:N}^{(\theta)}) - F(\theta) \right|^m \nonumber \\
    &= 2^m \E_{x_{1:N}^{(\theta)} \sim \Pb_{\theta}} \left| \hat{F}_{\kq}(\theta; x_{1:N}^{(\theta)}) - F(\theta) \right|^m \nonumber \\
    &\leq 2^m S_4^m \E_{x_{1:N}^{(\theta)} \sim \Pb_{\theta}} \left| \hat{J}_{\kq}(\theta; x_{1:N}^{(\theta)}) - J(\theta) \right|^m \nonumber \\
    &\stackrel{(ii)}{\leq} 2^m m! S_4^m S_1^m \mathfrak{C}^m N^{-m\frac{s_\calX}{d_\calX}} (\log N)^{m\frac{s_\calX+1}{d_\calX}} .
\end{align}
\endgroup
In the above chain of derivations, $(i)$ holds because $(a+ b)^m \leq 2^{m-1}(a^m + b^m)$. $(ii)$ holds because we know from \eqref{eq:F_L_g} and \eqref{eq:high_prob_hat_g_g} that $| \hat{J}_{\kq}(\theta; x_{1:N}^{(\theta)}) - J(\theta) | \leq \mathfrak{C} \tau N^{-\frac{s_\calX}{d_\calX}} (\log N)^{\frac{s_\calX+1}{d_\calX}} S_1$ holds with probability at least $1 - 4 e^{-\tau}$, and so $\E_{x_{1:N}^{(\theta)} \sim \Pb_{\theta}} | \hat{J}_{\kq}(\theta; x_{1:N}^{(\theta)}) - J(\theta)|^m$ can be bounded via \Cref{lem:prob_to_expectation}. Therefore, by comparing \eqref{eq:bernstein_noise} with \eqref{eq:mom}, we can see that the observation noise $r$ indeed satisfy the Bernstein noise moment condition with 
\begin{align*}
    \sigma = L = 2 S_4 S_1 \mathfrak{C} N^{-\frac{s_\calX}{d_\calX}}(\log N)^{\frac{s_\calX+1}{d_\calX}} = C_7 N^{-\frac{s_\calX}{d_\calX}} (\log N)^{\frac{s_\calX+1}{d_\calX}},
\end{align*} 
for $C_7 := 2 S_4 S_1 \mathfrak{C}$ a constant independent of $N,T$.
Before we employ \Cref{prop:krr_all}, we need to check the Assumptions \ref{as:kernel}---\ref{as:noise}. Assumption \ref{as:kernel} is satisfied for our choice of kernel $k_\Theta$. Assumption \ref{as:density} is satisfied due to Assumption \ref{as:equivalence}. 
Assumption \ref{as:bayes_predictor} is satisfied due to \eqref{eq:bar_F_norm}. Assumption \ref{as:noise} is satisfied for the Bernstein noise moment condition verified above.
Next, we compute all the constants in \Cref{prop:krr_all} in the current context. 
$\calN(\lambda_\Theta)$ is the effective dimension defined in \Cref{lem:dof} upper bounded by $D_\Theta \lambda_\Theta^{- d_\Theta / 2s_\Theta}$, $k_{\alpha}$ with $\alpha = \frac{2s_\Theta}{d_\Theta}$ defined in \Cref{lem:embedding} is upper bounded by a constant $M_\Theta$, 
$\| \Sigma_{\Qb}\|$ is the norm of the covariance operator defined in \eqref{eq:covariance_operator}.  Hence 
\begin{align*}
    g_{\lambda_\Theta} &:=\log \left(2 e \mathcal{N}(\lambda_\Theta) \frac{ \| \Sigma_{\Qb}\| + \lambda_\Theta }{ \| \Sigma_{\Qb}\| } \right), \qquad A_{\lambda_\Theta, \tau} := 8 k_{\alpha}^2 \tau g_{\lambda_\Theta} \lambda_\Theta^{-\frac{d_\Theta}{2s_\Theta} },  \\
    L_{\lambda_\Theta} &:= \max \left \{ L, \lambda_\Theta^{\frac{1}{2} - \frac{d_\Theta}{4s_\Theta} } \left( \| \bar{F}_{\kq} \|_{L_\infty(\Qb)} + k_{\alpha} \| \bar{F}_{\kq} \|_{s_\Theta, 2} \right) \right\} .
\end{align*}
Applying \Cref{prop:krr_all} shows that, for $T > A_{\lambda_\Theta, \tau}$, 
\begingroup\small
\begin{align}\label{eq:stage_ii_1}
     & \left\|\bar{F}_{\kq}-k\left(\cdot, \theta_{1: T}\right)\left(k_{\Theta}\left(\theta_{1: T}, \theta_{1: T}\right) + T \lambda_{\Theta} \Id_T\right)^{-1} \hat{F}_{\kq}\left(\theta_{1: T}\right)\right\|_{L_2(\Qb)}^2 \nonumber \\
     &\quad \leq \frac{576 \tau^2}{T}
     \left(
     L^2 D_\Theta \lambda_\Theta^{ -\frac{d_\Theta}{2s_\Theta} }
     + M_\Theta^2 \lambda_\Theta^{1 - \frac{d_\Theta}{2s_\Theta}} \left\| \bar{F}_{\kq} \right \|_{s_\Theta, 2}^2
     + 2 M_\Theta^2 \frac{L_{\lambda_\Theta}^2}{T} \lambda_\Theta^{-\frac{d_\Theta}{2s_\Theta}}
     \right) \nonumber\\
     &\qquad + \left\| \bar{F}_{\kq} \right\|_{s_\Theta, 2}^2 \lambda_\Theta ,
\end{align}
\endgroup
holds with probability at least $1 - 4 e^{-\tau}$.
We take $\lambda_\Theta \asymp T^{-2 \frac{s_\Theta}{d_\Theta}} (\log T)^{\frac{2s_\Theta+2}{d_\Theta}}$, then similar to the derivations from \eqref{eq:ratio_N_A},
\begin{align}\label{eq:T_large_enough}
    \lim_{T \to \infty} \frac{A_{\lambda_\Theta, \tau}}{T} \leq \lim_{T \to \infty} 16 (\log T)^{- \frac{s_\Theta+1}{s_\Theta}} k_{\alpha}^2 \tau \log \left( T\right) = 0 .
\end{align}
It means there exists a finite $T_0 > 0$ such that $T > A_{\lambda_\Theta, \tau}$ holds for any $T > T_0$. Notice that, with probability at least $1-4e^{-\tau}$,
\begin{align}\label{eq:C6_use}
    \| \bar{F}_{\kq} \|_{L_\infty(\Qb)} = \| \bar{F}_{\kq} \|_{L_\infty(\Theta)} \leq R_\Theta \| \bar{F}_{\kq} \|_{s_\Theta, 2} \leq R_\Theta C_6
\end{align} 
based on \eqref{eq:bar_F_norm} and the fact that $W_2^{s_\Theta}(\Theta) \hookrightarrow L_\infty(\Theta)$ with $\|W_2^{s_\Theta}(\Theta) \hookrightarrow L_\infty(\Theta)\| \leq R_\Theta$, we have
\begin{align*}
    L_{\lambda_\Theta} &\leq \max \{ L, T^{-\frac{s_\Theta}{d_\Theta} + \frac{1}{2}} (\log T)^{\frac{s_\Theta+1}{2s_\Theta} \frac{2s_\Theta-d_\Theta}{d_\Theta}} \left( R_\Theta + M_\Theta \right) C_6  \} \\
    &= \max \{ C_7 N^{-\frac{s_\calX}{d_\calX}} (\log N)^{\frac{s_\calX+1}{d_\calX}}, T^{-\frac{s_\Theta}{d_\Theta} + \frac{1}{2}} (\log T)^{\frac{s_\Theta+1}{2s_\Theta} \frac{2s_\Theta-d_\Theta}{d_\Theta}} \left( R_\Theta + M_\Theta \right) C_6 \}.
\end{align*}
So the above \eqref{eq:stage_ii_1} can be further upper bounded by 
\begin{align*}
    &\leq \frac{576 \tau^2}{T}
    \left(
    C_7^2 N^{- 2 \frac{s_\calX}{d_\calX}}
    (\log N)^{\frac{2s_\calX+2}{d_\calX}}
    D_\Theta T(\log T)^{-\frac{s_\Theta+1}{s_\Theta}}
    \right.\\
    &\hspace{45pt}\left.
    + M_\Theta^2 T^{- \frac{ 2 s_\Theta}{d_\Theta} + 1}
    (\log T)^{\frac{s_\Theta+1}{s_\Theta} \frac{2s_\Theta-d_\Theta}{d_\Theta}}
    C_6^2
    \right) \\
    &\quad\quad + \frac{576 \tau^2}{T} \cdot 2 M_\Theta^2
    \max \left \{
    C_7^2 N^{-2 \frac{s_\calX}{d_\calX}}
    (\log N)^{\frac{2s_\calX+2}{d_\calX}},
    \right.\\
    &\hspace{70pt}\left.
    T^{-2 \frac{s_\Theta}{d_\Theta} + 1}
    (\log T)^{\frac{s_\Theta+1}{s_\Theta} \frac{2s_\Theta-d_\Theta}{d_\Theta}}
    \left( R_\Theta + M_\Theta\right)^2 C_6^2
    \right\}
    (\log T)^{-\frac{s_\Theta+1}{s_\Theta}} \\
    &\qquad\qquad\qquad + C_6^2  T^{- 2 \frac{ s_\Theta}{d_\Theta}} (\log T)^{\frac{2s_\Theta+2}{d_\Theta}} \\
    &= 576 \tau^2
    \left(
    C_7^2 N^{- 2 \frac{s_\calX}{d_\calX}}
    (\log N)^{\frac{2s_\calX+2}{d_\calX}}
    D_\Theta(\log T)^{-\frac{s_\Theta+1}{s_\Theta}}
    \right.\\
    &\hspace{45pt}\left.
    + M_\Theta^2 T^{- \frac{ 2 s_\Theta}{d_\Theta}}
    (\log T)^{\frac{s_\Theta+1}{s_\Theta} \frac{2s_\Theta-d_\Theta}{d_\Theta}}
    C_6^2
    \right) \\
    &\quad\quad + 576 \tau^2 \cdot 2 M_\Theta^2
    \max \left \{
    C_7^2 N^{-2 \frac{s_\calX}{d_\calX}}
    (\log N)^{\frac{2s_\calX+2}{d_\calX}} T^{-1},
    \right.\\
    &\hspace{70pt}\left.
    T^{-\frac{2s_\Theta}{d_\Theta}}
    (\log T)^{\frac{s_\Theta+1}{s_\Theta} \frac{2s_\Theta-d_\Theta}{d_\Theta}}
    \left( R_\Theta + M_\Theta\right)^2 C_6^2
    \right\}
    (\log T)^{-\frac{s_\Theta+1}{s_\Theta}} \\
    &\qquad\qquad+ C_6^2  T^{- 2 \frac{ s_\Theta}{d_\Theta}} (\log T)^{\frac{2s_\Theta+2}{d_\Theta}} \\
    &\stackrel{(i)}{\leq} 576 \tau^2 C_7^2 N^{- 2 \frac{s_\calX}{d_\calX}} (\log N)^{\frac{2s_\calX+2}{d_\calX}} D_\Theta + 576 \tau^2 M_\Theta^2 T^{- \frac{ 2 s_\Theta}{d_\Theta}} (\log T)^{\frac{2s_\Theta+2}{d_\Theta}} C_6^2 \\
    &\quad\quad + 576 \tau^2 \cdot 2 M_\Theta^2 C_7^2
    N^{-2 \frac{s_\calX}{d_\calX}} (\log N)^{\frac{2s_\calX+2}{d_\calX}} \\
    &\quad\quad + 576 \tau^2 \cdot 2 M_\Theta^2
    T^{-\frac{2s_\Theta}{d_\Theta}}
    (\log T)^{\frac{2s_\Theta+2}{d_\Theta}}
    \left( R_\Theta + M_\Theta\right)^2 C_6^2 \\
    &\quad\quad + C_6^2
    T^{- 2 \frac{ s_\Theta}{d_\Theta}}
    (\log T)^{\frac{2s_\Theta+2}{d_\Theta}}\\
    &=: \tau^2 \big(
    C_8^2 N^{- \frac{ 2 s_\calX}{d_\calX}}
    (\log N)^{\frac{2s_\calX+2}{d_\calX}}
    \\
    &\quad
    + C_9^2 T^{- \frac{ 2 s_\Theta}{d_\Theta}}
    (\log T)^{\frac{2s_\Theta+2}{d_\Theta}}
    \big).
\end{align*}
$C_8, C_9$ are two constants independent of $N,T$.
In $(i)$, we use $\max\{ a_1, a_2\}\leq a_1 + a_2$, we also use the following
\begin{align*}
    (\log T)^{\frac{s_\Theta+1}{s_\Theta} \frac{2s_\Theta-d_\Theta}{d_\Theta}}
    &\leq
    (\log T)^{\frac{2s_\Theta+2}{d_\Theta}},\\
    (\log T)^{-\frac{s_\Theta+1}{s_\Theta}}
    &\leq 1 .
\end{align*}
Therefore, we have that, 
\begin{align}\label{eq:stage_2}
    \text{Stage II error} &:= \left\| \bar{F}_{\kq} - k(\cdot, \theta_{1:T}) (k_\Theta(\theta_{1:T}, \theta_{1:T}) + T \lambda_\Theta \Id_T )^{-1} 
    \hat{F}_{\kq}(\theta_{1:T}) \right\|_{L_2( \Qb ) } \nonumber \\
    &\leq \tau \left( C_8 N^{- \frac{ s_\calX}{d_\calX}} (\log N)^{\frac{s_\calX+1}{d_\calX}} + C_9 T^{- \frac{ s_\Theta}{d_\Theta}} (\log T)^{\frac{s_\Theta+1}{d_\Theta}} \right) ,
\end{align}
holds with probability at least $1 - 8 e^{-\tau}$.
\paragraph{Combine stage I and stage II error}
Combining the stage I error of \eqref{eq:stage_1} and the stage II error of \eqref{eq:stage_2}, we obtain
\begin{align*}
    \left| I - \hat{I}_{\nkq} \right| &\leq \text{Stage I error} + \text{Stage II error} \\
    &\leq C_3 N^{-\frac{ s_\calX}{d_\calX}} (\log N)^{\frac{s_\calX+1}{d_\calX}}
    + \tau \left( C_8 N^{- \frac{ s_\calX}{d_\calX} } (\log N)^{\frac{s_\calX+1}{d_\calX}} \right. \\
    &\hspace{100pt}\left. {}
    + C_9 T^{- \frac{ s_\Theta}{d_\Theta}} (\log T)^{\frac{s_\Theta+1}{d_\Theta}} \right) \\
    &\leq \tau \left( (C_8 + C_3) N^{- \frac{ s_\calX}{d_\calX} }
    (\log N)^{\frac{s_\calX+1}{d_\calX}} \right. \\
    &\hspace{100pt}\left. {}
    + C_9 T^{- \frac{ s_\Theta}{d_\Theta} } (\log T)^{\frac{s_\Theta+1}{d_\Theta}} \right) \\
    &=: \tau \left( C_1 N^{- \frac{ s_\calX}{d_\calX} }
    (\log N)^{\frac{s_\calX+1}{d_\calX}} \right. \\
    &\hspace{100pt}\left. {}
    + C_2 T^{- \frac{ s_\Theta}{d_\Theta} } (\log T)^{\frac{s_\Theta+1}{d_\Theta}} \right),
\end{align*}
holds with probability at least $1 - 8 e^{-\tau}$.
Here $C_1, C_2$ are two constants independent of $N,T$ so the proof concludes here.

\section{Further Background and Auxiliary Lemmas}\label{appendix:aux_lemmas}
All the results in this section are existing results in the literature. We provide them here and prove some of them in the specific context of Sobolev spaces explicitly for the convenience of the reader.

\paragraph{More technical notions of Sobolev spaces and the Sobolev embedding theorem}
In the main text, we provide in \eqref{eq:defi_sobolev} the standard definition of Sobolev spaces $W_2^s( \calX )$ when $s \in \N$.
Actually, Sobolev spaces $W_2^s( \calX )$
can be extended to $s$ that are positive real numbers. 
Such extension could be realized through 
real interpolation spaces (see \citep[Definition 1.7]{bennett1988interpolation}),
$W_2^{s}(\calX) := [W_2^{k}(\calX), W_2^{k+1}(\calX)]_{r, 2}$ where $k \in \mathbb{N}, s \in (k, k+1), r = s-\lfloor s\rfloor$.\footnote{Strictly speaking, the definition of \eqref{eq:defi_sobolev} extended to real numbers $s$ actually corresponds to the complex interpolation space of Sobolev spaces. Fortunately, complex interpolation spaces and real interpolation spaces coincide under Hilbert spaces~\citep[Corollary C.4.2]{hytonen2016analysis}, which is precisely our setting since $p=2$.}
Actually, such interpolation relations hold for any $0 \leq s , t$ and $0 < r < 1$~\citep[Section 7.32]{adams2003sobolev},
\begin{align}\label{eq:interpolation}
    W_{2}^k(\calX) = \left[W_2^s(\calX), W_{2}^t(\calX)\right]_{r, 2}, \quad k = (1-r) s + r t .
\end{align}
A special case of the above relation is $ W_{2}^s(\calX) = \left[L_2(\calX), W_{2}^t(\calX)\right]_{s/t, 2} $. 

The Sobolev embedding theorem~\citep{adams2003sobolev}, when applied to $W_2^s(\calX)$, states that if $s>\frac{d}{2}$ (where $d$ is the dimension of $\calX)$, then $W_2^s(\calX)$ can be continuously embedded into $C^0(\calX)$, the space of continuous and bounded functions. In other words, for every equivalence class $[f] \in W_2^s(\calX)$, there exists a unique continuous and bounded representative $f \in C^0(\calX)$, and the embedding map $I$ : $W_2^s(\calX) \rightarrow C^0(\calX)$, defined by $I([f])=f$, is continuous.
This continuous embedding $I$ can be written as $W_2^s(\calX) \hookrightarrow C^0(\calX)$. 
Since every continuous linear operator is bounded, we have $\| W_2^s(\calX) \hookrightarrow C^0(\calX)\|$ bounded by a constant that only depends on $s, \calX$.

\paragraph{More technical notions of reproducing kernel Hilbert spaces (RKHSs)}
For bounded kernels, $\sup_{x\in\calX} k(x, x) \leq \kappa$, its associated RKHS $\mathcal H$ can be canonically injected into $L_2(\Qb)$ using the operator $\iota_{\Qb} : \calH \to L_2(\Qb),\,f\mapsto f$ with its adjoint $ \iota_\Qb^\ast: L_2(\Qb) \rightarrow \calH$ given by $\iota_\Qb^\ast f(\cdot) = \int k(x,\cdot)f(x)d\Qb(x)$.
$\iota_{\Qb}$ and its adjoint can be composed to form a $L_2(\Qb)$ endomorphism 
$\calT_{\Qb} \coloneqq \iota_{ \Qb} \iota_{ \Qb}^\ast$ 
called the \emph{integral operator}, and a $\mathcal H$ endomorphism  
\begin{align}\label{eq:covariance_operator}
    \Sigma_{\Qb} \coloneq \iota_{ \Qb}^\ast \iota_{ \Qb}=\int k(\cdot, x) \otimes k(\cdot, x) d \Qb(x),
\end{align}
(where $\otimes$ denotes the tensor product such that $(a\otimes b)c \coloneqq \langle b,c \rangle_{\calH} a$ for $a,b,c\in \calH$) called the \emph{covariance operator}.
Both $\Sigma_\Qb$ and $\calT_{\Qb}$ are compact, positive, self-adjoint, and they have the same eigenvalues $\varrho_1 \geq \cdots \varrho_i \geq \cdots \geq 0$. Please refer to Section 2 of \citet{chen2024regularized} for more details. 

\begin{lem}[Effective dimension $\calN(\lambda)$]\label{lem:dof}
Let $\calX \subset \R^d$ be a compact domain, $\Qb$ be a probability measure on $\calX$ with density $p:\calX \to \R$. $k : \calX \times \calX \to \R$ is a Sobolev reproducing kernel of order $s > \frac{d}{2}$. 
$\{\varrho_m \}_{m \geq 0}$ are the eigenvalues of the integral operator $\calT_\Qb$.
Define the effective dimension $\mathcal{N}:(0, \infty) \rightarrow[0, \infty)$ as $\mathcal{N}(\lambda) \coloneq \sum_{m \geq 1} \frac{\varrho_m}{\varrho_m+\lambda}$. 
If $p(x) \geq G > 0$ for any $x \in \calX$, then $ \mathcal{N}(\lambda) \leq D \lambda^{- \frac{d}{2s} }$ with constant $D$ that only depends on $G$ and $\calX$.
\end{lem}
\begin{proof}
First, we study the asymptotic behavior of the eigenvalues $\left(\varrho_m \right)_{m \geq 1}$ of the integral operator $\calT_\Qb$. Theorem 15 of \cite{steinwart2009optimal} shows that the eigenvalues $\varrho_m$ share the same asymptotic decay rate as the squares of the entropy number $e_m^2\left( I_\Qb \right)$ of the embedding $I_\Qb: W_2^s(\calX) \rightarrow L_2(\Qb)$. 
Denote $\calL_\calX$ as the Lebesgue measure on $\calX$.
Since $p(x) \geq G$ for any $x \in \calX$, we know $\frac{d \calL_\calX}{d \Qb} \leq G^{-1} \text{Vol}(\calX)^{-1}$ so $\| L_2(\Qb) \hookrightarrow L_2(\calX) \| \leq G^{-1} \text{Vol}(\calX)^{-1}$, and consequently we have from Equation (A.38) of \citet{steinwart2008support} that
\begin{align*}
    e_m \left( I_\Qb \right) \leq e_m \left( I_{\calL_\calX} \right) \| L_2(\Qb) \hookrightarrow L_2(\calX) \| \leq G^{-1} \text{Vol}(\calX)^{-1} e_m \left( I_{\calL_\calX} \right) .
\end{align*}
Moreover, \citep[Equation 4 on p. 119]{edmunds1996function} shows that the entropy number $e_m\left( I_{\calL_\calX} \right) \leq \tilde{c} m^{-s / d}$ for some constant $\tilde{c}$, so we have $e_m \left( I_\Qb \right) \leq G^{-1} \text{Vol}(\calX)^{-1} \tilde{c} m^{-s / d}$ and consequently we have $\varrho_m \asymp e_m^2\left( I_\Qb \right) \leq G^{-2} \text{Vol}(\calX)^{-2} \tilde{c}^2 m^{-2s / d} =: c_2 m^{-2s / d}$.

Next, we have  
\begin{align*}
    \sum_{m \geq 1} \frac{\varrho_m}{\varrho_m+\lambda}
    &\leq \sum_{m \geq 1} \frac{1}{1+\lambda c_2^{-1} m^{2 s / d} }
    \leq \int_0^{\infty} \frac{c_2}{ c_2 + \lambda  t^{2 s / d} } dt \\
    &= \lambda^{- \frac{d}{2s} }
    \int_0^{\infty} \frac{c_2}{ c_2 + \tau^{2 s / d} } d \tau
    = \lambda^{- \frac{d}{2s} }
    \int_0^{\infty}
    \frac{ {c_2}^{\frac{d}{2s}} }{1 + u^{\frac{2s}{d}} } d u \\
    &= \lambda^{- \frac{d}{2s} } {c_2}^{\frac{d}{2s}}
    \frac{ \frac{\pi d}{2s} }{\sin \left( \frac{\pi d}{2s} \right)}
    =: D \lambda^{- \frac{d}{2s} },
\end{align*}
where $D$ is a constant that depends on the domain $\calX$ and $G$.
\end{proof}

\begin{lem}\label{lem:embedding}
Let $\calX \subset \R^d$ be a compact domain, $\Qb$ be a probability measure on $\calX$ with density $p:\calX \to \R$.
$k : \calX \times \calX \to \R$ is a Sobolev reproducing kernel of order $s > \frac{d}{2}$. 
$\{ \varrho_m, e_m \}_{m \geq 0}$ are the eigenvalues and eigenfunctions of the integral operator $\calT_\Qb$.
If there exists $G_0,G_1 > 0$ such that $G_0 \leq p(x) \leq G_1$ for any $x \in \calX$, then 
\begin{align}\label{eq:k_alpha}
    k_{\alpha} \coloneq \sup_{x\in\calX} \sum_{m \geq 1} \varrho_m^{\alpha } e_m^2(x) \leq M ,
\end{align}
holds for any $ \frac{d}{2s} < \alpha$.
Here, $M$ is a constant that depends on $\calX$ and $G_1, G_0$.
\end{lem}
\begin{proof}
If $t > \frac{d}{2}$, $W_2^t(\calX)$ can be continuously embedded into $L_\infty(\calX)$ the space of bounded functions \citep[Case A, Theorem 4.12]{adams2003sobolev}.
Hence, the operator $W_2^s(\calX) \hookrightarrow L_\infty(\calX) $ is a continuous linear operator between two normed vector spaces, hence a bounded operator. 
And $L_2(\Qb) $ is norm equivalent to $ L_2(\calX)$ because $G_0 \leq p(x) \leq G_1$ for any $x \in \calX$. 
Notice that $k_\alpha$ defined here is exactly $\left\|k_\nu^\alpha\right\|_{\infty}$ defined in Equation 16 of \cite{fischer2020sobolev}, so we know from Theorem 9 of \citet{fischer2020sobolev} that 
\begin{align*}
     \sup_{x\in\calX} \sum_{m \geq 1} \varrho_m^{\alpha} e_i^2(x) = \left\|  \left[L_2(\Qb), W_{2}^s(\calX) \right]_{\alpha, 2} \hookrightarrow L_{\infty}(\calX) \right\| .
\end{align*}
Notice that $\left[L_2(\Qb), W_{2}^s(\calX) \right]_{\alpha, 2} \cong \left[L_2(\calX), W_{2}^s(\calX) \right]_{\alpha, 2} \cong W_{2}^{s \alpha}(\calX)$, and notice the fact that $W_2^{s\alpha}(\calX) \hookrightarrow L_\infty(\calX)$ for any $s\alpha > \frac{d}{2}$, the right hand side of the above equation is bounded. Therefore, we have \eqref{eq:k_alpha} holds for any $\frac{d}{2s} < \alpha$.
\end{proof}

\begin{lem}\label{lem:sobolev_algebra}
    Let $\calX \subset \R^d$ be a bounded domain with Lipschitz continuous boundary and $W_2^s(\calX)$ be a Sobolev space with $s > \frac{d}{2}$. If functions $f:\calX \to \R$ and $g:\calX \to \R$ lie in $W_2^s(\calX)$, then their product $f \cdot g$ also lies in $W_2^s(\calX)$ and satisfies $\| f \cdot g \|_s \leq \| f \|_s \| g \|_s$.
\end{lem}
\begin{proof}
This is Theorem 7.4 of \citet{behzadan2021multiplication} with $s_1=s_2=s$ and $p_1=p_2=2$. 
\end{proof}
\begin{lem}\label{lem:prob_to_expectation}
For a positive valued random variable $R$, and $c > 0$ such that $\Pb(R \leq c \tau) \geq 1 - \exp(-\tau)$ for any positive $\tau$, it holds that $\E[R^m] \leq c_o m! $ for all integers $m \geq 1$. $c_o$ is some constant that only depends on $c, m$.
\end{lem}
\begin{proof}
Notice that $R$ is essentially a sub-exponential random variable.
Since a sub-exponential random variable is equivalent to the square root of a sub-Gaussian random variable, from Proposition 2.5.2 of \citet{vershynin2018high}, we have $\E[R^m] = \E[\sqrt R ^{2m}] \leq 2 c_o \Gamma(m+1) = 2 c_o m!$.
Here $\Gamma$ denotes the gamma function and $c_o$ is some constant that only depends on $c, m$.
\end{proof}

\begin{lem}\label{lem:integral_in_hilbert}
    For a mapping $F$ from a compact domain $\calX \subset \R^d$ to a Hilbert space $H$, given a measure $\mu$ on $\calX$, if $F$ is $\mu$-Bochner integrable, then $\int F(x) d\mu(x) \in H$ and additionally $ \| \int F(x) d\mu(x)\|_H \leq \int \| F(x)\|_H d\mu(x)$.
\end{lem}
\begin{proof}
    This is Definition A.5.20 of \citet{steinwart2008support}.
\end{proof}


\endgroup

\chapter{\texorpdfstring{Supplementary Material for \Cref{sec:drmmd}}{Supplementary Material for Minimum MMD for Generative Modelling}}
\label{app:drmmd}
\begingroup
\newcommand{\appsection}[2][]{%
  \refstepcounter{section}%
  \section*{Appendix \thesection: #2}%
  \if\relax\detokenize{#1}\relax
    \addcontentsline{toc}{section}{Appendix \thesection: #2}%
  \else
    \addcontentsline{toc}{section}{Appendix \thesection: #1}%
  \fi
}\section{Proofs} \label{sec:proofs}
In this section, we present the proofs of the results presented in \Cref{sec:drmmd}. 
\subsection{\texorpdfstring{Proof of \Cref{prop:mmd_chi2}}{Proof of the regularized MMD proposition}}\label{appsec:mmd_regularized_chi2}
Note that
\begin{align}
    \mmd(\mu \| \pi)^2 &= \left\| m_\mu - m_\pi \right\|_\calH^2 = \left\| \int k(\cdot, x) d\mu(x) - \int k(\cdot, x) d\pi(x) \right\|_\calH^2 \nonumber \\ 
    &=\left\| \int k(\cdot,x) \left(\frac{d\mu}{d\pi}(x)-1\right)\,d\pi(x) \right\|^2_\calH=\left\|\id^*_\pi \left(\frac{d\mu}{d\pi} - 1\right)\right\|^2_\calH\nonumber\\
    &=\left\langle \mathcal{T}_\pi\left(\frac{d\mu}{d\pi} - 1\right),\frac{d\mu}{d\pi} - 1 \right\rangle_{L^2(\pi)}
    = \left\| \calT_\pi^{\frac{1}{2} } \left(\frac{d \mu}{d \pi}-1\right)\right\|_{L^2(\pi)}^2 .\nonumber
\end{align}
Also recall that the $\chi^2$-divergence between $\mu$ and $\pi$ is
\begin{align*}
    \chi^2(\mu \| \pi) = \int \left( \frac{d\mu}{d\pi} - 1 \right)^2 d\pi = \left\|\mathrm{I}\left(\frac{d \mu}{d \pi}-1\right)\right\|_{L^2(\pi)}^2.\nonumber
\end{align*}

\subsection{\texorpdfstring{Proof of \Cref{prop:drmmd_representation_no_ratio}}{Proof of the DrMMD representation proposition}}\label{appsec:proof_chard_also}
Let $\mu \ll \pi$. 
In order to prove the alternative form of $\dmmd$ in \eqref{eq:drmmd_representation_no_ratio}, we start from \eqref{eq:drmmd_representation_no_ratio} and show that it recovers \eqref{eq:dmmd_with_inverse_operator}.
\begin{align}\label{eq:drmmd_derivations}
&\dmmd(\mu || \pi) = ( 1+\lambda) \left\|\left(\Sigma_\pi + \lambda \Id\right)^{-\frac{1}{2}}\left(m_\mu - m_\pi \right) \right\|_{\calH}^2 \nonumber \\
& =(1+\lambda)\left\|\left(\iota_\pi^\ast \iota_\pi + \lambda \Id \right)^{-\frac{1}{2}} \iota_\pi^\ast \left( \frac{d \mu}{ d \pi} - 1 \right) \right\|_{\calH}^2 \nonumber \\
& =(1+\lambda)\left\langle\left(\iota_\pi^\ast \iota_\pi + \lambda \Id \right)^{-\frac{1}{2}} \iota_\pi^\ast \left( \frac{d \mu}{ d \pi} - 1 \right),\left(\iota_\pi^\ast \iota_\pi + \lambda \Id \right)^{-\frac{1}{2}} \iota_\pi^\ast \left( \frac{d \mu}{ d \pi} - 1 \right)\right\rangle_{\calH}\nonumber \\
& =(1+\lambda)\left\langle \iota_\pi \left(\iota_\pi^\ast \iota_\pi + \lambda \Id \right)^{-1} \iota_\pi^\ast \left( \frac{d \mu}{ d \pi} - 1 \right), \frac{d \mu}{ d \pi} - 1 \right\rangle_{L^2(\pi)} \nonumber \\
&=(1+\lambda)\left\langle \iota_\pi \iota_\pi^\ast\left(\iota_\pi \iota_\pi^\ast+\lambda \Id \right)^{-1} \left( \frac{d \mu}{ d \pi} - 1 \right), \frac{d \mu}{ d \pi} - 1 \right\rangle_{L^2(\pi)},  
\end{align}
where the last equality follows from
\begin{align*}
    \iota_\pi \left( \iota_\pi^* \iota_\pi + \lambda \Id \right)^{-1} \iota_\pi^*
    &= \iota_\pi \iota_\pi^*
    \left( \iota_\pi \iota_\pi^* + \lambda \Id \right)^{-1} .
\end{align*}
Therefore,
\begin{align*}
\quad \dmmd(\mu || \pi) &= (1+\lambda)\left\langle \calT_\pi \left( \calT_\pi + \lambda \Id \right)^{-1} \left( \frac{d \mu}{ d \pi} - 1 \right),  \frac{d \mu}{ d \pi} - 1 \right\rangle_{L^2(\pi)} \\ 
&= (1+\lambda)\left\|\left( \left( \mathcal{T}_\pi+\lambda \mathrm{I}\right)^{-1} \mathcal{T}_\pi\right)^{1 / 2}\left(\frac{d \mu}{d \pi}-1\right)\right\|_{L^2(\pi)}^2,
\end{align*}
which follows from the positivity and self-adjointness of $\calT_\pi \left( \calT_\pi + \lambda \Id \right)^{-1}$.
So \eqref{eq:drmmd_representation_no_ratio} is proved. 
Next, we are going to prove the variational formulation in \eqref{eq:drmmd_variational}. Similarly, we start from \eqref{eq:drmmd_variational} and show it recovers \eqref{eq:dmmd_with_inverse_operator}. Consider
\begin{align}
    &\quad (1+\lambda) \sup _{h \in \calH}\left\{\int h \md \mu-\int\left(\frac{h^2}{4}+h \right) \md \pi - \frac{\lambda}{4}\|h\|_{\calH}^2\right\} \nonumber \\
    & =-(1+\lambda) \inf _{h \in \calH}\left\{\int \frac{h^2}{4} \md \pi - \int h (\md \mu - \md \pi) + \frac{\lambda}{4} \|h\|_{\calH}^2\right\}  \nonumber \\ 
    & = -(1+\lambda) \inf _{h \in \calH}\left\{\frac{1}{4} \left\langle h, \Sigma_\pi h \right \rangle_{\calH} - \left \langle h, m_\mu - m_\pi \right\rangle_{\calH}+\frac{\lambda}{4}\|h\|_{\calH}^2\right\}  \nonumber \\
    &= -(1+\lambda) \inf _{h \in \calH} \left\{\left\|\left(\frac{1}{4} \Sigma_\pi+\frac{\lambda}{4} \Id\right)^{1 / 2} h -\frac{1}{2}\left(\frac{1}{4} \Sigma_\pi+\frac{\lambda}{4} \Id\right)^{-1 / 2} (m_\mu - m_\pi) \right\|_{\calH}^2\right\}  \nonumber \\
    &\qquad\qquad+ \frac{(1+\lambda)}{4}\left\|\left(\frac{1}{4} \Sigma_\pi +\frac{\lambda}{4} \Id\right)^{-1 / 2} (m_\mu - m_\pi) \right\|_{\calH}^2 . \label{appeq:kernel_chi2_supremum}
\end{align}
The last equality follows from completing the squares, based on which it is easy to see that the infimum is achieved at $h_{\mu, \pi}^\ast = 2 \left(\Sigma_\pi + \lambda \Id\right)^{-1}\left( m_\mu - m_\pi \right)$. 
For $\mu \ll \pi$, following the same derivations in \eqref{eq:drmmd_derivations}, $h_{\mu, \pi}^\ast$ can be alternatively expressed as
\begin{align}\label{eq:witness_alternative}
    h_{\mu, \pi}^\ast = 2 \left( \mathcal{T}_\pi+\lambda \mathrm{I}\right)^{-1} \mathcal{T}_\pi \left( \frac{d\mu}{d \pi} - 1 \right)
\end{align}
Plugging $h_{\mu, \pi}^\ast$ back into \eqref{appeq:kernel_chi2_supremum} recovers \eqref{eq:dmmd_with_inverse_operator}, so \eqref{eq:drmmd_variational} is proved.

\subsection{\texorpdfstring{$\dmmd$ is $\mmd$ with a regularized kernel $\tilde{k}$}{DrMMD is MMD with a regularized kernel}}\label{appsec:drmmd_is_mmd}
Define the regularized kernel by
\begin{align*}
    \tilde{k}(x, x^\prime)
    &= \left\langle
    \left(\Sigma_\pi + \lambda \Id\right)^{-\frac{1}{2}} k(\cdot,x),
    \left(\Sigma_\pi + \lambda \Id\right)^{-\frac{1}{2}} k(\cdot,x^\prime)
    \right\rangle_\calH .
\end{align*}
It is clear that $\tilde{k}$ is symmetric and positive definite, so it has a unique associated reproducing kernel Hilbert space $\tilde{\calH}$~\citep[Theorem 4.21]{steinwart2008support} with canonical feature map $\tilde{k}(x, \cdot)$. Write $A_\lambda := (\Sigma_\pi + \lambda \Id)^{-1/2}$. Therefore,
\begingroup\small
\begin{align*}
&\quad \dmmd(\mu \| \pi) \\
& = (1 + \lambda) \left\Vert A_\lambda (m_\mu-m_\pi) \right\Vert^2_\calH \\
&= (1 + \lambda) \left\langle
\begin{aligned}
&A_\lambda \int k(x, \cdot) d\left( \pi - \mu \right)(x), \\
&A_\lambda \int k(x^\prime, \cdot) d\left( \pi - \mu \right)(x^\prime)
\end{aligned}
\right\rangle_\calH \\
&= (1 + \lambda) \left\langle
\begin{aligned}
&\int A_\lambda k(x, \cdot) d\left( \pi - \mu \right)(x), \\
&\int A_\lambda k(x^\prime, \cdot) d\left( \pi - \mu \right) (x^\prime)
\end{aligned}
\right\rangle_\calH \\
&= (1 + \lambda) \iint
\left\langle
\begin{aligned}
&A_\lambda k(x, \cdot), \\
&A_\lambda k(x^\prime, \cdot)
\end{aligned}
\right\rangle_\calH \\
&\qquad \dd \left( \pi- \mu \right) (x) \dd\left( \pi - \mu \right) (x^\prime) \\
&=(1 + \lambda) \iint \tilde{k}(x, x^\prime) \\
&\quad {}\times d\left( \pi- \mu \right) (x)
d\left( \pi - \mu \right) (x^\prime) \\
&= (1 + \lambda) \\
&\quad {}\times
\left\Vert \int \tilde{k}(x, \cdot) d(\mu-\pi)(x) \right\Vert^2_{\tilde{\calH}}.
\end{align*}
\endgroup
In the third and fourth equality above, we are using the fact that $\left(\Sigma_\pi + \lambda \Id\right)^{-\frac{1}{2} } k(x, \cdot) \in \calH$ is Bochner integrable and the Bochner integral preserves inner product structure. So $\dmmd$ is essentially $\mmd^2$ with a different kernel $\tilde{k}$ up to a multiplicative factor of $1 + \lambda$.

Next, we present the Mercer decomposition of $\tilde{k}$. 
Notice that $\{e_i\}_{i \geq 1}$ are the eigenfunctions of $\calT_\pi$, so $\{ \sqrt{\varrho_i} e_i\}_{i \geq 1}$ are the eigenfunctions of $\Sigma_\pi$.
For $x$ and $x^\prime$ in the support of $\pi$, $\tilde{k}$ also enjoys a pointwise convergent Mercer decomposition
\begin{align}\label{appeq:tilde_k_mercer}
    \tilde{k}(x, x^\prime) &= \left\langle \left(\Sigma_\pi + \lambda \Id\right)^{-\frac{1}{2} } k(\cdot,x),\left(\Sigma_\pi + \lambda \Id\right)^{-\frac{1}{2} } k(\cdot,x')\right\rangle_\calH \nonumber \\
    &= \left\langle \left(\Sigma_\pi + \lambda \Id\right)^{-\frac{1}{2} } \left( \sum_{i \geq 1} \varrho_i e_i(x) e_i \right), \left(\Sigma_\pi + \lambda \Id\right)^{-\frac{1}{2} } \left( \sum_{i \geq 1} \varrho_i e_i(x^\prime) e_i \right) \right\rangle_\calH \nonumber \\
    &= \left\langle \sum_{i \geq 1} \frac{\varrho_i}{\sqrt{\varrho_i + \lambda} } e_i(x) e_i , \sum_{i \geq 1} \frac{\varrho_i}{\sqrt{\varrho_i + \lambda} } e_i(x^\prime) e_i \right\rangle_\calH \nonumber \\
    &= \sum_{i \geq 1} \frac{\varrho_i}{\varrho_i+\lambda} e_i(x) e_i(x^{\prime}).
\end{align}
More properties of the regularized kernel $\tilde{k}$ are provided in \Cref{lem:k_tilde}.

\subsection{\texorpdfstring{Proof of \Cref{prop:chard_interpolation}}{Proof of the interpolation proposition}}\label{appsec:proof_chard_interpolation}
Given that $\frac{d \mu}{ d \pi} - 1 \in L^2(\pi)$, so 
\begingroup\small
\begin{align*}
    \dmmd(\mu \| \pi)
    &= (1+\lambda) \left\|\left(\left(\mathcal{T}_\pi + \lambda \mathrm{I} \right)^{-1} \mathcal{T}_\pi\right)^{1 / 2}
    \left(\frac{d \mu}{d \pi}-1\right)\right\|_{L^2(\pi)}^2 \\
    &\leq (1+\lambda) \left\|\frac{d \mu}{ d \pi} - 1 \right\|_{L^2(\pi)}^2,
\end{align*}
\endgroup
which is finite for any $\lambda \geq 0$. 
We are allowed to interchange the limit and integration according to the dominated convergence theorem~\citep{rudin1976principles} to achieve, 
\begin{align*}
    \lim_{\lambda \to 0} \dmmd(\mu \| \pi) &=  \left\| \lim_{\lambda \to 0} \left(\left(\mathcal{T}_\pi + \lambda \mathrm{I} \right)^{-1} \mathcal{T}_\pi\right)^{1 / 2}\left(\frac{d \mu}{d \pi}-1\right)\right\|_{L^2(\pi)}^2 \\
    &= \left\|  \frac{d \mu}{ d \pi} - 1 \right\|_{L^{2}(\pi)}^{2} = \chi^{2}(\mu \| \pi).
\end{align*}
From \eqref{eq:drmmd_representation_no_ratio}, we have that,
\begin{align}
    &\sqrt{\dmmd (\mu \| \pi)} 
    = \sqrt{1+\lambda} \left\|\left(\Sigma_{\pi} + \lambda \Id\right)^{-\frac{1}{2} }(m_{\mu} - m_{\pi})  \right\|_{\calH} \nonumber \\
    &\leq \sqrt{1+\lambda} \left\|\left(\Sigma_{\pi} + \lambda \Id\right)^{-\frac{1}{2} }\right\|_{op}\left\|m_{\mu} - m_{\pi}  \right\|_{\calH}
    \leq \sqrt{\frac{1+\lambda}{\lambda}} \times \mmd (\mu,\pi) , \label{appeq:dmmd_one_tide}
\end{align}
where the last inequality follows by noticing that $\Sigma_\pi = \iota_\pi^\ast \iota_\pi$ shares the same 
eigenvalues as $\calT_\pi = \iota_\pi \iota_\pi^\ast$, and hence the eigenvalues of $(\Sigma_\pi + \lambda \Id)^{-1}$ are ${(\varrho_i + \lambda)}^{-1}$ which all smaller than $\frac{1}{\lambda}$.
Therefore, $\dmmd(\mu \| \pi) \leq \frac{1+\lambda}{\lambda} \mmd^{2}(\mu \| \pi)$. 

On the other hand, using Lemma A.10 from \citep{hagrass2022spectral}, we have
\begin{align}
    &\mmd \left(\mu \| \pi \right) = \left\|m_\mu - m_\pi \right\|_{\calH} 
    \leq \left\| ( \Sigma_\pi+\lambda \Id\right ) ^{\frac{1}{2} } \|_{o p} \left\|\left(\Sigma_{\pi} + \lambda \Id\right)^{-\frac{1}{2} }(m_{\mu} - m_{\pi})  \right\|_{\calH} \nonumber  \\
    &= \sqrt{ \frac{1}{1+\lambda} }  \left\| ( \Sigma_\pi+\lambda \Id\right ) ^{\frac{1}{2} } \|_{o p} \sqrt{\dmmd \left(\mu \| \pi \right)} 
    \leq \sqrt{ \frac{K + \lambda}{1+\lambda} } \sqrt{ \dmmd \left(\mu \| \pi \right)} , \label{appeq:dmmd_tecond_tide}
\end{align}
where $K$ is the upper bound on kernel $k$ in {\color{black}\Cref{assumption:bounded_kernel}} and hence an upper bound on the operator norm of $\Sigma_\pi$.
Combining \eqref{appeq:dmmd_one_tide} and \eqref{appeq:dmmd_tecond_tide}, we have 
\begin{align*}
    \frac{1 + \lambda}{K + \lambda} \mmd^{2}(\mu \| \pi) \leq \dmmd \left(\mu \| \pi \right) \leq \frac{1 + \lambda}{\lambda} \mmd^{2}(\mu \| \pi) .
\end{align*}
Therefore, $\lim_{\lambda \to \infty} \dmmd(\mu \| \pi) = \mmd^2 (\mu \| \pi) $, and the proposition is proved.

\subsection{\texorpdfstring{Proof of \Cref{prop:topology_chard}}{Proof of the topology proposition}}\label{appsec:proof_of_topology_chard}
To show that $\dmmd$ is a probability divergence, we need to show that $\dmmd$ enjoys non-negativity and definiteness.
It is easy to see that $\dmmd(\mu \| \pi)$ is non-negative from its definition in \Cref{def:chard}.
Then, we prove definiteness, i.e., $\dmmd(\mu || \pi)=0 $ if and only if $ \mu = \pi$.
For the first direction, assume $\dmmd(\mu \| \pi)=0$, so $\| (\Sigma_\pi+\lambda \Id)^{-1 / 2} \left(m_\mu-m_\pi\right)\|_{\calH}^2=0$. 
Since $\left(\Sigma_\pi+\lambda \Id\right)^{-1 / 2}$ is a non-singular operator, we must have that $m_\mu=m_\pi$ which implies $\mu=\pi$ as $k$ is $c_0$-universal and hence characteristic~\citep{sriperumbudur2011universality}. 
For the other direction, when $\mu = \pi$, immediately we can see $\dmmd(\mu || \pi) = 0$.

Then we prove that $\dmmd$ metrizes weak convergence. 
For the first direction, we know from \eqref{appeq:dmmd_one_tide} that $\dmmd(\mu_n \| \pi) \leq \frac{1 + \lambda}{\lambda} \mmd^2(\mu_n \| \pi)$ and $\mmd^2(\mu_n \| \pi) \to 0$ as $\mu_n$ converges weakly to $\pi$~\citep{simon2023metrizing}.
For the converse direction, we assume that $\dmmd\left(\mu_n \| \pi \right) \to 0$.
From \eqref{appeq:dmmd_tecond_tide}, we know that $\mmd^2 \left(\mu \| \pi \right) \leq \frac{K + \lambda}{1+\lambda} \dmmd \left(\mu \| \pi \right) $, therefore $\dmmd\left(\mu_n \| \pi \right) \to 0$ implies $\mmd \left(\mu_n \| \pi \right) \to 0$, implying the weak convergence of $\mu_n$ to $\pi$, if $k$ is characteristic~\citep{simon2023metrizing}.

\subsection{\texorpdfstring{Proof of \Cref{prop:chard_gradient_flow}}{Proof of the DrMMD gradient-flow proposition}}\label{appsec:proof_chard_gradient_flow}
In order to show that $\calF_{\dmmd}(\cdot) = \dmmd(\cdot \|\pi)$ admits a well-defined gradient flow, we follow the same techniques in Proposition 7 of \cite{glaser2021kale} and Lemma B.2 of \cite{chizat2018global}, where the key is to show that $(1 + \lambda) \nabla h_{\mu, \pi}^\ast$ is the Fr\'{e}chet subdifferential of $\calF_{\dmmd}$ evaluated at $\mu$.\footnote{Although $\dmmd$ can be viewed as squared $\mmd$ with a regularized kernel $\tilde{k}$, we are not using the technique in \cite{arbel2019maximum} because it relies on Lemma 10.4.1 of \cite{ambrosio2005gradient} which only provides the Fr\'{e}chet subdifferential on probability measures $\mu$ that admit density functions. To construct the Wasserstein gradient flow of $\calF_{\dmmd}$ up to full generality, we resort to the techniques of \cite{glaser2021kale} and \cite{chizat2018global} instead.} 
According to Definition 10.1.1 of \cite{ambrosio2005gradient}, it is equivalent to prove that, for any $\mu \in \calP_2(\R^d)$ and $\phi \in C_c^\infty(\R^d)$, 
\begingroup\small
\begin{align}\label{appeq:target_subdiff}
&\dmmd \left( (\Id + \nabla \phi)_\# \mu \| \pi \right) - \dmmd \left(\mu \| \pi\right) \geq (1 + \lambda) \int \nabla \phi(x)^\top \nabla h_{\mu, \pi}^\ast(x) d \mu(x) \nonumber\\
&\qquad\qquad\qquad\qquad\qquad+ o\left( \| \nabla \phi \|_{L^2(\mu)} \right) .
\end{align}
\endgroup
Define $\rho_t = (\Id + t \nabla \phi)_\# \mu $, $\varphi_t : \R^d \to \R^d, x \mapsto x + t \nabla \phi(x)$ and $g(t) = \dmmd( \rho_t || \pi)$. Then from \Cref{lem:wass_gradient_hess_chard} we know that
$g(t)$ is continuous and differentiable with respect to $t$ and
\begin{align*}
    \frac{d}{dt}\Big|_{t=0} g(t)  = (1 + \lambda) \int \nabla \phi(x)^\top \nabla h_{\mu, \pi}^\ast(x) d\mu(x) .
\end{align*}
Since $t \mapsto g(t)$ is differentiable, using Taylor's theorem and mean value theorem \citep{rudin1976principles}, we know that there exists $0 < \kappa < 1$ such that
\begingroup\small
\begin{align*}
\dmmd \left( (\Id + \nabla \phi)_\# \mu \| \pi \right) - \dmmd \left(\mu \| \pi\right) &= g(1) - g(0) \\
&= \frac{d}{dt}\Big|_{t=0} g(t)  + \frac{d^2}{dt^2} \Big|_{t=\kappa} g(t).
\end{align*}
\endgroup
Therefore, to prove \eqref{appeq:target_subdiff}, the goal is to prove that $\frac{d^2}{dt^2} \big|_{t=\kappa} g(t) \geq o\left( \| \nabla \phi \|_{L^2(\mu)} \right)$. To this end, since we know from \Cref{lem:wass_gradient_hess_chard} that $t \mapsto \frac{d}{dt} g(t)$ is continuous and differentiable, we have 
\begingroup\small
\begin{align*}
    &\quad \frac{d^2}{d t^2} \dmmd(\rho_t||\pi) = 2 (1+\lambda) \iint \nabla \phi(x)^{\top}
    \nabla_1 \nabla_2 \tilde{k}\left(\varphi_t(x), \varphi_t(y)\right)
    \nabla \phi(y) d \mu(x) d \mu(y) \\
    &+ 2 (1+\lambda) \int \nabla \phi(x)^\top
    \left( \int\bH_1 \tilde{k} \left( \varphi_t(x), \varphi_t(y) \right) d \mu(y) \right. \\
    &\hspace{105pt}\left. {}
    - \int \bH_1 \tilde{k} \left( \varphi_t(x), y \right) d \pi(y) \right)
    \nabla \phi(x) d\mu(x),
\end{align*}
\endgroup
where $\tilde{k}$ is the regularized kernel defined in \eqref{eq:tilde_k} and $\tilde{\calH}$ is the associated RKHS.
Using \Cref{lem:k_tilde}, the first term above can be upper-bounded by,
\begingroup\small
\begin{align*}
    &\quad  \left|2(1+\lambda) \iint \nabla \phi(x)^{\top} \nabla_1 \nabla_2 \tilde{k}\left(\varphi_t(x), \varphi_t(y)\right) \nabla \phi(y) d \mu(x) d \mu(y) \right| \\
    &\leq 2(1+\lambda) \iint \left\| \nabla \phi(x) \right\| \left\|  \nabla_1 \nabla_2 \tilde{k}\left(\varphi_t(x), \varphi_t(y)\right) \right\|_F \left\| \nabla \phi(y) \right\| d \mu(x) d \mu(y) \\
    &\leq 2(1+\lambda) \frac{K_{1d}}{\lambda} \left( \int \left\| \nabla \phi(y) \right\| d \mu(y) \right)^2 
    \leq 2(1+\lambda) \frac{K_{1d}}{\lambda} \|\nabla \phi \|_{L^2(\mu)}^2 .
\end{align*}
\endgroup
Using \Cref{lem:k_tilde} again, the second term can be upper-bounded by
\begingroup\small
\begin{align*}
    &\quad \left| \int \nabla \phi(x)^\top \left( \int\bH_1 \tilde{k} \left( \varphi_t(x),  \varphi_t(y) \right) d \mu(y) -  \int \bH_1 \tilde{k} \left( \varphi_t(x), y \right) d \pi(y) \right) \nabla \phi(x) d\mu(x) \right| \\
    &\leq \left| \int \nabla \phi(x)^\top \left( \int \bH_1 \tilde{k} \left( \varphi_t(x),  \varphi_t(y) \right) d \mu(y) \right) \nabla \phi(x) d\mu(x) \right| \\
    &\qquad\qquad+ \left| \int \nabla \phi(x)^\top \left( \int \bH_1 \tilde{k} \left( \varphi_t(x), y  \right) d \pi(y) \right) \nabla \phi(x) d\mu(x) \right| \\
    &\leq \int \left\| \nabla \phi(x) \right\| \left( \int \left\|  \bH_1 \tilde{k} \left( \varphi_t(x),  \varphi_t(y) \right) \right\|_F d \mu(y) \right) \left\|\nabla \phi(x) \right\| d\mu(x) \\
    &\qquad\qquad+ \int \left\| \nabla \phi(x) \right\| \left( \int \left\|  \bH_1 \tilde{k} \left( \varphi_t(x), y \right) \right\|_F d \pi(y) \right) \left\|\nabla \phi(x) \right\| d\mu(x) \\
    &\leq 2 \frac{\sqrt{K K_{2d}}}{\lambda} \|\nabla \phi \|_{L^2(\mu)}^2 .
\end{align*}
\endgroup
Combining the above two inequalities to have
\begin{align}\label{appeq:dmmd_hessian_bounds}
    \left| \frac{d^2}{d t^2} \dmmd(\rho_t||\pi) \right| \leq 2(1+\lambda) \frac{K_{1d}}{\lambda} \|\nabla \phi \|_{L^2(\mu)}^2 + 4(1+\lambda) \frac{\sqrt{K K_{2d}}}{\lambda} \|\nabla \phi \|_{L^2(\mu)}^2 . 
\end{align}
Therefore, we have $\frac{d^2}{d t^2}\big|_{t=\kappa} g(t) = \calO(\|\nabla \phi \|_{L^2(\mu)}^2)=o(\|\nabla \phi \|_{L^2(\mu)})$ as $\|\nabla \phi \|_{L^2(\mu)} \to 0$.
So \eqref{appeq:target_subdiff} is proved, which means that $(1 + \lambda) \nabla h_{\mu, \pi}^\ast(x)$ is the Fr\'{e}chet subdifferential of $\calF_{\dmmd}$ evaluated at $\mu$.
According to Definition 11.1.1 of \cite{ambrosio2005gradient}, there exists a solution $(\mu_t)_{t\geq 0}$ such that the following equation holds in the sense of distributions,
\begin{align*}
    \partial_t \mu_t - (1 + \lambda) \nabla \cdot \left(\mu_t  \nabla h_{\mu_t, \pi}^\ast \right)=0,
\end{align*}
and such $(\mu_t)_{t\geq 0}$ is indeed the $\dmmd$ gradient flow, so existence is proved.

Next, we are going to prove uniqueness. 
\eqref{appeq:dmmd_hessian_bounds} indicates that $\calF_{\dmmd}$ is geodesically $-(1+\lambda) \frac{ 4 \sqrt{K K_{2 d}} + 2 K_{1 d}}{\lambda}$ semiconvex. Therefore the uniqueness of $(\mu_t)_t$ follows from Theorem 11.2.1 of \cite{ambrosio2005gradient}.
\eqref{eq:dmmd_semi_convex} is proved in \eqref{appeq:dmmd_hessian_bounds}.

\subsection{\texorpdfstring{Proof of \Cref{prop:geometric_prop}}{Proof of the geometric proposition}} \label{appsec:proof_geometric}
Given $\mu \in \calP_2\left(\mathbb{R}^d\right)$ and $\phi \in C_c^\infty\left(\mathbb{R}^d\right)$, consider the path $(\rho_t)_{0 \leq t \leq 1}$ from $\mu$ to $(\Id + \nabla \phi)_{\#} \mu$ given by $\rho_t = (\Id + t \nabla \phi)_{\#} \mu$. $(\rho_t)_{0 \leq t \leq 1}$ is a constant-time geodesic in the Wasserstein-2 space by construction (\Cref{appsec:wass_discussion}).
Define $\varphi_t: \R^d \to \R^d, x \mapsto x + t \nabla \phi(x)$. We know from \citep[Example 15.9]{villani2009optimal} (by taking $m=2$) that, 
\begin{align}\label{appeq:hess_chi2}
    \frac{d^2}{d t^2}\Big|_{t=0} \chi^2(\rho_t \| \pi) &= \int \frac{d \mu}{ d \pi} (x) \left( \nabla V(x)^\top \nabla \phi(x) - \nabla \cdot \nabla \phi(x) \right)^2 d\mu(x) \nonumber \\
    &\qquad+ \int \frac{d \mu}{ d \pi} (x) \nabla \phi(x)^\top \bH V(x) \nabla \phi(x) d\mu(x) \nonumber \\ 
    &\qquad\qquad+ \int \frac{d \mu}{ d \pi} (x) \left\| \bH \phi(x) \right\|_F^2 d \mu(x) .
\end{align}
$V$ is twice differentiable, so $\nabla V$ as a function from $\R^d$ to $\R^d$ is continuous.
$\phi \in C_c^\infty(\R^d)$ has compact support, so $x \mapsto \nabla V(x)^\top \nabla \phi(x) - \nabla \cdot \nabla \phi(x)$ is a continuous function over a compact domain, so its image is also compact and hence bounded. 
Using similar arguments, $x \mapsto \nabla \phi(x)^\top \bH V(x) \nabla \phi(x)$ and $x \mapsto \| \bH \phi(x) \|_F^2$ are also continuous functions over compact domains, so they are all bounded. 

Since $\frac{d \mu}{d \pi} - 1 \in \calH \subset L^2(\pi)$ and $\int \frac{d \mu}{ d \pi} (x) d \mu(x) = \| \frac{d \mu}{d \pi} - 1 \|_{L^2(\pi)}^2 + 1 < \infty$, we have
\begin{align}\label{appeq:hess_chi2_bounded}
    \frac{d^2}{d t^2}\big|_{t=0} \chi^2(\rho_t \| \pi) < \infty .
\end{align}
Next, from \Cref{lem:wass_chi2} we know that $\frac{d^2}{d t^2}\big|_{t=0} \chi^2(\rho_t \| \pi)$ can be alternatively expressed as,
\begin{align}\label{appeq:hessian_chi2_copy}
    \frac{d^2}{d t^2}\Big|_{t=0} \chi^2(\rho_t \| \pi) &= 2 \int \left( \nabla \cdot \left( \nabla \phi(x) \mu(x) \right) \frac{1}{\pi(x)} \right)^2 \pi(x) d x \nonumber \\
    & \qquad\qquad+ 2 \int \nabla \phi(x)^\top \bH \left(\frac{\mu(x) }{\pi(x)}\right) \nabla \phi(x) \mu(x) d x .
\end{align}
Recall from \Cref{lem:wass_gradient_hess_chard} that
\begin{align}\label{appeq:chard_wass_hessian_copy}
    &\quad \frac{d^2}{d t^2} \Big|_{t=0} \dmmd(\rho_t||\pi) \nonumber \\
    &=  2 (1+\lambda) \iint \nabla \phi(x)^{\top} \nabla_1 \nabla_2 \tilde{k}\left(x, y \right) \nabla \phi(y) d \mu(x) d \mu(y) \nonumber \\
    &\qquad+ 2 (1+\lambda) \int \nabla \phi(x)^\top
    \left( \int\bH_1 \tilde{k} \left( x, y \right) d \mu(y) \right. \nonumber \\
    &\hspace{105pt}\left. {}
    - \int \bH_1 \tilde{k} \left( x, y \right) d \pi(y) \right)
    \nabla \phi(x) d\mu(x) .
\end{align}
To prove \eqref{eq:hessian_lower_bound}, our aim then is to compare and bound the difference of $\frac{d^2}{d t^2}\big|_{t=0} \chi^2(\rho_t \| \pi)$ in \eqref{appeq:hessian_chi2_copy} and $\frac{1}{1+\lambda} \frac{d^2}{d t^2}\big|_{t=0} \dmmd(\rho_t ||\pi)$ in \eqref{appeq:chard_wass_hessian_copy}, so we compare and bound their first and second term separately. 

The first term of \eqref{appeq:hessian_chi2_copy} can be rewritten as
\begin{align}
&\int \left(\nabla \cdot( \nabla \phi(x) \mu(x) ) \frac{1}{\pi(x)} \right)^2 \pi(x) d x = \sum_{i \geq 1} \PSi{\nabla \cdot( \nabla \phi \mu ) \frac{1}{\pi}, e_i}{L^2(\pi)}^2 \nonumber \\
&= \sum_{i \geq 1} \left( \int \nabla \cdot \left( \nabla \phi(x) \mu(x) \right) e_i(x) dx \right)^2 
= \sum_{i \geq 1} \left( \int \nabla \phi(x)^\top \nabla e_i(x) \mu(x)  dx \right)^2,
\label{appeq:chi2_hessian_first_term}
\end{align}
where we use integration by parts in the last line since $\phi \in C_c^\infty(\R^d)$.
And the first term of \eqref{appeq:chard_wass_hessian_copy} after rescaling by $\frac{1}{1 + \lambda}$ can be rewritten as,
\begin{align}
&\quad 2 \iint \nabla \phi(x)^{\top} \nabla_1 \nabla_2 \tilde{k}\left(x , y \right) \nabla \phi(y) d \mu(x) d \mu(y) \nonumber \\
&= 2 \iint \nabla \phi(x)^{\top} \nabla_x \nabla_y \left( \sum_{i \geq 1} \frac{\varrho_i}{\varrho_i + \lambda} e_i(x) e_i(y) \right) \nabla \phi(y) d \mu(x) d \mu(y) \nonumber \\
&=  2 \iint \Big( \nabla \cdot \left( \nabla \phi(x) \mu(x)  \right) \Big) \Big( \nabla \cdot \left( \nabla \phi(y) \mu(y) \right) \Big) \sum_{i \geq 1} \frac{\varrho_i}{\varrho_i + \lambda} e_i(x) e_i(y) dx dy .
\label{appeq:infinite_tum_integral}
\end{align}
Since $\pi \ll \calL^d$ so \eqref{appeq:tilde_k_mercer} is true for all $x, y \in \R^d$, hence the second equality is true, and the last equality uses integration by parts since $\phi \in C_c^\infty(\R^d)$. Notice that
\begin{align*}
    &\quad \sum_{i\geq 1} \frac{\varrho_i}{\varrho_i + \lambda} \iint \left| e_i(x) e_i(y) \Big( \nabla \cdot ( \nabla \phi(x) \mu(x)) \Big) \Big( \nabla \cdot ( \nabla \phi(y) \mu(y) ) \Big) \right| dx dy \\
    &= \sum_{i\geq 1} \frac{\varrho_i}{\varrho_i + \lambda} \left(  \int \left| \Big( \nabla \cdot \left( \nabla \phi(x) \mu(x)  \right) \Big) e_i(x) \right| dx \right)^2 \\
    &\leq \frac{1}{\lambda} \sum_{i\geq 1} \varrho_i \left( \int \left( \frac{\nabla \cdot \left( \nabla \phi(x) \mu(x)  \right)}{\mu(x)} \right)^2 \mu(x) dx \right) \left( \int e_i(x)^2 \mu(x) dx \right)  \\
    &= \frac{1}{\lambda} \left( \int \left( \nabla \log \mu(x)^\top \nabla \phi(x) + \nabla \cdot \nabla \phi(x) \right)^2 d \mu(x) \right) \left( \int \sum_{i\geq 1} \varrho_i e_i(x)^2 d \mu(x) \right) \\
    &\leq \frac{K}{\lambda} \left( \int \left( \nabla \log \mu( x )^\top \nabla \phi(x) + \nabla \cdot \nabla \phi(x) \right)^2 d\mu(x) \right) 
    < +\infty .
\end{align*}
The first inequality uses Cauchy-Schwartz, the second inequality uses $\sum_{i \geq 1} \varrho_i (e_i(x) )^2 = k(x,x) \leq K$. 
The last quantity is finite because $x \to \nabla \log \mu( x )$ is a continuous function and $\phi \in C_c^\infty(\R^d)$ has compact support, hence the integral of a continuous function over a compact domain is always finite. 
Then, by using Fubini's theorem~\citep{rudin1976principles}, 
we are allowed to interchange the infinite sum and integration of \eqref{appeq:infinite_tum_integral} to reach,
\begin{align*}
    &\quad 2 \iint \Big( \nabla \cdot \left( \nabla \phi(x) \mu(x)  \right) \Big) \Big( \nabla \cdot \left( \nabla \phi(y) \mu(y) \right) \Big) \sum_{i \geq 1} \frac{\varrho_i}{\varrho_i + \lambda} e_i(x) e_i(y) dx dy \nonumber \\
    &= 2 \sum_{i \geq 1} \frac{\varrho_i}{\varrho_i + \lambda} \iint \Big( \nabla \cdot \left( \nabla \phi(x) \mu(x)  \right) \Big) \Big( \nabla \cdot \left( \nabla \phi(y) \mu(y) \right) \Big) e_i(x) e_i(y) dx dy \nonumber \\
    &= 2 \sum_{i \geq 1} \frac{\varrho_i}{\varrho_i + \lambda} \left(  \int \Big( \nabla \cdot \left( \nabla \phi(x) \mu(x)  \right) \Big) e_i(x)  dx \right)^2 \nonumber \\
    &= 2 \sum_{i \geq 1} \frac{\varrho_i}{\varrho_i + \lambda} \left( \int \nabla \phi(x)^\top \nabla e_i(x) \mu(x)  dx \right)^2,
\end{align*}
where the last equality uses integration by parts.

So the difference between the first term of \eqref{appeq:hessian_chi2_copy} and \eqref{appeq:chard_wass_hessian_copy} rescaled by $\frac{1}{1+\lambda}$ is,
\begin{align}
&\quad 2 \left| \sum_{i \geq 1}
\left( \int \nabla \phi(x)^\top \nabla e_i(x) \mu(x) dx \right)^2 \right. \nonumber \\
&\hspace{70pt}\left. {}
- \sum_{i \geq 1} \frac{\varrho_i}{\varrho_i + \lambda}
\left( \int \nabla \phi(x)^\top \nabla e_i(x) \mu(x) dx \right)^2 \right| \nonumber \\
&= 2 \sum_{i \geq 1} \frac{\lambda}{\varrho_i + \lambda} \left( \int \nabla \phi(x)^\top \nabla e_i(x) \mu(x)  dx \right)^2.
\label{appeq:diff_first_term}
\end{align}
Now we turn to the second term. 
The second term of \eqref{appeq:hessian_chi2_copy} can be rewritten as
\begin{align}
&\quad 2 \int \nabla \phi(x)^\top \bH \left( \frac{d \mu}{ d \pi} (x) \right) \nabla \phi(x) \mu(x) dx \nonumber \\
&= 2 \int \nabla \phi(x)^\top \bH \left( \frac{d \mu}{ d \pi} (x) - 1 \right) \nabla \phi(x) \mu(x) dx \nonumber \\
&= 2 \int \nabla \phi(x)^\top \bH \left( \sum_{i \geq 1} \PSi{\frac{d \mu}{d \pi}-1, e_i}{L^2(\pi)} e_i(x) \right) \nabla \phi(x) \mu(x) dx, \label{appeq:chi2_hessian_tecond_term}
\end{align}
and the second term of \eqref{appeq:chard_wass_hessian_copy} rescaled by $\frac{1}{1+\lambda}$ can be rewritten as,
\begin{align}
&\quad 2 \int \nabla \phi(x)^\top \left( \int\bH_1 \tilde{k} \left(x, y\right) d \mu(y) -  \int \bH_1 \tilde{k} \left(x, y\right) d \pi(y) \right) \nabla \phi(x) d\mu(x) \nonumber \\
&= 2 \int \nabla \phi(x)^\top \bH \left( \int \tilde{k} \left(x, y\right) d \mu(y) -  \int \tilde{k} \left(x, y\right) d \pi(y) \right) \nabla \phi(x) d\mu(x) \nonumber \\
&= 2 \int \nabla \phi(x)^\top  \bH \left( \sum_{i \geq 1} \frac{\varrho_i}{\varrho_i + \lambda} e_i(x) \int e_i(y)d (\mu- \pi)(y)  \right) \nabla \phi(x) d\mu(x) \nonumber \\
&= 2 \int \nabla \phi(x)^\top  \bH \left( \sum_{i \geq 1} \frac{\varrho_i}{\varrho_i + \lambda} \PSi{\frac{d \mu}{d \pi} - 1, e_i}{L^2(\pi)} e_i(x)  \right) \nabla \phi(x) d\mu(x) . \label{appeq:chard_hessian_tecond_term}
\end{align}
Since $\pi \ll \calL^d$, so \eqref{appeq:tilde_k_mercer} is true for all $x, y \in \R^d$ hence the third equality is true.
From  \Cref{lem:k_tilde}, we have $\int \tilde{k}(x,y) d\mu(y) \leq \frac{K}{\lambda}$, $x\mapsto \tilde{k}(x,y)$ is second-order differentiable, $\sup_{x} |\bH_1 \tilde{k}(x,y)| \leq \frac{\sqrt{K K_{2d}}}{\lambda}$. So we are allowed to interchange integration and Hessian in the second equality using the differentiation lemma~\citep[Theorem 6.28]{klenke2013probability}.
Consider the difference of \eqref{appeq:chard_hessian_tecond_term} and \eqref{appeq:chi2_hessian_tecond_term}, we have
\begin{align}
& \quad 2 \Bigg| \int \nabla \phi(x)^\top \bH \left( \sum_{i \geq 1} \PSi{\frac{d \mu}{d \pi}-1, e_i}{L^2(\pi)} e_i(x) \right) \nabla \phi(x) \mu(x) dx \nonumber \\
&\qquad\qquad- \int \nabla \phi(x)^\top  \bH \left( \sum_{i \geq 1} \frac{\varrho_i}{\varrho_i + \lambda} \PSi{\frac{d \mu}{d \pi} - 1, e_i}{L^2(\pi)} e_i(x)  \right) \nabla \phi(x) d\mu(x) \Bigg| \nonumber \\
&= 2 \left| \int \nabla \phi(x)^\top \bH \left( \sum_{i \geq 1}\frac{\lambda}{\varrho_i + \lambda} \PSi{\frac{d \mu}{d \pi}-1, e_i}{L^2(\pi)} e_i(x) \right) \nabla \phi(x) \mu(x) dx \right| \label{appeq:chi2_drmmd_diff}.
\end{align}
Given $\frac{d \mu}{d\pi} - 1 \in \calH$, there exists $q \in L^2(\pi)$ such that $\frac{d \mu}{d\pi} - 1 =\mathcal{T}^{1/2}_\pi q $ so that $\langle \frac{d \mu}{d\pi} - 1, e_i \rangle = \varrho_i^{1/2} \langle q, e_i \rangle$ for all $i$. 
For $j, r \in \{1, \cdots, d\}$, we have 
\begingroup\small
\begin{align*}
    g_{M_0}(x) &:= \left| \sum_{i \geq M_0} \frac{\lambda}{\varrho_i + \lambda} \PSi{\frac{d \mu}{d \pi} - 1, e_i}{L^2(\pi)} \partial_j \partial_r e_i(x) \right| \nonumber \\
    &=  \left| \sum_{i \geq M_0} \frac{\lambda}{\varrho_i + \lambda} \varrho_i^{1/2} \PSi{q, e_i}{L^2(\pi)} \partial_j \partial_r e_i(x) \right|  \nonumber \\
    & \leq \left( \sum_{i \geq M_0}
    \left( \frac{\lambda}{\varrho_i + \lambda} \right)^2
    \PSi{q, e_i}{L^2(\pi)}^2 \right)^{1/2} \nonumber \\
    &\quad \times \left( \sum_{i \geq M_0} \varrho_i
    \left( \partial_j \partial_r e_i(x) \right)^2 \right)^{1/2} \nonumber \\
    &\leq \left( \sum_{i \geq M_0} \PSi{q, e_i}{L^2(\pi)}^2 \right)^{1/2}
    \left( \sum_{i \geq M_0} \varrho_i
    \left( \partial_j \partial_r e_i(x) \right)^2 \right)^{1/2} \\
    &\leq \sqrt{K_{2d}}
    \left( \sum_{i \geq M_0} \PSi{q, e_i}{L^2(\pi)}^2 \right)^{1/2}.
\end{align*}
\endgroup
The final inequality holds because,
\begin{align}\label{appeq:sum_kernel_hessian}
    \sum_{j,r = 1}^d \sum_{i \geq 1} \varrho_i
    \left( \partial_j \partial_r e_i(x) \right)^2
    &\leq \sum_{j,r = 1}^d \sum_{i \geq 1} \varrho_i
    \PSi{\partial_j \partial_r k(x, \cdot), e_i}{\calH}^2 \nonumber \\
    &= \sum_{j,r = 1}^d \sum_{i \geq 1}
    \PSi{\partial_j \partial_r k(x, \cdot), \sqrt{\varrho_i} e_i}{\calH}^2 \nonumber \\
    &= \sum_{j,r = 1}^d
    \left\| \partial_j \partial_r k(x, \cdot) \right\|_\calH^2 \nonumber \\
    &= \left\| \bH_1 k(x,\cdot) \right\|_{\calH^{d \times d}}^2
    \leq K_{2d} .
\end{align}
Since $\left\| sdq \right\|_{L^2(\pi)}$ is bounded, so $\sum_{i \geq M_0} \PSi{q, e_i}{L^2(\pi)}^2$ converges to $0$ uniformly as $M_0 \to \infty$ and hence $g_{M_0}(x)$ converge to $0$ uniformly. Therefore, we are allowed to interchange the Hessian and the infinite sum~\citep{rudin1976principles} in \eqref{appeq:chi2_drmmd_diff} to achieve,
\begingroup\small
\begin{align}
 &\quad 2 \left| \int \nabla \phi(x)^\top \bH \left( \sum_{i \geq 1}\frac{\lambda}{\varrho_i + \lambda} \PSi{\frac{d \mu}{d \pi}-1, e_i}{L^2(\pi)} e_i(x) \right) \nabla \phi(x) \mu(x) dx \right| \nonumber \\
 &= 2 \left| \int \nabla \phi(x)^\top  \left( \sum_{i \geq 1}\frac{\lambda}{\varrho_i + \lambda} \varrho_i^{1/2} \PSi{q, e_i}{L^2(\pi)} \bH e_i(x) \right) \nabla \phi(x) \mu(x) dx \right| \nonumber \\
&\leq 2 \left( \sum_{i \geq 1}
\left( \frac{\lambda }{\varrho_i + \lambda}\right)^2
\PSi{q, e_i}{L^2(\pi)}^2 \right)^{1/2} \nonumber \\
&\quad {}\times \left( \sum_{i \geq 1} \varrho_i
\left( \int \nabla \phi(x)^\top \bH e_i(x)
\nabla \phi(x) \mu(x) dx \right)^2 \right)^{1/2} \nonumber \\
&\leq 2 \left( \sum_{i \geq 1}
\left( \frac{\lambda }{\varrho_i + \lambda}\right)^2
\PSi{q, e_i}{L^2(\pi)}^2 \right)^{1/2} \nonumber \\
&\quad {}\times \left( \sum_{i \geq 1} \varrho_i
\left\| \bH e_i(x) \right\|_{op}^2 \right)^{1/2}
\|\nabla \phi \|_{L^2(\mu)}^2 \nonumber \\
&\leq 2 \left( \sum_{i \geq 1}
\left( \frac{\lambda }{\varrho_i + \lambda}\right)^2
\PSi{q, e_i}{L^2(\pi)}^2 \right)^{1/2} \nonumber \\
&\quad {}\times \left( \sum_{i \geq 1} \varrho_i
\left\| \bH e_i(x) \right\|_{F}^2 \right)^{1/2}
\|\nabla \phi \|_{L^2(\mu)}^2 \nonumber \\
&\leq 2 \left( \sum_{i \geq 1}
\left( \frac{\lambda }{\varrho_i + \lambda}\right)^2
\PSi{q, e_i}{L^2(\pi)}^2 \right)^{1/2} \nonumber \\
&\quad {}\times \sqrt{K_{2d}} \|\nabla \phi \|_{L^2(\mu)}^2 .
\label{appeq:diff_tecond_term}
\end{align}
\endgroup
The first inequality uses Cauchy Schwartz, the second last inequality uses that matrix operator norm is smaller than matrix Frobenius norm, and the last inequality uses \eqref{appeq:sum_kernel_hessian}.
Combining together \eqref{appeq:diff_first_term} and \eqref{appeq:diff_tecond_term}, we reach
\begin{align*}
    &\quad \left| \frac{1}{1 + \lambda} \frac{d^2}{d t^2}\Big|_{t=0}  \dmmd(\rho_t||\pi) - \frac{d^2}{d t^2} \Big|_{t=0} \chi^2(\rho_t ||\pi) \right| \\
    &\leq 2 \sum_{i \geq 1} \frac{\lambda}{\varrho_i + \lambda} \left( \int \nabla \phi(x)^\top \nabla e_i(x) \mu(x)  dx \right)^2 \\
    &\qquad\qquad+ 2 \left( \sum_{i \geq 1} \left( \frac{\lambda }{\varrho_i + \lambda}\right)^2 \PSi{q, e_i}{L^2(\pi)}^2  \right)^{1/2} \sqrt{K_{2d}} \|\nabla \phi \|_{L^2(\mu)}^2  \\
    & =: \bar{R}(\lambda, \mu, \nabla \phi) .
\end{align*}
Therefore, 
\begin{align*}
    & \frac{d^2}{d t^2}\Big|_{t=0}  \dmmd(\rho_t ||\pi)  \geq (1+\lambda) \frac{d^2}{d t^2}\Big|_{t=0}  \chi^2(\rho_t || \pi) - (1+\lambda) \bar{R}(\lambda, \mu, \nabla \phi) \\
    & \geq (1+\lambda) \int \frac{d \mu}{ d \pi} (x) \nabla \phi(x)^{\top} \mathbf{H} V(x) \nabla \phi(x) d \mu(x) - (1+\lambda) \bar{R}(\lambda, \mu, \nabla \phi) \\
    &\geq (1+\lambda) \alpha \int \frac{d \mu}{d \pi}(x)\|\nabla \phi(x)\|^2 d \mu(x) - (1+\lambda) \bar{R} (\lambda, \mu, \nabla \phi) ,
\end{align*}
where the second inequality is using \eqref{appeq:hess_chi2} and the last inequality is using $\bH V \succeq \alpha \Id$. 
So \eqref{eq:hessian_lower_bound} is proved.

Define $R(\lambda, \mu, \nabla \phi) := (1+\lambda) \bar{R} (\lambda, \mu, \nabla \phi)$.
The final thing to check is $\lim_{\lambda \to 0} R(\lambda, \mu, \nabla \phi) = 0$, which is equivalent to check that $\lim_{\lambda \to 0} \bar{R}(\lambda, \mu, \nabla \phi) = 0$.
Since we know from \eqref{appeq:hess_chi2_bounded} and \eqref{appeq:chi2_hessian_first_term} that 
\begin{align*}
    \sum_{i \geq 1} \left( \int \nabla \phi(x)^\top \nabla e_i(x) \mu(x)  dx \right)^2  < \frac{d^2}{dt^2}\Big|_{t=0} \chi^2(\rho_t \| \pi) < \infty,
\end{align*}
using the dominated convergence theorem~\citep{rudin1976principles}, we are allowed to interchange infinite sum and taking limits,
\begin{align*}
    &\lim_{\lambda \to 0} \sum_{i \geq 1} \frac{\lambda}{\varrho_i + \lambda} \left( \int \nabla \phi(x)^\top \nabla e_i(x) \mu(x)  dx \right)^2 \\
    &= \sum_{i \geq 1} \lim_{\lambda \to 0} \frac{\lambda}{\varrho_i + \lambda} \left( \int \nabla \phi(x)^\top \nabla e_i(x) \mu(x)  dx \right)^2 = 0.
\end{align*}
Similarly, because $\sum_{i \geq 1} \left( \frac{\lambda }{\varrho_i + \lambda}\right)^2 \PSi{q, e_i}{L^2(\pi)}^2 < \left\| q \right\|_{L^2(\pi)}^2 < \infty$, using dominated convergence theorem~\citep{rudin1976principles} again, we have,
\begin{align*}
    \lim_{\lambda \to 0}  \sum_{i \geq 1} \left( \frac{\lambda }{\varrho_i + \lambda}\right)^2 \PSi{q, e_i}{L^2(\pi)}^2  = \sum_{i \geq 1} \lim_{\lambda \to 0} \left( \frac{\lambda }{\varrho_i + \lambda}\right)^2 \PSi{q, e_i}{L^2(\pi)}^2 = 0 .
\end{align*}
Therefore, we have that
\begin{align*}
    \lim_{\lambda \to 0} R(\lambda, \mu, \nabla \phi) = \lim_{\lambda \to 0} \bar{R}(\lambda, \mu, \nabla \phi) = 0 .
\end{align*}
And the proof of the proposition is finished.

\subsection{\texorpdfstring{Proof of \Cref{thm:continuous_time_convergence}}{Proof of the continuous-time convergence theorem}}\label{appsec:proof_continuous_time_convergence}
Considering the time derivative of $\kl(\mu_t \| \pi)$, we have
\begingroup\small
\begin{align}
&\quad \frac{d}{dt} \kl(\mu_t \| \pi) \nonumber \\
&= -(1+\lambda) \int \nabla h_{\mu_t, \pi}^\ast(x)^\top \nabla \log \frac{d \mu_t}{ d \pi} (x) \mu_t(x) dx \nonumber \\
&= -(1+\lambda) \int \nabla h_{\mu_t, \pi}^\ast(x)^\top \nabla \frac{d \mu_t}{ d \pi} (x) \pi(x) dx \nonumber \\
&= -(1+\lambda) \int \left( \nabla h_{\mu_t, \pi}^\ast(x) - 2 \nabla \frac{d \mu_t}{ d \pi} (x) \right)^\top \nabla \frac{d \mu_t}{ d \pi} (x) \pi(x) dx \nonumber \\
&\qquad  - 2(1+\lambda) \int \pi(x) \left\| \nabla \frac{d \mu_t}{ d \pi} (x) \right\|^2 dx \nonumber \\
&= -(1+\lambda) \int \left( \nabla h_{\mu_t, \pi}^\ast(x) - 2 \nabla \left( \frac{d \mu_t}{ d \pi} (x) -1 \right) 
 \right)^\top \nabla \frac{d \mu_t}{ d \pi} (x) \pi(x) dx \nonumber \\
&- 2(1+\lambda) \int \pi(x) \left\| \nabla \frac{d \mu_t}{ d \pi} (x) \right\|^2 dx . \label{eq:kl_d_dt}
\end{align}
\endgroup
\textit{Case one:} $\int \pi(x) \left\| \nabla \frac{d \mu_t}{ d \pi} (x) \right\|^2 dx < \infty$. 
We use integration by parts for the first term in \eqref{eq:kl_d_dt} and we can safely ignore the boundary term due to condition 5 that for $i = 1, \ldots, d$, $\lim\limits_{x\to \infty} \left( h_{\mu_t, \pi}^\ast(x) - 2 \frac{d \mu_t}{ d \pi} (x) \right) \left( \partial_i \frac{d \mu_t}{ d \pi} (x) \right) \pi(x) \to 0$. So, we obtain
\begin{align}
&\quad \frac{d}{dt} \kl(\mu_t \| \pi)
= (1+\lambda) \int \left( h_{\mu_t, \pi}^\ast(x)
- 2 \left( \frac{d \mu_t}{d \pi} (x) -1 \right) \right) \nonumber \\
&\hspace{80pt}{}\times \nabla \cdot
\left( \pi(x) \nabla \frac{d \mu_t}{d \pi} (x) \right) dx \nonumber \\
&\qquad\qquad- 2(1+\lambda) \int \pi(x) \left\| \nabla \frac{d \mu_t}{ d \pi} (x) \right\|^2 dx \nonumber \\
&= (1+\lambda) \int \left( h_{\mu_t, \pi}^\ast(x)
- 2 \left( \frac{d \mu_t}{d \pi} (x) -1 \right) \right) \nonumber \\
&\hspace{55pt}{}\times
\frac{ \nabla \cdot \left( \pi(x) \nabla \frac{d \mu_t}{d \pi} (x) \right) }{ \pi(x) }
\pi(x) dx \nonumber \\
&\qquad\qquad- 2(1+\lambda) \int \pi(x) \left\| \nabla \frac{d \mu_t}{ d \pi} (x) \right\|^2 dx \nonumber \\
&\leq (1+\lambda)
\left\| h_{\mu_t, \pi}^\ast - 2 \left( \frac{d \mu_t}{d \pi} -1 \right) \right\|_{L^2(\pi)} \nonumber \\
&\quad {}\times \left\|
\frac{ \nabla \cdot \left( \pi \nabla \frac{d \mu_t}{d \pi} \right) }{ \pi }
\right\|_{L^2(\pi)}
- \frac{2(1+\lambda)}{C_P} \kl(\mu_t \| \pi) .
\label{appeq:continous_term}
\end{align}
where the first part of the last inequality holds by using Cauchy Schwartz, and the second part holds by the fact that $\kl(\mu_t \| \pi) \leq \chi^2(\mu_t \| \pi)$~\citep{van2014renyi} and by applying the Poincar\'{e} inequality with $f = \frac{d \mu_t}{d \pi} - 1$ (notice that $ \| \nabla f \|_{L^2(\pi)} < \infty$ from \textit{Case one} and $\|f\|_{L^2(\pi)} < \infty$ from condition 3),
\begin{align*}
    \int \pi(x) \left\| \nabla \frac{d \mu_t}{ d \pi} (x) \right\|^2 dx \geq \frac{1}{C_P} \chi^2(\mu_t \| \pi) \geq \frac{1}{C_P} \kl(\mu_t \| \pi).
\end{align*}
Since $\frac{ d\mu_t}{d \pi} - 1 \in \operatorname{Ran} \left( \mathcal{T}_\pi^r \right)$ with $r > 0$, using \Cref{lem:h_difference} we have
\begin{align}\label{appeq:continous_term_1}
    \left\| h_{\mu_t, \pi}^\ast  - 2 \left( \frac{d \mu_t }{d \pi} -1 \right) \right\|_{L^2(\pi)} \leq 2 \lambda^{r} \left\|q_t \right\|_{L^2(\pi)}. 
\end{align}
Then, notice that
\begin{align}
&\left\| \frac{ \nabla \cdot \left( \pi \nabla \frac{d \mu_t}{ d \pi}  \right) }{ \pi } \right\|_{L^2(\pi)}^2 = \int \frac{ \left[ \nabla \cdot\left( \pi(x) \nabla \frac{d \mu_t}{ d \pi} (x)\right) \right]^2 }{\pi(x)^2} d \pi(x) \nonumber \\
& = \int \frac{ \left(  \nabla \pi(x)^\top \nabla \frac{d \mu_t}{ d \pi} (x) + \pi(x) \nabla \cdot \nabla \frac{d \mu_t}{ d \pi} (x) \right)^2 }{\pi(x)^2} d \pi(x) \nonumber \\
&= \int  \left( \nabla \log \pi(x)^\top \nabla \frac{d \mu_t}{ d \pi} (x) + \Delta \frac{d \mu_t}{ d \pi} (x) \right)^2 d \pi(x) \nonumber \\
&\leq \int 2 \left( \nabla \log \pi(x)^\top \nabla \frac{d \mu_t}{ d \pi} (x) \right)^2 + 2 \left( \Delta \frac{d \mu_t}{ d \pi} (x) \right)^2 d \pi(x) \nonumber \\
&= 2 \left\| \nabla \left(\log \pi\right)^\top \nabla \left( \frac{d \mu_t}{d\pi} \right) \right\|_{L^2(\pi)}^2 + 2 \left\| \Delta \left( \frac{d \mu_t}{d\pi} \right) \right\|_{L^2(\pi)}^2 
\leq 2 \mathcal{J}_t^2 + 2 \mathcal{I}_t^2 . \label{appeq:continous_term_2}
\end{align}
Therefore, plugging \eqref{appeq:continous_term_1} and \eqref{appeq:continous_term_2} back to \eqref{appeq:continous_term}, we have
\begin{align}\label{eq:kl_d_dt_final}
    \frac{d}{dt} \kl(\mu_t \| \pi) \leq 4 (1+\lambda)\lambda^{r} \left\| q_t \right\|_{L^2(\pi)} \left( \mathcal{J}_t +\mathcal{I}_t \right) - \frac{ 2 (1+\lambda)}{ C_P} \kl(\mu_t \| \pi).
\end{align}
\textit{Case two:} $\int \pi(x) \left\| \nabla \frac{d \mu_t}{ d \pi} (x) \right\|^2 dx = \infty$. 
The first term of \eqref{eq:kl_d_dt} remains the same, and the second term of \eqref{eq:kl_d_dt} now equals infinity which is larger than the finite $\frac{ 2 (1+\lambda)}{ C_P} \kl(\mu_t \| \pi)$, so we also obtain \eqref{eq:kl_d_dt_final} as in \textit{Case one}.

Therefore, both \textit{Case one} and \textit{Case two} result in \eqref{eq:kl_d_dt_final}. Using the Gronwall lemma, we obtain that for any $T > 0$,
\begin{align*}
\begin{aligned}
     \kl(\mu_T \| \pi) & \leq \exp\left( -\frac{ 2 (1+\lambda)}{ C_P} T\right) \kl(\mu_0 \| \pi) \nonumber \\
    &+ 4 (1+\lambda) \lambda^r \int_0^T \exp\left( -\frac{ 2(1+\lambda)}{C_P} (T-t) \right)\left\| q_t \right\|_{L^2(\pi)} (\mathcal{J}_t + \mathcal{I}_t )  dt,
    \end{aligned}
\end{align*}
which concludes the proof of \Cref{thm:continuous_time_convergence}.

\subsubsection{\texorpdfstring{Derivation of \eqref{eq:fourth_condition} under stronger range assumption $r = \frac{1}{2}$.}{Derivation under the stronger range assumption}}
Notice that for any $x \in \R^d$, since $k$ is differentiable
\begin{align*}
    \nabla \left( \frac{d \mu_t}{d \pi} -1 \right)(x) = \left\langle \nabla k(x, \cdot), \frac{d \mu_t}{d \pi} -1 \right\rangle_{\calH^d} \leq \sqrt{K_{1d}} \left\| \frac{d \mu_t}{d \pi} -1 \right\|_\calH .
\end{align*}
And since $\frac{d \mu_t}{d \pi} -1 \in \calH$, there exists $q_t \in L^2(\pi)$ such that $\left\langle\frac{d \mu_t}{d \pi}-1, e_i\right\rangle_{L^2(\pi)}=\varrho_i^{1/2} \left\langle q_t, e_i\right\rangle_{L^2(\pi)} $ for all $i$, so
\begin{align*}
    \left\| \frac{d \mu_t}{d \pi} -1 \right\|_\calH^2 = \sum_{i \geq 1} \frac{1}{\varrho_i} \left\langle \frac{d \mu_t}{d \pi} -1, e_i \right\rangle_{L^2(\pi)}^2 = \sum_{i \geq 1} \left\langle q_t, e_i \right\rangle_{L^2(\pi)}^2 = \| q_t \|_{L^2(\pi)}^2 .
\end{align*}
Combining the above two equations, we have
\begin{align}
    \left\| \nabla \left( \log \pi \right)^\top \nabla \left( \frac{d \mu_t}{d \pi} -1 \right) \right\|_{L^2(\pi)} \leq \sqrt{K_{1d}} \| q_t \|_{L^2(\pi)} \left\| \nabla  \log \pi \right\|_{L^2(\pi)} .\nonumber
\end{align}
Also, for the other one, notice that
\begin{align*}
    \Delta \left(\frac{d \mu_t}{d \pi} \right) (x) &\leq \left\| \bH \left(\frac{d \mu_t}{d \pi} -1 \right) (x) \right\|_F = \left\| \left\langle \bH k(x, \cdot ) , \frac{d \mu_t}{d \pi} -1 \right\rangle \right\|_F \\
    &\leq \left\| \bH k(x, \cdot ) 
    \right\|_{\calH^{d \times d}} \left\| \frac{d \mu_t}{d \pi} -1 \right\|_\calH \leq \sqrt{K_{2d} } \| q_t \|_{L^2(\pi)} .
\end{align*}
Therefore, 
\begin{align}
    \left\| \Delta \left(\frac{d \mu_t}{d \pi} \right)  \right\|_{L^2(\pi)} \leq \sqrt{K_{2d} } \| q_t \|_{L^2(\pi)} .\nonumber
\end{align}

\subsection{\texorpdfstring{Proof of \Cref{prop:descent_lemma_KL}}{Proof of the KL descent lemma}}\label{appsec:proof_kl_descent_lemma}
We know that $\mu_{n+1} = (\Id-\gamma (1+\lambda) \nabla h)_{\#} \mu_n$ where we drop the subscripts of the witness function $h_{\mu_n ,\pi}$ when it causes no ambiguity.
Denote $\rho_t = (\Id- t (1+\lambda) \nabla h)_{\#} \mu_n$, so $\rho_0 = \mu_n$, and $\rho_{\gamma} = \mu_{n+1}$. 
Consider the difference of KL divergence between the two iterates $\mu_{n+1}$ and $\mu_n$ along the time-discretized $\dmmd$ flow:
\begin{align}
    \kl(\mu_{n+1} \| \pi) - \kl(\mu_{n} \| \pi) &= \kl(\rho_{\gamma} \| \pi) - \kl(\rho_0 \| \pi)  \nonumber
    \\ 
    &= \frac{d }{d t}\Big|_{t=0} \kl(\rho_t \| \pi) \gamma + \int_0^{\gamma} (\gamma - t)\frac{d^2 }{d t^2} \kl(\rho_t \| \pi) dt . \label{appeq:discrete_all}
\end{align}
For the first term of \eqref{appeq:discrete_all},
\begingroup\small
\begin{align}\label{appeq:first_term}
    &\quad \frac{d }{d t}\Big|_{t=0} \kl(\rho_t \| \pi) \nonumber \\
    &= -(1 + \lambda)
    \E_{\mu_n} \left[ \nabla \log \frac{d \mu_n}{ d \pi} ^\top \nabla h \right] = - (1 + \lambda) \E_{\pi} \left[ \nabla \frac{d \mu_n}{ d \pi} ^\top \nabla h \right] \nonumber \\
    &= -2 (1 + \lambda) \left\| \nabla \frac{d \mu_n}{ d \pi}  \right\|_{L^2(\pi)}^2 + (1 + \lambda) \E_{\pi} \left[ \nabla \frac{d \mu_n}{ d \pi} ^\top \left( 2 \nabla \frac{d \mu_n}{ d \pi}  - \nabla h \right) \right] \nonumber \\
    &= -2 (1 + \lambda) \left\| \nabla \frac{d \mu_n}{ d \pi}  \right\|_{L^2(\pi)}^2 \nonumber \\
    &\qquad + (1 + \lambda) \int \left( \nabla \frac{d \mu_n}{ d \pi} (x) \right)^\top \left( 2 \nabla \left( \frac{d \mu_n}{ d \pi} (x) -1 \right) - \nabla h(x) \right) \pi(x) dx \nonumber \\
    &= -2 (1 + \lambda) \left\| \nabla \frac{d \mu_n}{ d \pi}  \right\|_{L^2(\pi)}^2 - \nonumber \\
    &\qquad (1 + \lambda) \int \left( 2 \left( \frac{d \mu_n}{ d \pi} (x) - 1 \right) - h(x) \right) \nabla \cdot \left( \pi(x) \nabla \frac{d \mu_n}{ d \pi} (x) \right) dx \nonumber \\
    &\leq - (1 + \lambda) \frac{2}{C_P} \chi^2(\mu_n \| \pi) + (1 + \lambda) \left\|  h - 2 \left( \frac{d \mu_n}{d \pi} -1 \right) \right\|_{L^2(\pi)} \left\| \frac{ \nabla \cdot \left( \pi \nabla \frac{d \mu_n}{ d \pi}  \right) }{ \pi } \right\|_{L^2(\pi)},
\end{align}
\endgroup
where the fourth equality uses an integration by parts, and the last inequality uses Poincar\'{e} inequality for the first term under similar arguments in \Cref{appsec:proof_continuous_time_convergence} and uses Cauchy-Schwarz for the second term. 
Using \Cref{lem:h_difference} and the derivations in \eqref{appeq:continous_term_2}, \eqref{appeq:first_term} can be further upper bounded by
\begin{align}\label{appeq:first_term_cont}
    \frac{d }{d t}\Big|_{t=0} \kl(\rho_t \| \pi) \leq -(1 + \lambda) \frac{2}{C_P} \chi^2(\mu_n \| \pi)  + 2(1 + \lambda) \lambda^{r} Q \left( \mathcal{J} +\mathcal{I} \right). 
\end{align}
Then, for the second term of \eqref{appeq:discrete_all}, we know from Example 15.9 of \cite{villani2009optimal} (taking $m=1$) that,
\begin{align*}
\frac{d^2 }{d t^2} \kl(\rho_t \| \pi) &= (1 + \lambda)^2 \int \nabla h( x )^\top \bH V( \varphi_t(x) ) \nabla h(x) d \mu_n(x) \\
&+ (1 + \lambda)^2 \int \left\| \bH h(x) \left(\Id - t (1 + \lambda) \bH h(x) \right)^{-1} \right\|_F^2 d\mu_n(x).
\end{align*}
Because $2(1+\lambda) \gamma \sqrt{\chi^2\left(\mu_0 \| \pi\right) \frac{K_{2 d}}{\lambda}} \leq \frac{\zeta-1}{\zeta}$ for $1 < \zeta < 2$, applying \Cref{lem:hessian_terms} we have,
\begin{align}\label{appeq:second_term}
    \frac{d^2 }{d t^2} \kl(\rho_t \| \pi) \leq 4 (1 + \lambda)^2 \beta \chi^2(\mu_n \| \pi) \frac{K_{1d}}{\lambda} + 4 (1 + \lambda)^2 \zeta^2 \chi^2(\mu_n \| \pi) \frac{K_{2d}}{\lambda}.
\end{align}
Combining the above two inequalities \eqref{appeq:first_term_cont} and \eqref{appeq:second_term} and plugging them back into \eqref{appeq:discrete_all}, we obtain
\begin{align*}
\kl(\mu_{n+1} \| \pi) - \kl(\mu_{n} \| \pi) &\leq -\frac{2}{C_P} (1 + \lambda ) \chi^2(\mu_n \| \pi) \gamma \\
&+ 2 (1 + \lambda )\gamma \lambda^{r} Q \left( \mathcal{J} +\mathcal{I} \right) \\
&+ 2(1 + \lambda )^2 \gamma^2 (\beta + \zeta^2) \chi^2(\mu_n \| \pi) \frac{K_{1d}+K_{2d} }{\lambda} \\
& \leq -\frac{2}{C_P} \chi^2(\mu_n \| \pi) \gamma \\
&+ 4 \gamma \lambda^{r} Q \left( \mathcal{J} +\mathcal{I} \right) \\
&+ 8 \gamma^2 (\beta + \zeta^2) \chi^2(\mu_n \| \pi) \frac{K_{1d}+K_{2d} }{\lambda},
\end{align*}
where the last inequality holds by using $0 < \lambda \leq 1$, and the result follows.

\subsection{\texorpdfstring{Proof of \Cref{thm:discrete_time_KL}}{Proof of the discrete-time KL theorem}}\label{appsec:proof_discrete_time_convergence}
In order to use \Cref{prop:descent_lemma_KL} in the proof, first we are going to show that \Cref{prop:descent_lemma_KL} holds under the conditions of \Cref{thm:discrete_time_KL}.
Notice that conditions 1-4 of \Cref{prop:descent_lemma_KL} are precisely the conditions 1-4 of \Cref{thm:discrete_time_KL}. To use \Cref{prop:descent_lemma_KL} in the proof of \Cref{thm:discrete_time_KL}, the only thing left is to check that the condition of step size $\gamma$ in \eqref{appeq:gamma_condition_1} is satisfied. 

In \Cref{thm:discrete_time_KL}, $\lambda_n$ is selected to be $\left( 2 \gamma \frac{\chi^2(\mu_n \| \pi) ( \beta + \zeta^2 ) (K_{1d} + K_{2d})}{ Q( \mathcal{J} +\mathcal{I}) } \right)^{\frac{1}{r+1}} \wedge 1$. 
If $\lambda_n$ is taken to be the former, then
\begin{align}
    &\quad 2 \gamma (1+\lambda_n) \sqrt{ \chi^2(\mu_n \| \pi) \frac{K_{2d}}{\lambda_n} } \leq 4 \gamma \sqrt{ \chi^2(\mu_n \| \pi) \frac{K_{2d}}{\lambda_n} } \nonumber \\
    &= 2 (2 \gamma)^{\frac{2r + 1}{2 r + 2 }} \chi^2\left(\mu_n \| \pi\right)^{\frac{r }{2 r +  2 } } \left( Q (\mathcal{J} + \mathcal{I} ) \right)^{\frac{1}{2r + 2 }} \left(  \frac{ 1 }{K_{1d} + K_{2d}}  \frac{ 1 }{\beta + \zeta^2 } \right)^{\frac{1}{2r + 2}} K_{2d}^{\frac{1}{2}} \nonumber  \\
    &\stackrel{(*)}{\leq} \left(8 \gamma\right)^{\frac{2r + 1}{2 r + 2 }} Q^{\frac{2r + 1}{2r + 2}} (\mathcal{J} + \mathcal{I} )^{\frac{1}{2r + 2 }} \left(\frac{ 1 }{\beta + \zeta^2 } \right)^{\frac{1}{2r + 2}} K_{2d}^{\frac{r}{2r + 2 }}  \nonumber  \\
    &\leq \frac{\zeta - 1 }{\zeta} \left(\frac{ 1 }{\beta + \zeta^2 } \right)^{\frac{r}{2r + 2}} \leq \frac{\zeta - 1 }{\zeta}  . \label{appeq:dummy_2}
\end{align}
$(*)$ holds because $\frac{K_{2d}}{K_{1d} + K_{2d}} \leq 1$ and
\begin{align}\label{appeq:KQ}
    \chi^2\left(\mu_n \| \pi\right) = \left\| \frac{d \mu_n}{d \pi} - 1\right\|_{L^2(\pi)}^2 = \left\| \calT_\pi^{r} q_n \right\|_{L^2(\pi)}^2 \leq K^{2r} Q^2 \leq Q^2.
\end{align}
The second last inequality of \eqref{appeq:dummy_2} holds due to the constraint on $\gamma$ in \eqref{eq:gamma_condition}, and the last inequality of \eqref{appeq:dummy_2} holds because $\beta + \zeta^2 \geq 1$.

On the other hand, if $\lambda_n$ is chosen to be $1$, similarly based on the constraint on $\gamma$ in \eqref{eq:gamma_condition}, we have
\begin{align*}
    2 \gamma (1+\lambda_n) \chi^2(\mu_n \| \pi) \frac{K_{2d}}{\lambda_n} &= 4 \gamma \chi^2(\mu_n \| \pi) K_{2d} \leq \frac{\zeta-1}{\zeta}.
\end{align*}
Therefore, all the conditions of \Cref{prop:descent_lemma_KL} have been verified. 
So, if we select
\begin{align*}
    \lambda_n =
    \left(
    2 \gamma
    \frac{\chi^2(\mu_n \| \pi) ( \beta + \zeta^2 ) (K_{1d} + K_{2d})}
    {Q( \mathcal{J} +\mathcal{I})}
    \right)^{\frac{1}{r+1}}
    \wedge 1,
\end{align*}
then 
\begin{align}\label{appeq:KL_descent_1}
\kl(\mu_{n+1} \| \pi) - \kl(\mu_{n} \| \pi) &\leq - \gamma \frac{2}{C_P} \chi^2(\mu_n \| \pi) \nonumber \\
&\hspace{-3em} + \underbrace{ 4 \gamma \lambda_n^{r} Q \left( \mathcal{J} +\mathcal{I} \right) }_{(\Delta_1)} + \underbrace{ 8 \gamma^2 (\beta + \zeta^2) \chi^2(\mu_n \| \pi) \frac{K_{1d}+K_{2d} }{\lambda_n} }_{(\Delta_2)}.
\end{align}
By observing \eqref{appeq:KL_descent_1}, the first term on the right-hand side $-\frac{2}{C_P} \chi^2(\mu_n \| \pi) \gamma $ is strictly negative and is decreasing KL divergence at each iteration $n$ of the $\dmmd$ gradient descent. 
In contrast, the second term $(\Delta_1) := 4 \gamma \lambda_n^{r} Q( \mathcal{J} +\mathcal{I})$ and the third term $(\Delta_2) := 8 \gamma^2 (\beta + \zeta^2) \chi^2(\mu_n \| \pi) \frac{K_{1d}+K_{2d} }{\lambda_n}$ are positive and prevent the KL divergence from decreasing. 
Denote $G(\lambda_n) = (\Delta_1) + (\Delta_2)$ and the optimal $\lambda_n$ is achieved by taking $\frac{d}{ d \lambda_n} G(\lambda_n) = 0$, which leads to $\lambda_n = \left( 2 \gamma \frac{\chi^2(\mu_n \| \pi) ( \beta + \zeta^2 ) (K_{1d} + K_{2d})}{ Q( \mathcal{J} +\mathcal{I}) } \right)^{\frac{1}{r+1}}$.
Plugging the value of $\lambda_n$ back to \eqref{appeq:KL_descent_1} to obtain,
\begingroup\small
\begin{align}\label{eq:kl_descent_2}
&\quad \kl(\mu_{n+1} \| \pi) - \kl(\mu_{n} \| \pi) \nonumber \\
&\le -\frac{2}{C_P} \chi^2(\mu_n \| \pi) \gamma + 4 \gamma \left( 2 \gamma\right)^{\frac{r}{r + 1} } \left( \chi^2(\mu_n \| \pi) ( \beta + \zeta^2 ) (K_{1d} + K_{2d}) \right)^{\frac{ r }{ r + 1 }}\nonumber\\
&\qquad {}\times \left( Q( \mathcal{J} +\mathcal{I}) \right)^{\frac{ 1 }{ r + 1 }} \nonumber \\
&\leq -\frac{2}{C_P} \chi^2(\mu_n \| \pi) \gamma + 8 \gamma^{1 + \frac{ r }{ r + 1 }} \left( \chi^2(\mu_n \| \pi) ( \beta + \zeta^2 ) (K_{1d} + K_{2d}) \right)^{\frac{ r }{ r + 1 }}\nonumber\\
&\qquad {}\times \left( Q( \mathcal{J} +\mathcal{I})\right)^{\frac{ 1 }{ r + 1 }} \nonumber \\
&\leq -\frac{2}{C_P} \chi^2(\mu_n \| \pi) \gamma + 8 \gamma^{1 + \frac{ r }{ r + 1 } } Q^{\frac{2r+1}{r+1}} \Big( (K_{1d} + K_{2d})(\beta + \zeta^2) \Big)^{\frac{ r }{ r + 1 }} ( \mathcal{J} + \mathcal{I})^{\frac{ 1 }{ r + 1 }} ,
\end{align}
\endgroup
where the last inequality holds because of \eqref{appeq:KQ}.
Since $ \chi^2(\mu_n \| \pi) \geq \kl(\mu_n \| \pi)$~\citep[Equation (7)]{van2014renyi}, we have
\begin{align*}
    \kl(\mu_{n+1} \| \pi)& \leq \left(1 - \gamma \frac{2}{C_P} \right) \kl(\mu_{n} \| \pi) \nonumber\\
    &\qquad\qquad+ 8 \gamma^{1 + \frac{ r }{ r + 1 } } Q^{\frac{2r+1}{r+1}} \Big( (K_{1d} + K_{2d})(\beta + \zeta^2) \Big)^{\frac{ r }{ r + 1 }} ( \mathcal{J} +\mathcal{I})^{\frac{ 1 }{ r + 1 }}  .
\end{align*}
After iterating, we obtain
\begin{align*}
    &\quad \kl(\mu_{n_{max} } \| \pi) \leq \left(1 - \gamma 
    \frac{2}{C_P} \right)^{n_{max}} \kl(\mu_0 \| \pi) \\
    & + 4 \gamma^{ \frac{ r }{ r + 1 } } C_P Q^{\frac{2r+1}{r+1}}
    \Big( (K_{1d} + K_{2d})(\beta + \zeta^2) \Big)^{\frac{ r }{ r + 1 }} \\
    &\hspace{80pt}{}\times ( \mathcal{J} +\mathcal{I})^{\frac{ 1 }{ r + 1 }} \\
    &\leq \exp\left(- \frac{ 2 n_{max} \gamma}{C_P} \right) \kl(\mu_0 \| \pi) 
    + 4 \gamma^{ \frac{ r }{ r + 1 } } C_P Q^{\frac{2r+1}{r+1}}
    \Big( (K_{1d} + K_{2d})(\beta + \zeta^2) \Big)^{\frac{ r }{ r + 1 }} \\
    &\hspace{80pt}{}\times ( \mathcal{J} +\mathcal{I})^{\frac{ 1 }{ r + 1 }}
\end{align*}
and the result follows.

\subsection{\texorpdfstring{Proof of \Cref{thm:discrete_time_KL_2}}{Proof of the second discrete-time KL theorem}}\label{appsec:proof_discrete_time_convergence_2}
The proof of \Cref{thm:discrete_time_KL_2} is also based on \Cref{prop:descent_lemma_KL} proved in the last section. 
Recalling \eqref{eq:kl_descent_2} yet with adaptive step size $\gamma_n$, we have
\begin{align*}
    &\quad \kl(\mu_{n+1} \| \pi) - \kl(\mu_{n} \| \pi) \\
    &\leq -\frac{2}{C_P} \chi^2(\mu_n \| \pi) \gamma_n
    + 8 \gamma_n^{1 + \frac{ r }{ r + 1 }} \\
    &\quad {}\times \left( \chi^2(\mu_n \| \pi)
    ( \beta + \zeta^2 ) (K_{1d} + K_{2d}) \right)^{\frac{ r }{ r + 1 }}
    \left( Q( \mathcal{J} +\mathcal{I})\right)^{\frac{ 1 }{ r + 1 }}.
\end{align*}
From \eqref{eq:gamma_condition_2}, we have 
\begin{align*}
    8 \gamma_n^{\frac{ r }{ r + 1 } } \Big( (K_{1d} + K_{2d})(\beta + \zeta^2) \Big)^{\frac{ r }{ r + 1 }} ( Q( \mathcal{J} +\mathcal{I}) )^{\frac{ 1 }{ r + 1 }} \leq \frac{1}{C_P} \chi^2(\mu_n \| \pi)^{\frac{1}{r + 1}},
\end{align*}
so that we have
\begin{align*}
    \kl(\mu_{n+1} \| \pi) - \kl(\mu_{n} \| \pi) \leq -\frac{1}{C_P} \chi^2(\mu_n \| \pi) \gamma_n \leq -\frac{1}{C_P} \kl(\mu_n \| \pi) \gamma_n.
\end{align*}
Hence
\begin{align}\label{eq:kl_monotone}
    \kl(\mu_{n+1} \| \pi) \leq \left( 1 - \frac{1}{C_P} \gamma_n \right) \kl(\mu_n \| \pi) .
\end{align}
After iterating $n$ from $1$ to $n_{\max}$, the theorem is proved.

\subsection{\texorpdfstring{Finite-sample expression for $\dmmd$}{Finite-sample expression for DrMMD}}\label{appsec:proof_empirical_chard}
By defining the following operators,
\begin{align*}
    &S_x: \calH \rightarrow \mathbb{R}^M, \quad f \rightarrow \frac{1}{\sqrt{M}}\left[ f ( x^{(1)}), \ldots, f ( x^{(M)} ) \right]^{\top}, \\
    &S_x^\ast : \R^M \rightarrow \calH , \quad \alpha \rightarrow \frac{1}{\sqrt{M} } \sum_{i=1}^M \alpha_i k( x^{(i)}, \cdot), \\
    &S_y: \calH \rightarrow \mathbb{R}^N, \quad f \rightarrow \frac{1}{\sqrt{N}}\left[ f ( y^{(1)}), \ldots, f ( y^{(N)} ) \right]^{\top}, \\
    &S_y^\ast : \R^N \rightarrow \calH , \quad \alpha \rightarrow \frac{1}{\sqrt{N}} \sum_{i=1}^N \alpha_i k( y^{(i)}, \cdot) .
\end{align*}
Then we have
\begin{align*}
    \Sigma_{\hat{\pi}} = S_x^\ast S_x, \quad K_{xx} = M S_x S_x^\ast, \quad K_{xy} = \sqrt{MN} S_x S_y^\ast.
\end{align*}
Using these, note that
\begingroup\small
\begin{align}
&h_{\hat{\mu}, \hat{\pi}}^\ast = 2 \left(\Sigma_{\hat{\pi}}+\lambda \Id\right)^{-1}\left( m_{\hat{\mu}} - m_{\hat{\pi}} \right) \nonumber \\
& =2\left(\frac{1}{M} \sum_{i=1}^M k\left(x^{(i)}, \cdot\right)
\otimes k\left(x^{(i)}, \cdot\right)+\lambda \Id\right)^{-1} \nonumber \\
&\quad {}\times \left(\frac{1}{N} \sum_{i=1}^N k\left(y^{(i)}, \cdot\right)
-\frac{1}{M} \sum_{i=1}^M k\left(x^{(i)}, \cdot\right)\right) \nonumber \\
& =2 \Big( S_x^\ast S_x + \lambda \Id \Big)^{-1}\left( \frac{1}{\sqrt{N} } S_y^\ast \one_N - \frac{1}{\sqrt{M}} S_x^\ast \one_M \right) . \label{eq:empirical_h_temp} 
\end{align}
\endgroup
From the Woodbury inversion lemma, we have that
\begin{align*}
    \Big( S_x^\ast S_x + \lambda \Id \Big)^{-1} = \frac{1}{\lambda} \Id - \frac{1}{\lambda} S_x^\ast (S_x S_x^\ast + \lambda\Id )^{-1} S_x .
\end{align*}
Plugging the above into \eqref{eq:empirical_h_temp}, we obtain
\begin{align}
    & h_{\hat{\mu}, \hat{\pi}}^\ast
    = 2 \left( \frac{1}{\lambda} \Id
    - \frac{1}{\lambda} S_x^\ast (S_x S_x^\ast + \lambda\Id )^{-1} S_x \right) \nonumber \\
    &\quad {}\times \left( \frac{1}{\sqrt{N}} S_y^\ast \one_N
    - \frac{1}{\sqrt{M}} S_x^\ast \one_M \right) \nonumber \\
    &= \frac{2}{\lambda} \left( \frac{1}{\sqrt{N}} S_y^\ast \one_N
    - \frac{1}{\sqrt{M}} S_x^\ast \one_M \right) \nonumber \\
    &\quad - \frac{2}{\lambda} S_x^\ast
    \left( \frac{1}{M} K_{xx} + \lambda\Id \right)^{-1}
    \frac{1}{N \sqrt{M}} K_{xy} \one_N \nonumber \\
    &\qquad\qquad+ \frac{2}{\lambda} S_x^\ast \left( \frac{1}{M} K_{xx} + \lambda\Id \right)^{-1} \frac{1}{M \sqrt{M} } K_{xx} \one_M \nonumber \\
    &=\frac{2}{N \lambda} k\left(\cdot, y^{1:N}\right) \one_N
    - \frac{2}{M \lambda} k\left(\cdot, x^{1:M}\right) \one_M \nonumber \\
    &\quad - \frac{2}{N \lambda} k\left(\cdot, x^{1:M}\right)
    \left( K_{xx} + M \lambda \Id \right)^{-1} K_{xy} \one_N \nonumber \\
    &\qquad\qquad +\frac{2}{M \lambda} k\left(\cdot, x^{1:M} \right)\left( K_{xx} + M \lambda \Id \right)^{-1} K_{xx} \one_M. \label{appeq:empirical_h}
\end{align}
Obtaining $\dmmd(\hat{\mu} \| \hat{\pi})$ is then easy with $h_{\hat{\mu}, \hat{\pi}}^\ast$ shown in \eqref{appeq:empirical_h}.
\begin{align}\label{appeq:dmmd_samples}
    &\quad \dmmd(\hat{\mu} \| \hat{\pi}) = (1+\lambda)\left\|\left(\Sigma_{\hat{\pi}} + \lambda \Id \right)^{-\frac{1}{2}}\left( m_{\hat{\mu}} - m_{\hat{\pi}} 
    \right)\right\|_{\calH}^2 \nonumber \\
    &= (1 + \lambda) \left\langle\frac{1}{2} h_{\hat{\mu}, \hat{\pi}}^*, m_{\hat{\mu}} - m_{\hat{\pi}} \right\rangle_{\calH} \nonumber \\
    & = (1 + \lambda) \left\langle\frac{1}{2} h_{\hat{\mu}, \hat{\pi}}^*, \frac{1}{N} \sum_{i=1}^N k\left(y^{(i)}, \cdot\right) - \frac{1}{M} \sum_{i=1}^M k\left(x^{(i)}, \cdot\right) \right\rangle_{\calH} \nonumber \\ 
    & =\frac{1 + \lambda }{\lambda} 
    \Bigg( \frac{1}{N^2} \one_N^{\top} K_{yy} \one_N + \frac{1}{M^2} \one_M^{\top} K_{xx} \one_M - \frac{2}{M N} \one_M^{\top} K_{xy}  \one_N \nonumber \\
    &- \frac{1}{N^2} \one_N^{\top} K_{xy}^\top \left(K_{xx} + M \lambda \Id\right)^{-1} K_{xy}  \one_N +\frac{2}{N M} \one_M^{\top} K_{xx}\left(K_{xx} + M \lambda \Id\right)^{-1} K_{xy}  \one_N \nonumber \\ 
    &\qquad-\frac{1}{M^2} \one_M^{\top} K_{xx}\left( K_{xx} + M \lambda \Id \right)^{-1} K_{xx} \one_M \Bigg) .
\end{align}

\appsection[Further Background on Wasserstein Space]{Further Background on\\ $(\calP_2(\R^d), W_2)$ }\label{appsec:wass_discussion}
Let $\mu$ and $\pi$ be two probability measures in $\cP_2(\R^d)$ and let $\Pi (\mu,\pi)$ denote the set of all admissible transport plans between $\mu$ and $\pi$, i.e., $\Pi (\mu, \pi) =\{ \Gamma \in \cP(\R^d\times \mathbb{R}^d); \left( \operatorname{proj}_{1}\right)_{\#} \Gamma=\mu,\left(\operatorname{proj}_{2}\right)_{\#} \Gamma=\pi\}$, where $\operatorname{proj}_{1}$ and $\operatorname{proj}_{2}$ respectively stand for projection maps $(x, y) \mapsto x$ and $(x, y) \mapsto y$, and $\#$ is the pushforward operator. The Wasserstein-2 distance between $\mu$ and $\pi$ is then defined as
\begin{align*}
W_2(\mu, \pi)=\left(\inf _{\Gamma \in \Pi(\mu, \pi)} \int \| x-y \|^2 d \Gamma(x, y)\right)^{\frac{1}{2}},
\end{align*}
and $(\cP_2(\R^d), W_2)$ is a metric space called the Wasserstein space \citep{panaretos2020wasserstein}.
Brenier's theorem guarantees that if $\mu$ is an absolutely continuous measure, then the optimal transport map is unique and is of the form $\Gamma^{\ast} = (\Id, T)_\# \mu$, i.e., $T_\# \mu = \pi$~\citep{santambrogio2017euclidean}. $T$ can also be expressed as $T(x) = x + \nabla \phi(x)$, where $\phi$ is known as the Kantorovich potential function and is differentiable $\mu-$a.e. 

For an absolutely continuous $\mu$ and the optimal transport plan $T$ such that $T_\# \mu = \pi$, the shortest path $(\rho_{t})_{0\leq t \leq 1}$ from $\mu$ to $\pi$ is called the (Wasserstein) geodesic given by the following form:
\begin{align*}
\rho_{t} = \left((1-t) \Id + t T \right)_{\#} \mu = \left( \Id + t \nabla \phi \right)_{\#} \mu. 
\end{align*}
Therefore, in this paper, we always use $(\Id + t \nabla \phi)_\# \mu$ with 
$\phi \in C_c^\infty(\R^d)$\footnote{$\phi$ is compactly supported because the tangent space of $\mu \in\calP_2(\R^d)$ is $\overline{ \{\nabla \psi, \psi \in C_c^{\infty} (\R^d)\} }^{L^2(\mu)}$~\citep[Definition 8.4.1]{ambrosio2005gradient}. } to define a geodesic curve that starts at $\mu$.
Define $\varphi_t(x) = x + t \nabla \phi(x)$ and $\omega_t(x) = \left[\nabla \phi \circ \varphi_t^{-1}\right](x)$.
Then $\omega_t$ becomes the optimal transport map from $\rho_t$ to $\pi$. Notice that $\| \omega_t \|_{L^2(\rho_t)} = \| \nabla \phi \|_{L^2(\mu)}$ for all $t\in [0,1]$, so
$(\rho_{t})_{0\leq t \leq 1}$ is also a constant-speed geodesic.
The notion of a constant-speed geodesic is crucial in the introduction of geodesic convexity below.

A functional $\mathcal{F}: \calP_2(\R^d) \to \R$ is geodesically convex if for any $\mu$ and $\pi$, the following inequality holds:
\begin{align}\label{appeq:displacement_convex_defi}
    \mathcal{F}\left(\rho_t\right) \leq(1-t) \mathcal{F}\left(\mu \right)+t \mathcal{F}\left(\pi \right), \quad \forall t \in[0,1],
\end{align}
where $\left(\rho_t\right)_{t \in[0,1]}$ is the constant-speed geodesic between $\mu$ and $\pi$. 
The geodesic convexity of $\calF$ can be equivalently characterized through the Wasserstein Hessian~\citep{villani2009optimal}. 
The geodesic convexity ought not to be confused with mixture convexity, which replaces displacement geodesic $\rho_t$ in \eqref{appeq:displacement_convex_defi} with the mixture geodesic $\nu_t = (1-t) \mu + t \pi$. 

\setcounter{equation}{0}
\appsection{Auxiliary Results}
In this appendix, we collect all technical results required to prove the main results of the paper.
\begin{lem}\label{lem:variation_chi2}
For $\mu \ll \pi$, $\chi^2$-divergence admits the following variational form:
\begin{align*}
    \chi^2(\mu \| \pi)=\sup _{h \in L^2(\pi)} \int h d \mu-\int\left(h+\frac{1}{4} h^2\right) d \pi,
\end{align*}
where it is sufficient to restrict the variational set to $L^2(\pi)$ in contrast to the set of all measurable functions for general $f$-divergences \citep{nowozin2016f, nguyen2010estimating}.
\end{lem}
\begin{proof}
For $\mu \ll \pi$, we have:
\begin{align*}
    &\chi^2(\mu \| \pi)  =\sup_h\left\{\int h d \mu-\int\left(\frac{h^2}{4} + h\right) d \pi\right\} \\ & =\sup_h\left\{\int h \frac{d \mu}{d \pi} d \pi-\int\left(\frac{h^2}{4}+h\right) d \pi\right\} 
     =\sup_h\left\{\int\left(h \frac{d \mu}{d \pi} -\frac{h^2}{4}-h\right) d \pi\right\} \\ 
    & =-\inf _h \left\{\int\left[\frac{h^2}{4}- h \left( \frac{d \mu}{d \pi} - 1 \right) \right] d \pi\right\} \\ & =-\inf _h \left\{\int\left[\frac{h}{2}- \left( \frac{d \mu}{d \pi} -1 \right) \right]^2 d \pi\right\} + \int \left( \frac{d \mu}{d \pi} - 1 \right)^2 d \pi .
\end{align*}
Clearly, the above equation is minimized at $h^*=2(\frac{d \mu}{d \pi} - 1)$ and $\chi^2(\mu \| \pi)=\int(\frac{d \mu}{d \pi} - 1)^2 d \pi$ which is finite if and only if $\frac{d \mu}{d \pi} -1 \in L^2(\pi)$. 
Therefore, it is sufficient to consider the above maximization over $L^2(\pi)$.
\end{proof}
\begin{lem}\label{lem:k}
Under {\color{black}\Cref{assumption:universal,assumption:bounded_kernel}}, the mappings  $x \mapsto k(x, \cdot)$ and $x \mapsto \nabla_1 k(x, \cdot)$ are differentiable and Lipschitz:
\begin{align*}
\begin{aligned}
\left\|k(x, \cdot) - k(y,\cdot) \right\|_\calH & \leq \sqrt{K_{1 d}}\|x-y\|, \\
\left\|\nabla_1 k(x, \cdot) - \nabla_1 k(y, \cdot) \right\|_{\calH^d} & \leq \sqrt{K_{2 d}}\|x-y\| . 
\end{aligned}
\end{align*}
\end{lem}
\begin{proof}
This is Lemma 7 from \cite{glaser2021kale}. 
\end{proof}

\begin{lem}\label{lem:k_tilde}
Under {\color{black}\Cref{assumption:universal,assumption:bounded_kernel}}, the regularized kernel $\tilde{k}$ defined in \eqref{eq:tilde_k} satisfies the following properties:\vspace{2mm}
\begin{enumerate}[itemsep=5.0pt,topsep=0pt,leftmargin=*]
\item $\big| \tilde{k}(x, y) \big| \leq \frac{K}{\lambda} $;
\item $\partial_i \tilde{k}(x, y)  =  \big\langle (\Sigma_\pi+\lambda \Id)^{-1} \partial_i k(x, \cdot), k(y, \cdot) \big\rangle_{\calH} $;
\item $ \big\| \nabla_1 \tilde{k}(x, y) \big\|^2 = \sum_{i=1}^d \partial_i \tilde{k}(x, y)^2 \leq \frac{ K K_{1d} }{\lambda^2}$;
\item $\partial_i \partial_{i+d} \tilde{k}(x, y) = \big\langle \left(\Sigma_\pi+\lambda \Id\right)^{-1} \partial_i k(x, \cdot), \partial_i k(y, \cdot) \big\rangle_{\calH} $; 
\item $ \big\| \nabla_1 \nabla_2 \tilde{k}(x, y) \big \|_F^2 = \sum_{i=1}^d \partial_i \partial_{i+d} \tilde{k}(x, y)^2 \leq \frac{K_{1d} }{\lambda^2 }$;
\item $\partial_i \partial_j \tilde{k}(x, y) = \big \langle \left(\Sigma_\pi+\lambda \Id\right)^{-1} \partial_i \partial_j k(x, \cdot), k(y, \cdot) \big\rangle_{\calH} $;
\item $\big\| \bH_1 \tilde{k}(x, y) \big\|_F^2 = \sum_{i,j = 1}^d \partial_i \partial_{j} \tilde{k}(x, y)^2 \leq \frac{ K K_{2d} }{\lambda^2}$; 
\item $\big\|\nabla_1 \tilde{k} \left(x, x^{\prime}\right) - \nabla_1 \tilde{k} \left(y, y^{\prime}\right) \big\| \leq \frac{\sqrt{K K_{2d}} }{\lambda}  \left( \left\| x - y \right\| + \left\| x^\prime - y^\prime \right\| \right)$.
\end{enumerate}
\end{lem}
\begin{proof}
Notice that
\begin{align*}
    \tilde{k}(x, y)
    &= \left\langle\left(\Sigma_\pi+\lambda \Id\right)^{-1}
    k(x, \cdot), k(y, \cdot) \right\rangle_{\calH} \\
    &\leq \frac{1}{\lambda} \left\| k(x, \cdot) \right\|_\calH
    \left\| k(y, \cdot) \right\|_\calH
    \leq \frac{K}{\lambda},
\end{align*}
so the first bullet point is proved.
Before we prove the second bullet point, we first prove the differentiability of $x \mapsto \tilde{k}(x,y)$.
For $i \in \{1, \cdots, d\}$, consider $h \in \R$ and denote $\Delta_i \in \R^d$ as a vector of all $0$ except the value at $i$ being equal to $h$. Then, for any $y \in \R^d$,
\begin{align*}
    &\lim_{h \to 0} \frac{ \tilde{k}(x + \Delta_i, y) - \tilde{k}(x, y)}{h} \\
    &= \lim_{h \to 0} \frac{1}{h}
    \left\langle (\Sigma_\pi + \lambda \Id)^{-1}
    \left( k(x + \Delta_i, \cdot) - k(x, \cdot)\right),
    k(y, \cdot) \right\rangle_\calH \\
    &\leq \lim_{h \to 0} \frac{\sqrt{K} }{\lambda}  \frac{\left\|k(x, \cdot) - k(x + \Delta_i,\cdot) \right\|_\calH}{h} 
    \leq \lim_{h \to 0} \frac{\sqrt{K} }{\lambda} \sqrt{K_{1d}} \frac{\| \Delta_i \| }{h} 
    = \frac{\sqrt{K} }{\lambda} \sqrt{K_{1d}}.
\end{align*}
So $x \mapsto \tilde{k}(x,y)$ is differentiable for any $y \in \R^d$.
Since the kernel $k$ is differentiable per {\color{black}\Cref{assumption:bounded_kernel}}, for any
$f \in \calH$, $\partial_{x_i} f(x) = \big \langle \partial_i k(x, \cdot) , f \big\rangle_\calH$. Hence,
\begin{align*}
    \partial_i \tilde{k}(x,y)
    &= \partial_{x_i} \left\langle k(x, \cdot),
    \left(\Sigma_\pi+\lambda \Id\right)^{-1} k(y, \cdot) \right\rangle_{\calH} \\
    &= \left\langle \partial_i k(x, \cdot),
    \left(\Sigma_\pi+\lambda \Id\right)^{-1} k(y, \cdot) \right\rangle_{\calH}.
\end{align*}
So the second bullet point is proved. 
Next, notice that
\begin{align*}
    \left\| \nabla_1 \tilde{k}(x, y) \right\|^2 &= \sum_{i=1}^d \left\langle \partial_i k(x, \cdot), \left(\Sigma_\pi+\lambda \Id\right)^{-1} k(y, \cdot) \right\rangle_{\calH}^2 
    \\
    &\leq \frac{1}{\lambda^2} \sum_{i=1}^d \left\| \partial_i k(x, \cdot) \right\|_\calH^2 \left\|  k(y, \cdot) \right\|_\calH^2 \\
    &\leq \frac{1}{\lambda^2} \left\| \nabla_1 k(x, \cdot) \right\|_{\calH^d}^2 \left\|  k(y, \cdot) \right\|_\calH^2 
    \leq \frac{K K_{1d}}{\lambda^2}.
\end{align*}
So the third bullet point is proved. 
Similar arguments above lead to bullet points $4$ to $7$. 

\noindent
Finally, to prove bullet point $8$, notice that
\begin{align*}
    &\left\|\nabla_1 \tilde{k} (x, x^{\prime})
    - \nabla_1 \tilde{k} (y, x^{\prime} ) \right\|^2 \\
    &\quad = \sum_{i=1}^d \left\langle
    \partial_i k(x, \cdot) - \partial_i k(y, \cdot),
    \left( \Sigma_\pi+\lambda \Id \right)^{-1} k(x^\prime,\cdot)
    \right\rangle_{\calH}^2 \\
    &\leq \frac{1}{\lambda^2} \sum_{j=1}^d \left\| \partial_i k(x, \cdot) - \partial_i k(y, \cdot)  \right\|_\calH^2 \left\| k(x^\prime,\cdot)  \right\|_\calH^2 
    \\
    &\leq \frac{1}{\lambda^2} \left\|\nabla_1 k(x, \cdot) - \nabla_1 k(y, \cdot)\right\|_{\calH^d}^2 \left\| k(x^\prime,\cdot)  \right\|_\calH^2 \\
    &\leq \frac{ K K_{2d} }{\lambda^2} \left\| x - y \right\|^2 ,
\end{align*}
where the last inequality uses \Cref{lem:k}.
Therefore,
\begin{align*}
    \left\|\nabla_1 \tilde{k} (x, x^{\prime})
    - \nabla_1 \tilde{k} (y, y^{\prime}) \right\|
    &\leq \left\|\nabla_1 \tilde{k} (x, x^{\prime})
    - \nabla_1 \tilde{k} (y, x^{\prime}) \right\| \\
    &\quad + \left\|\nabla_1 \tilde{k} (y, x^{\prime})
    - \nabla_1 \tilde{k} (y, y^{\prime}) \right\| \\
    &\leq \frac{\sqrt{K K_{2d}} }{\lambda}  \left( \| x - y \| + \| x^\prime - y^\prime \| \right) .
\end{align*}
So the bullet point $8$ is proved.
\end{proof}
\begin{lem}\label{lem:lipschitz_witness}
    For any two distributions $\mu_0, \mu_1 \in \mathcal{P}_2(\R^d) $, with associated $\dmmd$ witness functions $h_{\mu_0, \pi}^\ast, h_{\mu_1, \pi}^\ast$ defined in \Cref{prop:drmmd_representation_no_ratio}, we have
\begin{align*}
    \left\|h_{\mu_1, \pi}^\ast - h_{\mu_0, \pi}^\ast \right\|_\calH \leq \frac{2 \sqrt{K_{1d} }}{\lambda} W_2\left(\mu_0, \mu_1\right) ,\quad \text{and}\quad \left\|h_{\mu_0, \pi}^\ast \right\|_\calH \leq \frac{2 \sqrt{K}}{\lambda} .
\end{align*}
Both the witness function $h_{\mu, \pi}^\ast$ and its gradient $\nabla h_{\mu, \pi}^\ast$ are Lipschitz continuous, i.e.,
\begin{align*}
    \Big| h_{\mu_1, \pi}^\ast(x) - h_{\mu_0, \pi}^\ast(y) \Big| &\leq L  \Big( W_2(\mu_0, \mu_1) + \left\| x - y \right\| \Big); \\
    \Big\| \nabla h_{\mu_1, \pi}^\ast(x) - \nabla h_{\mu_0, \pi}^\ast(y) \Big\| &\leq L \Big( W_2(\mu_0, \mu_1) + \left\| x - y \right\| \Big),
\end{align*}
where the constant $L= \frac{ 1}{\lambda} \max \left\{ 2 \sqrt{K K_{1 d}}, 2 \sqrt{K K_{2 d}}, 2 K_{1d} \right\}$. 
\end{lem}
\begin{proof}
Let $\gamma \in \Gamma(\mu_0, \mu_1)$ be the optimal coupling between $\mu_0$ and $\mu_1$. Then
\begin{align*}
\begin{aligned}
    &\left\| h_{\mu_1, \pi}^\ast - h_{\mu_0, \pi}^\ast \right\|_\calH^2 = 4 \left\| \Big( \Sigma_\pi + \lambda \Id \Big)^{-1} \left(\int k(x, \cdot) d (\mu_1 - \mu_0) \right)\right\|^2_\calH \\
    &\leq \frac{4}{\lambda^2} \left\| \int k(x, \cdot) - k(y, \cdot) d \gamma(x,y) \right\|^2_\calH 
    \leq \frac{4 K_{1d} }{\lambda^2} \int \left\|x - y\right\|^2 d \gamma(x,y) 
    \\
    &= \frac{4 K_{1d} }{\lambda^2} W_2^2(\mu_0, \mu_1),
\end{aligned}
\end{align*}
where the first inequality holds because $\Sigma_\pi$ is a positive and self-adjoint operator, and the second inequality uses \Cref{lem:k}. Also note that
\begin{align*}
    \begin{aligned}
        \left\|h_{\mu_0, \pi}^\ast \right\|_\calH = \left\| 2 \Big( \Sigma_\pi + \lambda \Id \Big)^{-1} \left( \int k(x, \cdot) d \mu_0 - \int k(x, \cdot) d \pi \right) \right\|_\calH \leq \frac{2 \sqrt{K}}{\lambda}.
    \end{aligned}
\end{align*}
So the first part has been proved. Furthermore, note that
\begin{align*}
    &\Big| h_{\mu_1, \pi}^\ast(x) - h_{\mu_0, \pi}^\ast(y) \Big| \leq \Big| h_{\mu_1, \pi}^\ast(x) - h_{\mu_0, \pi}^\ast(x) \Big| + \Big| h_{\mu_0, \pi}^\ast(x) - h_{\mu_0, \pi}^\ast(y) \Big| \\ 
    &\leq \left\| k(x, \cdot)\right\|_\calH \left\| h_{\mu_1, \pi}^\ast - h_{\mu_0, \pi}^\ast \right\|_\calH + \left\|k(x,\cdot) - k(y,\cdot) \right\|_\calH \left\|h_{\mu_0, \pi}^\ast \right\|_\calH \\
    &\leq \frac{2\sqrt{K K_{1d} }}{\lambda} W_2(\mu_0,\mu_1) + \frac{2 \sqrt{K K_{1d} }}{\lambda} \left\| x - y \right\| \\
    &\leq \frac{2\sqrt{K K_{1d} }}{\lambda} \Big( W_2(\mu_0, \mu_1) + \left\| x - y \right\| \Big) 
    \leq L \Big( W_2(\mu_0, \mu_1) + \left\| x - y \right\| \Big)
\end{align*}
and
\begin{align*}
    &\quad \Big\| \nabla h_{\mu_1, \pi}^\ast(x) - \nabla h_{\mu_0, \pi}^\ast(y) \Big\| \\
    &\leq \Big\| \nabla h_{\mu_1, \pi}^\ast(x) - \nabla h_{\mu_0, \pi}^\ast(x) \Big\| + \Big\| \nabla h_{\mu_0, \pi}^\ast(x) - \nabla h_{\mu_0, \pi}^\ast(y) \Big\| \\ 
    &\leq \left\| \nabla_1 k(x, \cdot)\right\|_{\calH^d} \left\| h_{\mu_1, \pi}^\ast - h_{\mu_0, \pi}^\ast \right\|_\calH + \left\|\nabla_1 k(x,\cdot) - \nabla_1 k(y,\cdot) \right\|_{ \calH^d } \left\|h_{\mu_0, \pi}^\ast \right\|_\calH \\
    &\leq \frac{2 K_{1d} }{\lambda} W_2(\mu_0,\mu_1) + \frac{2 \sqrt{K K_{2d}} }{\lambda} \left\| x - y \right\| 
    \leq L \Big( W_2(\mu_0, \mu_1) + \left\| x - y \right\| \Big)
\end{align*}
and the result follows.
\end{proof}
\begin{lem}\label{lem:h_difference}
Given two probability measures $\mu \ll \pi$ that are both absolutely continuous with respect to Lebesgue measure. If $\frac{ d\mu }{d \pi} - 1 \in \operatorname{Ran} ( \mathcal{T}_\pi^{r} )$ with $r > 0$, i.e., there exists $q \in L^2(\pi)$ such that $\frac{d\mu}{d\pi} - 1 = \calT_\pi^{r} q$, then
\begin{align*}
    \left\| h  - 2 \left( \frac{d \mu}{d \pi} -1 \right) \right\|_{L^2(\pi)} \leq 2 \lambda^{r} \left\|q \right\|_{L^2(\pi)},
\end{align*}
where $h = 2(\Sigma_\pi + \lambda \Id )^{-1}(m_\pi - m_{\mu})$ is defined in \Cref{prop:drmmd_representation_no_ratio}.
\end{lem}
\begin{proof}
Given the assumption that $\frac{ d\mu }{d \pi} - 1 \in \operatorname{Ran} \left( \mathcal{T}_\pi ^{r} \right)$ with $r > 0$, there exists $q \in L^2(\pi)$ such that $\frac{d\mu}{d\pi} - 1 = \calT_\pi^{r} q$ and $\langle \frac{ d\mu }{d \pi} - 1, e_i \rangle_{L^2(\pi)} = \varrho_i^{r} \PSi{ q, e_i}{L^2(\pi)}$. 
Since $\mu \ll \pi$, the Mercer decomposition of $k$ holds for any $x$ in the support of $\mu$ and in the support of $\pi$,
\begin{align}\label{appeq:m_mu_m_pi}
    m_{\mu} - m_\pi &= \int k(x, \cdot) d (\mu - \pi)(x) = \sum_{i \geq 1} \varrho_i \left( \int e_i(x) d( \mu - \pi)(x) \right) e_i \nonumber \\
    &= \varrho_i  \PSi{\frac{d \mu}{d \pi}- 1, e_i}{L^2(\pi)} e_i.
\end{align}
So we have, 
\begin{align*}
    &\left\| h - 2 \left( \frac{d \mu}{d \pi} -1 \right) \right\|_{L^2(\pi)} = 2 \left\| \left(\Sigma_{\pi}+\lambda \Id\right)^{-1}\left(m_{\mu} - m_{\pi}\right) - \left( \frac{d \mu}{d \pi} -1 \right) \right\|_{L^2(\pi)} \nonumber \\
    &= 2 \left\| \sum_{i \geq 1} \frac{\varrho_i}{\varrho_i + \lambda} \PSi{\frac{d \mu}{d \pi}- 1, e_i}{L^2(\pi)} e_i - \sum_{i \geq 1} \PSi{\frac{d \mu}{d \pi}- 1, e_i}{L^2(\pi)} e_i \right\|_{L^2(\pi)} \nonumber \\
    &= 2 \left\| \sum_{i \geq 1} \frac{\lambda}{\varrho_i + \lambda} \PSi{\frac{d \mu}{d \pi}- 1, e_i}{L^2(\pi)} e_i \right\|_{L^2(\pi)} 
    = 2 \left\| \sum_{i \geq 1} \frac{\lambda \varrho_i^{r}}{\varrho_i + \lambda} \PSi{q, e_i}{L^2(\pi)} e_i \right\|_{L^2(\pi)} \nonumber \\
    &\leq 2 \lambda^{r} \left\| q \right\|_{L^2(\pi)}, 
\end{align*}
where the last inequality is obtained by using 
\begin{align*}
    \frac{\lambda \varrho_i^{r}}{\varrho_i + \lambda} = \left( \frac{\varrho_i}{\varrho_i + \lambda} \right)^{r} \left( \frac{\lambda}{\varrho_i + \lambda} \right)^{1 - r} \lambda^{r} \leq \lambda^{r}.
\end{align*}
Hence the proof.
\end{proof}

\begin{lem}\label{lem:wass_gradient_hess_chard}
Let $\rho \in \calP_2\left(\mathbb{R}^d\right)$ and $\phi \in C_c^\infty\left(\mathbb{R}^d\right)$. Consider the path $(\rho_s)_{0 \leq s \leq 1}$ from $\rho$ to $(\Id + \nabla \phi)_{\#} \rho$ given by $\rho_s = (\Id + s \nabla \phi)_{\#} \rho$. 
Define $\varphi_s: \R^d \to \R^d, x \mapsto x + s \nabla \phi(x)$. Let 
$\tilde{k}$ be the regularized kernel defined in \eqref{eq:tilde_k} along with its associated RKHS $\tilde{\calH}$. 
The mapping $s \mapsto \dmmd(\rho_s \| \pi)$ is continuous and differentiable, and its first-order time derivative is given by
\begin{align}\label{appeq:chard_wass_gradient}
     &\quad \frac{d}{d s} \dmmd(\rho_s||\pi) \nonumber \\
     &= 2(1+\lambda) \int \nabla \phi(x)^\top
     \left( \int \nabla_1 \tilde{k}(\varphi_s(x), \varphi_s(z)) d \rho(z) \right. \nonumber \\
     &\hspace{100pt}\left. {}
     - \int \nabla_1 \tilde{k}(\varphi_s(x), z) d \pi(z) \right) d \rho(x).
\end{align}
Moreover, 
\begin{align}\label{appeq:chard_wass_gradient_0}
     &\frac{d}{d s} \Big|_{s=0} \dmmd(\rho_s||\pi) \nonumber\\
     &= 2(1+\lambda) \int \nabla \phi(x)^\top \left( \int  \nabla_1 \tilde{k}(x, z) d \rho(z) - \int \nabla_1 \tilde{k}(x, z) d \pi(z) \right) d \rho(x) .
\end{align}
Additionally, the mapping $s \to \frac{d}{ds} \dmmd(\rho_s \| \pi)$ is continuous and differentiable, and the second-order time derivative of $\dmmd(\rho_s \| \pi)$ is given by
\begin{align}\label{appeq:chard_wass_hessian}
    &\quad \frac{d^2}{d s^2} \dmmd(\rho_s||\pi) \\
    &= 2 (1+\lambda) \iint \nabla \phi(x)^\top
    \nabla_1 \nabla_2 \tilde{k}(\varphi_s(x), \varphi_s(z))
    \nabla \phi(z) d\rho(x) d\rho(z) \nonumber \\
    &\quad + 2 (1+\lambda) \int \nabla \phi(x)^\top
    \left( \int\bH_1 \tilde{k} \left(\varphi_s(x), \varphi_s(z)\right) d \rho(z) \right. \nonumber \\
    &\hspace{105pt}\left. {}
    - \int \bH_1 \tilde{k} \left(\varphi_s(x), z\right) d \pi(z) \right)
    \nabla \phi(x) d\rho(x),
\end{align}
with
\begin{align}\label{appeq:chard_wass_hessian_0}
    &\quad \frac{d^2}{d s^2}\Big|_{s=0} \dmmd(\rho_s||\pi) \nonumber \\
    &= 2 (1+\lambda) \iint \nabla \phi(x)^\top
    \nabla_1 \nabla_2 \tilde{k}(x, z) \nabla \phi(z) d\rho(x) d\rho(z) \nonumber \\
    &\quad + 2 (1+\lambda) \int \nabla \phi(x)^\top
    \left( \int\bH_1 \tilde{k} \left(x, z\right) d \rho(z) \right. \nonumber \\
    &\hspace{105pt}\left. {}
    - \int \bH_1 \tilde{k} \left(x, z\right) d \pi(z) \right)
    \nabla \phi(x) d\rho(x) .
\end{align}

\end{lem}
\begin{proof}
Recall that $\dmmd$ is, up to a multiplicative factor of $(1+\lambda) $, $\mmd^2$ of the regularized kernel $\tilde{k}$ defined in \eqref{eq:tilde_k} and the associated RKHS $\tilde{\calH}$.
From \Cref{lem:k_tilde}, we know that assumptions (A) and (B) of \cite{arbel2019maximum} are satisfied for the regularized kernel $\tilde{k}$, so using Lemma 22 and Lemma 23 from \cite{arbel2019maximum}, \eqref{appeq:chard_wass_gradient} and \eqref{appeq:chard_wass_hessian} are proved.
Then \eqref{appeq:chard_wass_gradient_0} and \eqref{appeq:chard_wass_hessian_0} are subsequently proved by taking $s=0$.
\end{proof}

\begin{lem}\label{lem:hessian_terms}
For $\mu_0 \ll \pi$, define $h = 2(\Sigma_\pi + \lambda \Id )^{-1}(m_\pi - m_{\mu_0})$ and $\varphi_t: \R^d \to \R^d, x \mapsto x - t (1+\lambda) \nabla h(x)$. Suppose $\pi \propto \exp(-V)$, $\bH V \preceq \beta \Id$, and the step size $\gamma$ satisfies 
\begin{align*}
    2 (1+\lambda) \gamma \sqrt{ \chi^2(\mu_0 \| \pi) \frac{K_{2d}}{\lambda} } \leq \frac{\zeta - 1}{\zeta}
\end{align*}
for some constant $1 < \zeta < 2$. Then, for  $t \in [0, \gamma]$, the following inequalities hold,
\begin{align}
    \int \nabla h( x )^\top \bH V( \varphi_t(x) ) \nabla h(x) d \mu_0(x) & \leq 4 \beta \chi^2(\mu_0 \| \pi) \frac{K_{1d}}{\lambda}; \label{eq:hessian_terms_1} \\
    \int \left\| \bH h(x) \left(\Id -t (1 + \lambda) \bH h(x) \right)^{-1} \right\|_F^2 d\mu_0(x) &\leq 4 \zeta^2 \chi^2(\mu_0 \| \pi) \frac{K_{2d}}{\lambda} . \label{eq:hessian_terms_2} 
\end{align}
\end{lem}
\begin{proof}
First, recall \eqref{appeq:m_mu_m_pi},
\begin{align}
    \left\| \nabla h (x) \right\|^2 &= 4 \left\| \nabla (\Sigma_\pi + \lambda \Id)^{-1} (m_\pi - m_{\mu_0})(x) \right\|^2 \nonumber \\
    &= 4 \left\| \nabla \left( \sum_{i \geq 1} \frac{\varrho_i}{\varrho_i + \lambda} \PSi{\frac{ d \mu_0}{ d \pi} - 1, e_i}{L^2(\pi)} e_i(x) \right) \right\|^2. \label{appeq:nabla_h_1}
\end{align}
For any $j \in \{1, \cdots, d\}$, consider 
\begin{align*}
    g_{M_0}(x) &:= \sum_{i \geq M_0} \left| \varrho_i^{1/2} \left\langle \frac{d \mu_0}{ d \pi} - 1, e_i \right\rangle_{L^2(\pi)} \partial_j e_i(x)\right| \\
    &\leq \left( \sum_{i \geq M_0} \varrho_i \partial_j e_i(x)^2 \right)^{\frac{1}{2}} \left( \sum_{i \geq M_0} \PSi{ \frac{d \mu_0}{ d \pi} - 1 , e_i}{L^2(\pi)} ^2 \right)^{\frac{1}{2}} \\
    &\leq \left( \sum_{i \geq M_0} \varrho_i \partial_j e_i(x)^2 \right)^{\frac{1}{2}} \left\| \frac{d \mu_0}{ d \pi} - 1 \right\|_{L^2(\pi)}.
\end{align*}
Since
\begin{align}\label{appeq:sum_kernel_gradient}
    \sum_{i \geq 1} \sum_{j=1}^d \varrho_i \left( \partial_j e_i(x) \right)^2
    &= \sum_{i \geq 1} \sum_{j=1}^d \varrho_i
    \PSi{\partial_j k(x, \cdot), e_i}{\calH}^2 \nonumber \\
    &= \sum_{i \geq 1} \sum_{j=1}^d
    \PSi{\partial_j k(x, \cdot), \sqrt{\varrho_i} e_i}{\calH}^2 \nonumber \\
    &= \sum_{j=1}^d \left\| \partial_j k(x, \cdot) \right\|_{\calH}^2
    = \left\| \nabla_1 k(x,\cdot) \right\|_\calH^2 \leq K_{1d} ,
\end{align}
so $ \sum_{i \geq M_0} \varrho_i (\partial_j e_i(x))^2$ converges uniformly to $0$, and hence $g_{M_0}(x)$ also converges uniformly to $0$. Therefore, we are allowed to interchange the derivative and the infinite sum \citep{rudin1976principles} in \eqref{appeq:nabla_h_1} to achieve,
\begin{align}
    \left\| \nabla h (x) \right\|^2 &= 4 \left\| \sum_{i \geq 1} \frac{\varrho_i}{\varrho_i + \lambda} \PSi{\frac{d \mu_0}{ d \pi} - 1, e_i}{L^2(\pi)} \nabla e_i(x) \right\|^2 \nonumber \\
    &\leq 4 \left( \sum_{i \geq 1} \frac{\varrho_i}{(\varrho_i + \lambda)^2}
    \PSi{\frac{d \mu_0}{d \pi} - 1, e_i}{L^2(\pi)}^2 \right) \nonumber \\
    &\quad {}\times \left( \sum_{i\geq 1} \varrho_i
    \left\| \nabla e_i(x) \right\|^2 \right) \nonumber \\
    &\leq \frac{4}{\lambda} \left( \sum_{i \geq 1}
    \PSi{\frac{d \mu_0}{d \pi} - 1, e_i}{L^2(\pi)}^2 \right) \nonumber \\
    &\quad {}\times \left( \sum_{i\geq 1} \varrho_i
    \left\| \nabla e_i(x) \right\|^2 \right)
    \leq 4 \chi^2(\mu_0 \| \pi) \frac{K_{1d}}{\lambda} . \label{eq:nabla_h}
\end{align}
The first inequality follows from Cauchy Schwartz, the penultimate inequality follows by noting that $\frac{\varrho_i}{(\varrho_i + \lambda)^2} \leq \frac{1}{\lambda}$, and the last inequality follows from \eqref{appeq:sum_kernel_gradient}.
Given $\bH V \preceq \beta \Id$, \eqref{eq:hessian_terms_1} is proved by the following,
\begin{align*}
    \int \nabla h(x)^\top \bH V( \varphi_t(x) ) \nabla h(x) d \mu_0(x)
    &\leq \beta \int \left\| \nabla h(x) \right\|^2 d\mu_0(x) \\
    &\leq 4 \beta \chi^2(\mu_0 \| \pi) \frac{K_{1d}}{\lambda} .
\end{align*}
We now turn to proving the second statement. Similarly to \eqref{eq:nabla_h}, we have
\begin{align}\label{appeq:hess_h}
    \left\| \bH h(x) \right\|_F^2 \leq 4 \chi^2(\mu_0 \| \pi) \frac{K_{2d}}{\lambda} .
\end{align}
Using $2 (1+\lambda) \gamma \sqrt{ \chi^2(\mu_0 \| \pi) \frac{K_{2d}}{\lambda} } \leq \frac{\zeta - 1}{\zeta}$ for some constant $1 < \zeta < 2$, the inverse of $ \Id - t (1+\lambda) \bH h(x)$ can be represented by the Neumann series, and hence
\begingroup\small
\begin{align}\label{appeq:hessian_hs}
    \left\|(\Id - t (1+\lambda) \bH h(x))^{-1} \right\|_F &\leq \sum_{m \geq 0} \|t (1+\lambda) \bH h(x)\|_F^m \\
    &\leq \sum_{m \geq 0} \left( \gamma(1+\lambda) 2 \sqrt{ \chi^2(\mu_0 \| \pi) \frac{K_{2d}}{\lambda} } \right)^m \nonumber \\
    &\leq \sum_{m \geq 0} \left( \frac{\zeta - 1}{\zeta} \right)^m = \zeta.
\end{align}
\endgroup
Therefore, \eqref{eq:hessian_terms_2} is proved by combining \eqref{appeq:hess_h} and \eqref{appeq:hessian_hs},
\begin{align*}
    &\int \left\| \bH h(x) \left(\Id -t (1+\lambda) \bH h(x) \right)^{-1} \right\|_F^2 d\mu_0(x) \\
    &\leq \int \left\| \bH h(x) \right\|_F^2 \left\|(\Id - t (1+\lambda) \bH h(x))^{-1} \right\|_F^2 d\mu_0(x) 
    \leq 4 \zeta^2 \chi^2(\mu_0 \| \pi) \frac{K_{2d}}{\lambda} 
\end{align*}
and the result follows.
\end{proof}
\vspace{-15pt}

\begin{lem}\label{lem:hoeffding}
Let $\calH$ be a separable Hilbert space, $\xi_1, \ldots, \xi_n: \Omega \rightarrow \calH$ are $n$ identical independent $\calH$-valued random variables satisfying $\left\| \xi_i \right\|_\calH \leq B$. Then
\begin{align*}
    \E \left\|\frac{1}{n} \sum_{i=1}^n \xi_i - \E[\xi_1] \right\|_\calH \leq \frac{\sqrt{2 \pi} B}{2 \sqrt{n}} ,\quad\text{and} \quad \quad \E \left\|\frac{1}{n} \sum_{i=1}^n \xi_i - \E[\xi_1] \right\|_\calH^2 \leq \frac{B^2}{n} .
\end{align*}
\end{lem}
\begin{proof}
We know from Corollary 6.15 of \cite{steinwart2008support} that
\begin{align*}
    \Pb \left( \left\|\frac{1}{n} \sum_{i=1}^n \xi_i - \mathbb{E} \left[\xi_1 \right] \right\|_\calH \geq t\right) \leq 2 \exp \left(-2 n t^2 /B^2 \right) .
\end{align*}
Then, denote $R := \| \frac{1}{n} \sum_{i=1}^n \xi_i - \E[\xi_1] \|_\calH$, we know that
\begin{align*}
    \E[R] &= \int_0^\infty \Pb(R \geq t) dt  \leq 2 \int_0^\infty \exp(- 2 n t^2 / B^2 ) dt = \frac{\sqrt{2 \pi} B}{2 \sqrt{n}}.
\end{align*}
The other part is proved similarly.
\end{proof}
\vspace{-15pt}

\definecolor{c_1}{rgb}{1.0, 0.7, 0.8}
\definecolor{c_2}{rgb}{0.17254901960784313, 0.6274509803921569, 0.17254901960784313}
\definecolor{c_3}{rgb}{0.5803921568627451, 0.403921568627451, 0.7411764705882353}
\definecolor{c_4}{rgb}{0.5490196078431373, 0.33725490196078434, 0.29411764705882354}
\definecolor{c_5}{rgb}{0.0, 0.6, 0.8}
\definecolor{c_6}{rgb}{1.0, 0.6, 0.2}
\definecolor{c_7}{rgb}{1.0, 0.0, 0.0}
\definecolor{c_8}{rgb}{0.4, 0.0, 0.6}

\begin{lem}[Wasserstein Hessian of $\calF_{\chi^2}$]\label{lem:wass_chi2}
Let $\rho \in \mathcal{P}_2(\R^d) $ and $\phi \in C_c^\infty (\R^d) $. Consider the path $(\rho_s)_{0\leq s \leq 1}$ from $\rho$ to $(\Id + \nabla \phi)_{\#} \rho$ given by $\rho_s = (\Id + s \nabla \phi)_{\#} \rho$. 
Define $\varphi_s: \R^d \to \R^d, x \mapsto x + s \nabla \phi(x)$ and $\omega_s: \R^d \to \R^d, x \mapsto \left[\nabla \phi \circ \varphi_s^{-1}\right](x)$.
For $\pi(x) \propto \exp(-V(x))$, the second order derivative of $s \to \chi^2(\rho_s \| \pi)$ is given by
\begin{align}
    & \quad \frac{d^2}{d s^2} \chi^2(\rho_s \| \pi) = \int \frac{\rho_s(x)}{\pi(x)} \left( \nabla V(x)^\top \omega_s(x) - \nabla \cdot \omega_s(x) \right)^2 d\rho_s(x)\nonumber\\
    &\quad+  \int \frac{\rho_s(x)}{\pi(x)} \omega_s(x)^\top \bH V(x) \omega_s(x) d\rho_s(x) +  \int \frac{\rho_s(x)}{\pi(x)} \left\| \nabla \omega_s(x) \right\|_F^2 d \rho_s(x) . \label{appeq:wess_chi2_1}
\end{align}
Equivalently, the second order derivative of $s \to \chi^2(\rho_s \| \pi)$ can also be written as
\begin{align}\label{appeq:wess_chi2_2}
    \frac{d^2}{d s^2} \chi^2(\rho_s \| \pi) &= 2 \int\left( \nabla \cdot \left( \omega_s(x) \rho_s(x) \right) \frac{1}{\pi(x)} \right)^2 \pi(x) d x \nonumber\\
    &\qquad\qquad+ 2 \int \omega_s(x)^\top \bH \left(\frac{\rho_s(x) }{\pi(x)}\right) \omega_s(x) \rho_s(x) dx.
\end{align}
\end{lem}
\begin{proof}
\eqref{appeq:wess_chi2_1} is provided in the Example 15.9 of \cite{villani2009optimal} by taking $m=2$. Now, we are going to prove \eqref{appeq:wess_chi2_2}. For ease of notation in the following derivations, we are going to drop the function input $x$ in $\rho_s, \pi$, and $\omega_s$.
We introduce colors to picture grouping of terms that will carry over during chains of calculation.

In order to prove \eqref{appeq:wess_chi2_2}, we need to expand the terms in \eqref{appeq:wess_chi2_1} accordingly. We denote the three terms in the RHS of \eqref{appeq:wess_chi2_1} as $(A)$, $\textcolor{c_1}{(B)}$ and $\textcolor{c_2}{(C)}$. 
Consider 
 \begin{align*}
    (A) &= \int \frac{\rho_s^2}{\pi} \left( \nabla V^\top \omega_s  - \nabla \cdot \omega_s \right)^2 dx \\
    &= \int \frac{\rho_s^2}{\pi} \left( -\frac{1}{\pi} \nabla \pi^\top \omega_s  - \nabla \cdot \omega_s \right)^2 dx \\ 
    &= \textcolor{c_3}{\underbrace{\int \frac{\rho_s^2}{\pi^3}
    (\nabla \pi^\top \omega_s)^2 dx}_{(A_1)}} \\
    &\quad + \textcolor{c_4}{\underbrace{2 \int \frac{\rho_s^2}{\pi^2}
    (\nabla \pi^\top \omega_s) \nabla \cdot \omega_s dx}_{(A_2)}} \\
    &\quad + \textcolor{c_5}{\underbrace{\int \frac{\rho_s^2}{\pi}
    (\nabla \cdot \omega_s)^2 dx}_{(A_3)}}.
\end{align*}
Then we are going to use integration by parts for $\textcolor{c_4}{ (A_2)}$ and $\textcolor{c_5}{(A_3)}$. 
\begin{align*}
    \textcolor{c_4}{ (A_2) }
    &= 2 \int \frac{\rho_s^2}{\pi^2}
    (\nabla \pi^\top \omega_s) \nabla \cdot \omega_s dx \\
    &= -2 \int \omega_s^\top \nabla
    \left( \frac{\rho_s^2}{\pi^2} (\nabla \pi^\top \omega_s) \right) dx \\
    &= -4 \int (\omega_s^\top \nabla \rho_s) (\nabla \pi^\top \omega_s) \frac{\rho_s}{\pi^2} dx + 4 \int (\omega_s^\top \nabla \pi)^2 \frac{\rho_s^2}{\pi^3} dx \\
    &- 2 \int (\omega_s^\top \bH \pi \omega_s) \frac{\rho_s^2}{\pi^2} dx - 2 \int ( \omega_s^\top \nabla \omega_s \nabla \pi) \frac{\rho_s^2}{\pi^2} dx, \\
    \textcolor{c_5}{ (A_3) }
    &= \int \frac{\rho_s^2}{\pi} (\nabla \cdot \omega_s)^2 dx \\
    &= - \int \omega_s^\top \nabla
    \left( \nabla \cdot \omega_s \frac{\rho_s^2}{\pi} \right) dx \\
    &= \textcolor{c_6}{\underbrace{- \int \omega_s^\top \nabla (\nabla \cdot \omega_s)
    \frac{\rho_s^2}{\pi} dx}_{(A_{31})}} \\
    &\quad + \textcolor{c_7}{\underbrace{- \int \nabla \cdot \omega_s
    ( \omega_s^\top \nabla \rho_s) \frac{2 \rho_s}{\pi} dx}_{(A_{32})}} \\
    &\quad + \textcolor{c_8}{\underbrace{\int \nabla \cdot \omega_s
    (\omega_s^\top \nabla \pi) \frac{\rho_s^2}{\pi^2} dx}_{(A_{33})}} .
\end{align*}
Furthermore, we use integration by parts for $\textcolor{c_7}{ (A_{32}) }$ and $\textcolor{c_8}{ (A_{33}) }$, we have 
\begin{align*}
    \textcolor{c_7}{ (A_{32}) }
    &= -2 \int \nabla \cdot \omega_s
    ( \omega_s^\top \nabla \rho_s) \frac{\rho_s}{\pi} dx \\
    &= 2 \int \omega_s^\top \nabla
    \left( ( \omega_s^\top \nabla \rho_s) \frac{\rho_s}{\pi} \right) dx \\
    &= 2 \int (\omega_s^\top \nabla \omega_s \nabla \rho_s)
    \frac{\rho_s}{\pi} dx
    + 2 \int (\omega_s^\top \bH \rho_s \omega_s) \frac{\rho_s}{\pi} dx \\
    &\quad + 2 \int ( \omega_s^\top \nabla \rho_s)^2 \frac{1}{\pi} dx \\
    &- 2 \int ( \omega_s^\top \nabla \rho_s) (\omega_s^\top \nabla \pi) \frac{\rho_s}{\pi^2} dx, \\
    \textcolor{c_8}{ (A_{33}) }
    &= \int \nabla \cdot \omega_s (\omega_s^\top \nabla \pi)
    \frac{\rho_s^2}{\pi^2} dx \\
    &= -\int \omega_s^\top \nabla
    \left( (\omega_s^\top \nabla \pi) \frac{\rho_s^2}{\pi^2} \right) dx \\
    &= -\int ( \omega_s^\top \nabla \omega_s \nabla \pi)
    \frac{\rho_s^2}{\pi^2} dx
    - \int (\omega_s^\top \bH \pi \omega_s) \frac{\rho_s^2}{\pi^2} dx \\
    &\quad -2 \int ( \omega_s^\top \nabla \rho_s)
    (\omega_s^\top \nabla \pi) \frac{\rho_s}{\pi^2} dx \\
    &\qquad\qquad+2 \int (\omega_s^\top \nabla \pi)^2 \frac{\rho_s^2}{\pi^3} dx  .
\end{align*}
Having completed $(A)$, now we turn to $\textcolor{c_1}{ (B) }$.
\begin{align*}
    \textcolor{c_1}{ (B) }
    &= \int \frac{\rho_s^2}{\pi} \omega_s^\top \bH V \omega_s dx \\
    &= - \int \omega_s^\top \bH \pi \omega_s \frac{\rho_s^2}{\pi^2} dx
    + \int (\nabla \pi^\top \omega_s)^2 \frac{\rho_s^2}{\pi^3} dx .
\end{align*}
So, combining $(A)$, $\textcolor{c_1}{ (B) }$ and $\textcolor{c_2}{ (C) }$, we have 
\begin{align*}
    &\quad \frac{d^2}{ds^2} \chi^2(\rho_s\|\pi) = (A) + \textcolor{c_1}{ (B) } + \textcolor{c_2}{ (C) } \\
    &= \textcolor{c_3}{ (A_1)} + \textcolor{c_4}{ (A_2) }
    + \textcolor{c_6}{ (A_{31})} + \textcolor{c_7}{ (A_{32}) } \\
    &\quad + \textcolor{c_8}{ (A_{33}) } + \textcolor{c_1}{ (B) }
    + \textcolor{c_2}{ (C) } \\
    & = \textcolor{c_3}{ \int \frac{\rho_s^2}{\pi^3} (\nabla \pi^\top \omega_s)^2 dx } \\
    &\quad \textcolor{c_4}{-4 \int ( \omega_s^\top \nabla \rho_s)
    (\nabla \pi^\top \omega_s) \frac{\rho_s}{\pi^2} dx} \\
    &\quad \textcolor{c_4}{+ 4 \int (\omega_s^\top \nabla \pi)^2
    \frac{\rho_s^2}{\pi^3} dx
    - 2 \int (\omega_s^\top \bH \pi \omega_s) \frac{\rho_s^2}{\pi^2} dx} \\
    &\quad \textcolor{c_4}{- 2 \int ( \omega_s^\top \nabla \omega_s \nabla \pi)
    \frac{\rho_s^2}{\pi^2} dx}
    + \textcolor{c_6}{ (A_{31})} + \textcolor{c_7}{ (A_{32}) } + \textcolor{c_8}{ (A_{33}) } \\
    &\quad \textcolor{c_1}{- \int ( \omega_s^\top \bH \pi \omega_s)
    \frac{\rho_s^2}{\pi^2} dx
    + \int (\nabla \pi^\top \omega_s)^2 \frac{\rho_s^2}{\pi^3} dx} \\
    &\quad + \textcolor{c_2}{\int \frac{\rho_s^2}{\pi}
    \left\| \nabla \omega_s \right\|_F^2 dx} .
\end{align*}
Since $\omega_s = \nabla \phi \circ \varphi_s^{-1}$ is a mapping from $\R^d$ to $\R^d$, denote $\omega_{s, i} := \left[\nabla \phi \circ \varphi_s^{-1}\right]_i$ which is a mapping from $\R^d$ to $\R$.
Notice that $\omega_{s, i} \partial_{j} \omega_{s, i} $ vanishes at boundary because $\phi \in C_c^\infty$. Hence,
\begin{align*}
    0 &= \sum_{i,j} \int \partial_j \left( \omega_{s, i} \partial_{j} \omega_{s, i} \frac{\rho_s^2}{\pi} \right) dx \\
    &= \sum_{i,j} \int  \partial_{j} \omega_{s, i}  \partial_{j} \omega_{s, i} \frac{\rho_s^2}{\pi} dx + 
    \sum_{i,j} \int \omega_{s, i} \partial_{jj} \omega_{s, i} \frac{\rho_s^2}{\pi} dx \nonumber\\
    &\qquad\qquad+\sum_{i,j} \int \omega_{s, i} \partial_{j} \omega_{s, i} \left( \frac{2 \rho_s \partial_j \rho_s}{\pi} - \frac{\rho_s^2 \partial_j \pi}{\pi^2} \right) dx \\ 
    &= \textcolor{c_2}{\int \frac{\rho_s^2}{\pi}
    \left\| \nabla \omega_s \right\|_F^2 dx}
    + \int \frac{\rho_s^2}{\pi}
    ( \omega_s^\top \nabla(\nabla \cdot \omega_s)) dx \\
    &\quad + \int (\omega_s^\top \nabla \omega_s \nabla \rho_s)
    \frac{2 \rho_s}{\pi} dx \\
    &\qquad\qquad- \int ( \omega_s^\top \nabla \omega_s \nabla \pi) \frac{\rho_s^2}{\pi^2} dx .
\end{align*}
Therefore, replace
\begin{align*}
    \textcolor{c_2}{\int \frac{\rho_s^2}{\pi}
    \left\| \nabla \omega_s \right\|_F^2 dx},
\end{align*}
and notice that
\begin{align*}
    \textcolor{c_6}{-\int \frac{\rho_s^2}{\pi}
    ( \omega_s^\top \nabla(\nabla \cdot \omega_s)) dx}
    &= \textcolor{c_6}{(A_{31})}.
\end{align*}
We then have the following:
\begingroup\small
\begin{align*}
    &\quad \frac{d^2}{ds^2} \chi^2(\rho_s\|\pi) \\
    &= \textcolor{c_3}{ \int \frac{\rho_s^2}{\pi^3} (\nabla \pi^\top \omega_s)^2 dx}
    - \textcolor{c_4}{ 4 \int ( \omega_s^\top \nabla \rho_s) (\nabla \pi^\top \omega_s) \frac{\rho_s}{\pi^2} dx + 4 \int (\omega_s^\top \nabla \pi)^2 \frac{\rho_s^2}{\pi^3} dx } \\
    & \qquad\textcolor{c_4}{ - 2 \int (\omega_s^\top \bH \pi \omega_s) \frac{\rho_s^2}{\pi^2} dx - 2 \int ( \omega_s^\top \nabla \omega_s \nabla \pi) \frac{\rho_s^2}{\pi^2} dx} + \textcolor{c_6}{ (A_{31})} + \textcolor{c_7}{ (A_{32}) } + \textcolor{c_8}{ (A_{33}) } \\
    &\qquad \textcolor{c_1}{- \int ( \omega_s^\top \bH \pi  \omega_s) \frac{\rho_s^2}{\pi^2} dx + \int (\nabla \pi^\top \omega_s)^2 \frac{\rho_s^2}{\pi^3} dx } \\
    & \qquad\qquad-\int \frac{\rho_s^2}{\pi} ( \omega_s^\top  \nabla(\nabla \cdot \omega_s)) dx - \int (\omega_s^\top \nabla \omega_s \nabla \rho_s) \frac{2 \rho_s}{\pi} dx + \int ( \omega_s^\top \nabla \omega_s \nabla \pi) \frac{\rho_s^2}{\pi^2} dx .
\end{align*}
\endgroup
Next, we combine the terms and obtain
\begingroup\small
\begin{align*}
    &= 2\left( \textcolor{c_6}{ (A_{31})} + \textcolor{c_7}{ (A_{32}) }
    + \textcolor{c_8}{ (A_{33}) } \right)
    + 6 \int \frac{\rho_s^2}{\pi^3} (\nabla \pi^\top \omega_s)^2 dx \\
    &\quad - 4 \int ( \omega_s^\top \nabla \rho_s)
    (\nabla \pi^\top \omega_s) \frac{\rho_s}{\pi^2} dx
    - 3 \int (\omega_s^\top \bH \pi \omega_s) \frac{\rho_s^2}{\pi^2} dx \\
    &\quad - \int ( \omega_s^\top \nabla \omega_s \nabla \pi)
    \frac{\rho_s^2}{\pi^2} dx
    - 2 \int (\omega_s^\top \nabla \omega_s \nabla \rho_s) \frac{\rho_s}{\pi} dx \\
    &\quad - \textcolor{c_7}{ ( A_{32}) } - \textcolor{c_8}{ (A_{33}) }.
\end{align*}
\endgroup
Recall that $\textcolor{c_6}{ (A_{31})} + \textcolor{c_7}{ (A_{32}) } + \textcolor{c_8}{ (A_{33}) } = \textcolor{c_5}{ (A_3)}$, and replacing $\textcolor{c_7}{ (A_{32}) } $ and $\textcolor{c_8}{ (A_{33}) }$ with their expressions, we have
\begingroup\tiny
\begin{align*}
    &=2 \textcolor{c_5}{ \int \frac{\rho_s^2}{\pi} (\nabla \cdot \omega_s)^2 dx } + 6 \int \frac{\rho_s^2}{\pi^3} (\nabla \pi^\top \omega_s)^2 dx - 4 \int ( \omega_s^\top \nabla \rho_s) (\nabla \pi^\top \omega_s) \frac{\rho_s}{\pi^2} dx \\
    &-3 \int (\omega_s^\top \bH \pi \omega_s) \frac{\rho_s^2}{\pi^2} dx 
    - \int ( \omega_s^\top \nabla \omega_s \nabla \pi) \frac{\rho_s^2}{\pi^2} dx - 2 \int (\omega_s^\top \nabla \omega_s \nabla \rho_s) \frac{\rho_s}{\pi} dx \\
    &- \textcolor{c_7}{ \left( 2 \int (\omega_s^\top \nabla \omega_s \nabla \rho_s) \frac{\rho_s}{\pi} dx + 2 \int (\omega_s^\top \bH \rho_s \omega_s ) \frac{\rho_s}{\pi} dx + 2 \int ( \omega_s^\top \nabla \rho_s)^2 \frac{1}{\pi} dx -  2 \int ( \omega_s^\top \nabla \rho_s) (\omega_s^\top \nabla \pi) \frac{\rho_s}{\pi^2} dx \right) } \\
    &- \textcolor{c_8}{ \left( -  \int ( \omega_s^\top \nabla \omega_s \nabla \pi) \frac{\rho_s^2}{\pi^2} dx -  \int (\omega_s^\top \bH \pi \omega_s) \frac{\rho_s^2}{\pi^2} dx  -  2 \int ( \omega_s^\top \nabla \rho_s) (\omega_s^\top \nabla \pi) \frac{\rho_s}{\pi^2} dx + 2 \int (\omega_s^\top \nabla \pi)^2 \frac{\rho_s^2}{\pi^3} dx  \right)} \\
    &= \underbrace{\begin{aligned}
    &2\int (\omega_s^\top \bH \rho_s \omega_s ) \frac{\rho_s}{\pi} dx
    - 2\int (\omega_s^\top \bH \pi \omega_s) \frac{\rho_s^2}{\pi^2} dx \\
    &\quad - 4 \int ( \omega_s^\top \nabla \rho_s) (\omega_s^\top \nabla \pi) \frac{\rho_s}{\pi^2} dx
    + 4 \int \frac{\rho_s^2}{\pi^3} ( \nabla \pi^\top \omega_s )^2 dx
    \end{aligned}}_{(M)} \\
    &\qquad+ \underbrace{\begin{aligned}
    &2\int \frac{\rho_s^2}{\pi} (\nabla \cdot \omega_s)^2 dx
    - 2 \int ( \omega_s^\top \nabla \rho_s)^2 \frac{1}{\pi} dx \\
    &\quad -4 \int (\omega_s^\top \nabla \omega_s \nabla \rho_s) \frac{\rho_s}{\pi} dx
    \end{aligned}}_{(N_1)} \\
    &\qquad\qquad \underbrace{
    -4 \int (\omega_s^\top \bH \rho_s \omega_s ) \frac{\rho_s}{\pi} dx
    + 4 \int ( \omega_s^\top \nabla \rho_s) (\omega_s^\top \nabla \pi) \frac{\rho_s}{\pi^2}  dx}_{(N_2)} .
\end{align*}
\endgroup
Now we analyze $(M)$ and $(N_1) + (N_2)$ separately. Notice that
\begin{align*}
    (M) = 2\int \left( \omega_s^\top \bH \left( \frac{\rho_s}{\pi} \right) \omega_s \right) \rho_s dx ,
\end{align*} 
and 
\begingroup\small
\begin{align*}
    &\quad (N_1) + (N_2)
    = 2\int \frac{\rho_s^2}{\pi} (\nabla \cdot \omega_s)^2 dx + 2 \int ( \omega_s^\top \nabla \rho_s)^2 \frac{1}{\pi} dx     - 4 \int (\omega_s^\top \nabla \omega_s \nabla \rho_s) \frac{\rho_s}{\pi} dx \\
    &\qquad\qquad- 4\int (\omega_s^\top \bH \rho_s \omega_s ) \frac{\rho_s}{\pi} dx- 4 \int ( \omega_s^\top \nabla \rho_s)^2 \frac{1}{\pi} dx + 4 \int ( \omega_s^\top \nabla \rho_s) (\omega_s^\top \nabla \pi) \frac{\rho_s}{\pi^2} dx \\
    &= 2\int \frac{\rho_s^2}{\pi} (\nabla \cdot \omega_s)^2 dx + 2 \int ( \omega_s^\top \nabla \rho_s)^2 \frac{1}{\pi} dx -4 \int ( \omega_s^\top  \nabla ( \omega_s^\top \nabla \rho_s) ) \frac{\rho_s}{\pi} dx\\
    &\qquad\qquad- 4 \int ( \omega_s^\top \nabla \rho_s) ( \omega_s^\top  \nabla \frac{\rho_s}{\pi}) dx \\
    &= 2\int \frac{\rho_s^2}{\pi} (\nabla \cdot \omega_s)^2 dx + 2 \int ( \omega_s^\top \nabla \rho_s)^2 \frac{1}{\pi} dx + 4\int \nabla \cdot \omega_s ( \omega_s^\top \nabla \rho_s)  \frac{\rho_s}{\pi} dx \\
    &= 2 \int\left(\nabla \cdot( \omega_s \rho_s ) \frac{1}{\pi} \right)^2 \pi d x .
\end{align*}
\endgroup
Since $\frac{d^2}{ds^2} \chi^2(\rho_s\|\pi) = (M) + (N_1) + (N_2)$, \Cref{lem:wass_chi2} is proved.
\end{proof}

\endgroup

\chapter{\texorpdfstring{Supplementary Material for \Cref{sec:pgd}}{Supplementary Material for Minimum MMD for Statistical Inference}}
\label{app:pgd}
\begingroup

\Cref{sec:proof_np_descent} provides proof for main convergence theorem of the nonparametric descent scheme; \Cref{sec:proof_p_descent} provides proof for main convergence theorem of the parametric descent scheme. 
\Cref{sec:ngd} provides a detailed discussion on the connection of our projected gradient descent (PGD) scheme with natural gradient descent. 
\Cref{sec:aux} provides all the auxiliary results for the proof. 
\section{\texorpdfstring{Proof of \Cref{thm:np_rate}}{Proof of the nonparametric convergence theorem}}\label{sec:proof_np_descent}


From \Cref{prop:descent_np} below, we have for any $t\in\N$,
\begingroup\small
\begin{align*}
    \mmd_{\ell}^2(\Pb_{t+1}, \Qb) -\mmd_{\ell}^2(\Pb_{t}, \Qb) &\leq -2\gamma_t \left( \ell_t \ell^{-1} \right)^d \|\nabla f_{\ell, \Pb_t, \Qb}\|_{L_2^d(\Pb_t)}^2 \\
    &\hspace{-30pt}+ 4 \gamma_t \left( \ell_t \ell^{-1} \right)^d \frac{C_{d,s} \sqrt{\ell_t^2 - \ell^2} }{\ell^3} + {\frac{C_k \gamma_t^2}{\ell^4} + C_k \gamma_t^2\frac{(\ell_t^2-\ell^2)^2}{\ell^8}}.
\end{align*}
\endgroup
Under the assumption that $\left( \ell_t \ell^{-1} \right)^d \|\nabla f_{\ell, \Pb_t,\Qb}\|_{L_2^d(\Pb_t)}^2 \geq C \mmd_{\ell}^2(\Pb_{t}, \Qb) $, we have
\begingroup\small
\begin{align*}
    \mmd_{\ell}^2 (\Pb_{t+1}, \Qb) &\leq \left( 1 -2 C \gamma_t \right) \mmd_{\ell}^2(\Pb_{t}, \Qb) \\
    &\hspace{-20pt} + 4\gamma_t \left( \ell_t \ell^{-1} \right)^d \frac{C_{d,s} \sqrt{\ell_t^2 - \ell^2} }{\ell^3} + {\frac{C_k \gamma_t^2}{\ell^4} + C_k \gamma_t^2\frac{(\ell_t^2-\ell^2)^2}{\ell^8}}.
\end{align*}
\endgroup
By the discrete Gronwall's lemma~\citep[Appendix~B.2, Gronwall's inequality]{evans2022partial}, iterating the above relation from $t=0, \ldots, T-1$, we obtain
\begingroup\small
\begin{align}
    \mmd_{\ell}^2(\Pb_T, \Qb) &\leq \exp \left(-2C \sum_{s=0}^{T-1} \gamma_s \right) \mmd_{\ell}^2(\Pb_0, \Qb) \label{eq:term_1} \\
    &\quad + \sum_{t=0}^{T-1}
    \exp \left(-2 C \sum_{s=t+1}^{T-1} \gamma_s \right) \nonumber\\
    &\qquad \times
    \left[
    4 \gamma_t \left( \ell_t \ell^{-1} \right)^d \frac{C_{d,s} \sqrt{\ell_t^2 - \ell^2} }{\ell^3}
    + {\frac{C_k \gamma_t^2}{\ell^4}
    + C_k \gamma_t^2\frac{(\ell_t^2-\ell^2)^2}{\ell^8}}
    \right]\label{eq:term_2} .
\end{align}
\endgroup
The first term in Eq.~\eqref{eq:term_1} converges to $0$ as $T\to\infty$ by the condition that $\lim_{T\to\infty} \sum_{s=0}^{T-1} \gamma_s \to \infty$.
For the second term, the analysis is more involved.
Let $\varepsilon > 0$ be arbitrary small.
Since $\ell_t^2 - \ell^2 \to 0$, $\gamma_t\to 0$ and $\left( \ell_t \ell^{-1} \right)^d \to 1$ as $t\to\infty$ by design, there exists $t_0\in\N$ such that $4 \left( \ell_t \ell^{-1} \right)^d \frac{C_{d,s} \sqrt{\ell_t^2 - \ell^2} }{\ell^3}\leq \frac{\epsilon}{3}$, $\frac{C_k \gamma_t}{\ell^4} \leq \frac{\epsilon}{3}$, $C_k \gamma_t\frac{(\ell_t^2-\ell^2)^2}{\ell^8}\leq\frac{\epsilon}{3}$ all hold for all $t>t_0$.
Hence, we can split the second term in Eq.~\eqref{eq:term_2} into another two terms:
\begingroup\small
\begin{align*}
    \eqref{eq:term_2} &=
    \sum_{t=0}^{t_0} \exp \left(-2 C \sum_{s=t+1}^{T-1} \gamma_s \right)
    \\
    &\quad {}\times
    \left[
    4 \gamma_t \left( \ell_t \ell^{-1} \right)^d \frac{C_{d,s} \sqrt{\ell_t^2 - \ell^2} }{\ell^3}
    + \frac{C_k \gamma_t^2}{\ell^4}
    + C_k \gamma_t^2\frac{(\ell_t^2-\ell^2)^2}{\ell^8}
    \right] \\
    & + \sum_{t=t_0+1}^{T-1} \exp \left(-2 C \sum_{s=t+1}^{T-1} \gamma_s \right)
    \\
    &\quad {}\times
    \left[
    4 \gamma_t \left( \ell_t \ell^{-1} \right)^d \frac{C_{d,s} \sqrt{\ell_t^2 - \ell^2} }{\ell^3}
    + \frac{C_k \gamma_t^2}{\ell^4}
    + C_k \gamma_t^2\frac{(\ell_t^2-\ell^2)^2}{\ell^8}
    \right] .
\end{align*}
\endgroup
The first term is a finite sum which converges to $0$ as $T\to\infty$ since each term is multiplied by $\exp \left(-2 C \sum_{s=t+1}^{T-1} \gamma_s \right) \to 0$.
The second term, by the choice of $t_0$, is upper bounded by
$\epsilon \sum_{t=t_0+1}^{T-1} \exp\left(-2 C \sum_{s=t+1}^{T-1} \gamma_s \right)\gamma_t $. Set $w_{t,T}:=\exp(-2C\sum_{s=t+1}^{T-1}\gamma_s)$. Note that, $\gamma_t \leq \frac{1}{C}(1-e^{-2C \gamma_t})$ since $0<\gamma_t\leq\frac{1}{2C}$ by the condition imposed in the theorem, and we have,
\begin{align}\label{eq:epsilon}
    \epsilon \sum_{t=t_0+1}^{T-1} w_{t,T}\gamma_t
    &\leq \frac{\epsilon}{C} \sum_{t=t_0+1}^{T-1} w_{t,T} \nonumber \\
    &\quad {}\times \bigl(1-e^{-2C\gamma_t}\bigr) \nonumber \\
    &=
    \frac{\epsilon}{C} \sum_{t=t_0+1}^{T-1}
    \left[
    \exp \left(-2 C \sum_{s=t+1}^{T-1} \gamma_s \right)
    -
    \exp\left(-2 C \sum_{s=t}^{T-1} \gamma_s \right)
    \right] \nonumber \\
    &= \frac{\epsilon}{C} \\
    &\quad {}\times
    \left(1-\exp\left(-2C \sum_{s=t_0+1}^{T-1}\gamma_s\right)\right)
    \leq \frac{\epsilon}{C} .
\end{align}
The sum is bounded. Since $C$ is a constant independent of $t$ and $\epsilon$ is arbitrary, the second term also converges to zero.
Therefore, we have proved that $\lim_{T\to\infty} \mmd_{\ell}^2(\Pb_T, \Qb) = 0$.

\begin{prop}[Descent Lemma] \label{prop:descent_np}
Let kernel $k$ satisfy \textcolor{black}{\Cref{ass:kernel}}.
Let $t\in\N$ be fixed. 
Suppose $\Qb\in\calP_2(\R^d)$ and let 
$\Pb_{t+1} = (\Id - \gamma \nabla f_{\ell_t, \Pb_t,\Qb})_\# \Pb_t$ be the adaptive MMD descent defined in Eq.~\eqref{eq:adaptive_mmd_descent} with lengthscale $\ell_t>0$ and step size $\gamma >0$.
Let $0<\ell< \ell_t$ be another lengthscale. Then, 
\begin{align*}
    &\quad \mmd_{\ell}^2(\Pb_{t+1}, \Qb) -\mmd_{\ell}^2(\Pb_{t}, \Qb) \\
    &\leq -2\gamma \left( \ell_t \ell^{-1} \right)^d
    \|\nabla f_{\ell, \Pb_t, \Qb}\|_{L_2^d(\Pb_t)}^2 \\
    &\quad + 4\gamma \left( \ell_t \ell^{-1} \right)^d
    \frac{C_{d,s} \sqrt{\ell_t^2 - \ell^2} }{\ell^3}
    + \frac{C_k \gamma^2}{\ell^4} \\
    &\quad + C_k \gamma^2\frac{(\ell_t^2-\ell^2)^2}{\ell^8}.
\end{align*}
Here, $C_k$ is a constant depending only on the kernel $k$ and is independent of $\ell$; and $C_{d,s}$ is a constant depending only on the dimension $d$ and kernel smoothness $s$.
\end{prop}
\begin{proof}
In the proof of this proposition, since the time index is always fixed at $t$, we use the shorthand notation that $f_{\ell} := f_{\ell, \Pb_t, \Qb}$.
From \Cref{lem:descent}, we have
\begin{align}\label{eq:descent_lemma_1}
\mmd_{\ell}^2(\Pb_{t+1}, \Qb) -\mmd_{\ell}^2(\Pb_{t}, \Qb)
&\leq - 2\gamma \int \nabla f_{\ell}(x)^\top
\nabla f_{\ell_{t}}(x) \dd \Pb_t(x) \nonumber \\
&\quad + \frac{C_k\gamma^2}{\ell^2}
\|\nabla f_{\ell_{t}}\|^2_{L_2^d(\Pb_t)}.
\end{align}
The first term on the right hand side of Eq.~\eqref{eq:descent_lemma_1} can be further upper bounded by the following:
\begin{align*}
    &\quad -\int \nabla f_{\ell}(x)^\top \nabla f_{\ell_{t}}(x) \dd \Pb_t(x) \\
    &= - \left( \ell_t \ell^{-1} \right)^d \int \nabla f_{\ell}(x)^\top \E_{U\sim\nu} \Big[ \nabla f_{\ell}(x + U)\Big] \dd \Pb_t(x) \\
    &= - \left( \ell_t \ell^{-1} \right)^d
    \left[ \left\| \nabla f_{\ell} \right\|^2_{L_2^d(\Pb_t)} \right. \\
    &\hspace{60pt}\left. {}
    + \int \nabla f_{\ell}(x)^\top
    \left\{ \E_{U\sim\nu} \Big[ \nabla f_{\ell}(x + U)\Big]
    - \nabla f_{\ell}(x) \right\} \dd \Pb_t(x) \right] .
\end{align*}
The second last line holds by \Cref{lem:convolution}, where $\nu$  is a probability measure on $\R^d$ with $\E_{U\sim\nu}[\|U\|]\leq C_{d,s} \sqrt{{\ell_t}^2-{\ell}^2}$ for some constant $C_{d,s}$ depending only on the dimension $d$ and the smoothness parameter $s$. The last equality is just adding and subtracting the same term. 
Next, by the Cauchy-Schwarz inequality, we have
\begin{align*}
    &\leq -\left( \ell_t \ell^{-1} \right)^d
    \left[\left\| \nabla f_{\ell} \right\|^2_{L_2^d(\Pb_t)} \right. \\
    &\hspace{45pt}\left. {}
    - \int \left\| \nabla f_{\ell}(x)\right\|
    \left\| \E_{U\sim\nu} \Big[ \nabla f_{\ell}(x + U)\Big]
    - \nabla f_{\ell}(x) \right\| \dd \Pb_t(x)\right] \\
    &\leq -\left( \ell_t \ell^{-1} \right)^d
    \left[\left\| \nabla f_{\ell} \right\|^2_{L_2^d(\Pb_t)} \right. \\
    &\hspace{45pt}\left. {}
    - \int \left\| \nabla f_{\ell}(x)\right\|
    \E_{U\sim\nu} \left\| \nabla f_{\ell}(x + U)
    - \nabla f_{\ell}(x) \right\| \dd \Pb_t(x)\right] \\
    &\leq - \left( \ell_t \ell^{-1} \right)^d
    \left[ \left\| \nabla f_{\ell} \right\|^2_{L_2^d(\Pb_t)} \right. \\
    &\hspace{45pt}\left. {}
    - \frac{2 C_k}{\ell^2} \left\| \nabla f_{\ell} \right\|_{L_2^d(\Pb_t)}
    \E_{U\sim\nu}\left[ \|U\| \right]\right] \\
    &= - \left( \ell_t \ell^{-1} \right)^d \left[\left\| \nabla f_{\ell} \right\|^2_{L_2^d(\Pb_t)} - \left\| \nabla f_{\ell} \right\|_{L_2^d(\Pb_t)} \frac{C_{d,s}C_k \sqrt{\ell_t^2 - \ell^2} }{\ell^2}\right] .
\end{align*}
The second last inequality holds by the fact that $x\mapsto \nabla f_{\ell}(x)$ is \((2 C_k/\ell^2)\)-Lipschitz with a kernel-dependent constant $C_k$ for the Gaussian and the Mat\'ern kernel in \textcolor{black}{\Cref{ass:kernel}}.
The last inequality holds by the first moment bound on $U\sim\nu$. 
Then, for the second term on the right hand side of Eq.~\eqref{eq:descent_lemma_1}, we use the triangular-inequality and \Cref{lem:lengthscale_bound}:
\begin{align*}
    \|\nabla f_{\ell_{t}}\|^2_{L_2^d(\Pb_t)}
    &\leq 2\left\| \nabla f_{\ell} \right\|^2_{L_2^d(\Pb_t)}
    + 2\left\| \nabla f_{\ell_{t}} - \nabla f_{\ell}
    \right\|^2_{L_2^d(\Pb_t)} \\
    &\leq 2\left\| \nabla f_{\ell} \right\|^2_{L_2^d(\Pb_t)}
    + C_k \left(\frac{\ell_t^2 - \ell^2}{\ell^3}\right)^2 .
\end{align*}
Here, $C_k$ is a constant depending only on the kernel $k$ and is independent of $\ell$.
Combine the above two upper bounds for Eq.~\eqref{eq:descent_lemma_1}, we obtain
\begingroup\small
\begin{align*}
     \mmd_{\ell}^2(\Pb_{t+1}, \Qb) -\mmd_{\ell}^2(\Pb_{t}, \Qb)
    &\leq -2\gamma \left( \ell_t \ell^{-1} \right)^d
    \left[
    \|\nabla f_{\ell}\|_{L_2^d(\Pb_t)}^2
    \right.\\
    &\hspace{75pt}\left. {}
    -
    \left\| \nabla f_{\ell} \right\|_{L_2^d(\Pb_t)} \frac{C_{d,s} C_k \sqrt{\ell_t^2 - \ell^2} }{\ell^2}
    \right]\\
    &\quad + \frac{C_k\gamma^2}{\ell^2}
    \left[
    2\|\nabla f_{\ell}\|_{L_2^d(\Pb_t)}^2
    + C_k \left(\frac{\ell_t^2-\ell^2}{\ell^3}\right)^2 
    \right]
\end{align*}
\endgroup
Expanding, this gives,
\begingroup\small
\begin{align*}
    &\quad \mmd_{\ell}^2(\Pb_{t+1}, \Qb) -\mmd_{\ell}^2(\Pb_{t}, \Qb) \\
    &\leq -2\gamma \left( \ell_t \ell^{-1} \right)^d
    \|\nabla f_{\ell}\|_{L_2^d(\Pb_t)}^2 \\
    &\quad + 2\gamma \left( \ell_t \ell^{-1} \right)^d
    \|\nabla f_{\ell}\|_{L_2^d(\Pb_t)}
    \frac{C_{d,s} C_k \sqrt{\ell_t^2 - \ell^2} }{\ell^2} \\
    &\qquad + \frac{C_k\gamma^2}{\ell^2}
    \|\nabla f_{\ell} \|_{L_2^d(\Pb_t)}^2 + C_k \gamma^2\frac{(\ell_t^2-\ell^2)^2}{\ell^8} \\
    &\leq -2\gamma \left( \ell_t \ell^{-1} \right)^d
    \|\nabla f_{\ell}\|_{L_2^d(\Pb_t)}^2 \\
    &\quad + 4\gamma \left( \ell_t \ell^{-1} \right)^d
    \frac{C_{d,s} C_k \sqrt{\ell_t^2 - \ell^2} }{\ell^3}
    + \frac{C_k \gamma^2}{\ell^4}
    + C_k \gamma^2\frac{(\ell_t^2-\ell^2)^2}{\ell^8},
\end{align*}
\endgroup
where the last step holds by $\|\nabla f_{\ell}\|_{L_2^d(\Pb_t)}\leq C_k \ell^{-1}$ for some constant $C_k$ depending only on the kernel.
\end{proof}

\section{\texorpdfstring{Proof of \Cref{thm:p_rate}}{Proof of the parametric convergence theorem}}\label{sec:proof_p_descent}

\begin{proof}
In the proof, we drop the superscript $\pgd$ for notational simplicity.
Let $\widehat{\Pb}_{t+1} = (\Id - \gamma_t \nabla f_{\ell_t, \tildePthetanopgd{t},\Qb})_\# \tildePthetanopgd{t}$ be the next distribution under the nonparametric MMD update scheme.
Let $\Pb_{\theta_{t+1}} = \Pb_{\theta_t -\gamma_t \Delta \theta_{t, m}^\pgd}$ be the next distribution under the parametric update scheme in Eq.~\eqref{eq:parametric_mmd_update_new}.
For comparison purposes, let $\tildePthetanopgd{t+1} = \Pb_{\theta_t -\gamma_t \tildetheta{t}}$ be the next distribution under the parametric update scheme in Eq.~\eqref{eq:parametric_mmd_update_new_tilde}. $\widehat{\Pb}_{t+1}, \Pb_{\theta_{t+1}}$ are auxiliary quantities, while $\tildePthetanopgd{t+1}$ is our next iterate of interest in practice.
First, from the descent lemma proved in \Cref{prop:descent_np}, we have
\begingroup\small
\begin{align*}
    \mmd_{\ell}^2 (\widehat{\Pb}_{t+1}, \Qb)
    -\mmd_{\ell}^2(\tildePthetanopgd{t}, \Qb)
    &\leq -2\gamma_t \left( \ell_t \ell^{-1} \right)^d
    \|\nabla f_{\ell, \tildePthetanopgd{t}, \Qb}\|_{L_2^d(\tildePthetanopgd{t})}^2 \\
    &\quad + 4 \gamma_t \left( \ell_t \ell^{-1} \right)^d
    \frac{C_{d,s} \sqrt{\ell_t^2 - \ell^2} }{\ell^3}
    + \frac{C_k \gamma_t^2}{\ell^4} \\
    &\quad + C_k \gamma_t^2\frac{(\ell_t^2-\ell^2)^2}{\ell^8} .
\end{align*}
\endgroup
Then, from \Cref{prop:delta_theta_ast}, or more precisely from \Cref{prop:delta_theta_est_precise}, we have
\begingroup\small
\begin{align*}
    &\quad \mmd_{\ell}^2 (\Pb_{\theta_{t+1}}, \Qb) -\mmd_{\ell}^2(\widehat{\Pb}_{t+1}, \Qb) \\
    &\leq 2\gamma_t \mathfrak{R} \|\nabla f_{\ell,\tildePthetanopgd{t},\Qb}\|_{L_2^d(\tildePthetanopgd{t})}^2
    + C_k\gamma_t \mathfrak{R} \frac{\ell_t^2 - \ell^2}{\ell^4} \\
    &\quad + C_k d^{\frac{1}{2}} p^{\frac{2}{5}} \gamma_t^{\frac{6}{5}} \calU^{\frac{4}{5}} \ell^{-1}\left((1+\ell^{-1})\right)^{\frac{1}{5}} \ell_t^{-\frac{2}{5}}
    + C_k\gamma_t^2 \ell^{-2} \ell_t^{-2} .
\end{align*}
\endgroup
Finally, applying \Cref{prop:tilde_no_tilde} with evaluation lengthscale $\ell$ and update lengthscale $\ell_t$, and taking expectation over $\widehat{\Qb}_m$, gives
\begin{align*}
    \E_m\left[
    \mmd_{\ell}^2 ({\tildePthetanopgd{t+1}}, \Qb) - \mmd_{\ell}^2 (\Pb_{\theta_{t+1}}, \Qb)
    \right]
    \leq C_k \gamma_t \lambda_t^{-1} p \ell^{-1}\ell_t^{-1} \sqrt{d} \cdot \frac{1}{\sqrt{m}}.
\end{align*}
Adding the above three upper bounds and taking expectation over $\widehat{\Qb}_m$, we obtain
\begingroup\small
\begin{align}
    &\E_m\left[
    \mmd_{\ell}^2 (\tildePthetanopgd{t+1}, \Qb) - \mmd_{\ell}^2 (\tildePthetanopgd{t}, \Qb)
    \right] \nonumber\\
    &
    \leq -2(1-\mathfrak{R}) \gamma_t
    \E_m\left[
    \left( \ell_t \ell^{-1} \right)^d \|\nabla f_{\ell, \tildePthetanopgd{t}, \Qb}\|_{L_2^d(\tildePthetanopgd{t})}^2
    \right] \label{eq:descent_lemma_p} \\
    &\quad + C_k\gamma_t \mathfrak{R} \frac{\ell_t^2 - \ell^2}{\ell^4}
    + C_k d^{\frac{1}{2}} p^{\frac{2}{5}} \gamma_t^{\frac{6}{5}} \calU^{\frac{4}{5}} \nonumber \\
    &\qquad {}\times \ell^{-1}\left(1+\ell^{-1}\right)^{\frac{1}{5}}
    \ell_t^{-\frac{2}{5}}
    + C_k\gamma_t^2 \ell^{-2} \ell_t^{-2}
    \label{eq:mmd_parametric_descent_proof_residual_1} \\
    &\quad + 4 \gamma_t \left( \ell_t \ell^{-1} \right)^d
    \frac{C_{d,s} \sqrt{\ell_t^2 - \ell^2} }{\ell^3}
    + \frac{C_k \gamma_t^2}{\ell^4} \nonumber \\
    &\qquad + C_k \gamma_t^2\frac{(\ell_t^2-\ell^2)^2}{\ell^8}
    \label{eq:mmd_parametric_descent_proof_residual_2} \\
    &\quad + C_k \gamma_t^{\frac{3}{5}} p^{\frac{1}{5}}
    \ell^{-1}\left(1+\ell^{-1}\right)^{-\frac{2}{5}}
    \ell_t^{-\frac{1}{5}} \calU^{\frac{2}{5}} \cdot \frac{1}{\sqrt{m}}
    \label{eq:mmd_parametric_descent_proof_residual_3} .
\end{align}
\endgroup
Denote the sum of Eq.~\eqref{eq:mmd_parametric_descent_proof_residual_1} and Eq.~\eqref{eq:mmd_parametric_descent_proof_residual_2} as $\mathfrak{M}_t$, which is dependent on $t$ but independent of $m$.
Note that the term in Eq.~\eqref{eq:mmd_parametric_descent_proof_residual_3} except from the multiplier $\gamma_t^{\frac{3}{5}} m^{-1/2}$ can be upper bounded as
{$C_k p^{\frac{1}{5}} \ell^{-1}\left((1+\ell^{-1})\right)^{-\frac{2}{5}}\ell^{-\frac{1}{5}} \calU^{\frac{2}{5}} =: \mathfrak{N}$} which is independent of either $t$ or $m$.
For the well-specified case, under the pathwise gradient-dominance assumption stated in the theorem,
\begingroup\small
\begin{align*}
    \E_m\left[\mmd_{\ell}^2 (\tildePthetanopgd{t+1}, \Qb)\right]
    &\leq \Big(1 -2C(1-\mathfrak{R}) \gamma_t \Big)
    \E_m\left[\mmd_{\ell}^2 (\tildePthetanopgd{t}, \Qb)\right]
    \\
    &\quad + \mathfrak{M}_t
    + \mathfrak{N} \cdot \gamma_t^{\frac{3}{5}} \frac{1}{\sqrt{m}}.
\end{align*}
\endgroup
The following strategy will be similar to the proof in the nonparametric case in \Cref{sec:proof_np_descent}. 
Fix the optimization horizon $T\in\N$.
By the discrete Gronwall lemma ~\citep[Appendix~B.2, Gronwall's inequality]{evans2022partial}, we iterate this recursion over $t=0,\ldots,T-1$ and obtain
\begin{align}
    \E_m\left[\mmd_{\ell}^2(\tildePthetanopgd{T}, \Qb)\right]
    &\leq \exp \left(-2C(1-\mathfrak{R}) \sum_{s=0}^{T-1} \gamma_s \right)
    \mmd_{\ell}^2(\Pb_{\theta_0}, \Qb)
    \label{eq:parametric_term_1} \\
    &\quad + \sum_{t=0}^{T-1}
    \exp \left(-2C (1-\mathfrak{R}) \sum_{s=t+1}^{T-1} \gamma_s \right)
    \mathfrak{M}_t \label{eq:parametric_term_2} \\
    &\quad + \sum_{t=0}^{T-1}
    \exp \left(-2C (1-\mathfrak{R}) \sum_{s=t+1}^{T-1} \gamma_s \right) \\
    &\qquad {}\times \mathfrak{N} \gamma_t^{\frac{3}{5}} \frac{1}{\sqrt{m}}
    \label{eq:parametric_term_3} .
\end{align}
The first term in Eq.~\eqref{eq:parametric_term_1} converges to $0$ as $T\to\infty$ by the condition that $\lim_{T\to\infty} \sum_{s=0}^{T-1} \gamma_s \to \infty$.
For the second term in Eq.~\eqref{eq:parametric_term_2}, the analysis is more involved.
Let $\varepsilon > 0$ be arbitrary small. By definition of $\mathfrak{M}_t$ as the sum of Eq.~\eqref{eq:mmd_parametric_descent_proof_residual_1} and Eq.~\eqref{eq:mmd_parametric_descent_proof_residual_2}, it satisfies $\lim_{t\to\infty} \mathfrak{M}_t \gamma_t^{-1} = 0$. 
As a result, there exists $t_0\in\N$ such that $\mathfrak{M}_t \gamma_t^{-1} < \varepsilon$ for all $t>t_0$. 
We can split the second term in Eq.~\eqref{eq:parametric_term_2} into another two terms:
\begingroup\small
\begin{align*}
    \eqref{eq:parametric_term_2} &=
    \sum_{t=0}^{t_0} \exp \left(-2C (1-\mathfrak{R}) \sum_{s=t+1}^{T-1} \gamma_s \right)  \mathfrak{M}_t \\
    &\quad + \sum_{t=t_0+1}^{T-1} \exp \left(-2C (1-\mathfrak{R}) \sum_{s=t+1}^{T-1} \gamma_s \right) \mathfrak{M}_t.
\end{align*}
\endgroup
The first term in Eq.~\eqref{eq:parametric_term_2} is a finite sum  which converges to $0$ as $T\to\infty$ since each term is multiplied by $\exp (-2 C(1-\mathfrak{R})\sum_{s=t+1}^{T-1} \gamma_s ) \to 0$.
For the second term in Eq.~\eqref{eq:parametric_term_2}, by the choice of $t_0$, is upper bounded by 
$\epsilon \sum_{t=t_0+1}^{T-1} \exp(-2 C (1-\mathfrak{R})\sum_{s=t+1}^{T-1} \gamma_s )\gamma_t$.
Since $0<\gamma_t\leq\frac{1}{2C(1-\mathfrak{R})}$, following the same derivations as in Eq.~\eqref{eq:epsilon}, the second term in Eq.~\eqref{eq:parametric_term_2} is upper bounded by $\epsilon$. As $\epsilon$ is arbitrary, the second term also converges to zero as $T\to\infty$.
Finally, for the third term in Eq.~\eqref{eq:parametric_term_3}, recall that $\sqrt{m(T)} \geq (\sum_{s=0}^T \gamma_s^{0.6})  \cdot \log(T+1)$ assumed in the statement of the Theorem. 
\begin{align*}
    \eqref{eq:parametric_term_3} \leq \sum_{t=0}^{T-1}  \mathfrak{N} \cdot \gamma_t^{\frac{3}{5}} \frac{1}{\sqrt{m(T)}} \leq \frac{\mathfrak{N}}{\log(T+1)} \rightarrow 0, \quad \text{as  } T \to \infty.
\end{align*}
Therefore, we have proved that all the three terms in
Eq.~\eqref{eq:parametric_term_1}, Eq.~\eqref{eq:parametric_term_2} and Eq.~\eqref{eq:parametric_term_3} vanish as $T\to\infty$. {Denote the deterministic upper bound given by the right-hand side of Eq.~\eqref{eq:parametric_term_1}--\eqref{eq:parametric_term_3} as $r_T$.}
Hence,
\begin{align*}
    \E_m\left[\mmd_{\ell}^2(\tildePthetanopgd{T}, \Qb)\right] \leq r_T,
\end{align*}
This proves part \textup{(i)}. 

For the mis-specified case, under the alternative gradient-dominance condition in part \textup{(ii)}, Eq.~\eqref{eq:descent_lemma_p} gives the same recursion for the excess objective:
\begin{align*}
    &\quad \mmd_{\ell}^2 (\tildePthetanopgd{t+1}, \Qb)-{\min_{\theta\in\Theta}}\mmd_{\ell}^2(\Pb_\theta,\Qb) \\
    & \leq \Big(1 -2C(1-\mathfrak{R}) \gamma_t \Big) \\
    &\quad {}\times \left(\mmd_{\ell}^2 (\tildePthetanopgd{t}, \Qb)
    -{\min_{\theta\in\Theta}}\mmd_{\ell}^2(\Pb_\theta,\Qb) \right)
    +  \mathfrak{M}_t + \mathfrak{N} \cdot \gamma_t^{\frac{3}{5}} \frac{1}{\sqrt{m}} .
\end{align*}
{The same expected-recursion and discrete Gronwall argument therefore imply that}
\[
\E_m\left[
\mmd_{\ell}^2(\tildePthetanopgd{T},\Qb)-{\min_{\theta\in\Theta}}\mmd_{\ell}^2(\Pb_\theta,\Qb)
\right] \leq r_T,
\]
{where the corresponding deterministic remainder satisfies $r_T\to0$ as $T\to\infty$ under the same sample-size schedule.} This proves part \textup{(ii)}.
\end{proof}

\subsection{\texorpdfstring{Proof of \Cref{prop:delta_theta_ast}}{Proof of the parameter-update proposition}}\label{sec:proof_delta_theta}

\begin{prop}[Optimality of $\Delta \theta_t^\pgd$] \label{prop:delta_theta_est_precise}
Suppose \textcolor{black}{\Cref{ass:regularity_G,ass:residual}} hold.
Let both the nonparametric and parametric MMD descent schemes share the same initialization: $\Pb_0 = \Pb_{\theta_0}$.
Let $\gamma_0 > 0$.
Denote $\Pb_1$ as the next iterate under Eq.~\eqref{eq:adaptive_mmd_descent}. Denote $\Pb^\pgd_{\theta_{1}}$ as the next iterates under Eq.~\eqref{eq:mmd_descent} with a preconditioned update $\Delta \theta^\pgd$. 
Then, taking $\lambda_0 \propto (\gamma p^2 (1+\ell^{-1})  \ell_0^{-2})^{\frac{2}{5}}$, we have
\begingroup\small
\begin{align}\label{eq:precise_proposition}
    &\quad \mmd_{\ell}^2 (\Pb_{\theta^\pgd_{1}}, \Qb) -\mmd_{\ell}^2(\Pb_{1}, \Qb) \nonumber \\
    &\leq 2\gamma_0 \mathfrak{R} \|\nabla f_{\ell,\Pb_0,\Qb}\|_{L_2^d(\Pb_0)}^2
    + C_k\gamma_0 \mathfrak{R} \frac{\ell_0^2 - \ell^2}{\ell^4} \nonumber\\
    &\quad + C_k d^{\frac{1}{2}} p^{\frac{2}{5}} \gamma_0^{\frac{6}{5}} \calU^{\frac{4}{5}} \ell^{-1}\left((1+\ell^{-1})\right)^{\frac{1}{5}} \ell_0^{-\frac{2}{5}}
    + C_k\gamma_0^2 \ell^{-2} \ell_0^{-2}.
\end{align}
\endgroup
Here, $C_k$ depends only on the kernel profile. 
\end{prop}
\begin{proof}
Let $\Pb_0 = \Pb_{\theta_0}=(\Gmap_{\theta_0})_{\#}\rho$.
In the sequel, since the time index is fixed at $t=0$, we drop the subscripts in $\gamma_0, \lambda_0$.
Then the next iterate produced by the parametric MMD descent scheme and by the MMD gradient descent scheme is given by, respectively,
\begin{align*}
\Pb_{\theta_{1}} &= (\Gmap_{\theta_0 - \gamma \Delta \theta})_\# \rho, \\
\Pb_{1}
&= (\Id - \gamma \nabla f_{\ell_0, \Pb_0, \Qb})_\# \Pb_0 \\
&= (\Id - \gamma \nabla f_{\ell_0, \Pb_0, \Qb})_\#
(\Gmap_{\theta_0})_\# \rho .
\end{align*}
Note that here $\Delta \theta$ is a generic update in $\R^p$.
Next, we consider the squared MMD distance between the next parametric MMD descent and the target $\Qb$; as well as the squared MMD distance between the next nonparametric MMD descent and the target $\Qb$.
Specifically,
\begin{align}
\calI_1(\gamma) &:= \mmd_{\ell}^2\left(
(\Id-\gamma \nabla f_{\ell_0,\Pb_0,\Qb})_\# \Pb_0,\Qb\right) \nonumber \\
&=
\E_{z\sim\rho,z'\sim \rho}
\left[
\begin{aligned}
&k_{\ell}\Big(
\Gmap_{\theta_0}(z)-\gamma \nabla f_{\ell_0,\Pb_0,\Qb}(\Gmap_{\theta_0}(z)), \\
&\hspace{45pt}
\Gmap_{\theta_0}(z')-\gamma \nabla f_{\ell_0,\Pb_0,\Qb}(\Gmap_{\theta_0}(z'))
\Big)
\end{aligned}
\right] \label{eq:I_1_term_1} \\
&\quad
-2 \E_{z\sim \rho, y\sim \Qb}
\Big[
k_{\ell}\left(
\Gmap_{\theta_0}(z)-\gamma \nabla f_{\ell_0,\Pb_0,\Qb}(\Gmap_{\theta_0}(z)),
y
\right)
    \Big] \nonumber \\
    &\quad +\E_{y\sim\Qb,y'\sim \Qb}\left[k_{\ell}(y,y')\right]
    \label{eq:I_1_term_2} .
\end{align}
And
\begin{align}
\calI_2(\gamma) &:= \mmd_{\ell}^2\left(
(\Gmap_{\theta_0 - \gamma \Delta \theta})_\# \rho,\Qb\right) \nonumber \\
&= \E_{z,z'\sim \rho}
\Big[
k_{\ell} \left(
\Gmap_{\theta_0 - \gamma \Delta \theta}(z),
\Gmap_{\theta_0 - \gamma \Delta \theta}(z')
\right)
\Big] \label{eq:I_2_term_1} \\
&\quad
-2 \E_{z\sim \rho, y\sim \Qb}
\Big[
k_{\ell}\left(
\Gmap_{\theta_0 - \gamma \Delta \theta}(z),
y
\right)
\Big]
+\E_{y,y'\sim \Qb}\left[k_{\ell}(y,y')\right] \label{eq:I_2_term_2}
\end{align}
For the Gaussian kernel and for the Mat\'ern kernel with smoothness $s>1$, the map $(x,y)\mapsto k_\ell(x,y)$ is twice continuously differentiable with first- and second-order derivatives bounded by constants of order $\ell^{-1}$ and $\ell^{-2}$, respectively. Therefore, Taylor expansions with respect to $\gamma$ can be applied to both Eq.~\eqref{eq:I_1_term_1} and Eq.~\eqref{eq:I_1_term_2}.
We obtain
\begin{align*}
    \eqref{eq:I_1_term_1} &= \E_{z, z^{\prime}}\Big[k_{\ell}\left(\Gmap_{\theta_0}(z), \Gmap_{\theta_0}(z^{\prime})\right)\Big] \\
    &\quad - 2\gamma \E_{z, z^{\prime}}\left[\nabla_1 k_{\ell}\left(\Gmap_{\theta_0}(z), \Gmap_{\theta_0}(z^{\prime})\right)^{\top} \nabla f_{\ell_0,\Pb_0, \Qb}\left(\Gmap_{\theta_0}(z)\right)\right] + \frac{\kappa_1}{2} \gamma^2 \\
    \eqref{eq:I_1_term_2} &= -2 \E_{z,y}\left[ k_{\ell}\left(\Gmap_{\theta_0}(z),y\right)
\right]
+2\gamma \E_{z,y}\left[
\nabla_1 k_{\ell}\left(\Gmap_{\theta_0}(z),y\right)^\top
\nabla f_{\ell_0,\Pb_0,\Qb}\left(\Gmap_{\theta_0}(z)\right)
\right]
\\&\qquad + \frac{\kappa_2}{2} \gamma^2  + \E_{y,y'\sim \Qb}\left[k_{\ell}(y,y')\right] .
\end{align*}
Denote $y_s = \Gmap_{\theta_0}(z)-s\gamma \nabla f_{\ell_0,\Pb_0,\Qb}(\Gmap_{\theta_0}(z)) \in\R^d$ for $s\in[0,1]$; and similarly denote $y_s'$.
Here, $\kappa_1, \kappa_2 \in\R$ above is the remainder term that admits the following bound:
\begingroup\small
\begin{align*}
    \max\{\left|\kappa_1\right|, \left|\kappa_2 \right|\} &\leq \sup _{s \in[0,1]}\E_{z, z^{\prime}} \left[ \nabla f_{\ell_0,\Pb_0, \Qb}\left(\Gmap_{\theta_0}(z)\right)^{\top} \nabla_{11}^2 k_{\ell}(y_s, y_s') \nabla f_{\ell_0,\Pb_0, \Qb}\left(\Gmap_{\theta_s}(z)\right) \right] \\
    & + \sup _{s \in[0,1]} \E_{z, z^{\prime}} \left[ \nabla f_{\ell_0,\Pb_0, \Qb}\left(\Gmap_{\theta_s}(z)\right)^{\top} \nabla_{12}^2 k_{\ell}(y_s, y_s') \nabla f_{\ell_0,\Pb_0, \Qb}\left(\Gmap_{\theta_s}(z)\right) \right] \\
    &\leq {C_k \ell^{-2} \ell_0^{-2}} .
\end{align*}
\endgroup
The last inequality holds by the bounds
\[
\sup_{y,y'} \| \nabla_{11}^2 k_{\ell}(y,y') \|_{\mathrm{op}} \leq C_k\ell^{-2},\quad
\sup_{y,y'} \| \nabla_{12}^2 k_{\ell}(y,y') \|_{\mathrm{op}} \leq C_k\ell^{-2},
\]
and $\sup_{x}\|\nabla f_{\ell_0, \Pb_0, \Qb}(x)\| \leq C_k \ell_0^{-1}$.
Combining the above two equalities on Eq.~\eqref{eq:I_1_term_1} and Eq.~\eqref{eq:I_1_term_2}, we obtain
\begin{align}\label{eq:I_1_taylor}
\calI_1(\gamma)
&= \mmd_{\ell}^2(\Pb_{\theta_0}, \Qb)
- 2\gamma \cdot \E_z\left[
\nabla f_{\ell,\Pb_0,\Qb}\left(\Gmap_{\theta_0}(z)\right)^\top \right. \nonumber \\
&\hspace{90pt}\left. {}
\times \nabla f_{\ell_0,\Pb_0,\Qb}\left(\Gmap_{\theta_0}(z)\right)
\right] + \frac{\kappa_1+\kappa_2}{2} \gamma^2.
\end{align}
Similarly, we can perform the same analysis for $\calI_2(\gamma)$. Note that $\theta\mapsto \Gmap_\theta(z)$ is second-order continuously differentiable by \textcolor{black}{\Cref{ass:regularity_G}}.
Under Taylor expansion, we have
\begingroup\small
\begin{align*}
    \eqref{eq:I_2_term_1}
    &= \E_{z, z^{\prime}}\left[k_{\ell}\left(\Gmap_{\theta_0}(z),
    \Gmap_{\theta_0}(z^{\prime})\right)\right] \\
    &\quad - 2\gamma \E_{z, z^{\prime}}\left[
    \nabla_1 k_{\ell}\left(\Gmap_{\theta_0}(z), \Gmap_{\theta_0}(z^{\prime})\right)^{\top}
    \bJ_\theta \Gmap_{\theta_0}(z) \Delta\theta \right]
    + \kappa_3 \frac{\gamma^2}{2}, \\
    \eqref{eq:I_2_term_2}
    &= -2 \E_{z,y}\left[k_{\ell}\left(\Gmap_{\theta_0}(z),y\right) \right]
    + 2\gamma \E_{z,y}\left[
    \nabla_1 k_{\ell} \left(\Gmap_{\theta_0}(z),y\right)^\top
    \bJ_\theta \Gmap_{\theta_0}(z) \Delta\theta
    \right] + \kappa_4 \frac{\gamma^2}{2} .
\end{align*}
\endgroup
Denote $\theta_s = \theta_0 - s \gamma \Delta \theta {\in\Theta}$ for $s\in[0,1]$, then
$\kappa_3,\kappa_4\in\R$ above is the remainder term that admits the following bound:
\begin{align*}
    \max\{|\kappa_3|, |\kappa_4|\}
    &\leq \sup_{s\in[0,1]} \E_{z, z^{\prime}} \left[
    \begin{aligned}
    &\big(\bJ_\theta \Gmap_{\theta_s}(z)\Delta\theta\big)^\top
    \nabla_{11}^2 k_{\ell}\left(\Gmap_{\theta_s}(z),\Gmap_{\theta_s}(z')\right) \\
    &\qquad {}\times \big(\bJ_\theta \Gmap_{\theta_s}(z)\Delta\theta\big)
    \end{aligned}
    \right] \\
    &\quad + \sup_{s\in[0,1]} \E_{z, z^{\prime}} \left[
    \begin{aligned}
    &\big(\bJ_\theta \Gmap_{\theta_s}(z)\Delta\theta\big)^\top
    \nabla_{12}^2 k_{\ell}\left(\Gmap_{\theta_s}(z),\Gmap_{\theta_s}(z')\right) \\
    &\qquad {}\times \big(\bJ_\theta \Gmap_{\theta_s}(z')\Delta\theta\big)
    \end{aligned}
    \right] \\
    &\quad + \sup_{s\in[0,1]} \E_{z, z^{\prime}} \left[
    \begin{aligned}
    &\nabla_1 k_{\ell}\left(\Gmap_{\theta_s}(z),\Gmap_{\theta_s}(z')\right)^\top \\
    &\qquad {}\times \big(\Delta\theta^\top \nabla_\theta^2
    \Gmap_{\theta_s}(z)\Delta\theta\big)
    \end{aligned}
    \right] \\
    &\leq \sup_{x,x'}\|\nabla_{11}^2 k_{\ell}(x,x')\|_{op} \\
    &\quad {}\times \sup_{s\in[0,1]} \E_{z}
    \left[\|\bJ_\theta \Gmap_{\theta_s}(z)\|_{op}^2\right]
    \|\Delta\theta\|^2 \\
    &\quad + \sup_{x,x'}\|\nabla_{12}^2 k_{\ell}(x,x')\|_{op} \\
    &\quad {}\times \sup_{s\in[0,1]}
    \left[ \E_{z}\|\bJ_\theta \Gmap_{\theta_s}(z)\|_{op} \right]^2
    \|\Delta\theta\|^2 \\
    &\quad + \sup_{x,x'}\|\nabla_1 k_{\ell}(x,x')\| \\
    &\quad {}\times \sup_{s\in[0,1]} \E_{z}
    \left[\|\nabla_\theta^2 \Gmap_{\theta_s}(z)\|_{op} \right]
    \|\Delta\theta\|^2 \\
    &\leq C_k p\sqrt{d}(\ell^{-2}+\ell^{-1}) \|\Delta\theta\|^2.
\end{align*}
The second last step holds by \textcolor{black}{\Cref{ass:regularity_G}} along with the bounds that
\begingroup\small
\begin{align*}
    \sup_s \E_{z} \left[\|\bJ_\theta \Gmap_{\theta_s}(z)\|_{op}^2\right]
    &\leq {\sup_{\theta\in\Theta}} \E_{z\sim\rho}
    \left[\|\bJ_\theta \Gmap_{\theta}(z)\|_{op}^2\right] \\
    &\leq {\sup_{\theta\in\Theta}} \sum_{j=1}^p \E_{z\sim\rho}
    \left[ \|\partial_{\theta_j} \Gmap_\theta(z)\|^2 \right]
    \leq p, \\
    \sup_s \E_{z} \left[ \|\nabla_\theta^2 \Gmap_{\theta_s}(z)\|_{\mathrm{op}} \right]
    &\leq {\sup_{\theta\in\Theta}} \sqrt{\E_{z}
    \left[ \|\nabla_\theta^2 \Gmap_{\theta}(z)\|^2_{\mathrm{op}} \right]} \\
    &\leq {\sup_{\theta\in\Theta}}
    \sqrt{\sum_{i,j=1}^p \E_{z\sim\rho}
    \left[ \|\partial_{\theta_j}\partial_{\theta_i} \Gmap_\theta(z)\|^2 \right]}
    \leq p .
\end{align*}
\endgroup
and by that $\sup_{x,x'} \| \nabla_{1} k_{\ell}(x,x') \|\leq C_k\sqrt{d} \ell^{-1}$, $\sup_{x,x'} \| \nabla_{1,2}^2 k_{\ell}(x,x') \|_{op} \leq C_k\sqrt{d} \ell^{-2}$.
Here, $\nabla_\theta^2 \Gmap_{\theta_s}$ shall be interpreted as a bilinear mapping
\begingroup\small
\begin{align*}
    \nabla_\theta^2 \Gmap_\theta(z)[u, v]
    &:= \left(u^{\top} \nabla_\theta^2 [\Gmap_{\theta}(z)]_1 v, \ldots,
    u^{\top} \nabla_\theta^2 [\Gmap_{\theta}(z)]_d v\right) \\
    &\in\R^d, \qquad \forall u,v\in\R^p ,
\end{align*}
\endgroup
and its operator norm is
\begin{align*}
    \left\|\nabla_\theta^2 \Gmap_\theta(z)\right\|_{\mathrm{op}}
    &:= \sup _{\|u\|=1,\|v\|=1}
    \left\|\nabla_\theta^2 \Gmap_\theta(z)[u, v]\right\|.
\end{align*}

Combining the above two equalities on Eq.~\eqref{eq:I_2_term_1} and Eq.~\eqref{eq:I_2_term_2}, we obtain
\begin{align}\label{eq:I_2_taylor}
\calI_2(\gamma)
&= \mmd_{\ell}^2(\Pb_{\theta_0}, \Qb)
- 2\gamma \cdot \E_z\left[
\nabla f_{\ell,\Pb_0,\Qb}\left(\Gmap_{\theta_0}(z)\right)^\top \right. \nonumber \\
&\hspace{90pt}\left. {}
\times \bJ_\theta \Gmap_{\theta_0}(z) \Delta\theta
\right] + \frac{\kappa_3+\kappa_4}{2} \gamma^2 .
\end{align}
Now, from the Taylor expansion of $\calI_1(\gamma)$ in Eq.~\eqref{eq:I_1_taylor} and $\calI_2(\gamma)$ in Eq.~\eqref{eq:I_2_taylor}, we are going to study the difference of the following two squared MMD distances,
\begin{align}
&\quad \left| \mmd_{\ell}^2\left(
(\Id-\gamma \nabla f_{\ell_0,\Pb_0,\Qb})_\# \Pb_0,\Qb\right) \right. \nonumber \\
&\hspace{45pt}\left. {}
- \mmd_{\ell}^2\left((\Gmap_{\theta_0 - \gamma \Delta \theta})_\# \rho,\Qb\right)
\right| := | \calI_1(\gamma) - \calI_2(\gamma)|\nonumber \\
&= 2\gamma \left| \E_z\left[
\nabla f_{\ell,\Pb_0,\Qb} \left(\Gmap_{\theta_0}(z)\right)^\top
\Big( \nabla f_{\ell_0,\Pb_0,\Qb} \left(\Gmap_{\theta_0}(z)\right)
- \bJ_\theta \Gmap_{\theta_0}(z) \Delta\theta \Big) \right] \right| \nonumber \\
&\qquad + \left| \frac{\kappa_1+\kappa_2}{2} \gamma^2 -\frac{\kappa_3+\kappa_4}{2} \gamma^2  \right| . \nonumber \\
&\leq 2\gamma \|\nabla f_{\ell,\Pb_0,\Qb}\|_{L_2^d(\Pb_0)} \nonumber \\
&\quad {}\times \sqrt{\underbrace{\E_z \left[ \left\|
\nabla f_{\ell_0,\Pb_0,\Qb} \left(\Gmap_{\theta_0}(z)\right)
- \bJ_\theta \Gmap_{\theta_0}(z) \Delta\theta
\right\|^2 \right]}_{\calJ(\Delta \theta)}} \nonumber \\
&\qquad + C_k\gamma^2 \left( p\sqrt{d}(\ell^{-2}+\ell^{-1}) \|\Delta\theta\|^2 + \ell^{-2}\ell_0^{-2} \right) \label{eq:diff_I_1_I_2} .
\end{align}
The last inequality holds by applying Cauchy-Schwarz for the first term, and by plugging in the bounds on $|\kappa_1|, |\kappa_2|, |\kappa_3|, |\kappa_4|$ derived above.
Recall from \textcolor{black}{\Cref{ass:residual}} that $\min_{u\in\R^p}\calJ(u)
\le \mathfrak{R}^2 \|\nabla f_{\ell_0, \Pb_{\theta_0}, \Qb}\|_{L_2^d(\Pb_{\theta_0})}^2 = \mathfrak{R}^2 \|\nabla f_{\ell_0, \Pb_0, \Qb}\|_{L_2^d(\Pb_0)}^2$,
and also that $u_0^\ast:= \arg\min \calJ(u)$ and $\|u_0^\ast\|\leq\calU<\infty$.
These conditions imply that, there exists a $r\in L_2(\rho)$, such that $\nabla f_{\ell_0,\Pb_0,\Qb}(\Gmap_{\theta_0}( \cdot )) =\bJ_\theta \Gmap_{\theta_0}(\cdot) u_0^\ast + r(\cdot)$ with $\|r\|_{L_2(\rho)}^2 \leq \mathfrak{R}^2 \, \|\nabla f_{\ell_0, \Pb_0, \Qb}\|_{L_2^d(\Pb_0)}^2$.
Recall $\Delta \theta_0^\pgd$ defined in Eq.~\eqref{eq:parametric_mmd_update_new},
\begin{align*}
    \Delta \theta_0^\pgd
    &= \left\{ \E_{z\sim\rho}\left[
    \bJ_\theta \Gmap_{\theta_0}(z)^\top \bJ_\theta \Gmap_{\theta_0}(z)\right]
    + \lambda\Id \right\}^{-1} \\
    &\quad {}\times \E_{z\sim\rho}\left[
    \bJ_\theta \Gmap_{\theta_0}(z)^\top
    \nabla f_{\ell_0, \Pb_{\theta_0}, \Qb}(\Gmap_{\theta_0}(z)) \right] .
\end{align*}
From \Cref{lem:source_condition}, we have that
\begin{align*}
    \calJ(\Delta \theta^\pgd) &= \E_{z\sim\rho} \left[ \left\|\nabla f_{\ell_0, \Pb_0, \Qb}(\Gmap_{\theta_0}(z)) - \bJ_\theta \Gmap_{\theta_0}(z) \Delta \theta^\pgd \right\|^2 \right] \\
    &\leq \lambda \|u_0^\ast\|^2 + \mathfrak{R}^2 \, \|\nabla f_{\ell_0, \Pb_0, \Qb}\|_{L_2^d(\Pb_0)}^2 \\
    &\leq \lambda \calU^2 + \mathfrak{R}^2 \, \|\nabla f_{\ell_0, \Pb_0, \Qb}\|_{L_2^d(\Pb_0)}^2 .
\end{align*}
We also have that,
\begin{align*}
    \| \Delta \theta^\pgd \|^2 \leq \frac{1}{\lambda^2} \left\| \E_{z\sim\rho}\left[ \bJ_\theta \Gmap_{\theta_0}(z)^\top \nabla f_{\ell_0, \Pb_0, \Qb}(\Gmap_{\theta_0}(z)) \right]\right\|^2 \leq \frac{C_kp}{\lambda^2} \ell_0^{-2} .
\end{align*}
Denote $\Pb_{\theta^\pgd_{1}}:=(\Gmap_{\theta_0 - \gamma \Delta \theta^\pgd})_\# \rho$ as the next iterate under the update $\Delta \theta^\pgd$.
Therefore, continuing from Eq.~\eqref{eq:diff_I_1_I_2}, we obtain
\begin{align}
    &\quad \mmd_{\ell}^2 (\Pb_{\theta^\pgd_{1}}, \Qb) -\mmd_{\ell}^2(\Pb_{1}, \Qb) \label{eq:tradeoff} \\
    &\leq 2\gamma \|\nabla f_{\ell,\Pb_0,\Qb}\|_{L_2^d(\Pb_0)} \cdot \sqrt{\left( \lambda \calU^2 + \mathfrak{R}^2 \, \|\nabla f_{\ell_0, \Pb_0, \Qb}\|_{L_2^d(\Pb_0)}^2 \right)} \nonumber \\
    &\qquad + {C_k\gamma^2 \left( p\sqrt{d}(\ell^{-2}+\ell^{-1}) \|\Delta\theta^\pgd \|^2 + \ell^{-2}\ell_0^{-2} \right)} . \nonumber \\
    &\leq 2\gamma \mathfrak{R} \|\nabla f_{\ell,\Pb_0,\Qb}\|_{L_2^d(\Pb_0)} \cdot \|\nabla f_{\ell_0, \Pb_0, \Qb}\|_{L_2^d(\Pb_0)} + {C_k\gamma \sqrt{\lambda} \sqrt{d} \ell^{-1} \calU} \tag{Approximation error} \\
    &\qquad +  C_k\gamma^2 \left( p \sqrt{d}(\ell^{-2}+\ell^{-1}) \frac{p}{\lambda^2} \ell_0^{-2}  + \ell^{-2}\ell_0^{-2} \right) \tag{Taylor-expansion error}.
\end{align}
The last step holds by $\sqrt{a+b}\leq\sqrt{a}+\sqrt{b}$, $\sup_z\|\nabla f_{\ell,\Pb_0,\Qb}(z)\|\leq C_k\sqrt{d} \ell^{-1}$, and by plugging in the upper bound of $\|\Delta\theta^\pgd\|^2$.
One can observe a tradeoff on $\lambda$ in the above formula; see \Cref{rem:tradeoff} for details. To balance the two terms regarding $\lambda$ above, we pick $\lambda \asymp \left(\gamma p^2 (1+\ell^{-1}) \ell_0^{-2}\calU^{-1}\right)^{\frac{2}{5}}$. 
This results in
\begingroup\small
\begin{align*}
    &\quad \mmd_{\ell}^2 (\Pb_{\theta^\pgd_{1}}, \Qb) -\mmd_{\ell}^2(\Pb_{1}, \Qb) \\
    &\leq 2\gamma \mathfrak{R} \|\nabla f_{\ell,\Pb_0,\Qb}\|_{L_2^d(\Pb_0)} \cdot \|\nabla f_{\ell_0, \Pb_0, \Qb}\|_{L_2^d(\Pb_0)} \\
    &\quad + C_k d^{\frac{1}{2}} p^{\frac{2}{5}} \gamma^{\frac{6}{5}} \calU^{\frac{4}{5}} \ell^{-1}\left((1+\ell^{-1})\right)^{\frac{1}{5}} \ell_0^{-\frac{2}{5}}
    + C_k\gamma^2 \ell^{-2} \ell_0^{-2} .
\end{align*}
\endgroup
Then, we use \Cref{lem:lengthscale_bound} which shows that $\left\| \nabla f_{\ell_0, \Pb_0,\Qb}(x) - \nabla f_{\ell, \Pb_0,\Qb}(x) \right\| \leq C_k \frac{\ell_0^2 - \ell^2}{\ell^3}$. So we obtain,
\begingroup\small
\begin{align*}
    &\quad \mmd_{\ell}^2 (\Pb_{\theta^\pgd_{1}}, \Qb) -\mmd_{\ell}^2(\Pb_{1}, \Qb) \leq 2\gamma \mathfrak{R} \|\nabla f_{\ell,\Pb_0,\Qb}\|_{L_2^d(\Pb_0)}^2\\
    & \quad + C_k\gamma \mathfrak{R} \|\nabla f_{\ell,\Pb_0,\Qb}\|_{L_2^d(\Pb_0)} \cdot \frac{\ell_0^2 - \ell^2}{\ell^3} \\
    &\quad + C_k d^{\frac{1}{2}} p^{\frac{2}{5}} \gamma^{\frac{6}{5}} \calU^{\frac{4}{5}} \ell^{-1}\left((1+\ell^{-1})\right)^{\frac{1}{5}} \ell_0^{-\frac{2}{5}}
    + C_k\gamma^2 \ell^{-2} \ell_0^{-2} \\
    &\leq 2\gamma \mathfrak{R} \|\nabla f_{\ell,\Pb_0,\Qb}\|_{L_2^d(\Pb_0)}^2
    + C_k\gamma \mathfrak{R} \frac{\ell_0^2 - \ell^2}{\ell^4} \\
    &\quad + C_k d^{\frac{1}{2}} p^{\frac{2}{5}} \gamma^{\frac{6}{5}} \calU^{\frac{4}{5}} \ell^{-1}\left((1+\ell^{-1})\right)^{\frac{1}{5}} \ell_0^{-\frac{2}{5}}
    + C_k\gamma^2 \ell^{-2} \ell_0^{-2}.
\end{align*}
\endgroup
This concludes the proof.
\end{proof}


\begin{lem}[Residual error] \label{lem:source_condition}
Let $\theta\in\Theta$ and $\ell>0$ be fixed.
Suppose \textcolor{black}{\Cref{ass:regularity_G}} holds.
Suppose there exist $u^\dagger\in\R^p$ and $r\in L_2^d(\rho)$ such that
\[
\nabla f_{\ell,\Pb,\Qb}(\Gmap_\theta(z))
=
\bJ_\theta \Gmap_\theta(z)u^\dagger + r(z),
\qquad \rho\text{-a.e. } z\in\calZ. 
\]
Then, for $\Delta\theta^\pgd$ defined in Eq.~\eqref{eq:parametric_mmd_update_new},
\[
\Delta \theta^\pgd =  \left\{ \E_{z\sim\rho}\left[ \bJ_\theta \Gmap_{\theta}(z)^\top \bJ_\theta \Gmap_{\theta}(z)\right] + \lambda\Id \right\}^{-1} \E_{z\sim\rho}\left[ \bJ_\theta \Gmap_{\theta}(z)^\top \nabla f_{\ell, \Pb, \Qb}(\Gmap_{\theta}(z)) \right],
\]
there holds
\[
    \E_{z\sim\rho} \left[ \left\|\nabla f_{\ell, \Pb, \Qb}(\Gmap_{\theta}(z)) - \bJ_\theta \Gmap_{\theta}(z) \Delta \theta^\pgd \right\|^2 \right]
    \leq
    \E_{z\sim\rho}\big[\|r(z)\|^2\big] + \lambda\|u^\dagger\|^2.
\]
\end{lem}

\begin{proof}
Define an operator $A:\R^p\to L_2^d(\rho)$ by $(Au)(z):=\bJ_\theta \Gmap_\theta(z)u$.
Its adjoint operator $A^\ast:L_2^d(\rho)\to\R^p$ is given by
$A^\ast h=\E_{z\sim\rho}\big[\bJ_\theta \Gmap_\theta(z)^\top h(z)\big]$.
Under \textcolor{black}{\Cref{ass:regularity_G}}, both $A$ and $A^\ast$ are well-defined and bounded.
Then, $A^\ast A = \E_{z\sim\rho}\left[ \bJ_\theta \Gmap_{\theta}(z)^\top \bJ_\theta \Gmap_{\theta}(z)\right]\in\R^{p\times p}$ is a matrix.

In the following proof, write $f:=f_{\ell,\Pb,\Qb}$ for brevity, and define
$g:=\nabla f\circ \Gmap_\theta \in L_2^d(\rho)$.
Then, $g=Au^\dagger+r$ by assumption in the lemma. Note that $\Delta\theta^\pgd=(A^\ast A+\lambda I)^{-1}A^\ast g$.
Now observe that $\Delta\theta^\pgd$ is the unique minimizer of the functional
\[
u\mapsto \|g-Au\|_{L_2^d(\rho)}^2+\lambda\|u\|^2.
\]
Therefore, by optimality of $\Delta\theta^\pgd$, taking $u=u^\dagger$ yields
\[
\|g-A\Delta\theta^\pgd\|_{L_2^d(\rho)}^2+\lambda\|\Delta\theta^\pgd\|^2
\le
\|g-Au^\dagger\|_{L_2^d(\rho)}^2+\lambda\|u^\dagger\|^2.
\]
Since $g-Au^\dagger=r$, it follows that
\[
\|g-A\Delta\theta^\pgd\|_{L_2^d(\rho)}^2+\lambda\|\Delta\theta^\pgd\|^2
\le
\|r\|_{L_2^d(\rho)}^2+\lambda\|u^\dagger\|^2.
\]
Dropping the nonnegative term $\lambda\|\Delta\theta^\pgd\|^2$, we obtain
\[
\|g-A\Delta\theta^\pgd\|_{L_2^d(\rho)}^2
\le
\|r\|_{L_2^d(\rho)}^2+\lambda\|u^\dagger\|^2.
\]
Rewriting this in the original notation gives
\[
\E_{z\sim\rho} \left[ \left\|\nabla f_{\ell, \Pb, \Qb}(\Gmap_{\theta}(z)) - \bJ_\theta \Gmap_{\theta}(z) \Delta \theta^\pgd \right\|^2 \right]
\le
\E_{z\sim\rho}\big[\|r(z)\|^2\big]+\lambda\|u^\dagger\|^2.
\]
This completes the proof.
\end{proof}
\begin{rem}[Preconditioning comes from regression]
    The proof of \Cref{lem:source_condition} is based on the observation that $\Delta \theta^\pgd$ is the solution to the regularized least-square regression problem: $$\Delta \theta^\pgd=\arg\min_{\Delta \theta} \E_{z\sim \rho} \left[\left\| \bJ_{\theta} \Gmap_{\theta}(z) \Delta \theta - \nabla f_{\ell, \Pb,\Qb}(\Gmap_{\theta}(z)) \right\|^2 \right] + \lambda \|\Delta\theta\|^2 . $$
    The target of interest to be upper bounded corresponds to the ``approximation error'' in least-square regression; see \cite{fischer2020sobolev,shen2025nonparametric}.
    The proof of \Cref{lem:source_condition} is therefore a standard technique.
\end{rem}

\begin{prop}[Closeness of $\Pb_{\theta_{1}}^\pgd$ and $\tildePtheta{1}$]
\label{prop:tilde_no_tilde}
Let $\gamma>0$, $\lambda>0$, evaluation lengthscale $\ell>0$, and update lengthscale $\ell_0>0$ be fixed.
Suppose \textcolor{black}{\Cref{ass:kernel,ass:regularity_G}} hold.
Given a fixed parameter $\theta_0\in\Theta$, let $\Pb_{\theta_{1}}^\pgd = (\Gmap_{\theta_0 - \gamma \Delta \theta_0^\pgd})_\# \rho$ be the next distribution under the population PGD scheme in Eq.~\eqref{eq:parametric_mmd_update_new}, and let $\tildePtheta{1} = (\Gmap_{\theta_0 - \gamma \tildetheta{0}})_\# \rho$ be the next distribution under the empirical PGD scheme in Eq.~\eqref{eq:parametric_mmd_update_new_tilde}. 
Then
\begin{align*}
    \E_m \left[
    \left| \mmd_{\ell}^2 (\tildePtheta{1}, \Qb) -\mmd_{\ell}^2(\Pb_{\theta_{1}}^\pgd, \Qb) \right|
    \right] \leq {C_k \gamma \lambda^{-1} p \ell^{-1}\ell_0^{-1} \sqrt{d}} \cdot \frac{1}{\sqrt{m}} . 
\end{align*}
Here, $C_k$ depends only on the kernel profile.
\end{prop}
\begin{proof}
For the Gaussian kernel and for the Mat\'ern kernel with smoothness $s>1$, we have $\|k_\ell(x, \cdot)-k_\ell(y, \cdot) \|_{\mathcal{H}_\ell} \leq C_k\ell^{-1} \|x-y\|$.
Hence, we have
\begingroup\small
\begin{align*}
    &\quad \left| \mmd_{\ell} (\tildePtheta{1}, \Qb) -\mmd_{\ell}(\Pb_{\theta_{1}}^\pgd, \Qb) \right| \leq \mmd_{\ell} (\tildePtheta{1}, \Pb_{\theta_{1}}^\pgd) \\
    &\leq {C_k\ell^{-1}}\left\| \Gmap_{\theta_0 - \gamma \Delta \theta_0^\pgd}(\cdot) - \Gmap_{\theta_0 - \gamma \tildetheta{0}}(\cdot) \right\|_{L_2^d(\rho)} \leq {C_k\sqrt{p} \ell^{-1} \gamma} \left\| \Delta \theta_0^\pgd - \tildetheta{0} \right\|.
\end{align*}
\endgroup
The last step holds by $\|\Gmap_\theta-\Gmap_{\theta^{\prime}}\|_{L_2^d(\rho)} \leq \sqrt{p}\|\theta-\theta^{\prime}\|$ from \textcolor{black}{\Cref{ass:regularity_G}}.
By definition, recall that
\begingroup\small
\begin{align*}
    \Delta \theta^\pgd
    &= \left\{ \E_{z\sim\rho}\left[
    \bJ_\theta \Gmap_{\theta_0}(z)^\top \bJ_\theta \Gmap_{\theta_0}(z)\right]
    + \lambda \Id \right\}^{-1} \\
    &\quad {}\times \E_{z\sim\rho}\left[
    \bJ_\theta \Gmap_{\theta_0}(z)^\top
    \nabla f_{\ell_0, \Pb_{\theta_0}, \Qb}(\Gmap_{\theta_0}(z)) \right], \\
    \tildetheta{0}
    &= \left\{ \E_{z\sim\rho}\left[
    \bJ_\theta \Gmap_{\theta_0}(z)^\top \bJ_\theta \Gmap_{\theta_0}(z)\right]
    + \lambda \Id \right\}^{-1} \\
    &\quad {}\times \E_{z\sim\rho}\left[
    \bJ_\theta \Gmap_{\theta_0}(z)^\top
    \nabla f_{\ell_0, \Pb_{\theta_0}, \widehat{\Qb}_m}
    (\Gmap_{\theta_0}(z)) \right] .
\end{align*}
\endgroup
So we have
\begingroup\small
\begin{align*}
    \left\| \Delta \theta^\pgd - \tildetheta{0} \right\|
    &\leq \lambda^{-1} \E_{z\sim\rho}\left[
    \begin{aligned}
    &\bJ_\theta \Gmap_{\theta_0}(z)^\top
    \Big( \nabla f_{\ell_0, \Pb_{\theta_0}, \Qb}(\Gmap_{\theta_0}(z)) \\
    &\qquad - \nabla f_{\ell_0, \Pb_{\theta_0}, \widehat{\Qb}_m}
    (\Gmap_{\theta_0}(z)) \Big)
    \end{aligned}
    \right] \\
    &\leq \lambda^{-1} \sqrt{p} \E_{z\sim\rho} \left[
    \left\|
    \begin{aligned}
    &\nabla f_{\ell_0, \Pb_{\theta_0}, \Qb}(\Gmap_{\theta_0}(z)) \\
    &\qquad - \nabla f_{\ell_0, \Pb_{\theta_0}, \widehat{\Qb}_m}
    (\Gmap_{\theta_0}(z))
    \end{aligned}
    \right\|\right] \\
    &\leq C_k\lambda^{-1} \sqrt{p} \ell_0^{-1} \sqrt{d} \mmd_{\ell_0}(\Qb, \widehat{\Qb}_m).
\end{align*}
\endgroup
The second last step holds by \textcolor{black}{\Cref{ass:regularity_G}}; and
the last step holds by the following: for each coordinate $j\in\{1, \ldots, d\}$ and any $x\in\R^d$,
\begingroup\small
\begin{align*}
    \left|\int \partial_{1,j} k_{\ell_0}(x,y)
    \,\dd(\widehat{\Qb}_m - \Qb)(y)\right|
    &\leq \left\|\partial_{1, j} k_{\ell_0}(x,\cdot)\right\|_{\calH_{\ell_0}}
    \, \mmd_{{\ell_0}}(\widehat{\Qb}_m,\Qb) \\
    &\leq C_k\ell_0^{-1} \mmd_{{\ell_0}}(\widehat{\Qb}_m,\Qb).
\end{align*}
\endgroup
Since the kernel is bounded by $1$, the standard concentration bound gives
$\E_m[\mmd_{\ell_0}(\widehat{\Qb}_m,\Qb)]\leq C_k m^{-1/2}$~\citep[Theorem 3.4]{muandet2017kernel}.
Taking expectation in the preceding display therefore gives
\begin{align*}
    &\quad \E_m\left[
    \left| \mmd_{\ell}^2 (\tildePtheta{1}, \Qb) -\mmd_{\ell}^2(\Pb_{\theta_{1}}^\pgd, \Qb) \right|
    \right] \\
    &\leq
    4\E_m\left[
    \left| \mmd_{\ell} (\tildePtheta{1}, \Qb) -\mmd_{\ell}(\Pb_{\theta_{1}}^\pgd, \Qb) \right|
    \right]
    \leq C_k \gamma \lambda^{-1} p \ell^{-1}\ell_0^{-1} \sqrt{d} \cdot \frac{1}{\sqrt{m}}.
\end{align*}
\end{proof}

\section{Connection with natural gradient}\label{sec:ngd}
Natural gradient descent is a popular optimization method in probabilistic machine learning as an alternative to
standard Euclidean gradient descent~\citep{amari1998natural,hoffman2013stochastic,pascanu2014revisiting,martens2020new}.
Let $\Pb_\theta$ be a probabilistic model with parameter $\theta\in\R^p$.
Let $\gamma>0$ be the step size,
and $g\in\R^p$ be the standard Euclidean gradient with respect to some objective, then the natural gradient is defined as
\begin{align}\label{eq:natural_gradient}
    h = \arg\min_{h\in\R^p} - h^\top g + \frac{1}{2 \gamma} \kl(\Pb_\theta\|\Pb_{\theta + h}) .
\end{align}
Under the well-known approximation of the KL divergence and the Fisher information matrix $\kl(\Pb_\theta\|\Pb_{\theta + h}) \approx h^\top F h$~\citep{pascanu2014revisiting,martens2020new}, the solution to Eq.~\eqref{eq:natural_gradient} can be solved in closed form: $h = \gamma F^{-1} g \in \R^p$, hence natural gradient.
Fast convergence has been proved for natural gradient method when the objective is a convex quadratic~\citep[Theorem 5]{martens2020new} and when training wide two-layer neural networks~\citep{zhang2019fast}.

A straight-forward MMD analogue of the natural gradient descent would be defined as the following:
\begin{align}\label{eq:mmd_minimizer}
    h = \arg\min_{h\in\R^p} - h^\top g + \frac{1}{2\gamma} \mmd_{\ell}^2(\Pb_\theta, \Pb_{\theta + h}) .
\end{align}
For the Gaussian kernel and for the Mat\'ern kernel with smoothness $s>1$, the map $h \mapsto \mmd_{\ell}^2(\Pb_\theta, \Pb_{\theta + h})$ admits a second-order expansion at $h=0$, and hence
\begin{align*}
    \mmd_{\ell}^2(\Pb_\theta, \Pb_{\theta + h}) &= 0 + h^\top \frac{\dd }{\dd h}\Big|_{h=0}  \mmd_{\ell}^2(\Pb_\theta, \Pb_{\theta + h}) \\
    &\qquad + \frac{1}{2} h^\top \frac{\dd^2 }{\dd h^2}\Big|_{h=0}  \mmd_{\ell}^2(\Pb_\theta, \Pb_{\theta + h})h + {o(\|h\|^2)}.
\end{align*}
In the following derivation, we focus on the case where $\Pb_\theta=(\Gmap_\theta)_{\#} \rho$ for some simple base probability measure $\rho$, to better align with the settings considered in the main text. We have
\begin{align*}
    \frac{\dd }{\dd h}\Big|_{h=0} \frac{1}{2} \mmd^2(\Pb_\theta, \Pb_{\theta + h}) = 0 .
\end{align*}
And,
\begingroup\small
\begin{align*}
    \frac{\dd^2 }{\dd h^2}\Big|_{h=0}
    \frac{1}{2} \mmd_{\ell}^2(\Pb_\theta, \Pb_{\theta + h})
    &= \iint \bJ_\theta \Gmap_{\theta}(z)^\top
    \nabla_1 \nabla_2 k(\Gmap_\theta(z), \Gmap_\theta(z')) \\
    &\quad {}\times \bJ_\theta \Gmap_{\theta}(z')^\top
    \dd \rho(z) \dd \rho(z') \\
    &= \E_{z,z'\sim\rho}\Big[
    \bJ_\theta \Gmap_{\theta}(z)^\top
    \nabla_1 \nabla_2 k_\ell(\Gmap_{\theta}(z), \, \Gmap_{\theta}(z')) \\
    &\hspace{90pt}{}\times \bJ_\theta \Gmap_{\theta}(z') \Big].
\end{align*}
\endgroup
Therefore, we have
\begingroup\small
\begin{align*}
    \mmd_{\ell}^2(\Pb_\theta, \Pb_{\theta + h})
    &= h^\top \left\{ \E_{z,z'\sim\rho}\Big[
    \bJ_\theta \Gmap_{\theta}(z)^\top
    \nabla_1 \nabla_2 k_\ell(\Gmap_{\theta}(z), \, \Gmap_{\theta}(z')) \right. \\
    &\hspace{95pt}\left. {}
    \times \bJ_\theta \Gmap_{\theta}(z') \Big] \right\} h
    + o(\|h\|^2).
\end{align*}
\endgroup
The resulting minimizer of Eq.~\eqref{eq:mmd_minimizer} thus admits a closed-form \citep{briol2019statistical}
\begin{align}\label{eq:mmd_geometry}
    h = \gamma \left\{ \E_{z,z'\sim\rho}\Big[ \bJ_\theta \Gmap_{\theta}(z)^\top \nabla_1 \nabla_2 k_\ell(\Gmap_{\theta}(z), \, \Gmap_{\theta}(z'))\bJ_\theta \Gmap_{\theta}(z') \Big] \right\}^{-1} g,
\end{align}
where $g$ is the standard Euclidean gradient.
The preconditioning matrix here is distinct from the preconditioning matrix in Eq.~\eqref{eq:parametric_mmd_update_new} in the main text, which we recall below:
\begin{align*}
    h = \gamma \left\{ \E_{z\sim\rho}\Big[ \bJ_\theta \Gmap_{\theta_t}(z)^\top \bJ_\theta \Gmap_{\theta_t}(z)\Big] \right\}^{-1} g .
\end{align*}
Other preconditioning methods directly on $\calP_2(\R^d)$ have also been proposed
~\citep{dong2023particle,pfau2025wasserstein,bonet2024mirror}, which are also relevant yet distinct from our approaches.


\section{Auxiliary Results}\label{sec:aux}
\begin{lem}\label{lem:lengthscale_lip_bound}
Let $\varphi:[0, \infty) \rightarrow \mathbb{R}$ be either the Gaussian profile
$
\varphi(r)=\exp \left(-r^2 / 2\right),
$
or the Matérn profile with smoothness parameter $s>1$,
$\varphi_s(r)=\frac{2^{1-s}}{\Gamma(s)}(\sqrt{2 s} r)^s K_s(\sqrt{2 s} r)$, 
where $K_s$ is the modified Bessel function of the second kind. For $a>0$, define $\psi_a(t):=\varphi\left(\frac{\|t\|}{\sqrt{a}}\right)$.
Then for any $a_1>a_2>0$,
\begin{align*}
    \sup _{t \in \mathbb{R}^d}
    \left|\psi_{a_1}(t)-\psi_{a_2}(t)\right|
    &\leq C_0 \frac{\left|a_1-a_2\right|}{a_2}, \\
    \sup _{t \in \mathbb{R}^d}
    \left\| \nabla \psi_{a_1}(t)- \nabla \psi_{a_2}(t) \right\|
    &\leq C_1 \frac{\left|a_1-a_2\right|}{a_2^{3 / 2}} .
\end{align*}
Here, $C_0,C_1$ are two constants depending only on the kernel.
\end{lem}
\begin{proof}
Let $\varrho:=\frac{\|t\|}{\sqrt{a}}$.
We first control the zeroth-order term. Differentiating $\psi_a(t)$ with respect to $a$ gives
$\partial_a \psi_a(t) = -\frac{1}{2 a} \varrho \varphi^{\prime}(\varrho)$.
Therefore, $\left|\partial_a \psi_a(t)\right| \leq \frac{1}{a}\left(\frac{1}{2} \sup _{\varrho \geq 0} \varrho\left|\varphi^{\prime}(\varrho)\right|\right)$.
Define
$$
C_0:=\frac{1}{2} \sup _{\varrho \geq 0} \varrho\left|\varphi^{\prime}(\varrho)\right| .
$$
For the Gaussian profile, $C_0<\infty$ follows directly from
$\varrho\left|\varphi^{\prime}(\varrho)\right|=\varrho^2 e^{-\varrho^2 / 2}$.
For the Matérn profile with $s>1$, the standard asymptotics of $K_s$ give, as $\varrho\downarrow 0$,
\[
\varphi_s(\varrho)=1-\frac{s}{2(s-1)}\varrho^2+o(\varrho^2),
\]
and hence $\varrho|\varphi_s'(\varrho)|=O(\varrho^2)$ near the origin. At infinity, the exponential decay of $K_s$ gives exponential decay of $\varphi_s'$ up to polynomial factors, so $\varrho|\varphi_s'(\varrho)|$ is bounded. Hence $C_0<\infty$ in both cases.
By the mean value theorem, for any $a_1>a_2>0$,
$
\left|\psi_{a_1}(t)-\psi_{a_2}(t)\right| \leq\left|a_1-a_2\right| \sup _{a \in\left[a_2, a_1\right]}\left|\partial_a \psi_a(t)\right|$.
Using the previous bound,
$$
\left|\psi_{a_1}(t)-\psi_{a_2}(t)\right| \leq C_0\left|a_1-a_2\right| \sup _{a \in\left[a_2, a_1\right]} \frac{1}{a}=C_0 \frac{\left|a_1-a_2\right|}{a_2} .
$$
Taking the supremum over $t \in \mathbb{R}^d$ proves the first claim.

We next control the gradient. For $t \neq 0$, let $e_t:=\frac{t}{\|t\|}$. Then $\nabla_t \psi_a(t)=a^{-1 / 2} \varphi^{\prime}(\varrho) e_t$.
For the Gaussian profile and the Matérn profile with $s>1$, the expansion above gives $\varphi'(0)=0$, so this gradient extends continuously to $t=0$ by setting it equal to zero.
Differentiating with respect to $a$, and noting that $e_t$ is independent of $a$, gives
$$
\partial_a (\nabla_t \psi_a(t)) =-\frac{1}{2 a^{3 / 2}} \varphi^{\prime}(\varrho) e_t+a^{-1 / 2} \varphi^{\prime \prime}(\varrho)\left(-\frac{\varrho}{2 a}\right) e_t .
$$
Hence, $\partial_a (\nabla_t \psi_a(t)) = -\frac{1}{2 a^{3 / 2}}\left[\varphi^{\prime}(\varrho)+\varrho \varphi^{\prime \prime}(\varrho)\right] e_t$. Therefore,
$\left\|\partial_a (\nabla_t \psi_a(t)) \right\| \leq \frac{1}{a^{3 / 2}} (\frac{1}{2} \sup _{\varrho \geq 0}|\varphi^{\prime}(\varrho)+\varrho \varphi^{\prime \prime}(\varrho)| )$.
Define
$$
C_1:=\frac{1}{2} \sup _{\varrho \geq 0}\left|\varphi^{\prime}(\varrho)+\varrho \varphi^{\prime \prime}(\varrho)\right| .
$$
For the Gaussian profile, $\varphi^{\prime}(\varrho)=-\varrho e^{-\varrho^2 / 2}, \quad \varphi^{\prime \prime}(\varrho)=\left(\varrho^2-1\right) e^{-\varrho^2 / 2}$,
so $\varphi^{\prime}(\varrho)+\varrho \varphi^{\prime \prime}(\varrho)=\left(\varrho^3-2 \varrho\right) e^{-\varrho^2 / 2}$,
which is bounded on $[0, \infty)$. Hence $C_1<\infty$ for the Gaussian kernel.
For the Matérn profile, the assumption $s>1$ ensures bounded second-order behavior at the origin. More precisely, the expansion above gives
$
\varphi_s^{\prime}(\varrho)+\varrho \varphi_s^{\prime \prime}(\varrho)=O(\varrho) \quad \text { as } \varrho \downarrow 0$.
At infinity $\varrho\approx\infty$, the exponential decay of $K_s$ implies exponential decay of $\varphi_s$ and its derivatives. Therefore, $\sup _{\varrho \geq 0}\left|\varphi_s^{\prime}(\varrho)+\varrho \varphi_s^{\prime \prime}(\varrho)\right|<\infty$.
Thus $C_1<\infty$ for the Matérn kernel with smoothness $s>1$.
Applying the mean-value theorem again concludes the proof.
\end{proof}

\begin{lem}[Strong convexity implies gradient dominance]\label{lem:strong_convex}
    Suppose that $\theta \mapsto \mmd^2(\Pb_\theta, \Qb)$ is $\alpha$ strongly convex. Suppose \textcolor{black}{\Cref{ass:regularity_G}} holds. Then,
    \[
    \mmd_\ell^2(\Pb_\theta, \Qb) - {\min_{\theta\in\Theta}} \mmd_{\ell}^2(\Pb_{\theta}, \Qb)
    \leq \frac{2p}{\alpha}  \E_{z \sim \rho}\left[\left\|\nabla f_{\ell, \mathbb{P}_\theta, \mathbb{Q}}\left(\Gmap_\theta(z)\right)\right\|^2\right].
    \]
\end{lem}
\begin{proof}
Since $\alpha$ strongly convex implies gradient dominance~\citep{karimi2016linear}, we have
\begingroup\small
\begin{align*}
    &\quad \mmd_\ell^2(\Pb_\theta, \Qb)
    - {\min_{\theta\in\Theta}} \mmd_{\ell}^2(\Pb_{\theta}, \Qb) \\
    &\leq \frac{1}{2\alpha}
    \|\nabla_\theta\mmd_\ell^2(\Pb_\theta, \Qb) \|^2 \\
    &= \frac{2}{\alpha} \left\| \E_{z\sim\rho}\left[
    \bJ_\theta \Gmap_{\theta}(z)^\top
    \nabla f_{\ell, \Pb_{\theta}, \Qb}(\Gmap_{\theta}(z)) \right]\right\|^2 \\
    &= \frac{2}{\alpha} \sum_{j=1}^p\left|\E_{z \sim \rho}\left[
    \partial_{\theta_j} \Gmap_\theta(z)^\top
    \nabla f_{\ell, \mathbb{P}_\theta, \mathbb{Q}}(\Gmap_\theta(z))
    \right]\right|^2 \\
    &\leq \frac{2}{\alpha} \sum_{j=1}^p
    \E_{z \sim \rho}\left[\left\|\partial_{\theta_j} \Gmap_\theta(z)\right\|^2\right] \\
    &\quad {}\times \E_{z \sim \rho}\left[\left\|
    \nabla f_{\ell, \mathbb{P}_\theta, \mathbb{Q}}(\Gmap_\theta(z))
    \right\|^2\right] \\
    &\leq \frac{2p}{\alpha} \E_{z \sim \rho}\left[\left\|
    \nabla f_{\ell, \mathbb{P}_\theta, \mathbb{Q}}(\Gmap_\theta(z))
    \right\|^2\right].
\end{align*}
\endgroup
The last inequality holds by \textcolor{black}{\Cref{ass:regularity_G}}.
\end{proof}

\begin{figure}[t]
    \centering
    \includegraphics[width=0.3\linewidth]{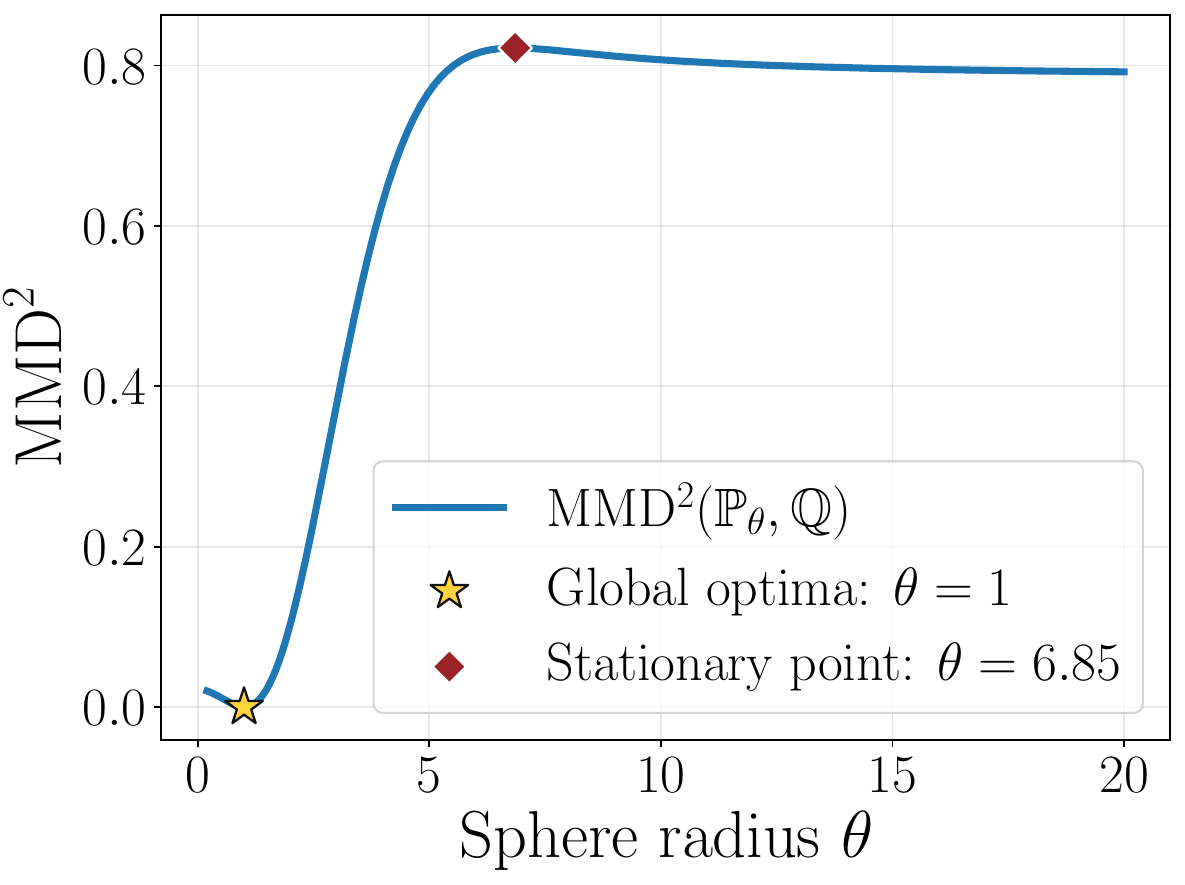}
    \caption[Stationary points on the sphere]{The example of $\Qb= \operatorname{Unif}(\mathbb{S}^{d-1})$ has stationary points that are not global minima.}
    \label{fig:sphere}
\end{figure}

\begin{lem}[Local minima of parametric and nonparametric descent]
\label{lem:residual_excludes_spurious_minima}
Let $\theta^\star$ be a stationary point of the parametric objective $\theta\mapsto \mmd_\ell^2(\Pb_\theta, \Qb)$. 
Then, for \textcolor{black}{\Cref{ass:residual}} to hold, $\Pb_{\theta^\star}$ must also be a stationary point of the objective $\Pb\mapsto \mmd_\ell^2(\Pb, \Qb)$, with respect to the Wasserstein-2 gradient.
\end{lem}
\begin{proof}
Since $\theta^\star$ is a stationary point, then, for any direction $u\in\R^p$, the first order condition gives
\begin{equation*}
    0 = \mathbb{E}_{z\sim\rho}\left[\nabla f_{\ell,\Pb_{\theta^\star},\Qb}(\Gmap_{\theta^\star}(z))^\top \bJ_\theta \Gmap_{\theta^{\star}}(z) u \right] .
\end{equation*}
In other words, at such a stationary point, the function $z\mapsto \nabla f_{\ell,\Pb_{\theta^\star},\Qb} (\Gmap_{\theta^\star}(z))$ is orthogonal in $L_2(\rho)$ to every tangent direction of the form $z\mapsto \bJ_\theta \Gmap_{\theta^\star}(z) u$ with $u\in\R^p$.
Write $a_u(z):=\bJ_\theta\Gmap_{\theta^\star}(z)u$ and $b(z):=\nabla f_{\ell,\Pb_{\theta^\star},\Qb}(\Gmap_{\theta^\star}(z))$.
The least-squares residual from \textcolor{black}{\Cref{ass:residual}} at $\theta^\star$ is
\begingroup\small
\begin{align*}
    \mathbb{E}_{z\sim\rho}\left[\left\|a_u(z)-b(z)\right\|^2\right]
    &= \mathbb{E}_{z\sim\rho}\left[\left\|a_u(z)\right\|^2\right]
    + \mathbb{E}_{z\sim\rho}\left[\left\|b(z)\right\|^2\right] \\
    &\quad - 2\mathbb{E}_{z\sim\rho}\left[b(z)^\top a_u(z)\right] \\
    &= \mathbb{E}_{z\sim\rho}\left[\left\|a_u(z)\right\|^2\right]
    + \mathbb{E}_{z\sim\rho}\left[\left\|b(z)\right\|^2\right].
\end{align*}
\endgroup
where the cross term vanishes due to the first-order condition from above. Thus the minimiser of the residual is obtained by any direction $u$ which satisfies $\mathbb{E}_{\rho}\left[\left\| \bJ_\theta \Gmap_{\theta^\star}(z) u\right\|^2 \right]=0$, in which case
\begingroup\small
\begin{align*}
    \min_{u\in\R^p}\mathbb{E}_{z\sim\rho}
    \left[\left\|a_u(z)-b(z)\right\|^2\right]
    &= \left\|\nabla f_{\ell,\Pb_{\theta^\star},\Qb}
    \right\|_{L_2(\Pb_{\theta^\star})}^2
\end{align*}
\endgroup
For \textcolor{black}{\Cref{ass:residual}} to hold, we would then require
\begin{align*}
    \left\|  \nabla f_{\ell,\Pb_{\theta^\star},\Qb} \right\|_{L_2(\Pb_{\theta^\star})}^2   \leq \mathfrak{R}^2 \left\|  \nabla f_{\ell,\Pb_{\theta^\star},\Qb} \right\|_{L_2(\Pb_{\theta^\star})}^2
\end{align*}
for some fixed $0<\mathfrak{R}<1$. This can only occur when $\left\|  \nabla f_{\ell,\Pb_{\theta^\star},\Qb} \right\|_{L_2(\Pb_{\theta^\star})}=0$, meaning that $\Pb_{\theta^\star}$ is a stationary point, in the sense of Wasserstein gradient, of the objective $\Pb\mapsto \mmd^2_\ell(\Pb,\Qb)$~\citep{chen2025stationary}.
\end{proof}

\begin{lem}[Example for \textcolor{black}{\Cref{ass:residual}}]\label{lem:example}
    Let $\Qb$ be a radial distribution on $\R^d$. Let $\Pb_\theta = (\Gmap_\theta)_\# \operatorname{Unif} (\mathbb{S}^{d-1})$ be a a parametric family with a pushforward map $\Gmap_\theta: z\mapsto \theta z$ and $\theta\in\R$. 
    Let $k_\ell$ be a Gaussian kernel with lengthscale $\ell>0$. Then, for any fixed $\theta\in\R$, $\nabla f_{\ell, \Pb_\theta, \Qb}(\Gmap_\theta(\cdot )) \in \mathrm{span}\{ \bJ_\theta \Gmap_\theta(\cdot) u : u\in\R\}$. Moreover, the representing coefficient is uniformly bounded over $\theta\in\R$. 
\end{lem}
\begin{proof}
    Since the Gaussian kernel is radial, $k_\ell(x,y)=\exp(-(2\ell^2)^{-1} \|x-y\|^2)$ with lengthscale $\ell>0$, the witness function, $f_{\ell, \Pb_{\theta}, \Qb}(x)=\int k_{\ell}(x, y) \; \dd ( \Pb_{\theta}- \Qb)(y)$ is a convolution of two radial function, and hence also radial. Therefore, there exists a scalar function $\varphi_\theta:[0, \infty) \rightarrow \mathbb{R}$ such that $f_{\ell, \Pb_\theta, \Qb}(x) = \varphi_\theta(\|x\|)$. 
    For $\theta\neq0$ and $z\in\mathbb{S}^{d-1}$,
    \[
        \nabla f_{\ell, \Pb_\theta, \Qb}(\Gmap_\theta(z))
        = \varphi_\theta^{\prime}(|\theta|)
        \frac{\theta z}{|\theta|}
        = \operatorname{sign}(\theta)\varphi_\theta^{\prime}(|\theta|) z.
    \]
    When $\theta=0$, $f_{\ell,\Pb_0,\Qb}$ has zero gradient at the origin, so the conclusion holds trivially by taking the representing coefficient to be zero. 
    On the other hand, we have $\bJ_\theta \Gmap_\theta(z) = z$. 
    Hence, there exists a $u=\operatorname{sign}(\theta)\varphi_\theta^{\prime}(|\theta|)\in\R$, 
    depending only on $\theta$, such that $\nabla f_{\ell, \Pb_\theta, \Qb}(\Gmap_\theta(z)) = \bJ_\theta \Gmap_\theta(z) u$ holds for any $z\in\mathbb{S}^{d-1}$.
    Finally, the representing coefficient satisfies
    $|u|\leq \sup_x \|\nabla f_{\ell,\Pb_\theta,\Qb}(x)\| \leq 2 \ell^{-1}$, since the Gaussian-kernel gradient is uniformly bounded by $\ell^{-1}$. 
    The proof is thus concluded.
\end{proof}

\begin{rem}
    In \Cref{example:ass}, even when in the simplest case where $\Qb=\operatorname{Unif}(\mathbb{S}^{d-1})$ and $\Pb_\theta = (\Gmap_\theta)_\# \operatorname{Unif} (\mathbb{S}^{d-1})$ with $\theta$ representing the radius, the objective $\theta\mapsto \mmd^2(\Pb_\theta,\Qb)$ still has a local minima that are not the global minima for a fixed lengthscale $\ell>0$; see \Cref{fig:sphere}. This further illustrates the importance of adaptive kernel lengthscale as done in our PGD scheme.
\end{rem}

\begin{lem}[Continuity in lengthscale]\label{lem:lengthscale_bound}
Suppose \textcolor{black}{\Cref{ass:kernel}} holds. Let lengthscale $\ell > 0$.
Let $f_{\ell, \Pb,\Qb}(\cdot) = \int k_\ell(x, \cdot) \dd (\Pb-\Qb)(x)$ be the associated witness function.
Then, for $\ell_1 > \ell_2 > 0$, we have
\begin{align*}
    \left\| \nabla f_{\ell_1, \Pb,\Qb}(x) - \nabla f_{\ell_2, \Pb,\Qb}(x) \right\| \leq 2 C_k \frac{\ell_1^2 - \ell_2^2}{\ell_2^3} .
\end{align*}
Here, $C_k$ is a constant depending only on the kernel but not on the lengthscale. 
\end{lem}

\begin{proof}
From \Cref{lem:lengthscale_lip_bound}, we have for any $x,y\in\R^d$,
\begin{align*}
    \left\| \nabla_2 k_{\ell_1}(x,y) - \nabla_2 k_{\ell_2}(x,y) \right\| \leq C_k \frac{\ell_1^2 - \ell_2^2}{\ell_2^3} ,
\end{align*}
where $C_k$ is a constant depending only on the kernel but not on the lengthscale.
So we have
\begingroup\small
\begin{align*}
    &\quad \left\| \nabla f_{\ell_1, \Pb,\Qb}(y)
    - \nabla f_{\ell_2, \Pb,\Qb}(y) \right\| \\
    &= \left\| \int \left(\nabla_2 k_{\ell_1}(x,y)
    - \nabla_2 k_{\ell_2}(x,y)\right) \dd (\Pb - \Qb)(x) \right\| \\
    &\leq \int \left\| \nabla_2 k_{\ell_1}(x,y)
    - \nabla_2 k_{\ell_2}(x,y) \right\| \dd |\Pb - \Qb|(x) \\
    &\leq C_k \frac{\ell_1^2 - \ell_2^2}{\ell_2^3}
    \int \dd |\Pb - \Qb|(x)
    \leq 2C_k \frac{\ell_1^2 - \ell_2^2}{\ell_2^3} .
\end{align*}
\endgroup
In the above derivations, $|\Pb-\Qb|$ denote the total variation measure associated with the signed measure $\Pb-\Qb$. The last factor of $2$ appears because the total variation norm is always bounded by $2$. 
\end{proof}


\begin{lem}[One-step descent inequality for MMD]\label{lem:descent}
Let kernel $k_\ell$ satisfy \textcolor{black}{\Cref{ass:kernel}} with lengthscale $\ell>0$.
Let $G:\R^d\to\R^d$ and assume $G\in L_2^d(\Pb_0)$. 
Let $\Pb_0\in\calP_2(\R^d)$ and let the next iterate be $\Pb_1 = (\Id - \gamma G)_\# \Pb_0$.
Let $\ell >0$.
Then, we have
\begingroup\small
\begin{align*}
    \mmd_\ell^2(\Pb_1 , \Qb) \leq \mmd_\ell^2(\Pb_0, \Qb) - 2\gamma \int \nabla f_{\ell, \Pb_0, \Qb}(x)^\top G(x) \dd \Pb_0(x) + \frac{C_k\gamma^2}{\ell^2} \|G\|^2_{L_2^d(\Pb_0)} .
\end{align*} 
\endgroup
Here, $C_k$ depends only on the kernel profile and not on the lengthscale. 
\end{lem}
\begin{proof}
    Define, for \(t\in[0,1]\),
    \[
        G_t(x):=x-t\gamma G(x), \qquad \Pb_t:=(G_t)_\#\Pb_0,
    \]
    so that \(\Pb_0={T_0}_{\#} \Pb_0\) and \(\Pb_1=(\mathrm{Id}-\gamma G)_\#\Pb_0\).
    Moreover, let \(X_0\sim \Pb_0\), and set
    \[
        X_t:=G_t(X_0)=X_0-t\gamma G(X_0),
    \]
    which gives us \(X_t\sim \Pb_t\) and \(\dot X_t=-\gamma G(X_0)\).
    Using the definition,
\begingroup\small
    \begin{align*}
        \mmd_\ell^2(\Pb_t, \mathbb{Q})
        &= \E\left[k_\ell(X_t,X_t')\right]
        -2\,\E\left[k_\ell(X_t,Z)\right]
        +\E\left[k_\ell(Z,Z')\right], \\
        &\hspace{45pt} X_t, X_t' \sim \Pb_t,
        \qquad Z, Z' \sim \Qb.
    \end{align*}
\endgroup
    We can differentiate with respect to $t$ and use that $\dot X_t = -\gamma G(X_0)
    \text{ and } \dot X_t' = -\gamma G(X_0')$ to obtain
\begingroup\small
    \begin{align*}
        \frac{\dd}{\dd t} \mmd_\ell^2(\Pb_t, \mathbb{Q}) =-2\gamma\,\E_{X_0\sim \Pb_0} \left[
        \nabla f_{\ell,\Pb_t,\Qb}(X_0-t\gamma G(X_0) )^\top G(X_0)\right] .
    \end{align*}
\endgroup
    Then, by the fundamental theorem of calculus~\citep{rudin1976principles}, we have
    \begingroup\small
    \begin{align*}
    &\quad \mmd_\ell^2(\Pb_1,\Qb)-\mmd_\ell^2(\Pb_0,\Qb) \\
    &=-2\gamma\int_0^1 \E_{X_0\sim \Pb_0}\left[
    \nabla f_{\ell,\Pb_t,\Qb}(X_0-t\gamma G(X_0))^\top G(X_0)
    \right] \dd t.
    \end{align*}
    \endgroup
    By writing \(\nabla f_{\ell,\Pb_t,\Qb}(X_0-t\gamma G(X_0))
    =
    \nabla f_{\ell,\Pb_0,\Qb}(X_0)
    +
    (
    \nabla f_{\ell,\Pb_t,\Qb}(X_0-t\gamma G(X_0))
    -
    \nabla f_{\ell,\Pb_0,\Qb}(X_0)
    ),\) the previous expression gives
    \begingroup\small
    \begin{align}
    &\quad \mmd_\ell^2(\Pb_1,\Qb) =\mmd_\ell^2(\Pb_0,\Qb)
    -2\gamma\int \nabla f_{\ell,\Pb_0,\Qb}(x)^\top G(x)\,\dd \Pb_0(x)
    \nonumber\\
    &-2\gamma\int_0^1
    \E_{X_0\sim \Pb_0} \left[
    \Bigl(
    \nabla f_{\ell,\Pb_t,\Qb}(X_0-t\gamma G(X_0))
    -
    \nabla f_{\ell,\Pb_0,\Qb}(X_0)
    \Bigr)^\top G(X_0)
    \right] \dd t.
    \label{eq:quad}
    \end{align} 
    \endgroup
    For the last term of the right hand side, by adding and subtracting \(
    \nabla f_{\ell,\Pb_t,\Qb}(X_0)\) inside \((\nabla f_{\ell,\Pb_t,\Qb}(X_0-t\gamma G(X_0))-\nabla f_{\ell,\Pb_0,\Qb}(X_0)),\)
    \begin{align}
    &\quad -2\gamma\int_0^1
    \E_{X_0\sim \Pb_0}\left[
    \Bigl(
    \nabla f_{\ell,\Pb_t,\Qb}(X_0-t\gamma G(X_0))
    -
    \nabla f_{\ell,\Pb_0,\Qb}(X_0)
    \Bigr)^\top G(X_0)
    \right] \dd t \nonumber \\
    &=
    -2\gamma \int_0^1
    \E_{X_0\sim \Pb_0} \left[
    \Bigl(\nabla f_{\ell,\Pb_t,\Qb}(X_0-t\gamma G(X_0))-\nabla f_{\ell,\Pb_t,\Qb}(X_0)\Bigr)^\top G(X_0)
    \right] \dd t
    \label{eq:first_term_lemma} \\
    &\qquad -2\gamma \int_0^1
    \E_{X_0\sim \Pb_0} \left[
    \Bigl(\nabla f_{\ell,\Pb_t,\Qb}(X_0)-\nabla f_{\ell,\Pb_0,\Qb}(X_0)\Bigr)^\top G(X_0)
    \right] \dd t. \label{eq:second_term_lemma}
    \end{align}
    Since \(x\mapsto \nabla f_{\ell,\Pb_t,\Qb}(x)\) is \((C_k/\ell^2)\)-Lipschitz for the Gaussian kernel and for the Mat\'ern kernel with smoothness $s>1$,
    \begin{align*}
    \bigl\|
    \nabla f_{\ell,\Pb_t,\Qb}(X_t)
    -
    \nabla f_{\ell,\Pb_t,\Qb}(X_0)
    \bigr\|
    \leq
    \frac{C_k}{\ell^2} \|X_t-X_0\|
    =
    \frac{C_k t\gamma}{\ell^2}\|G(X_0)\|.
    \end{align*}
    $C_k$ depends only on the kernel profile and not on the lengthscale. 
    Therefore,
    \begin{align*}
    \eqref{eq:first_term_lemma}
    &\le 2\gamma \int_0^1 \E_{X_0\sim \Pb_0} \Big[
    \|\nabla f_{\ell,\Pb_t,\Qb}(X_t)
    -\nabla f_{\ell,\Pb_t,\Qb}(X_0)\| \\
    &\hspace{105pt}{}\times \|G(X_0)\| \Big]dt \\
    &\le 2\gamma \int_0^1 \frac{C_k\gamma t}{\ell^2} \,
    \E_{X_0\sim \Pb_0}\left[\|G(X_0)\|^2\right]dt \\
    &= \frac{2 C_k\gamma^2}{\ell^2} \,\|G\|_{L^2(\Pb_0)}^2.
    \end{align*}
    Moreover, using the coupling $Y_t=Y_0-\gamma tG(Y_0)$ with \(Y_0\sim \Pb_0\) and the \((C_k/\ell^2)\)-Lipschitzness of the derivative of $k_\ell$, for every \(x\in\mathbb{R}^d\), we have
    \begin{align*}
    \bigl\|
    \nabla f_{\ell,\Pb_t,\Qb}(x)
    -
    \nabla f_{\ell,\Pb_0,\Qb}(x)
    \bigr\|
    &= \E_{Y_0,Y_t} \bigl[
    \|\nabla_1 k_\ell(x, Y_t)-\nabla_1 k_\ell(x, Y_0)\|\bigr] \\
    &\leq \frac{C_k t\gamma}{\ell^2} \,
    \E\bigl[\|G(Y_0)\|\bigr].
    \end{align*}
    Hence, by Cauchy-Schwarz, integrating over \(t\in[0,1]\),
    \begin{align*}
        \eqref{eq:second_term_lemma}
        &\le 2\gamma\int_0^1 \frac{C_k t\gamma}{\ell^2} \,
        \E\bigl[\|G(X_0)\|^2\bigr] \dd t \\
        &= \frac{C_k\gamma^2}{\ell^2} \|G\|_{L_2^d(\Pb_0)}^2.
    \end{align*}
    Combining the last two upper bounds on Eq.~\eqref{eq:first_term_lemma} and Eq.~\eqref{eq:second_term_lemma}, along with inequalities with \eqref{eq:quad}, we obtain
\begingroup\small
\begin{align*}
    \mmd_\ell^2(\Pb_1 , \Qb)
    &\leq \mmd_\ell^2(\Pb_0, \Qb)
    - 2\gamma \int \nabla f_{\ell, \Pb_0, \Qb}(x)^\top G(x) \dd \Pb_0(x) \\
    &\quad + \frac{C_k\gamma^2}{\ell^2} \|G\|^2_{L_2^d(\Pb_0)} .
\end{align*}
\endgroup
\end{proof}

\begin{lem}[Bandwidth adjustment via convolution with noise]\label{lem:convolution}
Let \(k_\ell\) be either the Gaussian kernel or the Mat\'ern kernel with smoothness $s$ stated in \textcolor{black}{\Cref{ass:kernel}}.
Let $f_{\ell, \Pb,\Qb}(\cdot) = \int k_\ell(x, \cdot) \dd (\Pb-\Qb)(x)$ be the associated witness function.
Let $\ell' > \ell >0$.
There exists a probability measure $\nu$ on $\R^d$ such that, for any $x\in\R^d$,
\begin{align}
    \E_{U\sim\nu}\big[\nabla f_{\ell, \Pb, \Qb}(x+ U) \big]
    =
    \left( \ell \ell'^{-1}
    \right)^{d}
    \nabla  f_{\ell', \Pb, \Qb}(x), \quad \E_{U\sim\nu}[\|U\|]\leq C_{d,s} \sqrt{{\ell^\prime}^2-{\ell}^2}.
\end{align}
Here, $C_{d,s}$ is a constant depending only on the dimension $d$ and the smoothness parameter $s$.
\end{lem}
\begin{proof}
Notice that
\begingroup\small
\begin{align*}
    \E_{U\sim\nu} \bigl[\nabla f_{\ell,\Pb,\Qb}(x+ U) \bigr]
    &= \int_{\mathbb R^d} \int_{\mathbb R^d}\nabla_1 k_\ell(x+u,y)\,\dd (\Pb-\Qb)(y) \, \dd \nu(u)
    \\
    &= \int_{\mathbb R^d}\nabla_x \left(\int_{\mathbb R^d} k_\ell(x+u,y)\,\, \dd \nu(u) \right) \dd(\Pb-\Qb)(y),
\end{align*}
where the second equality is from using Fubini's theorem~\citep{rudin1976principles} given that $k_\ell$ is a Gaussian or a Mat\'ern kernel hence its gradient is uniformly bounded. Now, for fixed $y\in\mathbb R^d$, based on \Cref{lem:noise_pert}, the inner integral becomes
\begin{align*}
     \int_{\mathbb R^d} k_\ell(x+u,y)\,\dd \nu(u)
     &= \left( \ell \ell'^{-1} \right)^d k_{\ell'}(x,y) .
\end{align*}
And, we obtain
\begin{align*}
    \E_{U\sim\nu} \bigl[\nabla f_{\ell,\Pb,\Qb}(x+ U) \bigr]
    &=\int_{\mathbb R^d}
    \left( \ell \ell'^{-1} \right)^d
    \nabla_1 k_{\ell'} (x,y) \,
    \dd (\Pb-\Qb)(y) \\
    &=\left( \ell \ell'^{-1} \right)^d \nabla f_{\ell',\,\Pb,\Qb}(x).
\end{align*}
\endgroup
The first moment bound on $U\sim\nu$ is proved in \Cref{lem:noise_pert}.
\end{proof}

\noindent
\textbf{Notations:} We denote $\calF[f]:\R^d\to\R$ as the Fourier transform of a function $f:\R^d\to\R$.

\begin{lem}[Noise perturbation]\label{lem:noise_pert}
    Let \(k_\ell\) be either the Gaussian kernel or the Mat\'ern kernel with smoothness $s$ stated in \textcolor{black}{\Cref{ass:kernel}}. Then, for any \(\ell'>\ell>0\), there exists a probability measure \(\nu\) on \(\mathbb{R}^d\)  such that $\E_{U\sim \nu}[k_\ell(x+U, y)] = \left( \ell \ell'^{-1} \right)^d k_{\ell'}(x,y)$ holds for any $x,y\in\R^d$.
    Moreover, $\E_{U\sim\nu}[\|U\|]\leq C_{d,s} \sqrt{{\ell^\prime}^2-{\ell}^2}$ for some constant $C_{d,s}$ depending only on the dimension $d$ and the smoothness parameter $s$.
\end{lem}
\begin{proof}
    Write \(k_\ell(x,y)=\psi_\ell(x-y)\), and denote \(S_\ell=\calF[\psi_\ell]:\R^d\to\R\) as the Fourier transform of $\psi_\ell$.
    It is enough to find a probability measure \(\nu\) with finite first moment, symmetric with respect to $0$, and its characteristic function satisfies
    \begin{align}\label{eq:goal}
        \widehat \nu(\omega) := \int e^{-i \omega^{\top} u} \, \dd \nu(u) = \left(\ell \ell'^{-1}\right)^d \frac{S_{\ell'}(\omega)}{S_\ell(\omega)}.
    \end{align}
    Indeed, based on the convolution theorem for Fourier transforms (see \cite{katznelson2004introduction}, Chapter VI, Section 2, pp. 144--145), this gives
    \begin{align*}
        \mathcal{F}\left[ \E_{U\sim\nu}\psi_\ell(\cdot+U) \right](\omega) &= \calF\left[ \int_{\mathbb{R}^d} \psi_{\ell}(\cdot + u) \dd \nu(u) \right] = \calF\left[ \int_{\mathbb{R}^d} \psi_{\ell}(\cdot - u) \dd \nu(u) \right] \\
        &= S_\ell(\omega) \cdot \widehat \nu(\omega) = \left(\ell \ell'^{-1}\right)^d S_{\ell'}(\omega),
    \end{align*}
    and hence we obtain the desired result:
    $\E_{U\sim \nu}\big[k_\ell(x+U,y)\big] = \left(\ell \ell'^{-1}\right)^d k_{\ell'}(x,y).$

    Next, we are going to prove Eq.~\eqref{eq:goal} for the Gaussian and Mat\'{e}rn kernels. For the Gaussian kernel, we have
    $S_\ell(\omega) = C_d \ell^d \exp(-\frac{\ell^2\|\omega\|^2}{2})$.
    Therefore \[ \left(\ell \ell'^{-1}\right)^d \frac{S_{\ell'}(\omega)}{S_\ell(\omega)} = \exp\left( -\frac{(\ell'^2-\ell^2)\|\omega\|^2}{2} \right), \] which is the characteristic function of a normal distribution $\nu=\calN(0,(\ell'^2-\ell^2)\Id)$. It is symmetric with respect to $0$ and has bounded first moment $\E_{U\sim\nu}[\|U\|]\leq \sqrt{\E_{U\sim\nu}[\|U\|^2]} = \sqrt{d\left(\ell'^2-\ell^2\right)}$.
    Thus the claim follows in the Gaussian case.

    Next, for the Mat\'ern kernel with smoothness parameter $s$, its Fourier transform is of the following form~\citep{bach2024learning}:
    \[
    S_\ell(\omega) = C_{s,d}\ell^d \left( 1+\frac{\ell^2\|\omega\|^2}{2s} \right)^{-(s+d/2)} . \] 
    Hence \[ \left(\ell \ell'^{-1}\right)^d \frac{S_{\ell'}(\omega)}{S_\ell(\omega)} = \left( \frac{1+\frac{\ell^2\|\omega\|^2}{2s}} {1+\frac{\ell'^2\|\omega\|^2}{2s}} \right)^{s+d/2}. \] Since \(\ell'>\ell\), we claim that the function
\begin{align}\label{eq:g_function}
    g:r\mapsto 
    \left(
    \frac{1+\frac{\ell^2 r}{2s}}
         {1+\frac{\ell'^2 r}{2s}}
    \right)^{s+d/2},
    \qquad r\geq 0,
\end{align}
is completely monotone\footnote{
A function \(f:(0,\infty)\to\mathbb{R}\) is completely monotone if 
\(f\in C^\infty(0,\infty)\) and 
\((-1)^n f^{(n)}(\lambda)\geq 0\) for all 
\(n\in\mathbb{N}\cup\{0\}\) and \(\lambda>0\); see 
\cite[Definition~1.3]{schilling2012bernstein}.
}
and satisfies \(g(0)=1\). 
To prove the claim, set $h(r):=-\log g(r)$. Then, $h(r) = \left(s+\frac d2\right) \left[
\log\left(1+\frac{\ell'^2 r}{2s}\right) - \log\left(1+\frac{\ell^2 r}{2s}\right)
\right]$, and hence $h'(r) = \left(s+\frac d2\right)
\frac{(\ell'^2-\ell^2)/(2s)}
{\left(1+\frac{\ell^2 r}{2s}\right)
\left(1+\frac{\ell'^2 r}{2s}\right)}$.
 The prefactor is positive because \(\ell'>\ell\). Moreover, for every \(a>0\),
the map \(r\mapsto (1+ar)^{-1}\) is completely monotone, since
\[
    (-1)^n \frac{\dd^n}{\dd r^n}(1+ar)^{-1}
    = n!a^n(1+ar)^{-n-1} \geq 0 , \quad \forall n\in\N. 
\]
Therefore both factors
\[
    r\mapsto \left(1+\frac{\ell^2 r}{2s}\right)^{-1},
    \qquad
    r\mapsto \left(1+\frac{\ell'^2 r}{2s}\right)^{-1}
\]
are completely monotone. Since products of completely monotone functions are
completely monotone~\citep[Corollary~1.6]{schilling2012bernstein}, it follows
that \(h'\) is completely monotone.

Thus \(h\) is a Bernstein function\footnote{
Equivalently, a nonnegative function \(h\in C^\infty(0,\infty)\) is a Bernstein
function if \(h'\) is completely monotone; see 
\cite[Definition~3.1]{schilling2012bernstein}.
}.
Indeed, \(h(0)=0\), \(h'\geq 0\), and \(h'\) is completely monotone. By the
standard Bernstein-function characterization, if \(h\) is Bernstein, then
\(\exp(-u h)\) is completely monotone for every \(u>0\)
\citep[Theorem~3.7]{schilling2012bernstein}. Taking \(u=1\), we conclude that
\[
    g(r)=\exp(-h(r))
\]
is completely monotone. Also \(g(0)=1\) follows immediately from
Eq.~\eqref{eq:g_function}. 
By Schoenberg's theorem~\citep[Theorem~3]{schoenberg1938metric}, every
completely monotone function \(g\) with \(g(0)=1\) gives rise to a
positive-definite radial function
$\omega\mapsto g(\|\omega\|^2)$
on \(\mathbb{R}^d\). Consequently,
\[
    \hat{\nu}(\omega)
    :=
    g(\|\omega\|^2)
    =
    \left(
    \frac{1+\frac{\ell^2\|\omega\|^2}{2s}}
         {1+\frac{\ell'^2\|\omega\|^2}{2s}}
    \right)^{s+d/2}
\]
is positive definite on \(\mathbb{R}^d\). 
And consequently, via the Bochner's theorem~\citep{bochner2005harmonic}, it is a valid characteristic function of a probability measure on \(\mathbb{R}^d\).
    Moreover, notice that the first order derivative at the origin of the mapping in Eq.~\eqref{eq:g_function} equals $\left(s+\frac{d}{2}\right) \frac{\ell^2-{\ell^\prime}^2}{2 s}$ which is finite.
    And
    \begin{align*}
        \E_{U\sim\nu}[\|U\|^2]= -\operatorname{Tr} \left[ \frac{\partial^2 }{\partial^2 \omega} \hat{\nu}(\omega) \Big.\mid_{\omega=0} \right] = -2 g^\prime(0) d = d \left(s+\frac{d}{2}\right) \frac{{\ell^\prime}^2-{\ell}^2}{s} < \infty.
    \end{align*}
    so the associated probability measure $\nu$ has finite second moment. Hence its first moment satisfies $\E_{U\sim\nu}[\|U\|]\leq \sqrt{d \left(1+\frac{d}{2s}\right)} \cdot \sqrt{{\ell^\prime}^2-\ell^2}$. This concludes the proof of the Mat\'ern case.
\end{proof}

\begin{rem}[Extension to more general kernels]\label{rem:extension_general_kernel}
    \Cref{lem:convolution} shows that, for the Gaussian or Mat\'ern kernel, MMD flow with noise injection is equivalent to MMD flow with an increased kernel lengthscale. This observation was first made in \cite{glaser2020regularizing}.
    The above equivalence can be generalized to other translational invariant kernels $k_\ell(x,y) = \psi_\ell(x-y)$ and mean-centered symmetric probability measure $\nu$, as long as there exists a $\ell'>0$ such that $\mathcal{F}[\psi_\ell] \times \mathcal{F}[p] = \mathcal{F}[\psi_{\ell'}]$.
    \cite{nishiyama2016characteristic} provides a more general family of kernels and distributions where such relation holds, which they refer to as ``conjugacy''.
    In this paper, we focus explicitly on the Gaussian and Mat\'ern kernel, for their being widely used in minimum MMD estimation~\citep{briol2019statistical,cherief2022finite}, leaving other kernels as future work.
\end{rem}

\endgroup
 

\cleardoublepage
\phantomsection
\addcontentsline{toc}{chapter}{Bibliography}

\bibliography{main}

\end{document}